%% file: arxiv_clean.tex
\documentclass{article}

\usepackage[numbers]{natbib}
\usepackage[eandd, final]{neurips_2026}

\usepackage[utf8]{inputenc} 
\usepackage[T1]{fontenc}    
\usepackage{hyperref}       
\usepackage{url}            
\usepackage{booktabs}       
\usepackage{amsfonts}       
\usepackage{nicefrac}       
\usepackage{microtype}      
\usepackage{xcolor}         
\usepackage[pdftex]{graphicx}
\usepackage[most]{tcolorbox}
\usepackage{enumitem}
\usepackage{wrapfig}
\usepackage{caption}
\usepackage{subcaption}
\usepackage{multirow}
\usepackage{threeparttable}
\usepackage{array}
\usepackage{graphicx}
\usepackage{pifont}
\usepackage{titletoc}
\usepackage{xspace}

\newcommand{\modelicon}[1]{%
  \raisebox{-0.2\height}{\includegraphics[height=0.9em]{images/company-logos/#1}}%
}

\definecolor{teal}{RGB}{0,170,170}
\definecolor{blue}{RGB}{0,0,255}
\newcommand{\positive}[1]{\textcolor{blue}{#1}}
\newcommand{\negative}[1]{\textcolor{red}{#1}}

\newcommand{\camready}[1]{\textcolor{black}{#1}}

\newcommand{\xhdr}[1]{%
  \noindent\textbf{#1.}
}
\newcommand{\dashboard}{\raisebox{-1.5pt}{\includegraphics[height=1.05em]{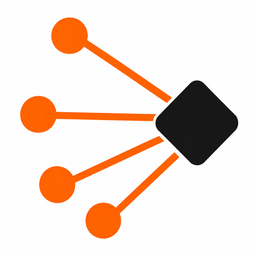}}\xspace}
\newcommand{\huggingface}{\raisebox{-1.5pt}{\includegraphics[height=1.05em]{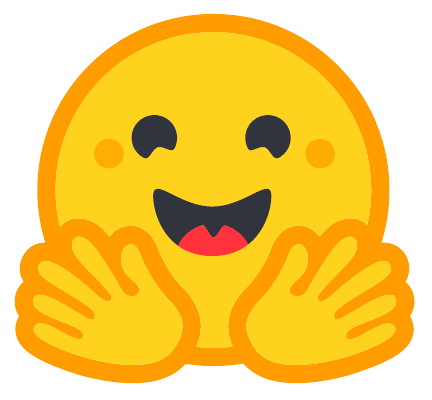}}\xspace}
\newcommand{\github}{\raisebox{-1.5pt}{\includegraphics[height=1.05em]{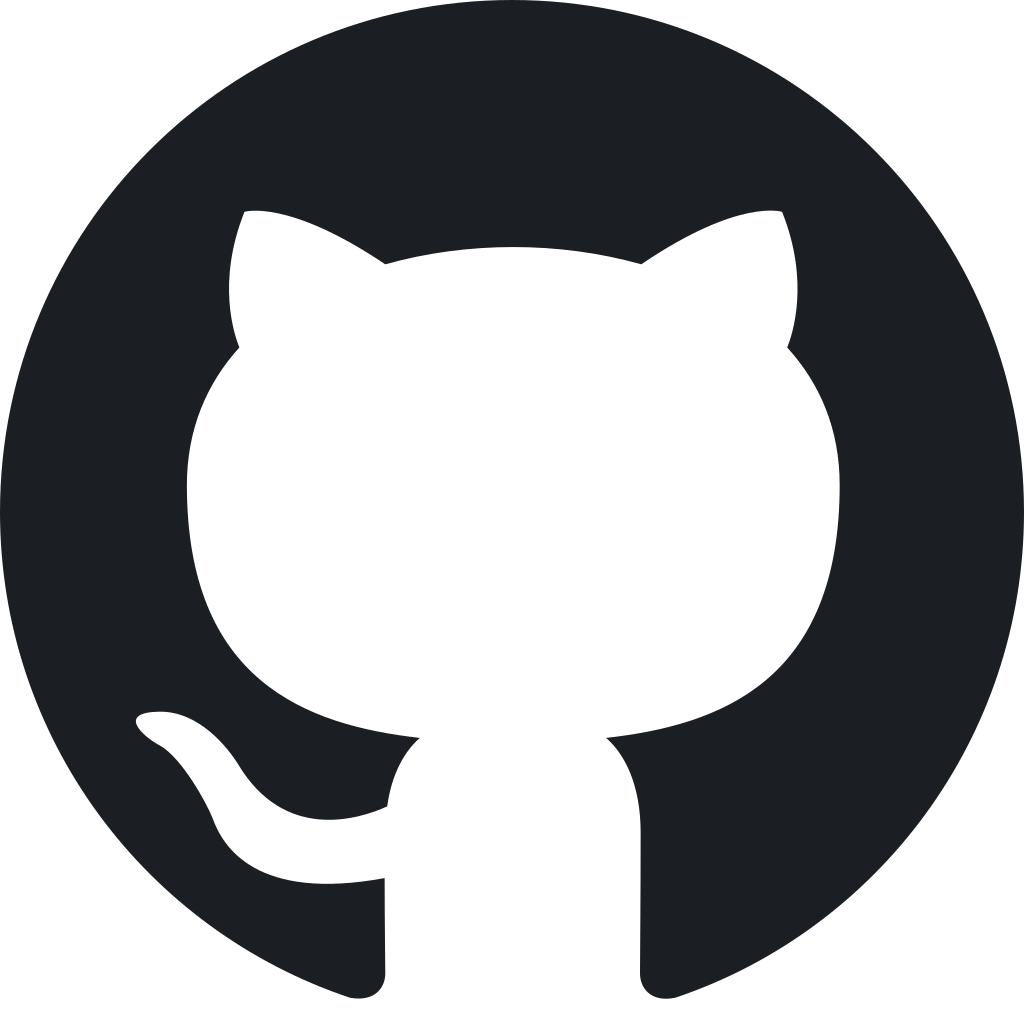}}\xspace}

\newcommand{\cmark}{\textcolor{green!60!black}{\ding{51}}}
\newcommand{\xmark}{\textcolor{red!70!black}{\ding{55}}}

\title{Can AI Agents Synthesize Scientific\\Conclusions in Health?}

\author{
    \textbf{Hayoung Jung$^{\spadesuit}$} \quad
    \textbf{Pedro Viana Diniz$^{\clubsuit}$} \quad
    \textbf{José Reinaldo Corrêa Roveda$^{\clubsuit}$} \\
    \textbf{Abner Fernandes da Silva$^{\clubsuit}$} \quad
    \textbf{Haeun Jung$^{\heartsuit}$} \quad 
    \textbf{Enoch Tsai$^{\diamondsuit}$} \quad  \\
    \textbf{Aleksandra Korolova$^{\spadesuit}$\thanks{Jointly advised this work.}} \quad
    \textbf{Manoel Horta Ribeiro$^{\spadesuit}$\footnotemark[1]} \\
    \textnormal{$^{\spadesuit}$Princeton University \quad $^{\clubsuit}$}Universidade Federal de Minas Gerais \\ \textnormal{$^{\heartsuit}$Stony Brook University \quad $^{\diamondsuit}$Hackensack Meridian School of Medicine}
    \vspace{0.2em} \\
    \texttt{\{hayoung, korolova, manoel\}@cs.princeton.edu}
    \vspace{0.2em} \\
    \github \textbf{Code}: \url{https://github.com/hayoungjungg/SciConBench}
    \vspace{0.2em} \\
    \huggingface \textbf{\textsc{SciConBench} Dataset}:~\href{https://huggingface.co/datasets/hayoungjung/SciConBench}{\texttt{hayoungjung/SciConBench}}
    \vspace{0.2em} \\
    \dashboard \textbf{Dashboard}: \url{https://sciconbench.cs.princeton.edu/}
} 

\begin{document}

\maketitle
\setcounter{footnote}{0}

\begin{abstract}
Scientific AI agents increasingly retrieve evidence, reason across sources, and synthesize conclusions used in consequential decisions. Yet, their ability to do so in high-stakes domains such as health remains unclear. We introduce \textsc{SciConBench}, a large-scale \textit{live} benchmark of \camready{9.21K} questions and expert-written conclusions from systematic reviews to evaluate open-domain scientific conclusion synthesis \camready{in health and medical domains}. The benchmark draws on an expert-validated automated evaluation pipeline that decomposes conclusions into atomic facts and measures correctness and comprehensiveness via factual precision and recall. To mitigate data leakage, we further introduce \textsc{SciConHarness}, a \textit{clean-room} evaluation harness that equips agents with controlled web interaction to ensure valid measurement. Evaluating 8 frontier models and deep research agents, we find that factual quality remains low: under clean-room settings, the best agent achieves only a factual F1 of 0.337. Our clean-room setting \textit{consistently} reduces performance relative to unconstrained evaluation, suggesting that leakage \textit{inflates} estimates of models' true synthesis capabilities. Finally, we audit consumer-facing agents (e.g., Google AI Overview, OpenEvidence) and find they frequently generate incomplete and sometimes contradictory conclusions, even when the ground-truth answer is available. Overall, our results show that reliable synthesis of scientific conclusions \camready{in health} remains an open challenge, and that \textit{clean-room} evaluation is essential for assessing open-domain AI agents.
\end{abstract}

\section{Introduction}

AI agents are transforming how individuals and institutions access and act on scientific knowledge \cite{anthropic2026advancing, openai2026_ai_healthcare_ally, yang2025adoptionusageaiagents}. Unlike traditional search engines (e.g., Google Search), which retrieve relevant documents and leave the synthesis to users, agentic systems from Anthropic \cite{anthropic_claude_research_blog}, Google DeepMind \cite{google_gemini_deep_research}, OpenAI \cite{openai_deep_research_guide}, and Perplexity \cite{perplexity_deep_research_2025} increasingly synthesize conclusions from scientific evidence. They retrieve relevant evidence from the open web, filter irrelevant sources, reconcile conflicting findings, assess the evidence quality, and produce a long-form expert-level conclusion. This long-horizon task of \textit{scientific conclusion synthesis} is increasingly delegated to such systems, accelerating decision-making and shaping decisions in health, science, and policy \cite{openai2026_ai_healthcare_ally, yang2025adoptionusageaiagents}.

One of the most consequential areas for scientific synthesis is health, and its impact is already evident in practice. OpenAI reports \textit{billions} of weekly ChatGPT messages concern healthcare, with 40 million daily users, including the general public, many of whom trust AI-generated health information, as well as physicians, who rely on AI for symptom and treatment exploration \cite{openai2026_ai_healthcare_ally}. More recently, specialized platforms like OpenEvidence serve as clinical AI copilots for high-stakes decision-making, reporting over 200 million AI-powered health consultations and widespread use among U.S. clinicians \cite{openevidence_about}. 

\begin{figure}[t]
    \centering
    \small
    \includegraphics[width=\linewidth]{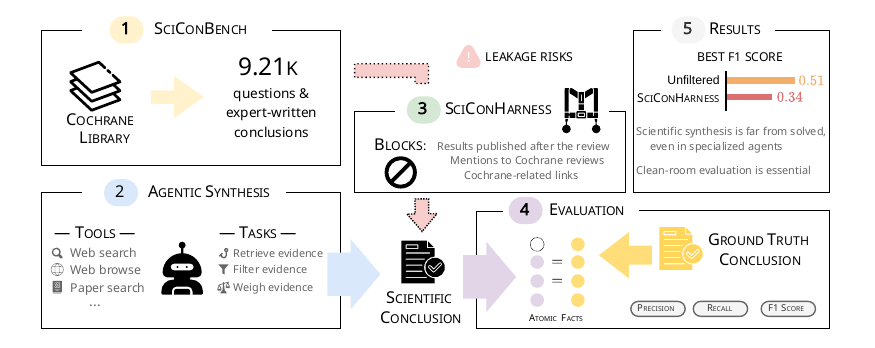}
    \caption{
    \small{
    \textbf{Overview.} (1) We construct \textsc{SciConBench}, a live benchmark of \camready{9.21K} \camready{health and medical} questions and expert-written conclusions. (2) The benchmark evaluates AI agents' capability for scientific synthesis by using web tools. (3) \textsc{SciConHarness} enforces clean-room evaluation by blocking ground-truth artifacts. (4) Generated conclusions are evaluated against ground-truth references using an expert-validated pipeline that decomposes both into atomic facts and computes factual precision, recall, and F1. (5) Results suggest that frontier systems achieve low factual F1 under clean-room evaluation, highlighting the difficulty of reliable scientific conclusion synthesis.}
    }
    \label{fig:placeholder}
\end{figure}

However, prior work falls short in evaluating AI agents on the full \textit{long-horizon} task of synthesizing \textit{long-form} scientific conclusions from the \textit{open web}. Existing works focus on intermediate artifacts, such as retrieval and citation grounding \cite{ajith-etal-2024-litsearch, gao-etal-2023-enabling, liu-etal-2023-evaluating}, summarization \citep{joseph-etal-2024-factpico, takeshita-etal-2024-aclsum, zhang-etal-2025-massw}, short-form factuality \cite{wei2024measuring, wei2025browsecomp}, or multiple-choice QA \cite{jin-etal-2019-pubmedqa, pampari-etal-2018-emrqa, wadden-etal-2020-fact, wadden-etal-2022-scifact}---rather than \textit{scientific conclusions}. 
As such, they fail to capture the core challenges and real-world complexity of scientific conclusion synthesis. Recent work moves closer by using expert-curated datasets to evaluate open-web synthesis \cite{Asai2024OpenScholarSS, du2025deepresearch, li2025reportbenchevaluatingdeepresearch, mialon2023gaia, patel2025deepscholar, ruan2025expertlongbench}. However, these benchmarks remain limited: they are often small due to the high cost of expert curation ($N \leq 100$), become outdated as new information emerges, and fail to address benchmark leakage, where models may be pre-trained on or retrieve ground-truth artifacts.

In this work, we introduce \textsc{SciConBench}, a \textit{live} benchmark of \camready{9.21K health and medical} questions and expert-written conclusions, derived from the \textit{Cochrane Database of Systematic Reviews} (\texttt{CDSR}). \textsc{SciConBench} evaluates whether agents can synthesize scientific conclusions from open-web evidence, and is updated monthly with new \texttt{CDSR} reviews to reduce benchmark leakage. To further mitigate leakage, we introduce \textsc{SciConHarness}, a \textit{clean-room} evaluation harness with controlled web search and browsing tools.
Finally, we develop a factual evaluation pipeline that decomposes generated conclusions into atomic facts\footnote{Defined as ``statements containing single pieces of information'' \cite{min-etal-2023-factscore, nenkova-passonneau-2004-evaluating}.} and uses LLM-based judges to measure \textit{factual precision} (correctness), \textit{factual recall} (coverage), and F1 (overall quality), showing strong agreement with expert judgments.

Evaluating 8 frontier models and deep research agents on \textsc{SciConBench}, we find scientific conclusion synthesis remains an open challenge under \textit{clean-room} evaluation: the best system, \texttt{o3-deep-research}, achieves only F1$=0.337$. Across systems, \textit{clean-room} evaluation reduces factual F1 by $0.020$--$0.172$ relative to unconstrained settings (agents can access ground-truth artifacts), indicating that much of the apparent performance arises from retrieving ground-truth artifacts rather than from genuine synthesis. This highlights the importance of \textit{clean-room} evaluation for valid measurement of open-domain AI agent capabilities. Finally, we audit consumer-facing agents (e.g., Google AI Overview, OpenEvidence) increasingly used in 
health contexts \cite{annenberg_asaph_2024, openevidence2026million, nytimes_google_ai_health_2024}. Despite access to ground-truth artifacts, these systems remain unreliable (F1$=0.361$--$0.522$), often generating incomplete and sometimes contradictory conclusions. Our main contributions are:

\vspace{-1mm}
\begin{enumerate}[leftmargin=1.5em]
    \item We introduce \textbf{\textsc{SciConBench}}, a \textit{live} benchmark of \camready{9.21K} questions paired with expert-written scientific conclusions, capturing real-world, open-domain scientific synthesis tasks \camready{in health}.
    
    \item We develop \textbf{\textsc{SciConHarness}}, a \textit{clean-room} evaluation harness that provides controlled web tools for AI agents, mitigating leakage and enabling valid measurement of synthesis capabilities. 
    
    \item Using our \textit{expert-validated} \textbf{factual evaluation pipeline} that decomposes conclusions into facts and measures \textit{factual precision} and \textit{recall}, our benchmark evaluation of frontier models and deep research agents reveals that \textbf{reliable scientific conclusion synthesis remains unresolved.}
    
    \item We \textit{audit} widely-deployed consumer-facing agents, including Google AI Overview and OpenEvidence, and find \textbf{they synthesize incomplete and sometimes contradictory scientific conclusions}, raising concerns for high-stakes decision-making relying on them in real-world health contexts.    
\end{enumerate}

\section{The \textsc{SciConBench} Dataset}\label{sec:sciconbench}

\textsc{SciConBench} leverages the \textit{Cochrane Database of Systematic Reviews} to evaluate models on the long-horizon task of scientific conclusion synthesis \camready{in health}: retrieving sources, appraising evidence quality, and integrating heterogeneous evidence to construct a long-form, expert-level conclusion. 

\xhdr{Background}\label{sec:background}
The \textit{Cochrane Database of Systematic Reviews (\texttt{CDSR})} is a peer-reviewed collection of systematic reviews that synthesize evidence to answer well-defined clinical and public health questions \citep{cochrane_cdsr_about}. Each review identifies and evaluates a body of relevant studies---from a few to hundreds of publications---to answer a well-defined clinical or public health question.
The review appraises the quality of the evidence, reconciles conflicting findings, and synthesizes the overall evidence into a concise, paragraph-long conclusion \citep{hevia-etal-2025-roboto2, cochrane_methodology_guide_2025, cochrane2026gradeing}. To ensure conclusions remain current as new scientific evidence emerges, \texttt{CDSR} re-evaluates the literature every two years and updates the conclusions \citep{CochraneHandbook2024ChapterIV}, though \citet{french2005investing} found that most conclusions remain stable over time. As the ``gold-standard'' for evidence-based synthesis \cite{smith2013cochrane}, \texttt{CDSR}'s expert-written conclusions inform real-world clinical decisions and health policy, making it a valuable data source for evaluating AI agents' ability to synthesize conclusions from the scientific literature.

\xhdr{Automated Data Collection: A \textit{Live} Benchmark}\label{sec:data-collection}
We construct a \textit{live} benchmark by drawing from systematic reviews from the regularly updated \texttt{CDSR}. As of January 1st, 2026, we collected 9,531 systematic reviews, of which 424 were withdrawn, yielding 9,107 valid reviews\camready{, that form the benchmark snapshot used in our experiments}.
Given growing concerns around benchmark leakage during the pre-training of frontier models \cite{xu2024benchmarking}, we design our benchmark to be continuously updated as new \texttt{CDSR} reviews are released.
In contrast to static benchmarks \citep{arora2025healthbench, li2025reportbenchevaluatingdeepresearch, mialon2023gaia}, this ensures timely evaluations of the latest agents, while mitigating leakage. \camready{As of October 2026, \textsc{SciConBench} has grown to 9,214 questions through monthly updates. See \S\ref{appendix:monthly-update} for details on the update process.}

\xhdr{Data Preprocessing}\label{sec:preprocessing}
We convert the expert-authored systematic reviews into structured question–answer (QA) evaluation units. For each review, we use the \textit{Objectives} as the basis for the question and employ the \textit{Authors' Conclusions} as the answer. See Figure \ref{fig:cochrane-example} for an example. The  \textit{Objectives} define the research question and scope the review aims to address---typically structured around the population, intervention, comparator, and outcomes (PICO) framework---while the \textit{Conclusions} provide the corresponding evidence-based synthesis, including key findings and their certainty.

\textit{Question Generation.}
Since \textit{Objectives} are typically written as declarative statements rather than questions, we transform them into clinically grounded questions with the PICO framework (Populations, Interventions, Comparisons, Outcomes), which is widely used to formulate clinical research questions and guide evidence retrieval~\cite{pico_framework}.
Using \texttt{gpt-5-chat}, we convert each \textit{Objective} into a sentence-style question, consistent with prior work showing that users prefer sentence-style queries over keyword-based inputs when using LLMs \cite{caramancion2024large, zhang2025source}. This formulation aligns with real-world usage, where clinicians and scientists increasingly ask scientific and medical questions to AI systems for decision-making~\cite{stanford_chatbot_physician_2025, goh2025gpt}. Applying this pipeline, we construct a QA-style benchmark comprising 9,214 samples \camready{(as of October 2026)} spanning nearly 30 years of systematic reviews across diverse scientific and clinical domains, from neonatal care to kidney disease. 
We provide additional details in \S\ref{appendix:question-gen-technical}, including prompts and an example question (Figure~\ref{fig:question-generation-prompt}).

\textit{Validation.} We validate whether the generated questions faithfully reflect the intent and scope of each \texttt{CDSR} review with annotations by two medical students with extensive clinical research experience. Given the generated question, \textit{Objectives}, and  \textit{Background} of the \texttt{CDSR} review, annotators evaluate question quality along three dimensions grounded in prior works \cite{kaur2025whosaskingsimulatingrolebased, mir-etal-2019-evaluating, park-etal-2024-valuescope}: Faithfulness, PICO Completeness, and Clarity \& Answerability. In the calibration phase, the annotators label 10 questions to validate the task guidelines and resolve disagreements. They then independently label an additional 10 questions to assess reliability, measured using Gwet's AC1, which is robust to skewed label distributions \cite{ohyama2021statistical, wongpakaran2013comparison}. Agreement is high across dimensions (AC1: 0.76--1.00; see Table~\ref{tab:interrater}), comparable to or exceeding prior work \cite{kaur2025whosaskingsimulatingrolebased, park-etal-2024-valuescope}. Given high agreement, each annotator independently labeled 40 questions ($N=100$ total). We find generated questions to be faithful (92\%), PICO-complete (92\%), and clear \& answerable (96\%) \camready{(95\% margin of error $\approx\pm$4--5 pp)}. Appendix \S\ref{appendix:question-gen-validation} provides details, including annotation guidelines (Figure~\ref{fig:annotation_guidelines}) and interface (Figures~\ref{fig:generated-question-interface-part1}--\ref{fig:generated-question-interface-part2}).

\section{\textsc{SciConHarness}: Controlled Evaluation in the Clean-Room}\label{sec:clean-room}

Synthesizing \textit{scientific conclusions} with open-domain information access is challenging, as models may find sources that contain the already synthesized information wholesale, thus turning the harder synthesis task into a simpler retrieval task.
We address this challenge with \textsc{SciConHarness}, an MCP-based harness that provides controlled access to web search and browsing tools.
\textsc{SciConHarness} enforces a ``clean-room setting'' preventing models' access to ground-truth systematic reviews, while retrieving, integrating, and reasoning over scientific evidence from the open web \cite{hou2025model}.

\xhdr{Overview}
\textsc{SciConHarness} orchestrates the full model-tool interactions, executing iterative model-driven tool calls with web access, and appending tool outputs to the model context. 
The harness was built using Ai2's open-source \texttt{dr-agent-lib} \cite{shao2025drtulureinforcementlearning}, and supports iterative model-tool interaction to discover and obtain scientific evidence from the open web, following prior work~\cite{gao2025beyond, li2025webweaver, liu2025webexplorer, shao2025drtulureinforcementlearning}: (1) \texttt{google\_search} (query $\rightarrow$ top search results) via Serper API, (2) \texttt{web\_browse} (URL $\rightarrow$ page text) via Jina Browsing API, and (3) \texttt{paper\_search} (query $\rightarrow$ relevant paragraphs from open-access papers) via Semantic Scholar Full-Text Search API.\footnote{Links: \url{https://serper.dev/}, \url{https://jina.ai/reader/}, \url{https://api.semanticscholar.org/api-docs}} Like prior works \cite{shao2025drtulureinforcementlearning}, \texttt{google\_search} and \texttt{paper\_search} retrieve the top 10 results by default, though models may request more. For \texttt{web\_browse}, we use \texttt{gpt-5-mini} to summarize the raw webpage text, reducing context usage and cost \cite{li2025webweaver, shao2025drtulureinforcementlearning}.
For brevity, we leave implementation details in \S\ref{appendix:sciconharness-implementation}. 

\xhdr{The Clean-Room}
Analogous to controlled environments in the life sciences to prevent contamination \cite{achengineering_biotech_cleanroom}, we design \textsc{SciConHarness} to support a \textit{clean-room} evaluation protocol that mitigates benchmark leakage. To prevent access to the ground-truth artifacts (e.g., \texttt{CDSR} articles), we implement a filtering middleware in \textsc{SciConHarness} that inspects all tool outputs and enforces explicit \textit{clean-room} protocols before returning them to the model. Concretely, we filter outputs from \texttt{google\_search} and \texttt{paper\_search} using: (1) URLs from \texttt{CDSR} domains, (2) result titles containing ``cochrane'' or matching a \texttt{CDSR} review title, or (3) results published after the ground-truth review's publication date (to prevent indirect leakage from derivative content).
For \texttt{web\_browse}, we filter content containing either ``cochrane'' or the ground-truth \texttt{CDSR} review title. Our \textit{clean-room} protocol is conservative: it may occasionally filter legitimate content, but it ensures that the benchmark measures synthesis rather than memorization or shortcut retrieval.

\camready{Since \texttt{CDSR} periodically updates existing reviews, earlier versions may have different titles and thus occasionally evade the clean-room filters, exposing models to conclusions that may partially overlap with the reference answer. While this can be ecologically valid, as human experts updating a review also build on its preceding versions, we acknowledge that these versions and subsequent studies citing them can make the synthesis task easier, and such leakage cannot be \textit{fully} eliminated on the open web. As such, we treat the clean-room as an \textit{adaptive}, living rulebook.\footnote{We unexpectedly observed DeepSeek-V4-Pro using \texttt{web\_browse} to query the Europe PMC API with a memorized review DOI, retrieving the ground-truth article. We patched this loophole; the clean-room must evolve as agents find new shortcuts.}}

\xhdr{Validation} We validate \textsc{SciConHarness}'s \textit{clean-room} filtering with a random sample of $N{=}150$ tool outputs from logged tool calls from the full benchmark evaluations (\S\ref{sec:benchmark-result})
($50$ per tool; e.g., search result for \texttt{google\_search}, page text for \texttt{web\_browse}). These are stratified evenly between filtered and unfiltered cases.
We manually annotated each output, measuring false positives (over-filtering benign content) and false negatives (missed leakage), with the latter being more critical as they allow access to the reference conclusions. As shown in Table~\ref{tab:filtering-validation}, the filtering achieves high precision (0.88--1.00; avg.\ 0.933) and recall (0.957--1.00; avg.\ 0.972) across tools, indicating effective leakage mitigation with minimal over-filtering. Notably, \textit{all} ground-truth \texttt{CDSR} articles were successfully removed across all tools, demonstrating robust prevention of direct leakage. The remaining false negatives arise from \textit{indirect} leakage (e.g., news coverage regarding the ground-truth \texttt{CDSR} article).

\section{Measuring Factual Quality}\label{sec:metrics}

How to evaluate the \textit{factual quality} of scientific conclusions?
We decompose both generated and reference conclusions into \textit{atomic facts} (\S\ref{sec:atomic-fact-gen}).
Then, we compare the extent to which the generated claims: (1) are supported and non-contradictory to the reference (\textit{factual precision}) and (2) support facts necessary to answer the question (\textit{factual recall}; see  \S\ref{sec:factual-metrics}).
Finally, we scale our evaluation procedure using LLM-based judges (\S\ref{sec:factual-llm-judge}), finding strong agreement with domain experts.

\subsection{Decomposing Conclusions into Atomic Facts}\label{sec:atomic-fact-gen}

\xhdr{Pipeline Overview} We design a modular pipeline that transforms long-form scientific conclusions 
into high-quality, self-contained atomic facts. Building on prior works on long-form factuality evaluations \cite{liu-etal-2025-verifact, min-etal-2023-factscore, 10.5555/3737916.3740483}, our pipeline comprises six steps.
We tokenize paragraph-length conclusions into sentences (step 0: preprocessing) and decompose each into atomic facts using \texttt{gpt-5.1} (step 1:~decomposition).
Then, we decontextualize each fact to resolve implicit references (step 2: decontextualization) and ensure self-containment by rewriting incomplete facts to fill in missing references, comparisons, and conditions (step 3: incomplete fact rewriting).
Finally, we filter facts for relevance to the original question, retaining only those that directly contribute to answering it or provide necessary context (step 4:  relevance filtering) and
filter out redundant facts within each sentence (step 5: redundancy filtering).
For details, see \S\ref{appendix:afg-pipeline} and Figures \ref{fig:decomposition-prompt}--\ref{fig:redundant-fact-prompt}.

\textit{Cost Consideration.} Since generating atomic facts from paragraph-length conclusions is financially and computationally expensive at scale, the pipeline is designed to be \textit{modular}, enabling per-step model selection and hyperparameter selection to balance quality and cost. We allocate higher-capability models (e.g., \texttt{gpt-5.1}) to early, formative stages (steps 1--2) to ensure high-quality initial facts, and smaller models (e.g., \texttt{gpt-5-mini}) to later classification-style steps (steps 3--5). On \texttt{CDSR} conclusions, this reduces cost from \$0.35 to \$0.13 per instance. Full design decisions are in \S\ref{appendix:afg-cost}.

\xhdr{Validation} We evaluate the quality of generated facts via expert annotations from two medical doctors with substantial clinical practice and research experience. Given a source sentence, its paragraph context, and its extracted atomic facts, annotators assess fact quality along dimensions grounded in prior work~\cite{kaur2025whosaskingsimulatingrolebased, liu-etal-2025-verifact, mir-etal-2019-evaluating, park-etal-2024-valuescope}: \textit{Faithfulness} and \textit{Completeness} at the fact-level (i.e., per fact), and \textit{Comprehensiveness} and \textit{Redundancy} at the sentence-level (i.e., per sentence and its set of atomic facts). See \S\ref{appendix:afg-validation} for details, including annotation guidelines (Figure~\ref{fig:annotation-guidelines-atomic-facts}) and interface (Figure~\ref{fig:afg-interface}).
We find agreement to be high across all dimensions (AC1: 0.597–0.955), comparable to prior work (Table~\ref{tab:inter-annotator-afg}). Given high agreement, annotators independently label 90 sentences each, stratified-sampled from generated and reference conclusions, yielding a total of $N_{\text{sent}}{=}200$ sentences with $N_{\text{facts}}{=}469$. We find that generated facts are largely faithful (96.4\%), complete (96.0\%), comprehensive (98.0\%), and non-redundant (90.5\%); see Table~\ref{tab:annotation-quality-afg}.

\subsection{Factual Precision and Recall Metrics}\label{sec:factual-metrics}
Scientific conclusions should be accurate and comprehensive \cite{tang2023evaluating}. Thus we define two metrics to measure factual quality, capturing correctness and coverage \cite{liu-etal-2025-verifact, min-etal-2023-factscore}. Following prior work \cite{liu-etal-2025-verifact, min-etal-2023-factscore, wadden-etal-2022-scifact}, we adopt a source-grounded view of factuality, defining correctness of statements with respect to a trusted reference---here, \texttt{CDSR} reviews, a ``gold standard'' in evidence-based science~\cite{smith2013cochrane}.

\textit{Factual precision} measures the extent to which facts from generated conclusions are supported and non-contradictory with respect to the \texttt{CDSR} review,~$R$. Let $\varepsilon_x$ represent the extracted facts from a generated conclusion~$x$. Each fact $e \in \varepsilon_x$ is labeled as \textsc{Contradicted}, \textsc{Supported}, or \textsc{Not Supported}. We compute the factual precision of generated conclusion $x$ as: $(\frac{1}{|\varepsilon_x|} \sum_{e \in \varepsilon_x} \mathbf{1}[e \text{ is } \textsc{Supported by } R]) \cdot (1 - \frac{1}{|\varepsilon_x|} \sum_{e \in \varepsilon_x} \mathbf{1}[e \text{ is } \textsc{Contradicted by } R]).$ This formulation rewards conclusions whose facts are supported, while explicitly penalizing contradictions, consistent with prior work \cite{cheng2025facts, liu-etal-2025-verifact, min-etal-2023-factscore}.

\textit{Factual recall} measures the extent to which generated conclusions cover facts from the \textit{Authors’ Conclusions} of \texttt{CDSR} reviews, treated as the authoritative set required to answer the question. Let $x$ represent the generated conclusion and $\varepsilon_A'$ be the corresponding reference facts from the \textit{Authors’ Conclusions}, A. Each fact $e' \in \varepsilon_A'$ is labeled as \textsc{Supported} or \textsc{Not Supported}. We compute the factual recall of the generated conclusion $x$ as: $(\frac{1}{|\varepsilon_A'|} \sum_{e' \in \varepsilon_A'} \mathbf{1}[e' \text{ is } \textsc{Supported by } x])$. This formulation rewards conclusions that cover the reference facts necessary to answer the question.

\textit{Factual F1} measures the overall factual quality of the generated conclusions as the harmonic mean of factual precision and recall; thus, a high factual F1-score requires both high correctness (precision) and strong coverage (recall) of the generated conclusion.

\subsection{Measuring Factual Precision and Recall at Scale}\label{sec:factual-llm-judge}

Annotating for factual precision and recall is challenging, requiring annotators to understand \texttt{CDSR} systematic reviews and reason over complex clinical evidence with substantial domain expertise. However, with dozens of atomic facts per conclusion, manual evaluation is expensive and infeasible. We therefore carefully construct annotation guidelines, develop an expert-annotated gold-standard dataset, and validate LLM-based judges against it. Full details are provided in \S\ref{appendix:details-measuring-fact}.

\xhdr{Creating the Gold-Standard Dataset} 
For both factual precision and recall tasks, we develop annotation guidelines (\S\ref{appendix:fact-eval-annotation-task};  Figures~\ref{fig:annotation_guidelines-fact-precision}--\ref{fig:annotation_guidelines-fact-recall}) and conduct multiple rounds of annotation with two medical doctors, with a third independently adjudicating disagreements to produce consensus labels (\S\ref{appendix:fact-eval-annotation-procedure}). 
This process resulted in a gold-standard dataset of $N=129$ facts for precision and $N=119$ for recall, representing a substantial annotation effort, exceeding or matching the scale of prior expert-annotated evaluations of LLM judges \cite{cheng2025facts, 10.1093/jla/laae003, dammu-etal-2024-uncultured, jung-etal-2025-mythtriage, phutane2025ableistintersectionaldisabilitybias}. Experts reported an average annotation time of 6 minutes per fact. See annotation interfaces in Figures~\ref{fig:fact-eval-interface1}--\ref{fig:fact-eval-interface3}.

\begin{table}[t]
\centering
\footnotesize
\caption{
Percentage agreement (\%), Cohen's $\kappa$, and Gwet's AC1 between experts and the LLM judge for factual precision and recall using \texttt{gpt-5.4-mini}, the best-performing judge.
}
\label{tab:factual-agreement}

\setlength{\tabcolsep}{4pt}

\begin{tabular}{lcccccc}
\toprule
\textbf{Pairwise Annotators} 
& \multicolumn{3}{c}{\textbf{Factual Precision}} 
& \multicolumn{3}{c}{\textbf{Factual Recall}} \\
\cmidrule(lr){2-4} \cmidrule(lr){5-7}
& \% & $\kappa$ & AC1 & \% & $\kappa$ & AC1 \\
\midrule
\texttt{Expert A} -- \texttt{Expert B} 
& 0.691 & 0.519 & 0.545 
& 0.832 & 0.658 & 0.670 \\
\texttt{Expert A} -- \texttt{LLM}
& 0.707 & 0.526 & 0.579 
& 0.885 & 0.753 & 0.785 \\
\texttt{Expert B} -- \texttt{LLM}
& 0.683 & 0.497 & 0.541 
& 0.823 & 0.637 & 0.660 \\
\texttt{Avg.} \texttt{Expert} -- \texttt{LLM}
& 0.695 & 0.512 & 0.560 
& 0.854 & 0.695 & 0.723 \\
\bottomrule
\end{tabular}
\end{table}

\textit{Agreement and resulting dataset.}
Between the two experts, we observe moderate agreement for factual precision (Cohen's $\kappa=0.519$, Gwet's AC1$=0.545$) and substantial agreement for factual recall (Cohen's $\kappa=0.658$, Gwet's AC1$=0.670$) \cite{8d20e0b8-89d8-3d65-bcf5-8c19d56ec4ab}. 
Agreement is comparable to or exceeds prior work \cite{cheng2025facts, liu-etal-2025-verifact, min-etal-2023-factscore}, indicating that our annotation task is well-defined and yields reliable labels. See Table~\ref{tab:iaa-fact-eval} for the full list of agreement scores across annotation rounds.\footnote{Expert disagreement can reflect \textit{legitimate differences} in evaluations that reflect background and expertise~\cite{hwang2026deepresearchshallowevaluation}.} The resulting gold-standard dataset contains 19 \textsc{Contradicted}, 54 \textsc{Supported}, and 56 \textsc{Not Supported} labels for factual precision, and 48 \textsc{Supported} and 71 \textsc{Not Supported} labels for factual recall.

\xhdr{Validating LLM Judge} Using the gold-standard dataset, we validate LLM-based judges through extensive prompt design and systematic evaluation across three models (\texttt{gpt-5.4-mini}, \texttt{claude-haiku-4.5}, and \texttt{gemini-3-flash}), varying reasoning levels, temperature settings, and prompts (zero-shot, few-shot). In the few-shot setting, we include six annotated examples from our gold-standard dataset,\footnote{For each task, we exclude few-shot examples from evaluation to avoid data leakage.} following prior work \cite{Mittal_Jung_ElSherief_Mitra_De_Choudhury_2025}. We detail the input features and prompt design in \S\ref{appendix:llm-judge-features}--\ref{appendix:llm-judge-prompt} and show prompts in Figures~\ref{fig:fact-precision-prompt}--\ref{fig:fact-recall-prompt}.

\textit{Validation Results.} 
For both tasks, \texttt{gpt-5.4-mini} achieves the strongest performance in its best configurations (macro F1 of 0.837 for precision and 0.868 for recall), which we use in our downstream evaluation.\footnote{Factual precision: few-shot prompt, no reasoning, temp 0.2; factual recall: zero-shot prompt, no reasoning, temp 1.0.} In addition, \texttt{gpt-5.4-mini} demonstrates strong alignment with the expert annotators. As shown in Table~\ref{tab:factual-agreement}, agreement between \texttt{gpt-5.4-mini}  as a judge and individual experts is comparable to---and in some cases exceeds---the agreement between the experts themselves. \texttt{gpt-5.4-mini} also passes the \textit{Alternative Annotator Test} \cite{calderon-etal-2025-alternative}, a leave-one-out statistical test for evaluating substitute annotators, with a winning rate of 1.0 on both tasks, statistically supporting its use as a reliable substitute annotator. Together, high task performance and strong agreement with experts validate both the quality of our prompts and the use of frontier LLMs as reliable evaluators for factual precision and recall. To identify common failure modes and areas to improve LLM judge accuracy and reliability, we conduct an error analysis of the LLM judge in \S\ref{appendix:llm-judge-error-analysis}. Tables~\ref{tab:fact-precision-full}--\ref{tab:fact-recall-full} present the full evaluation results for both factual precision and recall, with details in \S\ref{appendix:llm-judge-result}.

\section{Evaluation Details}\label{sec:evaluation-details}

\xhdr{Overview} We benchmark eight state-of-the-art models and deep research agents equipped with \textsc{SciConHarness} on  \textsc{SciConBench}. We evaluate three frontier LLMs at the time of our experiment: \texttt{gpt-5.1} \cite{openai_gpt51_model}, \texttt{claude-sonnet-4.5} \cite{anthropic_system_prompts_release_notes}, and \texttt{gemini-3-pro} \cite{google_gemini3pro_vertexai}---under three \textsc{SciConHarness} settings: (1) \textit{base} (parametric-only, no tools), (2) \textsc{SciConHarness} \textit{tools} (without the clean-room protocol, allowing retrieval access to ground-truth artifacts), and (3) \textsc{SciConHarness} \textit{tools + clean-room} (filters ground-truth leakage). These settings enable comparisons among parametric models, nonparametric models with unrestricted retrieval, and nonparametric models under our controlled \textit{clean-room} retrieval. We also evaluate other models and deep research agents, including Ai2's open-source \texttt{DR Tulu} \cite{shao2025drtulureinforcementlearning}, OpenAI's \texttt{o3-deep-research} \cite{openai_o3_deep_research} and \texttt{o4-mini-deep-research} \cite{openai_o4mini_deep_research}, and Perplexity's \texttt{sonar-deep-research} \cite{perplexity_sonar_deep_research} and \texttt{sonar-reasoning-pro} \cite{perplexity_sonar_reasoning_pro}.\footnote{We only consider closed-source deep research agents that provide an API and support integration with \textsc{SciConHarness} at the time of writing, excluding Gemini Deep Research and Grok Deep Research.} As these agents natively use tools, we evaluate them under \textit{tools} and \textit{tools + clean-room} settings.

\xhdr{Setup} Across all evaluations, we use the same system prompt (Figure~\ref{fig:model-instruction-prompt}), except for \texttt{DR Tulu}, for which we adapt their default system prompt (see Figure~\ref{fig:dr-tulu-instruction-prompt}). All systems must produce a paragraph-length synthesized conclusion. We use default, recommended hyperparameters (e.g., temperature) and the highest available reasoning level. To support long-form synthesis across potentially hundreds of web sources, we do not limit the number of tool calls.%
\footnote{Except for \texttt{o3-deep-research} and \texttt{o4-mini-deep-research}, which were capped at $N{=}30$ tool calls per query due to cost constraints; only 2\% of their queries reached this limit during evaluation.} 
Some provider-hosted agents 
do not support direct integration with \textsc{SciConHarness}, requiring alternative strategies to enforce our \textit{clean-room} protocol, e.g., remote MCP endpoints. See \S\ref{appendix:full-evaluation-details} for details.

\xhdr{Data} To mitigate data contamination and benchmark leakage from pretraining, we obtain $N=268$ samples from \textsc{SciConBench} \camready{(January 1st, 2026 snapshot of 9,107 samples)}, \camready{comprising \textit{all}} \texttt{CDSR} reviews published after the latest model knowledge cutoff \camready{among evaluated models} (i.e., the end of January 2025 for \texttt{gemini-3-pro})\camready{, so that no evaluated model could have been pre-trained on them}. For benchmark performance comparison, this sample size provides sufficient power to detect a statistically significant difference between two model performances in factual F1-scores of at least $\Delta \approx 0.037$ at $\alpha=0.05$ with power $0.8$ (see \S\ref{appendix:power-analysis} for the power analysis). After generating conclusions, we decompose both reference and generated conclusions into atomic facts (\S\ref{sec:atomic-fact-gen}) and evaluate factual precision, recall, and F1 using our expert-validated LLM judge. See \S\ref{appendix:cost-analysis} for cost analysis and Tables~\ref{tab:cost-breakdown-querying}--\ref{tab:fact-cost-breakdown} for the cost breakdowns of the end-to-end benchmark evaluation.

\begin{table*}[t]
\centering
\small
\caption{Benchmark performance of models and deep research (\textit{DR}) across \textsc{SciConHarness} settings. We report the macro factual precision, recall, and F1 (variance in parentheses). $\dagger$ indicates evaluation on a subset ($N=100$ of 268) due to high cost (>\$2/query). \textbf{Bold} denotes the best performance without clean-room, and \underline{underline} with clean-room. \textbf{$\Delta_{\text{Tools}}$F1} is the F1 change from Base to \textsc{SciConHarness}, and \textbf{$\Delta_{\text{Clean}}$F1} denotes the change when adding clean-room to \textsc{SciConHarness}.}
\setlength{\tabcolsep}{5pt}
\resizebox{\textwidth}{!}{
\begin{tabular}{lccccc}
\toprule
 & \textbf{Factual Precision (Var)} & \textbf{Factual Recall (Var)} & \textbf{Factual F1 (Var)} & \textbf{$\Delta_{\text{Tools}}$ F1} & \textbf{$\Delta_{\text{Clean}}$ F1} \\
\midrule

\textcolor{gray}{\footnotesize \textit{Base Models (No Tools)}} \\
\modelicon{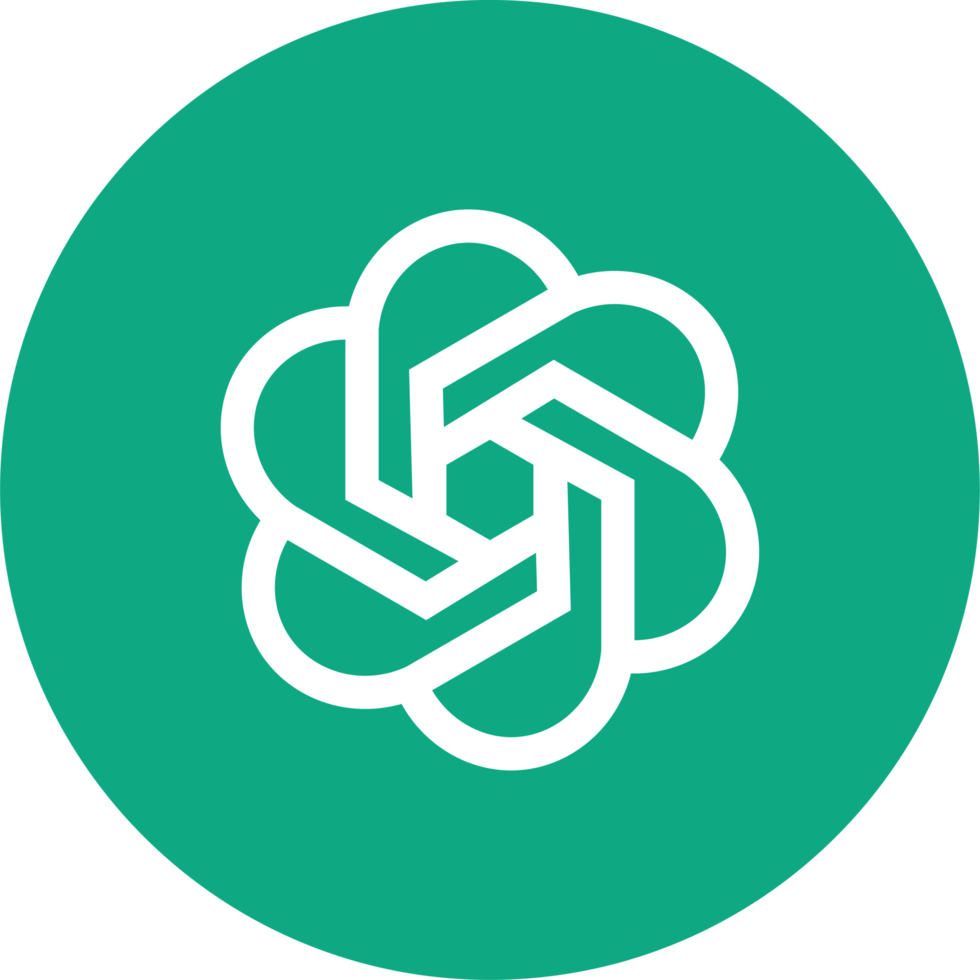}~\texttt{gpt-5.1} & 0.366 (0.019) & 0.382 (0.057) & 0.332 (0.027) & -- & -- \\
\modelicon{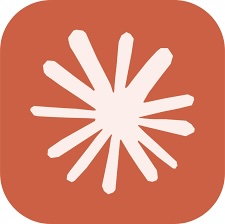}~\texttt{claude-sonnet-4.5} & 0.464 (0.025) & 0.270 (0.049) & 0.291 (0.037) & -- & -- \\
\modelicon{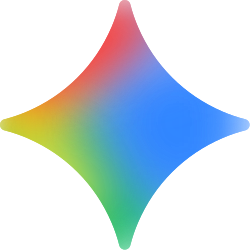}~\texttt{gemini-3-pro} & 0.339 (0.032) & 0.246 (0.045) & 0.239 (0.029) & -- & -- \\

\midrule
\textcolor{gray}{\footnotesize\textit{Models (\textsc{SciConHarness})}} \\
\modelicon{openai_icon}~\texttt{gpt-5.1} & 0.329 (0.017) & 0.446 (0.062) & 0.344 (0.026) & \positive{+0.012} & -- \\
\modelicon{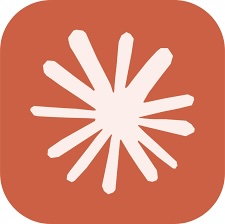}~\texttt{claude-sonnet-4.5} & 0.435 (0.043) & 0.409 (0.080) & 0.382 (0.054) & \positive{+0.091} & -- \\
\modelicon{gemini_icon}~\texttt{gemini-3-pro} & 0.311 (0.035) & 0.222 (0.048) & 0.213 (0.035) & \negative{$-$0.025} & -- \\
\modelicon{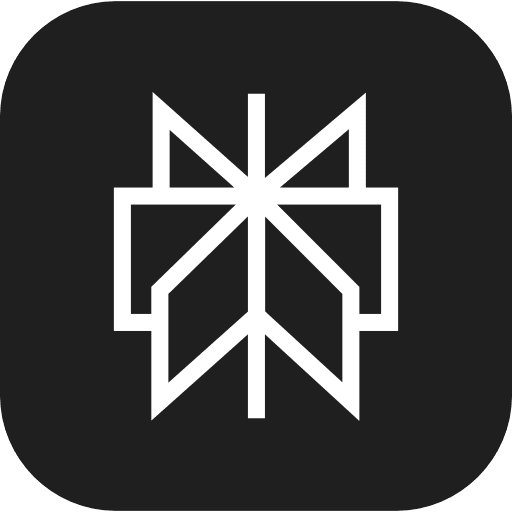}~\texttt{sonar-reasoning-pro} & 0.547 (0.064) & 0.363 (0.084) & 0.392 (0.075) & -- & -- \\

\midrule
\textcolor{gray}{\footnotesize\textit{Models (\textsc{SciConHarness} + clean-room)}} \\
\modelicon{openai_icon}~\texttt{gpt-5.1} & 0.294 (0.017) & \underline{0.408} (0.065) & 0.300 (0.024) & -- & \negative{$-$0.044} \\
\modelicon{claude_icon}~\texttt{claude-sonnet-4.5} & 0.350 (0.020) & 0.329 (0.057) & 0.297 (0.030) & -- & \negative{$-$0.085} \\
\modelicon{gemini_icon}~\texttt{gemini-3-pro} & 0.294 (0.034) & 0.206 (0.042) & 0.194 (0.029) & -- & \negative{$-$0.020} \\
\modelicon{perplexity_icon}~\texttt{sonar-reasoning-pro} & 0.384 (0.032) & 0.205 (0.044) & 0.220 (0.035) & -- & \negative{$-$0.172} \\

\midrule
\textcolor{gray}{\footnotesize\textit{DR (\textsc{SciConHarness})}} \\
\modelicon{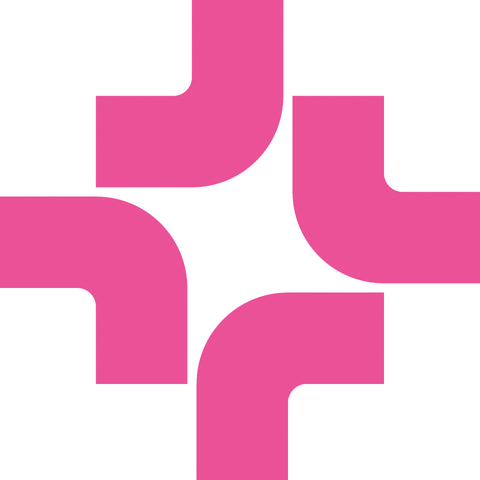}~\texttt{DR Tulu} & 0.308 (0.042) & 0.178 (0.034) & 0.175 (0.027) & -- & -- \\
\modelicon{perplexity_icon}~\texttt{sonar-deep-research$\dagger$} & 0.383 (0.028) & 0.396 (0.066) & 0.351 (0.041) & -- & -- \\
\modelicon{openai_icon}~\texttt{o4-mini-deep-research} & 0.593 (0.044) & 0.386 (0.063) & 0.427 (0.054) & -- & -- \\
\modelicon{openai_icon}~\texttt{o3-deep-research} & \textbf{0.628} (0.045) & \textbf{0.483} (0.069) & \textbf{0.508} (0.051) & -- & -- \\

\midrule
\textcolor{gray}{\footnotesize\textit{DR (\textsc{SciConHarness} + clean-room)}} \\
\modelicon{ai2_icon}~\texttt{DR Tulu} & 0.259 (0.038) & 0.168 (0.034) & 0.145 (0.023) & -- & \negative{$-$0.030} \\
\modelicon{perplexity_icon}~\texttt{sonar-deep-research$\dagger$} & 0.357 (0.036) & 0.243 (0.047) & 0.237 (0.034) & -- & \negative{$-$0.115} \\
\modelicon{openai_icon}~\texttt{o4-mini-deep-research} & \underline{0.467} (0.028) & 0.298 (0.051) & 0.315 (0.039) & -- & \negative{$-$0.113} \\
\modelicon{openai_icon}~\texttt{o3-deep-research} & 0.441 (0.033) & 0.342 (0.054) & \underline{0.337} (0.035) & -- & \negative{$-$0.170} \\
\bottomrule
\end{tabular}}
\label{tab:full-benchmark}
\end{table*}

\section{Results}\label{sec:benchmark-result}

\textbf{Models and deep research agents have substantial room for improvement.} Table \ref{tab:full-benchmark} shows the benchmark performance of models and deep research agents across \textsc{SciConHarness} settings. Across all systems, factual F1-score remains far from reliable for scientific conclusion synthesis \camready{in health}. Even in the favorable setting without the clean-room, no system exceeds 0.63 on any metric. The best-performing \texttt{o3-deep-research} achieves the highest precision (0.628), recall (0.483), and F1 (0.508), significantly outperforming the second-best \texttt{o4-mini-deep-research} (F1 0.427; paired t-test: $t(267)=6.1$, $p<0.001$, Cohen’s $d=0.37$). Under clean-room evaluation, which better isolates true synthesis capability, \texttt{o4-mini-deep-research} achieves the highest precision (0.467), \texttt{gpt-5.1} the highest recall (0.408), and \texttt{o3-deep-research} the highest F1 (0.337), again significantly outperforming \texttt{o4-mini-deep-research} (F1 0.315; $t(267)=2.36$, $p<0.05$, $d=0.14$). \texttt{DR Tulu}, despite being the most cost-efficient fully open agent, shows the weakest performance. \camready{An approximate oracle upper bound, which submits each \textsc{CDSR} review's own \textit{Authors' Conclusions} as the model response, achieves a factual F1 of 0.95 (\S\ref{appendix:oracle}). While not an independent human baseline, this upper bound indicates a large gap between current agents and expert-written synthesis.} 

\textbf{At the conclusion level, factual quality issues were pervasive across models and deep research agents}: 44.8--84.0\% of generated conclusions contained at least one fact contradicting the reference \texttt{CDSR} review, and nearly all contained at least one fact not supported by the reference review (Table \ref{tab:response-level-precision}). The generated conclusions were also incomplete, failing to support and cover 55.4--84.7\% of reference facts from \texttt{CDSR} reviews (Table~\ref{tab:label-distribution}). These findings indicate that current models and agents often produce scientific conclusions that are incomplete, contradictory, or not supported by the reference review. Our failure mode analysis shows that models often invert treatment effects, mischaracterize evidence quality, and generate overly broad conclusions lacking outcome-level specificity (\S\ref{appendix:failure-mode}--\ref{appendix:diagnostic-analysis}). As AI agents are increasingly used to synthesize evidence for clinical and scientific decision-making, such errors may distort high-stakes judgments \camready{in patient care and health}.

\textbf{Tool augmentations require precise use and effective evidence integration to improve synthesis.} Among base models, \texttt{gpt-5.1} achieves the highest F1 score (0.332) compared to \texttt{claude-sonnet-4.5} (0.291) and \texttt{gemini-3-pro} (0.239), highlighting limited synthesis capability from parametric knowledge alone. Adding \textsc{SciConHarness} tools (no clean-room) improves performance in most cases, but unevenly: \texttt{claude-sonnet-4.5} achieves large gains in recall (+0.139) and F1 (+0.091) with relatively efficient tool use (8.97 calls/query), despite a drop in precision ($-0.029$), while \texttt{gpt-5.1} shows only a marginal gain (+0.012 F1) despite heavier tool usage (14.05 calls/query; see Table \ref{tab:tool-calls-detailed}). \texttt{gemini-3-pro} degrades with tools and uses them least (5.59 calls/query), indicating poor integration. This demonstrates the challenges of \textit{scientific conclusion synthesis}: even without clean-room constraints---where agents can access ground-truth artifacts---strong performance requires not just access to tools, but disciplined, intentional tool use and effective evidence integration.

\textbf{Our clean-room evaluation \textit{consistently} attenuates performance.} Applying our clean-room protocol consistently reduces F1 by $0.020$–$0.172$ across all systems, even eliminating gains from unconstrained tool use. For example, \texttt{claude-sonnet-4.5} drops by $0.085$, nearly offsetting its $+0.091$ gain with tools, while \texttt{sonar-reasoning-pro} shows the largest decline ($-0.172$). Even top-performing \texttt{o3-deep-research} (F1 0.508 without clean-room) degrades sharply by $0.17$. These results indicate that much of the observed performance without clean-room constraints is driven by \textit{retrieval} of ground-truth artifacts rather than genuine \textit{synthesis}. In our experiment, we observe that agents actively exploit open-web access to \textit{retrieve} the ground-truth \texttt{CDSR} conclusion---even when instructed not to---shortcutting synthesis 
and creating leakage \cite{xu2024benchmarking}. By mitigating leakage with our clean-room evaluation, we enforce controlled evaluation that prevents conflating retrieval with synthesis, avoiding overestimation of true capability and maintaining construct validity \cite{zhou2026general}. \camready{The same pattern holds for four new open-weight models, where clean-room evaluation attenuates performance gains from unconstrained tool use (\S\ref{appendix:open-weight}).}

\textbf{\camready{Findings generalize to newer models.}} \camready{Open-weight models are competitive: with tools, \texttt{glm-5.2} (F1 0.469) and \texttt{deepseek-v4-pro} (0.445) outperform all systems in Table~\ref{tab:full-benchmark} except \texttt{o3-deep-research} (\S\ref{appendix:open-weight}). The latest frontier models show progress, with \texttt{gpt-6.1-sol} reaching a clean-room F1 of 0.411. However, under clean-room, no model exceeds 0.49 on any metric (\S\ref{appendix:latest-models}), indicating that reliable scientific conclusion synthesis remains far from solved.}

\subsection{\camready{Additional Analyses}}  For brevity, we summarize additional analyses from Appendix \S\ref{appendix:additional-analysis}, which are organized into four themes: (1) \camready{\textit{additional models}, including newer open-weight models (\S\ref{appendix:open-weight}) and the latest frontier models (\S\ref{appendix:latest-models});} (2) \textit{contextualizing performance}, \camready{including an approximate oracle upper bound (\S\ref{appendix:oracle})}, Pareto frontiers of performance vs.\ time and cost (\S\ref{appendix:performance-tradeoffs}), and label distribution (\S\ref{appendix:label-distribution}); (3) \textit{performance breakdowns} by conclusion length (\S\ref{appendix:length-metric-tradeoff}) \camready{, along with the stratification of benchmark performance by clinical domain and review type (\S\ref{appendix:stratification})}; and (4) \textit{agent behavior and failure modes}, including \textsc{SciConHarness} tool usage patterns (\S\ref{appendix:tool-usage}), failure mode analysis (\S\ref{appendix:failure-mode}), \camready{and diagnostic analysis of tool use, retrieval, and synthesis (\S\ref{appendix:diagnostic-analysis})}.

Tool usage varies substantially across AI agents, with OpenAI agents relying heavily on \texttt{google\_search} and \texttt{web\_browse}, while Claude and Gemini use \texttt{paper\_search} more. Across systems that use \texttt{google\_search} (excluding \texttt{DR Tulu}), \texttt{google\_search} exhibits the highest clean-room filtering rates (49.6\%--81.8\%; \S\ref{appendix:tool-usage}), highlighting the importance of clean-room evaluation to mitigate benchmark leakage. Failure mode analysis further reveals that models often invert treatment effects, mischaracterize evidence quality, and generate overly broad conclusions (\S\ref{appendix:failure-mode}). \camready{Most failures arise in synthesis, not retrieval: in a qualitative analysis of agent logs (\S\ref{appendix:diagnostic-analysis}), 18 of 20 logs show synthesis failures, such as overstating the certainty of ``very uncertain'' evidence or fabricating effect sizes absent from any retrieved source. Retrieval and source-selection errors (6 of 20) are less common but propagate into overly broad conclusions unsupported by the reference review.} Finally, longer conclusions generally trade higher recall for lower precision, indicating that simply generating longer outputs does not improve factual F1 and that agent quality, rather than verbosity, drives performance.

\subsection{Auditing Consumer-Facing Agents} 

Using \textsc{SciConBench}, we audit proprietary, consumer-facing agents increasingly used by laypeople and clinicians to synthesize scientific conclusions in high-stakes health contexts \cite{annenberg_asaph_2024, openevidence2026million, nytimes_google_ai_health_2024}. We evaluate Google AI Overview, Google AI Mode, and OpenEvidence on the same $N=268$ benchmark samples without the clean-room protocol, decompose the conclusions into facts, and evaluate them (\S\ref{sec:evaluation-details}). See \S\ref{appendix:audit-details} for details on data collection. 

\begin{wrapfigure}{r}{0.45\columnwidth}
    \vspace{-8pt}
    \centering
    \footnotesize
    \setlength{\tabcolsep}{2pt}
    \resizebox{\linewidth}{!}{
    \begin{tabular}{lccc}
    \toprule
    & \textbf{Precision} & \textbf{Recall} & \textbf{F1} \\
    \midrule
    Google AI Mode 
    & 0.443 (0.048) & 0.380 (0.077) & 0.361 (0.054) \\
    Google AI Overview  
    & 0.508 (0.044) & 0.367 (0.061) & 0.384 (0.048) \\
    OpenEvidence 
    & \textbf{0.580} (0.028) & \textbf{0.541} (0.070) & \textbf{0.522} (0.042) \\
    \bottomrule
    \end{tabular}
    }
    \caption{Performance of consumer-facing AI agents (precision, recall, F1; variance in parentheses).}
    \label{tab:consumer-agents}
    \vspace{-2pt}
    \end{wrapfigure}

\textbf{Consumer-facing agents generate unreliable scientific conclusions despite access to ground-truth artifacts.} As shown in Figure~\ref{tab:consumer-agents}, Google AI Mode (F1: 0.361) and Google AI Overview (F1: 0.384) perform poorly, with weak recall (0.380 and 0.367) indicating limited coverage of key reference facts. More concerningly, a substantial share of generated conclusions containing at least one fact contradicting the \texttt{CDSR} review: 56.3\% for Google AI Overview and 59\% for Google AI Mode (Table~\ref{tab:response-level-precision})---despite access to the ground-truth \texttt{CDSR} review. This suggests failures from these consumer-facing agents come from unreliable synthesis, not missing information. Given these agents' widespread use in health contexts \cite{nytimes_google_ai_health_2024}, these contradiction rates are particularly concerning. OpenEvidence performs better (F1: 0.522), achieving the highest precision (0.580) and recall (0.541). However, it remains far from reliable: 50.8\% of generated conclusions from OpenEvidence contain at least one fact contradicting the \texttt{CDSR} review (Table~\ref{tab:response-level-precision}) and cover only half (51.7\%) of the reference facts (Table~\ref{tab:label-distribution}). Thus, even the strongest agent, which is specialized for health, frequently omits critical information and occasionally contradicts established conclusions, which may pose a risk for high-stakes decision-making in real-world health contexts.

\section{Related Work}

\begin{table*}[t]
\centering
\small
\caption{
Comparison of \textsc{SciConBench} with prior benchmarks.}
\setlength{\tabcolsep}{5pt}
\resizebox{\textwidth}{!}{
\begin{tabular}{lcccccccc}
\toprule
\textbf{Benchmark} 
& \textbf{Size} 
& \textbf{Domain} 
& \textbf{Agentic}
& \textbf{Open-Web} 
& \textbf{Long-Form} 
& \textbf{Long-Horizon}
& \textbf{Live} 
& \textbf{Clean-Room} \\
\midrule

PubMedQA \cite{jin-etal-2019-pubmedqa} 
& 1K 
& Medical
& \xmark
& \xmark 
& \xmark 
& \xmark
& \xmark 
& \xmark \\

SciFact \cite{wadden-etal-2020-fact}
& 1.4K 
& Scientific Facts
& \xmark
& \xmark 
& \xmark 
& \xmark
& \xmark 
& \xmark \\

MedQA \cite{jin2021disease}
& 12.7K 
& Medical
& \xmark
& \xmark 
& \xmark 
& \xmark
& \xmark 
& \xmark \\

GAIA \cite{mialon2023gaia}
& 466
& General
& \cmark
& \cmark 
& \xmark 
& \cmark
& \xmark 
& \xmark \\

Humanity's Last Exam \cite{phan2025humanity}
& 2.5K
& Expert
& \xmark 
& \xmark 
& \xmark 
& \xmark
& \cmark 
& \xmark \\

HealthBench \cite{arora2025healthbench}
& 5K 
& Health 
& \xmark
& \xmark 
& \cmark 
& \xmark
& \xmark 
& \xmark \\

ExpertLongBench \cite{ruan2025expertlongbench}
& 1.05K 
& Expert 
& \xmark 
& \xmark 
& \cmark 
& \xmark
& \xmark 
& \xmark \\

ScholarQABench \cite{Asai2024OpenScholarSS}
& 1K 
& Literature Review
& \cmark
& \cmark 
& \cmark 
& \cmark
& \xmark 
& \xmark \\

ResearcherBench \cite{xu2025researcherbench}
& 65
& Research
& \cmark
& \cmark 
& \cmark 
& \cmark
& \xmark 
& \xmark \\

ReportBench \cite{li2025reportbenchevaluatingdeepresearch}
& 100 
& Research Report
& \cmark
& \cmark 
& \cmark 
& \cmark
& \xmark 
& \xmark \\

DeepResearch Bench \cite{du2025deepresearch}
& 100 
& Research Report
& \cmark
& \cmark 
& \cmark 
& \cmark
& \xmark 
& \xmark \\

DeepScholar-Bench \cite{patel2025deepscholar}
& 200 
& Literature Review
& \cmark
& \cmark 
& \cmark 
& \cmark
& \cmark 
& \xmark \\

\midrule

\textbf{\textsc{SciConBench} (Ours)}
& \textbf{\camready{9.21K}} 
& \textbf{Scientific Synthesis} 
& \cmark
& \cmark 
& \cmark 
& \cmark
& \cmark 
& \cmark \\

\bottomrule
\end{tabular}
}
\label{tab:related-work-comparison}
\end{table*}

\xhdr{Evaluations for Science, Health, and Factuality} 
LLMs are increasingly deployed in science and health contexts~\cite{costa2026public, tang2023evaluating}. To evaluate their capabilities, prior work has developed benchmarks for biomedical and scientific question answering~\cite{dasigi-etal-2021-dataset, jin-etal-2019-pubmedqa, vladika-etal-2024-medreqal}, clinical and scientific reasoning~\cite{10.5555/3618408.3619195, li2024mediq, sun-etal-2025-reasonmed}, scientific text summarization and simplification \cite{bakker-kamps-2024-cochrane, devaraj-etal-2021-paragraph, takeshita-etal-2024-aclsum}, risk of bias assessment~\cite{huang2025large, marshall-etal-2017-automating, https://doi.org/10.1002/cesm.70044, wang-etal-2025-measuring}, factuality evaluation~\cite{joseph-etal-2024-factpico, kaur-etal-2024-evaluating, kwiatkowski-etal-2019-natural, wadden-etal-2020-fact, wadden-etal-2022-scifact}, citation grounding and verifiability~\cite{funkquist-etal-2023-citebench, gao-etal-2023-enabling, liu-etal-2023-evaluating}, and literature review generation~\cite{antu2023using, cao2025automation, doi:10.1073/pnas.2411962122, mostafapour2024evaluating}. Concurrently, small-scale studies comparing AI-assisted and expert-written scientific summaries suggest that LLMs may help scale evidence communication and systematic review workflows~\cite{cadiente2024evaluating, DEVANE2025111894}. Other works examine whether LLMs can support scientific discovery and research ideation~\cite{gupta2025all, si2025ideation, si2024can}. However, existing benchmarks largely evaluate intermediate artifacts (e.g., risk of bias, citation quality) rather than the final scientific conclusion. They also fail to capture the realistic long-horizon task of scientific synthesis \camready{for health and medical questions} with current AI agents, which iteratively retrieve sources from the open web, reason across them, and integrate heterogeneous evidence to produce long-form, multi-source scientific conclusions.

\xhdr{Long-Horizon Synthesis Benchmarks for AI Agents}
Recent work has introduced increasingly capable deep research agents, including \texttt{OpenScholar} \cite{Asai2024OpenScholarSS}, \texttt{DR Tulu} \cite{shao2025drtulureinforcementlearning},  \texttt{OpenResearcher} \cite{zheng-etal-2024-openresearcher}, and \texttt{WebThinker} \cite{Li2025WebThinker}. In parallel, prior benchmarks have studied long-form QA and factuality \cite{fan-etal-2019-eli5, liu-etal-2025-verifact, min-etal-2023-factscore, stelmakh-etal-2022-asqa} and open-domain QA \cite{amouyal-etal-2023-qampari, 10.5555/3737916.3740483}, but are largely non-agentic or fail to reflect realistic long-horizon settings where complex open-ended questions require open-web tool use and synthesis across multiple sources. More recent agentic benchmarks \cite{cheng2025facts, dammu2026iagentbench, polzak2025can, wan2026deep, wei2024measuring, wei2025browsecomp, wu2025webwalker} begin to address this gap. However, as shown in Table \ref{tab:related-work-comparison}, many benchmarks remain small-scale, static, and poorly suited to evaluating robust synthesis from noisy open-web evidence. They often do not test whether agents can filter irrelevant or unreliable sources and produce high-quality long-form conclusions from heterogeneous evidence. Existing benchmarks also rarely control for benchmark leakage, where web-enabled agents retrieve ground-truth artifacts directly from the open web \cite{anthropic2026evalawareness}.

\section{Conclusion}

We introduce \textsc{SciConBench}, a large-scale \textit{live} benchmark for the long-horizon task of open-domain scientific conclusion synthesis \camready{in health}; \textsc{SciConHarness}, a \textit{clean-room} evaluation harness with controlled web tools to mitigate benchmark leakage; and an expert-validated factual evaluation pipeline based on atomic facts, factual precision, and recall. Through our benchmark evaluation and audits of deployed consumer-facing systems, we show that current frontier models and AI agents still struggle to synthesize accurate and comprehensive scientific conclusions. We hope this work guides the development of  AI agents that can more reliably synthesize scientific evidence from the open web and support trustworthy scientific and medical decision-making in high-stakes contexts.

\begin{ack}
This work was supported in part by the National Science Foundation grants CNS-1956435 and CNS-2344925, and by the Alfred P. Sloan Research Fellowship for A. Korolova. We thank Francesco Salvi, Alejandro Cuevas, Max Springer, Chung Peng Lee, Elijah Fullerton, Jane Castleman, Blossom Metevier, Jason Greenfield, the members of the Center for Information Technology Policy at Princeton University, and the anonymous reviewers for their insightful feedback and discussions.
\end{ack}

\bibliography{neurips_2026}
\bibliographystyle{neurips_2026}

\newpage
\appendix
\startcontents[appendix]


{\LARGE\bfseries Appendices\par}
\vspace{0.4cm}

{\small\printcontents[appendix]{}{1}{\setcounter{tocdepth}{2}}}
\newpage


\setcounter{table}{0}
\setcounter{figure}{0}
\renewcommand{\thetable}{S\arabic{table}}
\renewcommand{\thefigure}{S\arabic{figure}}
\renewcommand{\theHtable}{S\arabic{table}}   
\renewcommand{\theHfigure}{S\arabic{figure}}

\section{Discussions}\label{appendix:discussion}

\subsection{Limitations}

While \textsc{SciConBench} addresses current pitfalls of agentic benchmarks, some limitations remain: 

\textit{Focus on Science and Health.} Because \textsc{SciConBench} is derived from the \texttt{CDSR}, our benchmark primarily reflects scientific and health-related conclusion synthesis tasks grounded in evidence-based medicine. While these clinical and health domains are consequential and increasingly targeted by AI agents \cite{nytimes_google_ai_health_2024, openevidence2026million}, they do not fully capture the diversity of synthesis tasks across other disciplines (e.g., social science, legal). As such, performance on \textsc{SciConBench} may not generalize to other long-form synthesis in different domains. Nevertheless, \texttt{CDSR} provides a large-scale corpus of expert-written scientific conclusions spanning nearly 30 years and over \camready{9.21K} benchmark samples, enabling substantially broader and more realistic evaluations than many existing synthesis benchmarks.

\textit{Dependence on Cochrane Reviews.} Our benchmark treats \texttt{CDSR} reviews as the reference standard for scientific conclusions. While Cochrane reviews are widely regarded as a gold standard for evidence synthesis \cite{smith2013cochrane}, they are still products of human judgment and may inherit biases, including emphasis on Western clinical research practices. While systematic reviews themselves can become outdated as new evidence emerges \cite{doi:10.7326/0003-4819-147-4-200708210-00179}, French et al. \cite{french2005investing} found that only a small portion of updated reviews change conclusions, suggesting that stale conclusions are a small fraction of the \textsc{SciConBench} dataset. To mitigate this issue, \texttt{CDSR} recommends periodic literature updates approximately every two years and withdraws reviews deemed excessively outdated \cite{CochraneHandbook2024ChapterIV}. Our benchmark continuously updates with new reviews, and our evaluations focus on recent reviews published after model knowledge cutoffs to reduce data contamination from pre-training and conclusion staleness.

\textit{Residual benchmark leakage may still exist.} Although \textsc{SciConHarness} introduces a clean-room protocol to mitigate retrieval of ground-truth artifacts (in our validation, all ground-truth \texttt{CDSR} articles were filtered, with filtering recall of 0.957--1.00 across tools; Table \ref{tab:filtering-validation}), leakage may still occur indirectly in the noisy open-web environment. For example, derivative webpages (e.g., news coverage) or summaries may paraphrase Cochrane conclusions without explicitly mentioning the original review. As such, occasional residual leakage may still exist. 

\textit{Evolving Agentic Frameworks and Tooling.} \textsc{SciConHarness} uses the same set of agentic tools as prior works \cite{shao2025drtulureinforcementlearning}. While this enables controlled and reproducible evaluation, AI agents and tool ecosystems are evolving rapidly. Future systems may employ substantially different retrieval infrastructures, reasoning paradigms, memory mechanisms, or tool orchestration strategies beyond our current setup. \camready{Summarizing \texttt{web\_browse} outputs with \texttt{gpt-5-mini} also adds a layer of model dependence; we adopt this compaction because raw webpage content can quickly exhaust models' context windows, consistent with common practice in agentic systems~\cite{li2025webweaver, shao2025drtulureinforcementlearning}.} At the same time, our benchmark evaluation results (e.g., \texttt{o3-deep-research} scoring 0.508 F1) are broadly consistent with our audit of consumer-facing agents using proprietary harnesses (e.g., \texttt{OpenEvidence} scoring 0.522 F1), suggesting that the relatively low performance observed in \textsc{SciConBench} is unlikely to be solely an artifact of our harness design. We make \textsc{SciConHarness} fully open-source to support extensibility and enable future practitioners to improve upon the framework with new tools, reasoning processes, \camready{context-management and compaction methods,} and orchestration protocols.

\textit{LLM Error Cascades.} We use LLMs to generate \textsc{SciConBench} questions, decompose conclusions into atomic facts, and evaluate factual quality. While domain experts largely validate the quality of the generated questions (\S\ref{appendix:question-gen-validation}) and atomic facts (\S\ref{appendix:afg-validation}), the small error rates may potentially influence our downstream evaluations. In addition, our factual evaluation pipeline relies on expert-validated LLM judges (\S\ref{appendix:llm-judge-result}), but may still introduce errors that may propagate into downstream evaluations. To better understand this error, we conduct an error analysis of the LLM judge in \S\ref{appendix:llm-judge-error-analysis}. \camready{This analysis reveals two recurring failure modes, namely missing clinical synonymy, which could systematically bias scores differently across AI agents, e.g., if some paraphrase clinical terminology more freely. However, such ambiguities are not unique to LLM judges; for example, experts may also disagree when determining whether clinically synonymous outcomes are equivalent~\cite{hwang2026deepresearchshallowevaluation}. Despite this, our validation efforts show high agreement between the LLM judge and medical doctors, even exceeding agreement among the medical doctors themselves (Table~\ref{tab:factual-agreement}). These results suggest that the identified failure modes may not substantially undermine the overall validity of the evaluation.}

\camready{\textit{Limited Validation Sample Sizes.} Our expert validation sets are modest in size (e.g., $N{=}100$ generated questions), often due to the substantial time and financial costs of manual annotation from clinical experts. Each task requires clinical expertise from medical students and physicians to reason over complex clinical evidence, with our annotation efforts spanning months. Moreover, larger validation samples yield diminishing gains in precision. For example, with $N{=}100$ questions and observed rates of 92--96\% across the evaluation dimensions (Table~\ref{tab:generated-question-validation}), the 95\% margin of error is about $\pm$4--5 percentage points, sufficient to distinguish high quality from markedly lower rates (e.g., 85\%); quadrupling to $N{=}400$ would narrow it only to about $\pm$2--3 points. Given these costs and the diminishing gains in statistical precision, we therefore view our validation sizes as a reasonable balance between statistical rigor and feasibility.}

\camready{\textit{No Independent Human-Expert Baseline.} Recruiting clinical experts to independently re-synthesize Cochrane-level conclusions is difficult, as producing a Cochrane review is a lengthy, specialized process. We therefore do not report a human-expert baseline. Instead, we report an approximate oracle upper bound using the reviews' own \textit{Authors' Conclusions} (factual F1 0.95; \S\ref{appendix:oracle}), which serves as a reference point for expert-level performance but is not an independent human baseline.}

\textit{Decision Loss.} While \textsc{SciConBench} measures factual precision and recall, these metrics do not directly capture downstream decision impact. In practice, missing, incomplete, or contradictory information becomes most consequential when it \textit{changes} scientific, clinical, or policy decisions. For example, omissions in treatment effectiveness, uncertainty, or harms may lead users toward different conclusions or actions despite only modest changes in factual F1. The impact of synthesis errors may differ across populations or user contexts. Future work should investigate how factual errors translate into downstream \textit{decision loss}, particularly in personalized or high-stakes health contexts.

\textit{Focus on Scientific Conclusion Synthesis.} We focus on scientific conclusion synthesis, a consequential output that shapes decisions in health, science, and policy. However, our benchmark does not directly evaluate intermediate reasoning processes, such as multi-hop reasoning, evidence selection, or assessment of evidence quality. Although many proprietary, frontier models do not expose detailed reasoning traces, \textsc{SciConHarness} logs reasoning summaries and tool calls. We leave the evaluation of intermediate reasoning artifacts to future work.

\textit{Alternative Clean-Room Evaluations.} Some provider-hosted deep research agents do not support direct integration with \textsc{SciConHarness}, requiring alternative clean-room protocols for comparability with other evaluated systems (\S\ref{appendix:closed-sourced}). In particular, Perplexity agents rely on provider-native search APIs without custom MCP support, for which we use provider-side search filters as a ``best-effort'' clean-room protocol. As a result, these evaluations may not be perfectly comparable to direct \textsc{SciConHarness} integration. Nevertheless, both Perplexity and OpenAI deep research agents consistently exhibit substantial performance decreases under clean-room settings (\S\ref{sec:benchmark-result}), suggesting that our central finding on leakage-driven performance inflation is robust across implementations.

\textit{Real-World Query Formulation.} \textsc{SciConBench} questions are derived from the Objectives sections of published \texttt{CDSR} reviews, reflecting real-world scientific synthesis tasks. However, these questions may not fully capture how diverse users---such as patients, caregivers, and the general public \cite{kaur2025whosaskingsimulatingrolebased}---naturally formulate queries in health contexts. Real-world information seeking is often underspecified, conversational, personalized, or shaped by incomplete domain knowledge, whereas our benchmark questions are grounded in well-defined review objectives and PICO-style framing. While the current study focuses on the task of scientific conclusion synthesis rather than noisy or ambiguous real-world queries, evaluating agent behavior under such settings represents an important direction for future work, as AI agents are increasingly deployed in real-world health and scientific contexts.

\subsection{Ethical Considerations \& Broader Impact}\label{appendix:ethics}

As AI agents are increasingly used to access and synthesize scientific conclusions, \textsc{SciConBench} can guide the development of more reliable, transparent, and evidence-grounded AI agents for health and science in open-domain settings. By evaluating and identifying weaknesses in current frontier agents, our work can help researchers and practitioners better understand the limitations of existing AI agents before deployment in high-stakes health settings. Broadly, our benchmark may support the development of safer benchmark evaluation practices for AI agents on the open web and encourage future work on trustworthy scientific synthesis, evidence integration, and uncertainty communication.

At the same time, our work carries ethical considerations. As a publicly available and open-source benchmark, \textsc{SciConBench} may itself become a target for benchmark contamination or overfitting, potentially inflating future agent performance through memorization or retrieval of benchmark artifacts rather than genuine synthesis capability. To mitigate this risk, we design \textsc{SciConBench} as a continuously updated live benchmark and introduce \textsc{SciConHarness}, an open-source clean-room evaluation harness that restricts access to ground-truth artifacts. In addition, improvements on \textsc{SciConBench} may be misinterpreted as evidence that AI-generated scientific conclusions are sufficiently reliable for real-world deployment. However, our results show that current systems frequently omit necessary facts and sometimes generate contradictory conclusions, raising concerns about overtrust and automation bias in high-stakes health contexts. We therefore emphasize that \textsc{SciConBench} is intended as an evaluation benchmark rather than a deployment endorsement.

Our benchmark does not assign new licenses to \texttt{CDSR}-authored content, which remains copyrighted by the original authors and/or Cochrane and governed by Cochrane/Wiley terms and applicable Creative Commons licensing terms. We claim no ownership over Cochrane-authored content. The benchmark was developed through non-commercial academic research and constructed from publicly accessible \texttt{CDSR} components, including Abstract sections (e.g., \textit{Objectives}, \textit{Authors’ Conclusions}) and Plain Language Summaries. These components are available without paywall access and have been used in prior research \cite{bakker-kamps-2024-cochrane, DEVANE2025111894, polzak2025can}.

\newpage

\section{Details on Question Generation}
Here, we discuss the technical details of the question generation pipeline (\S \ref{appendix:question-gen-technical}), the validation results of the generated questions (\S \ref{appendix:question-gen-validation}), and the monthly update process of \textsc{SciConBench} (\S\ref{appendix:monthly-update}).

\subsection{Technical Details and Methods}\label{appendix:question-gen-technical}

For each systematic review, we generate a single question from the \textit{Objectives} section, using the \textit{Background} section as additional context. We employ \texttt{gpt-5-chat} via Microsoft Azure AI Foundry\footnote{\url{https://ai.azure.com/}} with a few-shot prompt (Figure~\ref{fig:question-generation-prompt}). The prompt includes three examples of objective-to-question conversions and instructs the model to output a JSON object containing the generated question and a brief step-by-step justification, which have been shown to improve performance \cite{dammu-etal-2024-uncultured, 10.5555/3600270.3602070}. We set the temperature to 0 and a maximum of 1{,}024 tokens.

To ensure well-formed questions, we apply a rules-based check: 1) strip whitespace, 2) enforce a trailing question mark, 3) verify the presence of interrogative structure (e.g., \textit{what}, \textit{how}, \textit{which}, \textit{is}, etc), and 4) discard outputs that are empty or shorter than 10 characters. Invalid outputs are regenerated until they pass validation. See Figure~\ref{fig:question-generation-prompt} for an example of a generated question.

\subsection{Validation of Generated Questions}\label{appendix:question-gen-validation}

\xhdr{Overview} We validate the quality of generated questions through domain-expert annotations conducted by two medical student annotators. The task is to assess whether each question faithfully reflects the intent and scope of the underlying systematic review. These medical students are from established medical schools in the U.S. with extensive clinical and scientific research experiences. Annotators are provided with the \textit{Objectives}, generated question, and supporting context from the \textit{Background} and \textit{Authors' Conclusions}, with optional access to the full article.

\xhdr{Evaluation Dimensions} We evaluate question quality along three dimensions grounded in prior works \cite{kaur2025whosaskingsimulatingrolebased, mir-etal-2019-evaluating, park-etal-2024-valuescope, pico_framework}:
\begin{itemize}[leftmargin=0.6cm]
    \item \textbf{Faithfulness:} Whether the generated question preserves the meaning of the \textit{Objectives} without distortions. \textsc{Answer Choices: Faithful, Unfaithful}
    \item \textbf{PICO Completeness:} Whether the generated question captures all key PICO components present in the \textit{Objectives}. \textsc{Answer Choices: Complete, Partially Complete, Incomplete}
    \item \textbf{Clarity \& Answerability:} Whether the generated question is clearly phrased and answerable by a systematic review. \textsc{Answer Choices: Clear and Answerable, Unclear / Unanswerable}
\end{itemize}

\begin{figure*}[t]
    \centering

    \begin{subfigure}[t]{0.48\textwidth}
        \centering
        \includegraphics[width=6cm, height=8cm, keepaspectratio]{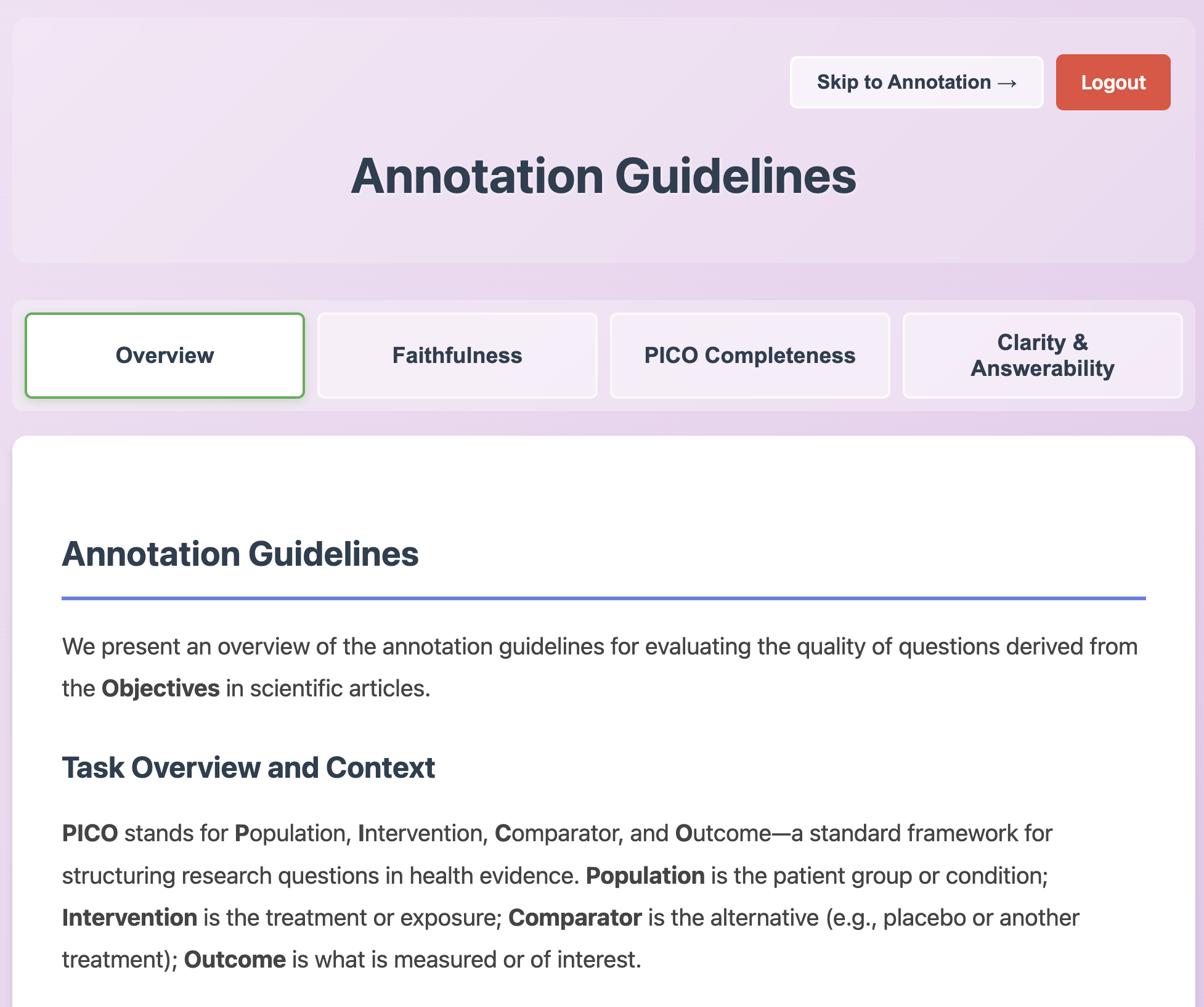}
        \caption{Overview Page}
    \end{subfigure}
    \hfill
    \begin{subfigure}[t]{0.48\textwidth}
        \centering
        \includegraphics[width=\linewidth, trim={0cm 9cm 0cm 0cm}, clip]{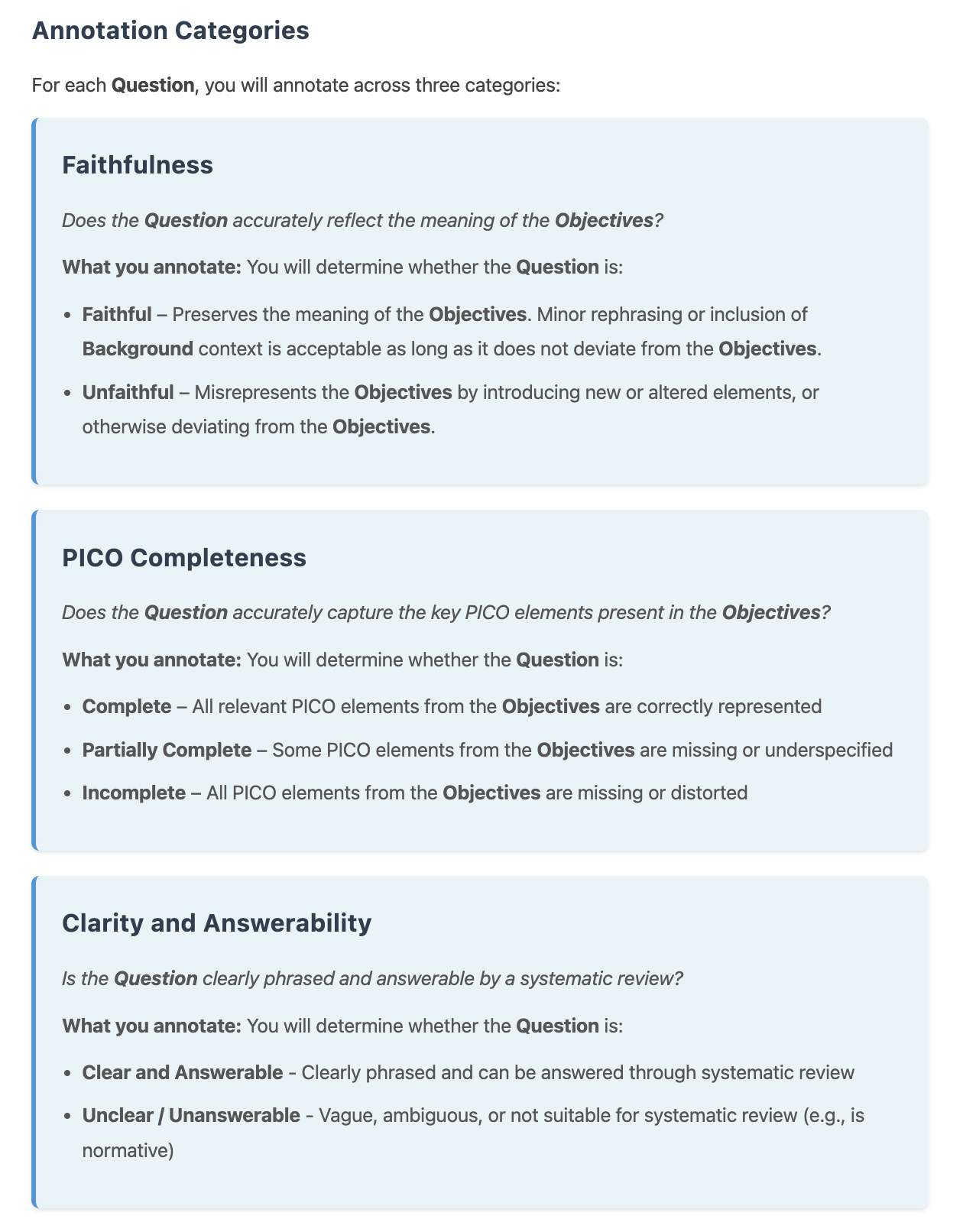}
        \caption{Introduction to Evaluation Dimensions}
    \end{subfigure}
    \vspace{0.75em}

    \begin{subfigure}[t]{0.48\textwidth}
        \centering
        \includegraphics[width=6cm, keepaspectratio]{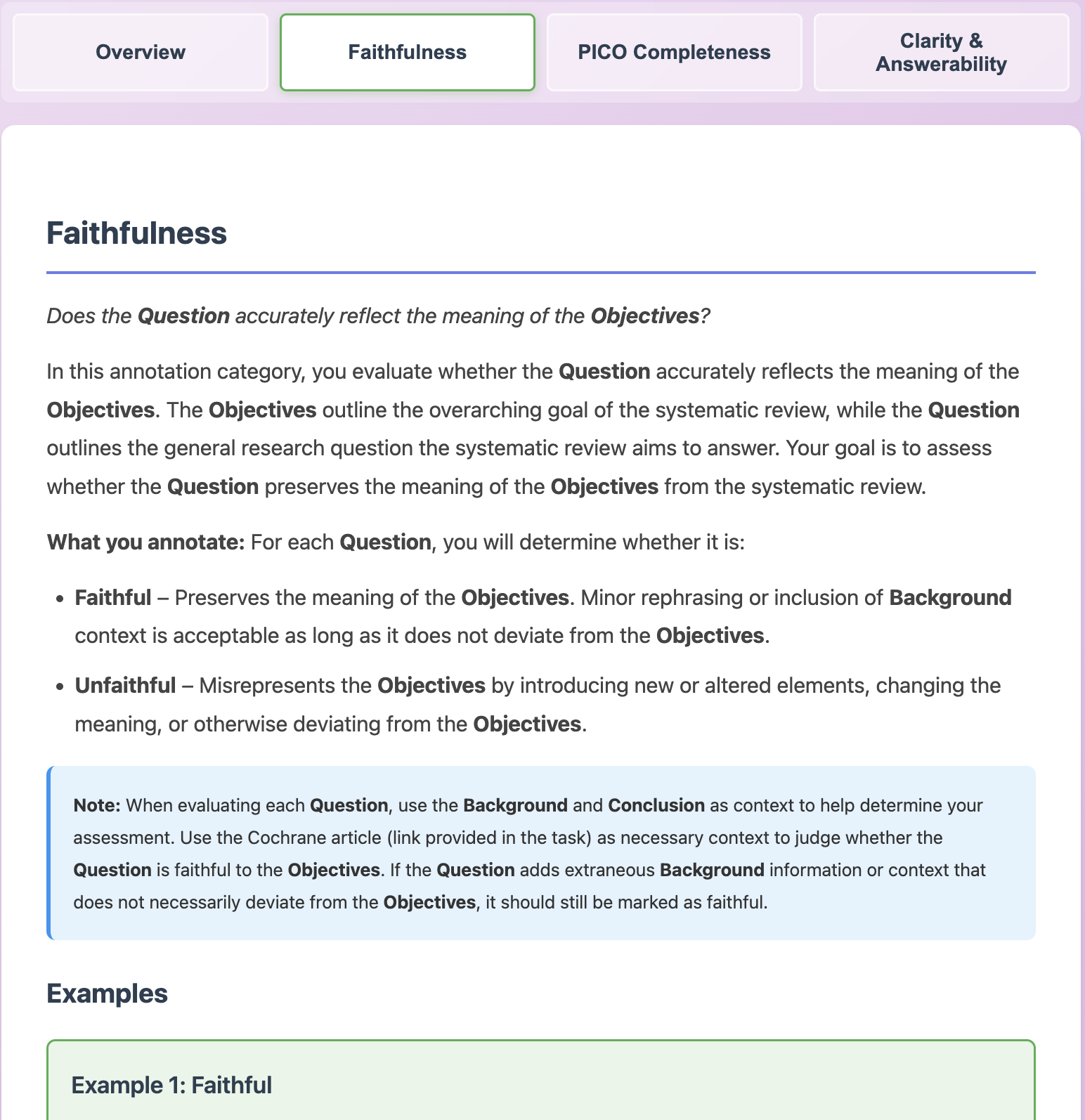}
        \caption{Faithfulness Definition}
    \end{subfigure}
    \hfill
    \begin{subfigure}[t]{0.48\textwidth}
        \centering
        \includegraphics[width=\linewidth, trim={0cm 5.75cm 0cm 0cm}, clip]{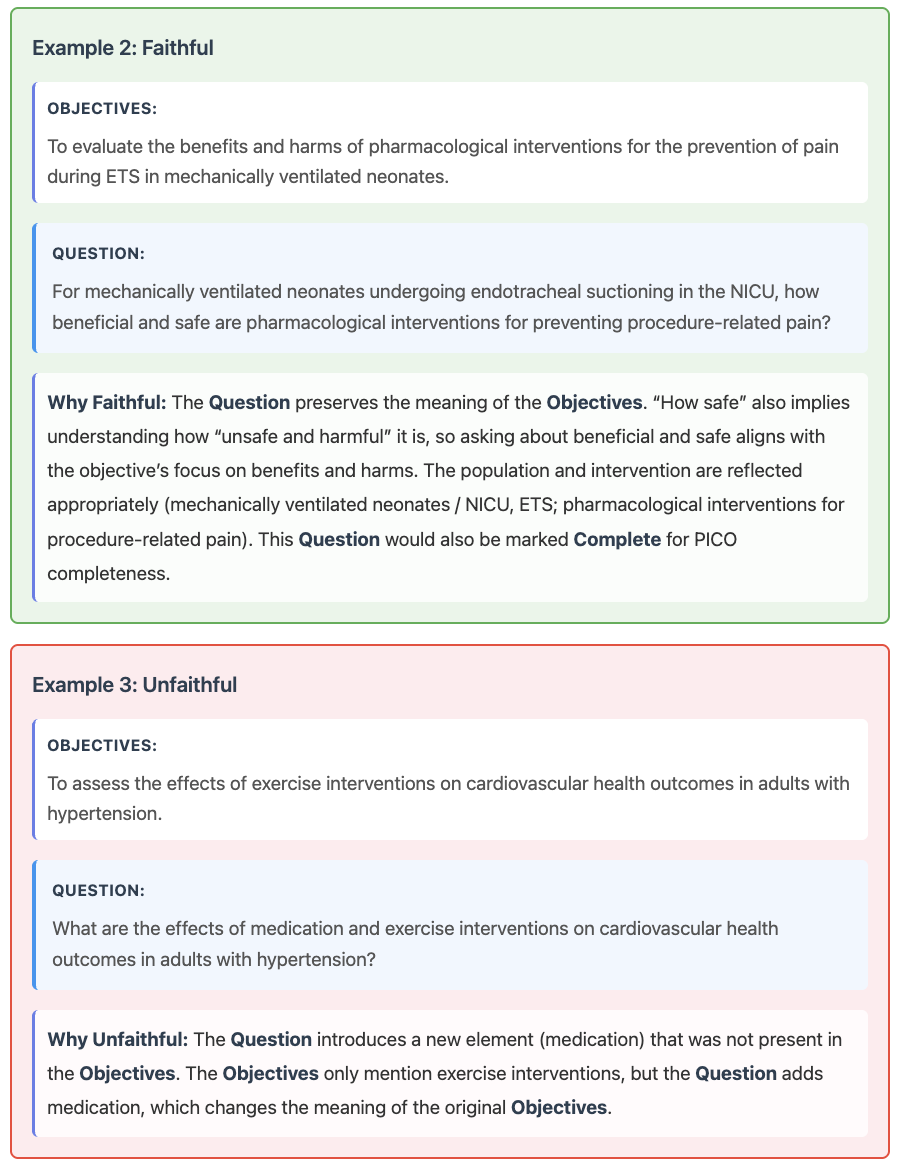}
        \caption{Labeled Examples of Faithfulness}
    \end{subfigure}

    \caption{Overview of the annotation interface for evaluating generated questions. Panels show the task introduction and guidelines: (a) overview page, (b) evaluation dimensions, (c) definition of faithfulness, and (d) labeled examples of faithfulness. Panels are cropped or resized for space. Additional panels of the annotation interface are in Figure \ref{fig:generated-question-interface-part2}.}
    \label{fig:generated-question-interface-part1}
\end{figure*}

\begin{figure*}[t]
    \centering

    \begin{subfigure}[t]{0.48\textwidth}
        \centering
        \includegraphics[width=6cm, height=8cm, keepaspectratio]{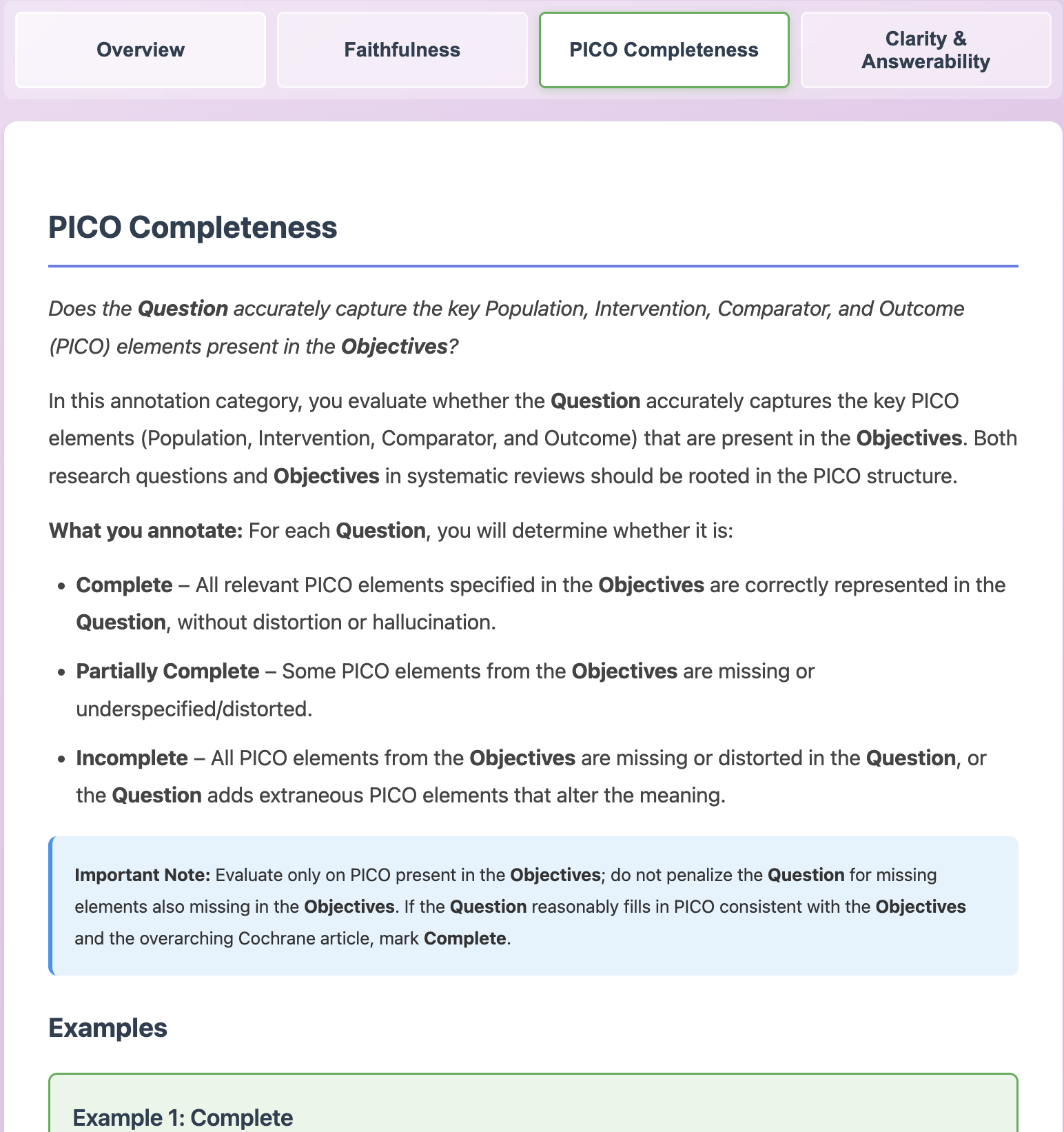}
        \caption{PICO Completeness Definition}
    \end{subfigure}
    \hfill
    \begin{subfigure}[t]{0.48\textwidth}
        \centering
        \includegraphics[width=\linewidth, trim={0cm 12.3cm 0cm 0cm}, clip]{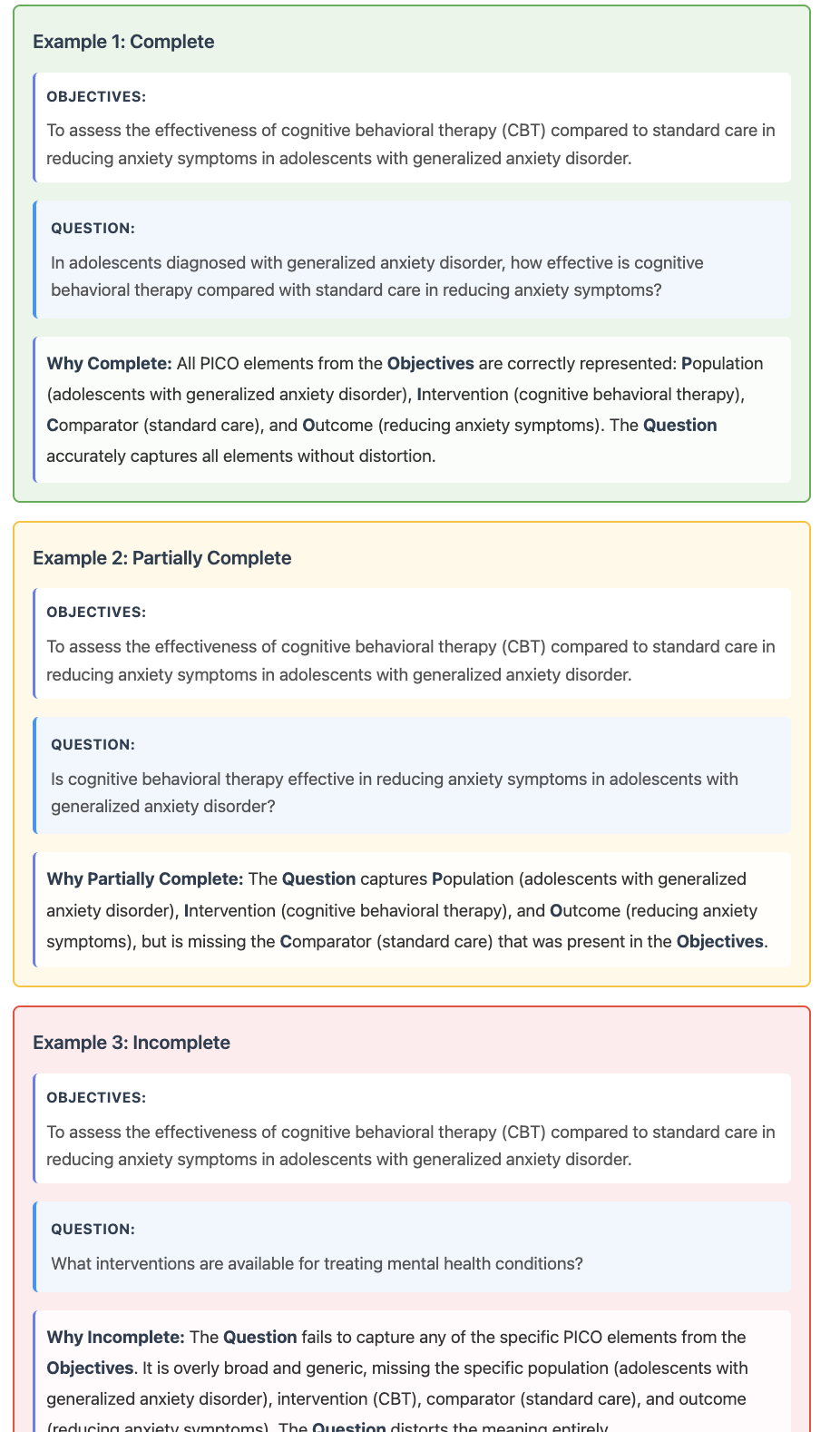}
        \caption{Labeled Examples of PICO Completeness}
    \end{subfigure}
    \vspace{0.75em}

    \begin{subfigure}[t]{0.48\textwidth}
        \centering
        \includegraphics[width=6cm, height=8cm, keepaspectratio]{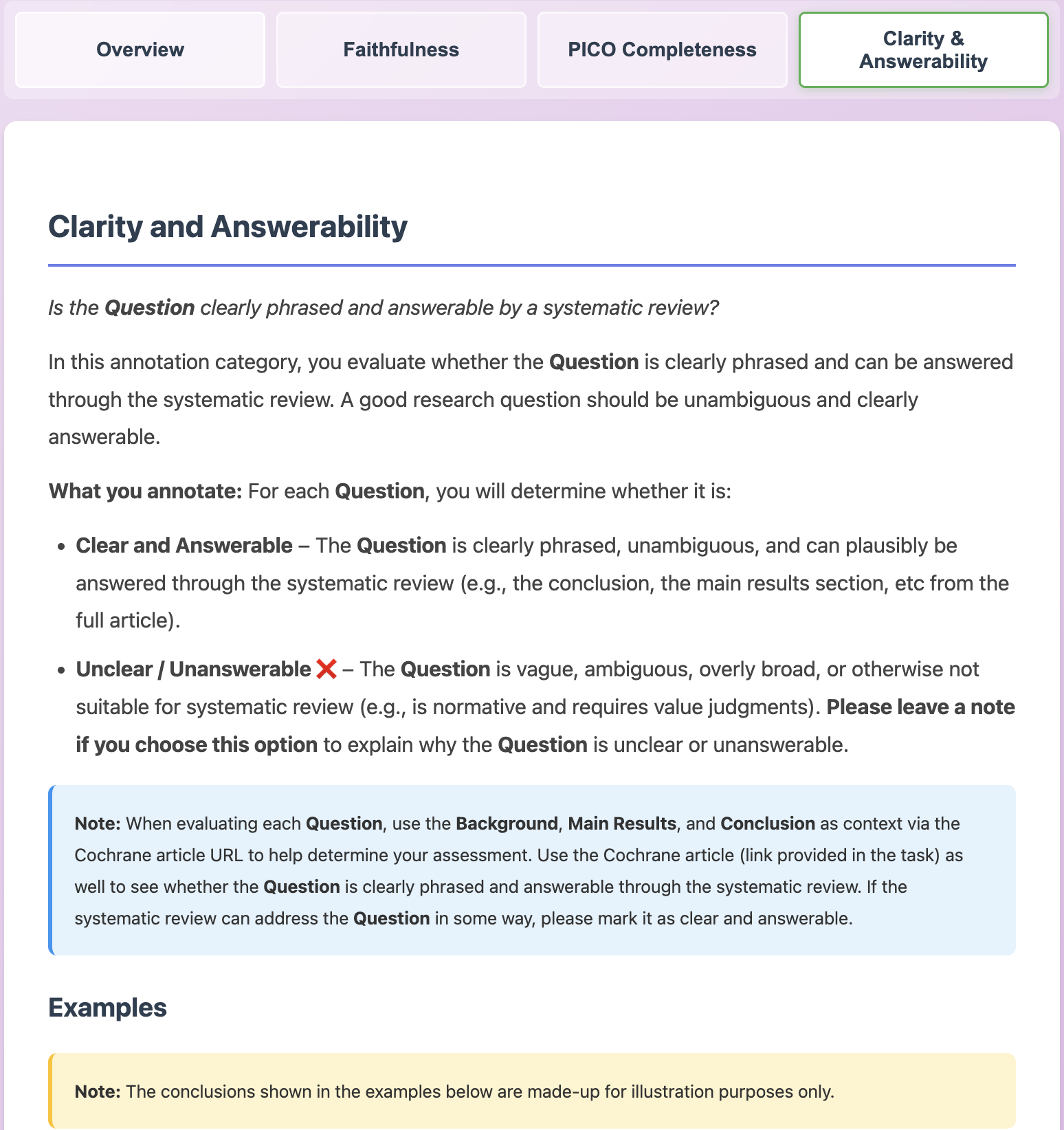}
        \caption{Clarity \& Answerability Definition}
    \end{subfigure}
    \hfill
    \begin{subfigure}[t]{0.48\textwidth}
        \centering
        \includegraphics[width=6.2cm, height=7cm, keepaspectratio]{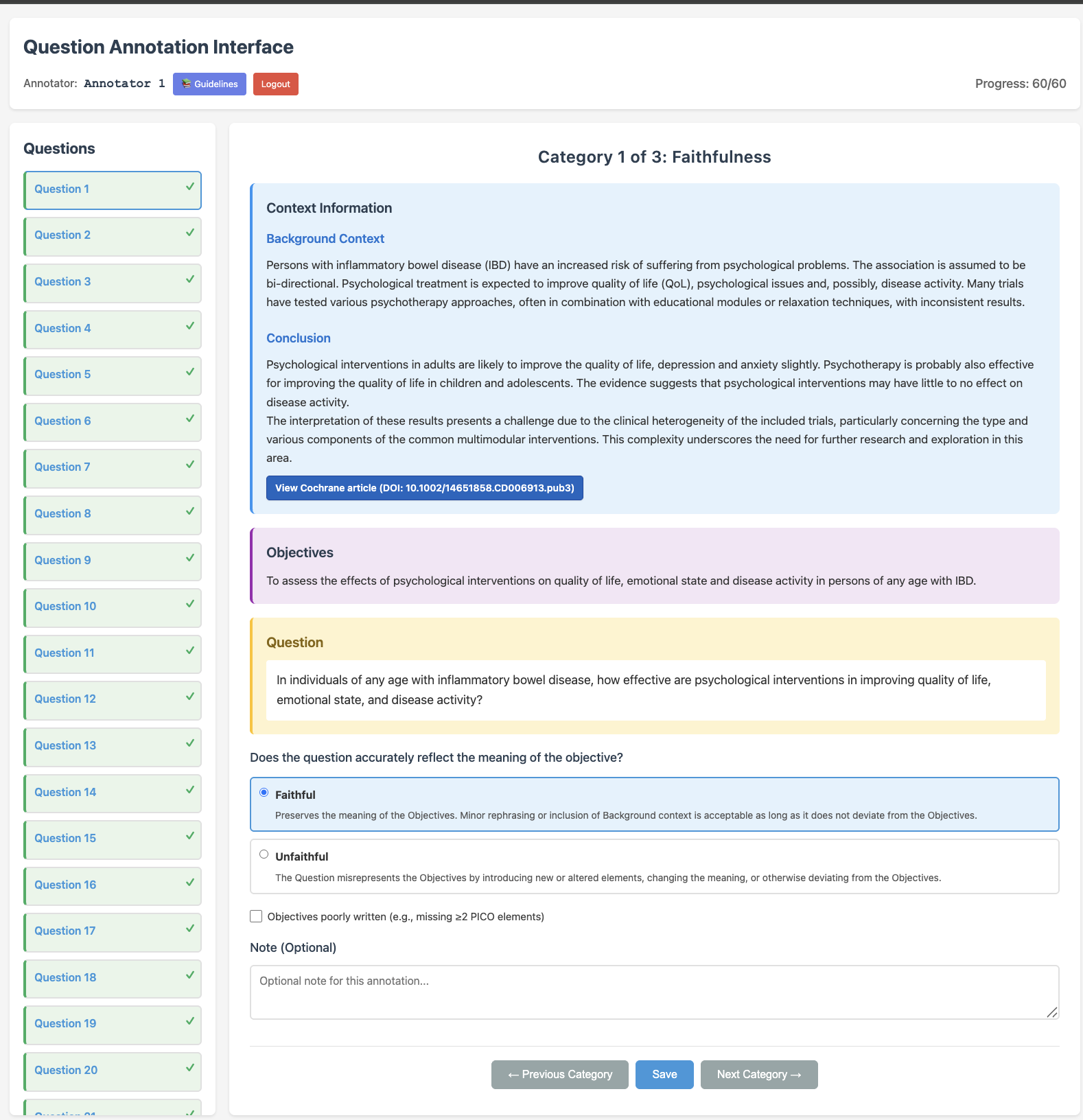}
        \caption{Annotation Interface}
    \end{subfigure}

    \caption{Additional panels of the annotation interface for evaluating generated questions. Panels present the interface defining the evaluation dimensions and labeling annotations: (a) definition of PICO completeness, (b) labeled examples of PICO completeness, (c) definition of clarity \& answerability, and (d) annotation interface. Panels are cropped or resized for space.}
    \label{fig:generated-question-interface-part2}
\end{figure*}

The full annotation guidelines are provided in Figure~\ref{fig:annotation_guidelines}, and the annotation interface used by annotators is shown in Figures~\ref{fig:generated-question-interface-part1} and \ref{fig:generated-question-interface-part2}.

\begin{wraptable}{r}{0.45\columnwidth}
\vspace{-12pt}
\centering
\small
\caption{Agreement between two expert annotators on evaluating generated questions (N=10) across three dimensions: Faithfulness (\textbf{Faith.}), PICO Completeness (\textbf{PICO}), and Clarity \& Answerability (\textbf{Clar.}).}
\setlength{\tabcolsep}{3.5pt}
\begin{tabular}{lccc}
\toprule
\textbf{Agreement} & \textbf{Faith.} & \textbf{PICO} & \textbf{Clar.} \\
\midrule
Gwet's AC1   & 0.76 & 0.78 & 1.00 \\
\% Agreement & 0.80  & 0.80 & 1.00 \\
\bottomrule
\end{tabular}
\label{tab:interrater}
\vspace{-7pt}
\end{wraptable}

\xhdr{Annotation Procedure} Based on the annotation guidelines, we first conduct a calibration phase in which annotators jointly label 10 generated questions to refine the task, validate the evaluation dimensions, and resolve disagreements. During this process, we clarify edge cases (e.g., allowing additional background context in the generated questions when it does not alter meaning for Faithfulness) and update the codebook accordingly. After refining and validating the annotation task, annotators then independently label an additional 10 generated questions to assess reliability. To assess their reliability, we compute the inter-annotator agreement using Gwet's AC1, which is more robust than Cohen's $\kappa$ under a skewed label distribution \cite{wongpakaran2013comparison, ohyama2021statistical}.

\xhdr{Results}
As shown in Table \ref{tab:interrater}, the inter-annotator agreement is high across all three evaluation dimensions (Gwet's AC1: 0.76--1.00; Percentage Agreement: 0.80--1.00), comparable to or exceeding prior work \cite{kaur2025whosaskingsimulatingrolebased, park-etal-2024-valuescope}. After measuring agreement, the expert annotators resolved disagreements through discussion. Following high agreement, the two annotators independently labeled 40 generated questions each, yielding $N=100$ annotated questions total. 

\begin{table}[t!]
\centering
\small
\caption{Label distribution across evaluation dimensions for the generated questions ($N=100$). Abbreviations: Partial = Partially Complete; Clarity = Clarity \& Answerability; Clear = Clear and Answerable; Unclear = Unclear / Unanswerable.}
\setlength{\tabcolsep}{4pt}
\begin{tabular}{lccccc c c}
\toprule
\textbf{Metric} 
& \multicolumn{2}{c}{\textbf{Faithfulness}} 
& \multicolumn{3}{c}{\textbf{PICO Completeness}} 
& \multicolumn{2}{c}{\textbf{Clarity}} \\
\cmidrule(lr){2-3} \cmidrule(lr){4-6} \cmidrule(lr){7-8}
& Faithful & Unfaithful & Complete & Partial & Incomplete & Clear & Unclear \\
\midrule
\textbf{Overall} 
& 92.0\% & 8.0\% 
& 92.0\% & 6.0\% & 2.0\% 
& 96.0\% & 4.0\% \\
\bottomrule
\end{tabular}
\label{tab:generated-question-validation}
\end{table}

Table~\ref{tab:generated-question-validation} summarizes the validation results, in which the annotators evaluated the generated questions to be largely faithful (92\%), complete with key PICO elements (92\%), and clear and answerable (96\%). These results indicate that the question generation pipeline reliably produces faithful, complete, and answerable research questions that accurately reflect the intent and scope of the underlying systematic reviews, validating their use in downstream evaluation. 

\xhdr{\camready{Failure Mode Analysis}} \camready{We identify two main failure modes in question generation: (1) incorporating non-PICO content into the question, and (2) merging distinct objectives into a single multi-part question. Figure~\ref{fig:qg-failure-example} shows an example of both: the review's secondary objective (maintaining evidence currency via a living systematic review) is a methodological statement rather than PICO content and should be excluded, yet the pipeline merges it with the primary objective into a multi-part question. Expert annotators rated this question as \textsc{Unfaithful} and \textsc{Unclear / Unanswerable}. More broadly, although \texttt{CDSR} objectives are expert-written, a small fraction are under- or over-specified, and generated questions may inherit their level of specificity. As shown in our expert validation (Table~\ref{tab:generated-question-validation}), such failures are rare.}

\subsection{\camready{Monthly Benchmark Updates}}\label{appendix:monthly-update}
\camready{\textsc{SciConBench}\footnote{\url{https://sciconbench.cs.princeton.edu/}} is maintained through an automated monthly pipeline that: (1) identifies newly published \texttt{CDSR} reviews; (2) retrieves their content from the Wiley Online Library Text and Data Mining API using institutional access; (3) generates benchmark questions and reference atomic facts using the same pipelines described in \S\ref{appendix:question-gen-technical} and \S\ref{appendix:afg-pipeline}; and (4) replaces superseded reviews, adds new reviews, and publishes the updated dataset on HuggingFace\footnote{\url{https://huggingface.co/datasets/hayoungjung/SciConBench}}. This ensures that \textsc{SciConBench} reflects the latest expert-written scientific conclusions each month.}

\begin{figure}[h!]
\centering
\begin{small}
\begin{tcolorbox}[colback=gray!5, colframe=red, width=\textwidth, title=\textbf{Question Generation Failure Example}]
\textsc{Objectives:} ``To assess the benefits and harms of stem cell injections for people with osteoarthritis of the knee. \textbf{A secondary objective is to maintain the currency of the evidence, using a living systematic review approach.}''\\[2pt]
\textsc{Generated Question:} ``For people with osteoarthritis of the knee, what are the benefits and harms of stem cell injections, \textbf{and how can a living systematic review approach help ensure that the evidence remains current?}''\\[2pt]
\textsc{Expert Labels:} Faithfulness: \textsc{Unfaithful}; Clarity \& Answerability: \textsc{Unclear / Unanswerable}
\end{tcolorbox}
\end{small}
\caption{\camready{Example question generation failure. The pipeline incorporates a methodological (non-PICO) secondary objective (\textbf{bold}) and merges it with the primary objective into a multi-part question.}}
\label{fig:qg-failure-example}
\end{figure}

\begin{figure}[h!]
\begin{small}
\centering
\begin{tcolorbox}[
    colback=gray!5,
    colframe=black,
    width=\linewidth,
    arc=2mm,
    boxrule=0.5pt,
    title=\textbf{Question Generation Prompt}
]
\textbf{System Prompt:} \textcolor{teal}{You are an expert specialized in converting research objectives into their corresponding research question format.}\\

\textbf{Instruction:} Given the following research objective, please convert it into a single question, using the provided study background as context to inform your answer.\\

\texttt{***RESEARCH OBJECTIVE STARTS HERE***}\\
\textcolor{blue}{To assess the benefits and risks of medical treatments prior to surgery for uterine fibroids.}\\
\texttt{***RESEARCH OBJECTIVE ENDS HERE***}\\

\texttt{***STUDY BACKGROUND CONTEXT STARTS HERE***}\\
\textcolor{blue}{Uterine fibroids occur in up to 40\% of women over 35 years of age. Up to 50\% of uterine fibroids cause symptoms that warrant treatment: anaemia caused by heavy menstrual bleeding, pelvic pain, dysmenorrhoea, infertility and poor quality of life. Surgery is the first choice of treatment...}\\
\texttt{***STUDY BACKGROUND CONTEXTS ENDS HERE***}\\

\texttt{***EXAMPLE 1 STARTS HERE***}\\
Research Objective: ...\\
Generated Question: ...\\
\texttt{***EXAMPLE 1 ENDS HERE***}\\

...\\

\texttt{***EXAMPLE 3 STARTS HERE***}\\
Research Objective: ...\\
Generated Question: ...\\
\texttt{***EXAMPLE 3 ENDS HERE***}\\

Now, given what you learned from the examples and using the study background as context, please convert the provided objective into a single research question, justifying your answer and thinking step-by-step about your answer. Use the guidelines below to inform your output question and justification.\\

\texttt{***GUIDELINES FOR GENERATING RESEARCH QUESTIONS STARTS HERE:***}\\
- The generated question should be specific, answerable, and capture the main research focus in the objective.\\
- Do not include any extraneous information or context in your answer that were not provided in the study background\\
- Do not add additional information beyond what is stated in the objective. Only use the background as context to inform your answer.\\
- Avoid copying the objective verbatim and aim to generate a question that is Google-Proof. Preserve the objective's semantics and the objective to maintain the full population, intervention, comparison, and outcome (PICO) of the study and objective in the generated question.\\
\texttt{***GUIDELINES FOR GENERATING RESEARCH QUESTIONS ENDS HERE.***}\\

Output should be in JSON format with the following structure:\\
- ``question'': string research question\\
- ``justification'': string justification for your answer regarding the converted question.\\
\end{tcolorbox}
\begin{tcolorbox}[
    colback=gray!5,
    colframe=black,
    width=\linewidth,
    arc=2mm,
    boxrule=0.5pt,
    title=\textbf{Example Generated Question}
]
For women with uterine fibroids undergoing surgery, what are the benefits and risks of using medical treatments before surgery?
\end{tcolorbox}

\caption{Few-shot prompt used to convert \texttt{CDSR} \textit{Objectives} into equivalent research questions. The \textit{Background} section is included to provide additional context during generation. Text in \textcolor{blue}{blue} shows an example \textit{Objective} and \textit{Background} from a \texttt{CDSR} systematic review, along with the corresponding generated question below.}
\label{fig:question-generation-prompt}
\end{small}
\end{figure}

\begin{figure*}[h!]
\centering
\begin{small}
\begin{tcolorbox}[colback=gray!5, colframe=black, width=\textwidth, title=\textbf{Annotation Guideline for Evaluating Generated Questions}, fonttitle=\bfseries]

\textbf{Task Overview.}
\textsc{PICO} stands for \textsc{Population}, \textsc{Intervention}, \textsc{Comparator}, and \textsc{Outcome}—a standard framework for structuring research questions in health evidence. Population is the patient group or condition; Intervention is the treatment or exposure; Comparator is the alternative (e.g., placebo or another treatment); Outcome is what is measured or of interest.\\

Your task is to evaluate the quality of questions generated from the \textbf{Objectives} section in Cochrane review articles. According to prior work, both research questions and the \textbf{Objectives} section in systematic review articles are closely related and should be rooted in the PICO framework. Since research questions are often framed within PICO, we convert \textbf{Objectives} into a \textbf{Question} format.

\vspace{0.5em}
\textbf{Inputs.}
You will be provided with the following:

\begin{itemize}[leftmargin=0.6cm]
    \item \textbf{Question}: Unit of evaluation. An overarching research question of the review.
    \item \textbf{Objectives}: Direct source of the \textbf{Question}. Outlines the overarching goal of the systematic review.
    \item \textbf{Background and Conclusion}: Paragraph(s) from the Background section and the Conclusion paragraph to provide context and motivation. \textbf{Use these as context when evaluating the Question.}
    \item \textbf{Cochrane Article Link}: Provided for additional context if needed.
\end{itemize}

\vspace{0.5em}
\textbf{Evaluation Criteria.}
When evaluating each \textbf{Question}, use the Background and Conclusion as context.

\begin{enumerate}[leftmargin=0.6cm]
    \item \textbf{Faithfulness.} \textit{Does the generated \textbf{Question} accurately reflect the meaning of the \textbf{Objectives}?} 
    \vspace{-0.2em}
    \begin{itemize}[leftmargin=0.5cm]
    \item\textsc{Faithful}: Preserves the meaning of the Objectives. Minor rephrasing or inclusion of Background context is acceptable as long as it does not deviate from the \textbf{Objective}. 
    \item\textsc{Unfaithful}: Misrepresents the Objectives by introducing new or altered elements or changing the meaning.
    \end{itemize}
    \text{Note}: Additional background context in the generated question is fine as long as they do not change the meaning of the original \textbf{Objective}. 

    \vspace{0.5em}
    \item \textbf{PICO Completeness.} \textit{Does the generated \textbf{Question} capture the key Population, Intervention, Comparator, and Outcome (PICO) elements present in the \textbf{Objectives}?}
    \vspace{-0.2em}
    \begin{itemize}[leftmargin=0.5cm]
    \item\textsc{Complete}: All relevant PICO elements are correctly represented without distortion or hallucination. 
    \item\textsc{Partially Complete}: Some PICO elements are missing or underspecified/distorted. 
    \item\textsc{Incomplete}: PICO elements are missing, distorted, or extraneous elements are introduced that alter meaning.
    \end{itemize}
    \text{Note}: Evaluate only PICO present in the Objectives; do not penalize PICO elements that are also missing from the Objectives. 

    \vspace{0.5em}
    \item \textbf{Clarity and Answerability.} \textit{Is the \textbf{Question} clearly phrased and answerable by a systematic review?} 
    \vspace{-0.2em}
    \begin{itemize}[leftmargin=0.5cm]
    \item\textsc{Clear and Answerable}: Clearly phrased, unambiguous, and can be answered through a systematic review. 
    \item\textsc{Unclear / Unanswerable}: Vague, ambiguous, overly broad, or not suitable for systematic review (e.g., requires primary data, is normative, or lacks operational clarity).
    \end{itemize}
\end{enumerate}
\end{tcolorbox}
\caption{Annotation Guideline for evaluating generated questions derived from \textit{Objectives} in \texttt{CDSR} systematic reviews.}
\label{fig:annotation_guidelines}
\end{small}
\end{figure*}

\clearpage

\clearpage

\begin{figure*}[t]
    \centering
    \begin{tcolorbox}[colback=gray!10, colframe=gray!60!black, boxrule=0.5pt, arc=2mm,
        left=8pt, right=8pt, top=8pt, bottom=8pt,
        title=\textbf{Example \texttt{CDSR} Review Article}]
    \centering
    {\setlength{\fboxsep}{0pt}\fcolorbox{gray!50}{white}{\includegraphics[width=0.92\linewidth, clip]{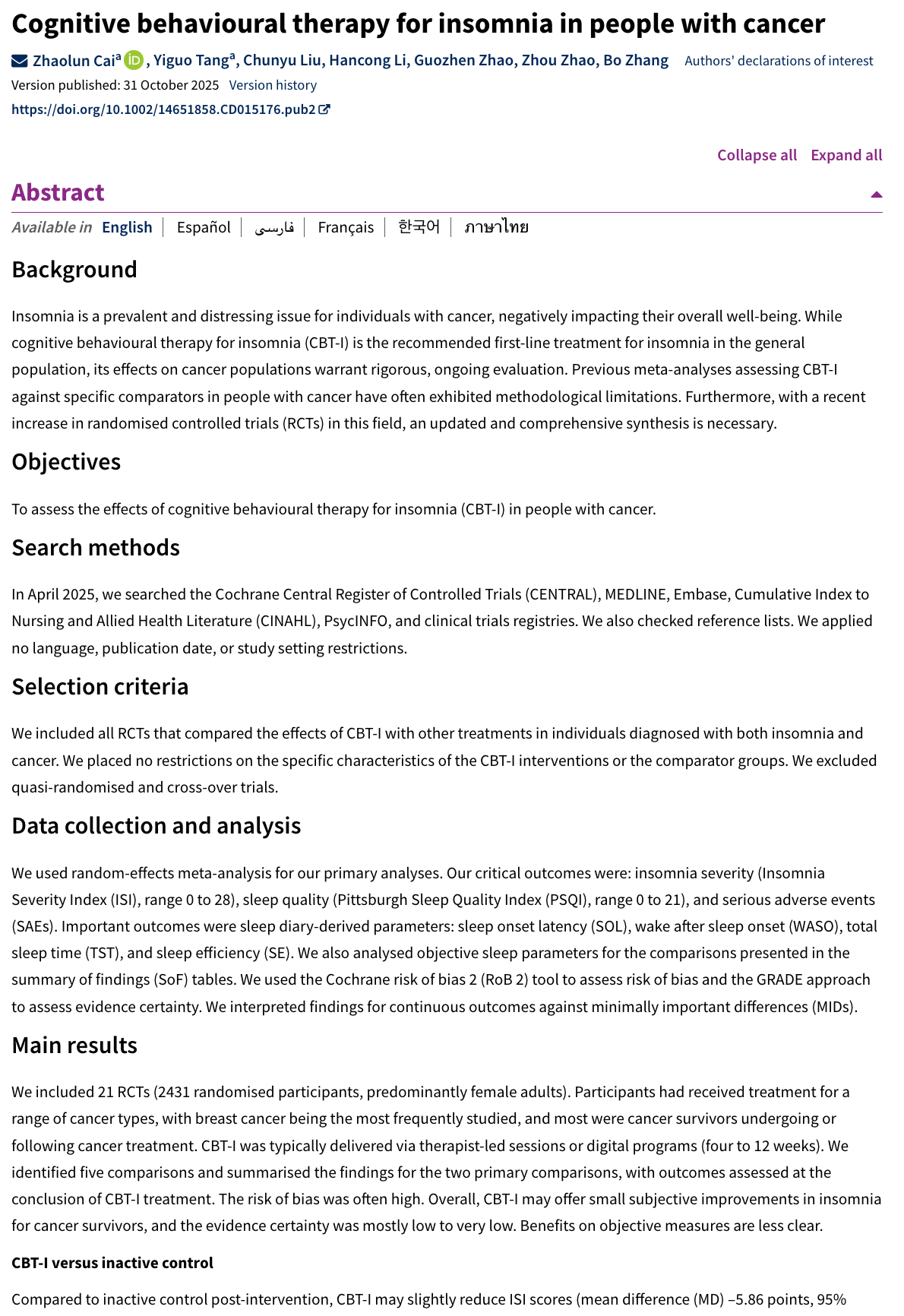}}}
    \end{tcolorbox}
    \caption{Example of a \texttt{CDSR} systematic review article used in \textsc{SciConBench}. We derive benchmark questions from the publicly available \textit{Objectives} section and reference scientific conclusions from the \textit{Authors’ Conclusions} in the \texttt{CDSR} review. Note that the \textit{Authors’ Conclusions} section, which follows \textit{Main results}, is not shown in the screenshot due to the figure crop, but is fully available in the original review.}
    \label{fig:cochrane-example}
\end{figure*}

\clearpage

\section{\textsc{SciConHarness} Implementation Details}\label{appendix:sciconharness-implementation}

\textsc{SciConHarness} is implemented as a unified MCP-based control layer that separates \emph{orchestration} from \emph{tool execution}. The MCP client side handles any LLM providers, managing the model context and tool-calling state. The MCP server side exposes API-based web search and browsing tools over MCP, and, if specified, enforces \textit{clean-room} filtering protocols before evidence is returned to the model. This separation enables consistent evaluation across heterogeneous model APIs while preserving a common tool interface. Below, we describe the core \textsc{SciConHarness} design: a unified client--server framework for orchestrating tool use, executing retrieval, and enforcing \textit{clean-room} filtering during evaluation.

\xhdr{MCP Client: The Orchestration Engine}
The MCP client is the orchestration engine of \textsc{SciConHarness}. The client abstracts LLM provider-specific semantics (e.g., OpenAI, Anthropic, etc) into a unified execution loop, formats tool schemas, and maintains intermediate reasoning and structured tool outputs in the context across iterations. The client also handles retries, error handling, and logs detailed metadata (e.g., iterations, tool invocations, token usage). This design provides a consistent interface for evaluating tool-using models.

\xhdr{MCP Server: The Data Plane for Tool Executions}
The MCP server acts as the data plane for API-based web search and browsing tools. The server is built on \texttt{FastMCP}\footnote{\url{https://github.com/prefecthq/fastmcp}} and Ai2's \texttt{dr-agent-lib} \cite{shao2025drtulureinforcementlearning}, which features global caching and asynchronous request handling \cite{jiang2025verltool}. As described in \S\ref{sec:clean-room}, the server exposes \texttt{google\_search}, \texttt{paper\_search}, and \texttt{web\_browse}, returning standardized outputs consumable by any client-supported model.

To enforce the clean-room protocol, \texttt{paper\_search} is temporally constrained to return only papers published before the ground-truth review, preventing access to post-publication evidence. For \texttt{web\_browse}, the page text output can be very long---sometimes exceeding the context limits of models such as \texttt{gemini-3-pro}. Following prior work \cite{li2025webweaver, shao2025drtulureinforcementlearning}, we summarize retrieved content using \texttt{gpt-5-mini}\footnote{Configured with medium verbosity and reasoning effort.} to reduce context usage and cost while preserving relevant information. See Figure \ref{fig:summarization-prompt} for the summarization prompt. Under the \textit{clean-room} protocol, filtering is applied server-side before results are returned to the MCP client. Centralizing filtering within the MCP server ensures that models cannot access the reference conclusions via retrieval leakage.

\xhdr{MCP Client--Server Interaction}
At runtime, the client initializes a session with the server and provides the schemas of available tools to the model. The client then facilitates an iterative loop: (1) model inference (including tool calls), (2) tool-call parsing, (3) tool execution, and (4) appending results back into context. This continues until the model produces a final conclusion wrapped in triple brackets. All interactions are logged (e.g., tool calls, outputs, reasoning traces). Figure~\ref{fig:model-instruction-prompt} shows the system prompt used to evaluate all models on \textsc{SciConBench}.

\begin{table}[h]
\centering
\footnotesize
\setlength{\tabcolsep}{4pt}
\centering
\caption{Validation results of \textsc{SciConHarness}'s \textit{clean-room} setup. All ground-truth \texttt{CDSR} articles (\textbf{GT}) were successfully filtered across all tools.}
\label{tab:filtering-validation}
\begin{tabular}{lccccc}
\toprule
\textbf{Tools} & \textbf{Precision} & \textbf{Recall} & \textbf{F1} & \textbf{Accuracy} & \textbf{\% GT Filtered} \\
\midrule
\texttt{google\_search} & 0.920 & 1.000 & 0.958 & 0.960 & 100\% \\
\texttt{web\_browse} & 0.880 & 0.957 & 0.917 & 0.920 & 100\% \\
\texttt{paper\_search} & 1.000 & 0.962 & 0.980 & 0.980 & 100\% \\
\midrule
\textbf{Overall} & 0.933 & 0.972 & 0.952 & 0.953 & 100\% \\
\bottomrule
\end{tabular}

\end{table}

\newpage




\section{Details on Atomic Fact Generation}

\subsection{Pipeline Details}\label{appendix:afg-pipeline} Our atomic fact generation pipeline decomposes long-form scientific conclusions (e.g., model-generated conclusions, \textit{Authors' Conclusions} in \texttt{CDSR} reviews) into a set of self-contained, complete atomic facts. Our design draws from prior works on long-form factuality evaluations \cite{min-etal-2023-factscore, 10.5555/3737916.3740483, liu-etal-2025-verifact}, comprising six sequential steps: (0) preprocessing, (1) decomposition, (2) decontextualization, (3) incomplete fact rewriting, (4) Relevance Filtering, and (5) Redundancy Filtering. 

\xhdr{Step 0: Preprocessing Module} Each paragraph-length conclusion is first tokenized into sentences using NLTK's \texttt{sent\_tokenize} \cite{bird-loper-2004-nltk}. Following Min et al. \cite{min-etal-2023-factscore}, we apply multi-pass corrections to fix common tokenization errors: 1) merge spurious splits from initials (e.g., ``J.K. Rowling''), and (2) combine bullet lists into a single sentence to preserve context. We then filter out non-informative sentences using simple heuristics, removing those that are too short ($<$ 10 characters or $<2$ words), contain only punctuation or formatting characters, or consist of conversational or meta-commentary (e.g., ``Happy to help,'' ``Here is a breakdown'').

\xhdr{Step 1: Decomposition Module} After preprocessing, we decompose each sentence into \textit{atomic facts} using \texttt{gpt-5.1}. For each sentence, the model is given the sentence, its parent paragraph, and the generated question, and outputs a JSON object containing a set of atomic facts with brief justifications for its outputs. The few-shot prompt contained detailed guidelines instructing \texttt{gpt-5.1} to (i) produce independent, self-contained facts; (ii) avoid inferred or indirect facts; (iii) avoid writing review-specific meta-statements (e.g., ``The review found...''); (iv) reframe authorial judgments into direct claims (e.g., ``The certainty of the evidence was downgraded due to imprecision'' becomes ``Imprecision reduced the certainty of the evidence on [TOPIC]''); and (v) skip non-content elements (conversational markers, formatting). The prompt includes five worked examples from \texttt{CDSR} \textit{Authors’ Conclusions}, illustrating correct decomposition across simple and multi-clause sentence examples. See Figure~\ref{fig:decomposition-prompt} for the prompt.

\xhdr{Step 2: Decontextualization Module} We then \textit{decontextualize} each atomic fact using \texttt{gpt-5.1} so it can be self-contained without additional context. Given an atomic fact, its parent paragraph, and the generated question, the model outputs a JSON object containing the decontextualized fact and a brief justification. Following prior work \cite{liu-etal-2025-verifact, 10.5555/3737916.3740483}, this step replaces vague references (e.g., pronouns, shortened names, review-specific mentions) with the specific entities they denote, using the parent paragraph and generated question as context while preserving the original claim. We adapt the prompt from Wei et al. \cite{10.5555/3737916.3740483} and use a few-shot setup with three worked examples. See Figure~\ref{fig:decontextualization-prompt} for the prompt.

\xhdr{Step 3: Incomplete Fact Rewriting Module} Following Liu et al. \cite{liu-etal-2025-verifact}, we identify \textit{incomplete facts}---those that require additional context or other facts to be correctly interpreted. 
Given the original fact and its parent paragraph, \texttt{gpt-5-mini} classifies each fact as \textit{Independent} (self-contained) or \textit{Dependent} (context-dependent). Dependent facts are further categorized into three subtypes: (1) \textit{Ambiguous Concepts/Pronouns}, e.g., ``this method,” “they” without clear referents, (2) \textit{Missing Comparison}, implies comparison (e.g., ``better'') without an explicit comparison target, and (3) \textit{Lack of Condition} indicates a missing temporal, hypothetical, or qualifying context (e.g., missing ``if'' conditions). For dependent facts, the model rewrites the fact by incorporating the missing context from the source paragraph. See Figure~\ref{fig:incomplete-fact-prompt} for the prompt.

\xhdr{Step 4: Relevance Filtering Module} Using \texttt{gpt-5-mini}, we filter atomic facts for relevance with respect to the generated question, following \cite{liu-etal-2025-verifact, 10.5555/3737916.3740483}. Following the SAFE approach \cite{10.5555/3737916.3740483}, we use substitute labels ``Foo'' and ``Not Foo'' instead of ``Relevant'' and ``Irrelevant'' to force the model to follow our definition of relevance instead of the model relying on its own prior notion of relevance. A fact is labeled ``Foo'' (relevant) if it directly contributes to answering the question or provides useful background context; otherwise (e.g., generic responses, indirect inferences, or meta-commentary), it is labeled ``Not Foo.'' The prompt includes three worked examples from \texttt{CDSR} reviews. Given a fact, the question, and its parent paragraph as context, \texttt{gpt-5-mini} outputs a classification with justification; only ``Foo'' facts are retained. This module is disabled for \textit{Authors’ Conclusions} from \texttt{CDSR} reviews, as they are expert-written to directly address the question. See Figure~\ref{fig:irrelevant-fact-prompt} for the prompt.

\xhdr{Step 5: Redundancy Filtering Module} After preprocessing and context restoration, some atomic facts became highly redundant. This module removes redundancy among the set of facts extracted from a single sentence. For sentences with more than one atomic fact, \texttt{gpt-5-mini} takes the sentence and its full fact set as input and returns a maximally non-redundant subset, preserving the most atomic and specific facts. See Figure~\ref{fig:redundant-fact-prompt} for the prompt.

\subsection{Cost \& Quality Considerations}\label{appendix:afg-cost}

We design the pipeline to be \textit{modular}, enabling per-component model assignment to balance quality and cost. Atomic fact generation is financially and computationally expensive at our \textit{scale}---processing paragraph-length conclusions with several sentences---so careful allocation is critical. We route early, high-impact stages to more capable models (e.g., \texttt{gpt-5.1}) to ensure high-quality facts, and later well-defined, classification-style stages to smaller models (e.g., \texttt{gpt-5-mini}) to reduce cost.

Specifically, the decomposition (Step 1) and decontextualization (Step 2) modules use \texttt{gpt-5.1} (without reasoning), as we observed higher fact quality without any reasoning. The remaining steps (Steps 3–5) use \texttt{gpt-5-mini} with minimal reasoning effort to efficiently handle classification and filtering while maintaining quality. All components use low verbosity with automatic reasoning summaries. The pipeline is implemented via the Azure OpenAI Responses API.

Across the entire six-step pipeline, for a conclusion with $S$ content sentences that decompose into a total of $F$ atomic facts, the total number of LLM calls is $S + 3F + S'$, where $S'$ is the number of sentences with more than one fact after prior filtering stages. When applying the atomic fact generation pipeline to \textit{Authors’ Conclusions} from \texttt{CDSR} reviews, a \texttt{gpt-5.1}-only pipeline with medium reasoning effort costs \$0.35 per conclusion, while our cost-optimized pipeline reduces this to \$0.13 per conclusion.

\subsection{Validating the Pipeline}\label{appendix:afg-validation}


\xhdr{Overview} We validate the quality of the atomic fact generations through domain-expert annotations by two medical doctors. The task assesses whether each atomic fact represents a single, independent piece of information from its source sentence. Annotators are based at a leading federal university medical school in Brazil with extensive research experience. As inputs, they are given the source sentence, its set of atomic facts, and the full paragraph containing the sentence for context.

\xhdr{Evaluation Dimensions} We evaluate the atomic fact quality using four dimensions across two granularities (e.g., fact-level, sentence-level), grounded in prior works \cite{kaur2025whosaskingsimulatingrolebased, liu-etal-2025-verifact, mir-etal-2019-evaluating, park-etal-2024-valuescope}. 

At the fact-level (per atomic fact), we evaluate:
\begin{itemize}[leftmargin=0.6cm]
    \item \textbf{Faithfulness:} Whether the fact accurately reflects the meaning of the source sentence. \textsc{Answer Choices: Faithful, Unfaithful}
    \item \textbf{Completeness:} Whether the fact represents a single complete piece of information. \textsc{Answer Choices: Complete Fact, Incomplete Fact, Compound Fact}
\end{itemize}

\begin{table}[t]
\centering
\footnotesize
\setlength{\tabcolsep}{4pt}
\caption{Agreement between two expert annotators on evaluating generated atomic facts ($N_{sent}=20$, $N_{facts}=46$) across four dimensions: Faithfulness (Faith.), Completeness (Compl.), Comprehensiveness (Compr.), and Redundancy (Redun.).}
\label{tab:inter-annotator-afg}

\begin{tabular}{lcccc}
\toprule
\textbf{Metric} & \textbf{Faith.} & \textbf{Compl.} & \textbf{Compr.} & \textbf{Redun.} \\
\midrule
Gwet's AC1 
& 0.955 & 0.836 & 0.597 & 0.680 \\
\% Agreement 
& 0.957 & 0.848 & 0.700 & 0.750 \\
\bottomrule
\end{tabular}
\end{table}

At the sentence-level (per sentence and its set of atomic facts), we evaluate:
\begin{itemize}[leftmargin=0.6cm]
    \item \textbf{Comprehensiveness:} Whether the set of facts completely and accurately captures the meaning of the source sentence. \textsc{Answer Choices: Comprehensive, Partially Comprehensive, Not Comprehensive}
    \item \textbf{Redundancy:} Whether any facts are duplicated or substantially overlap in content. \textsc{Answer Choices: No Redundancy, Redundancy}
\end{itemize}

\begin{figure*}[t]
    \centering

    \begin{subfigure}[t]{0.48\textwidth}
        \centering
        \includegraphics[width=6cm, height=8cm, keepaspectratio]{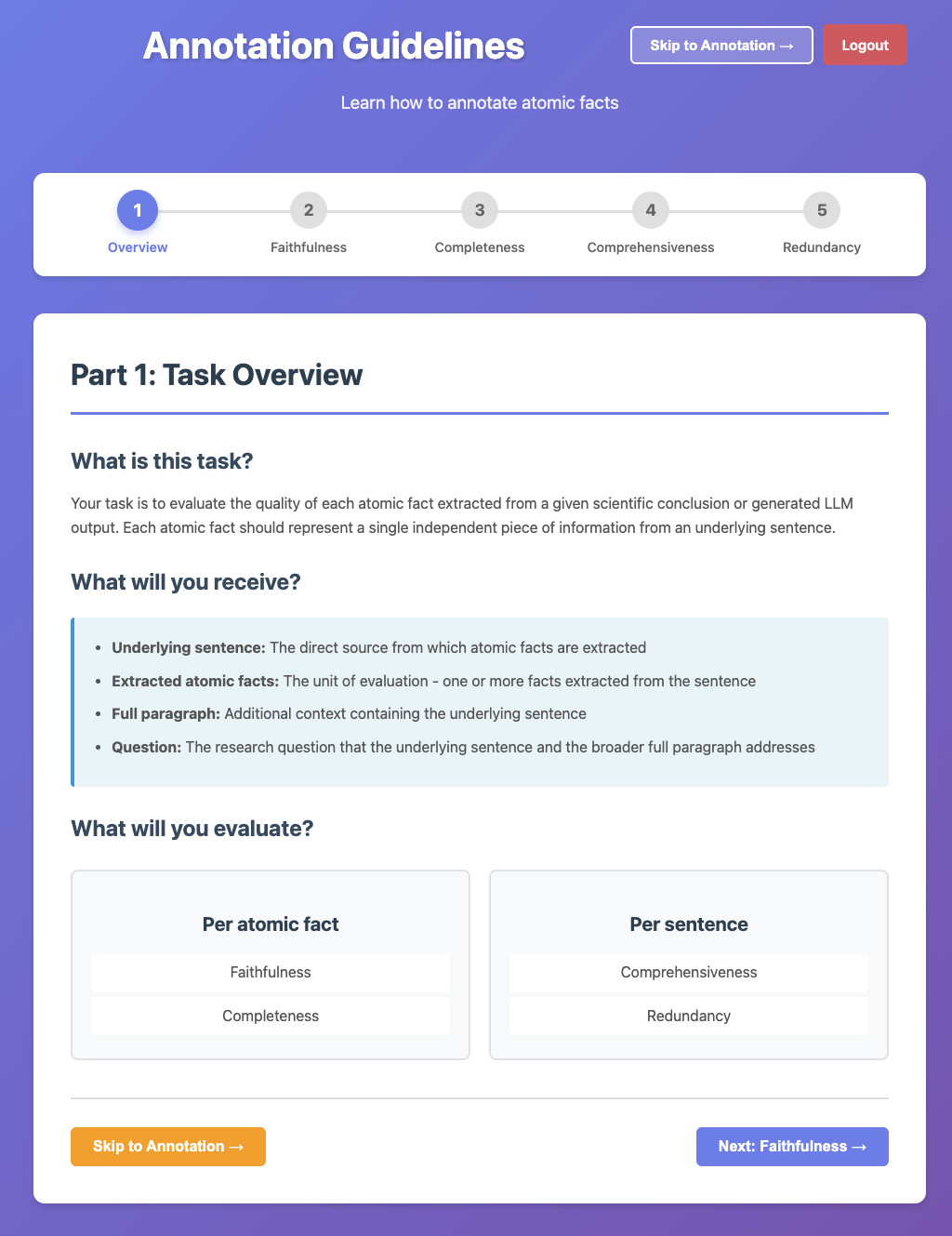}
        \caption{Task Overview Page}
    \end{subfigure}
    \hfill
    \begin{subfigure}[t]{0.48\textwidth}
        \centering
        \includegraphics[width=\linewidth, trim={0cm 23cm 0cm 0cm}, clip]{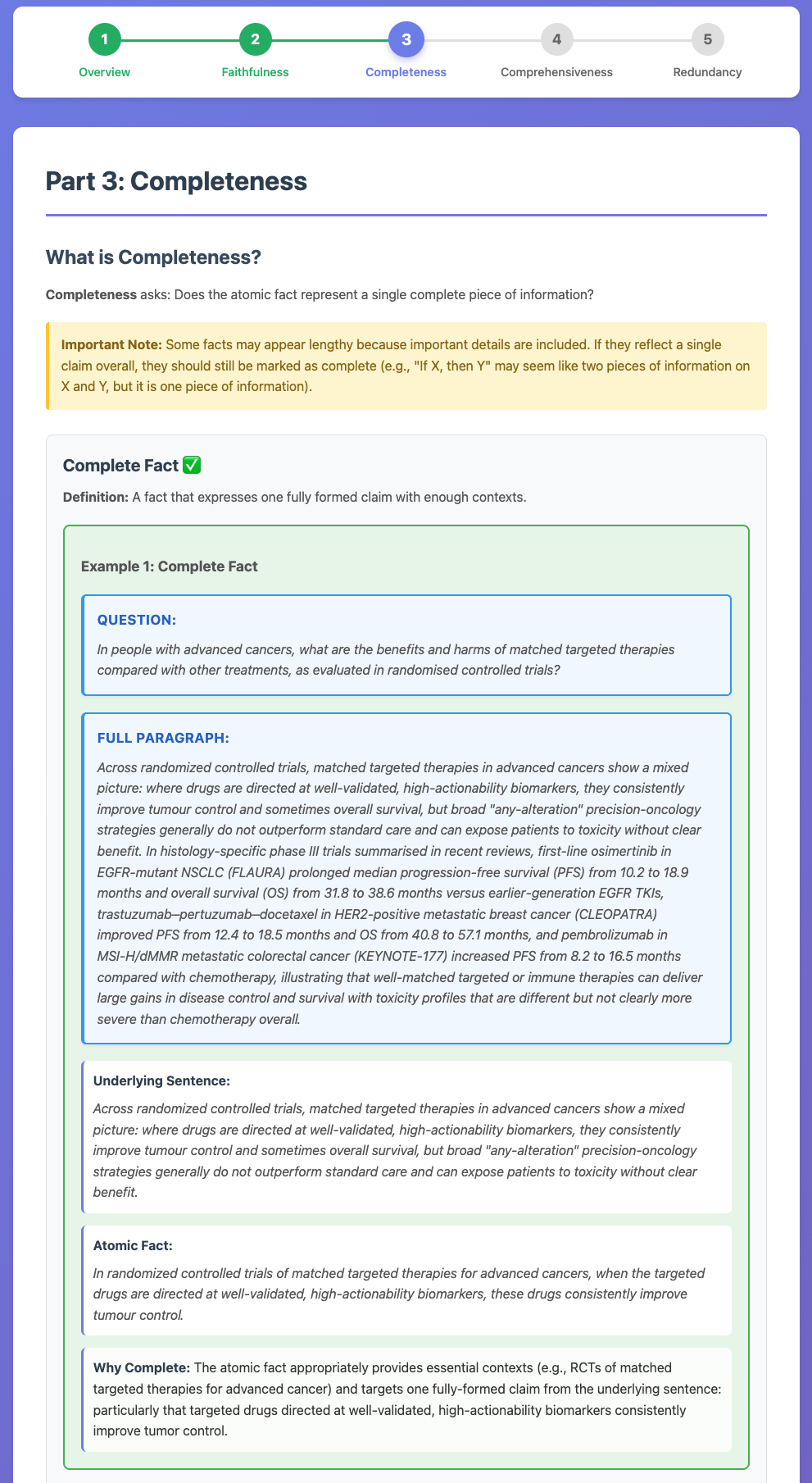}
        \caption{Definition and Examples of Completeness}
    \end{subfigure}
    \vspace{0.75em}

    \begin{subfigure}[t]{0.48\textwidth}
        \centering
        \includegraphics[width=6cm, trim={0cm 13cm 0cm 0cm}, clip]{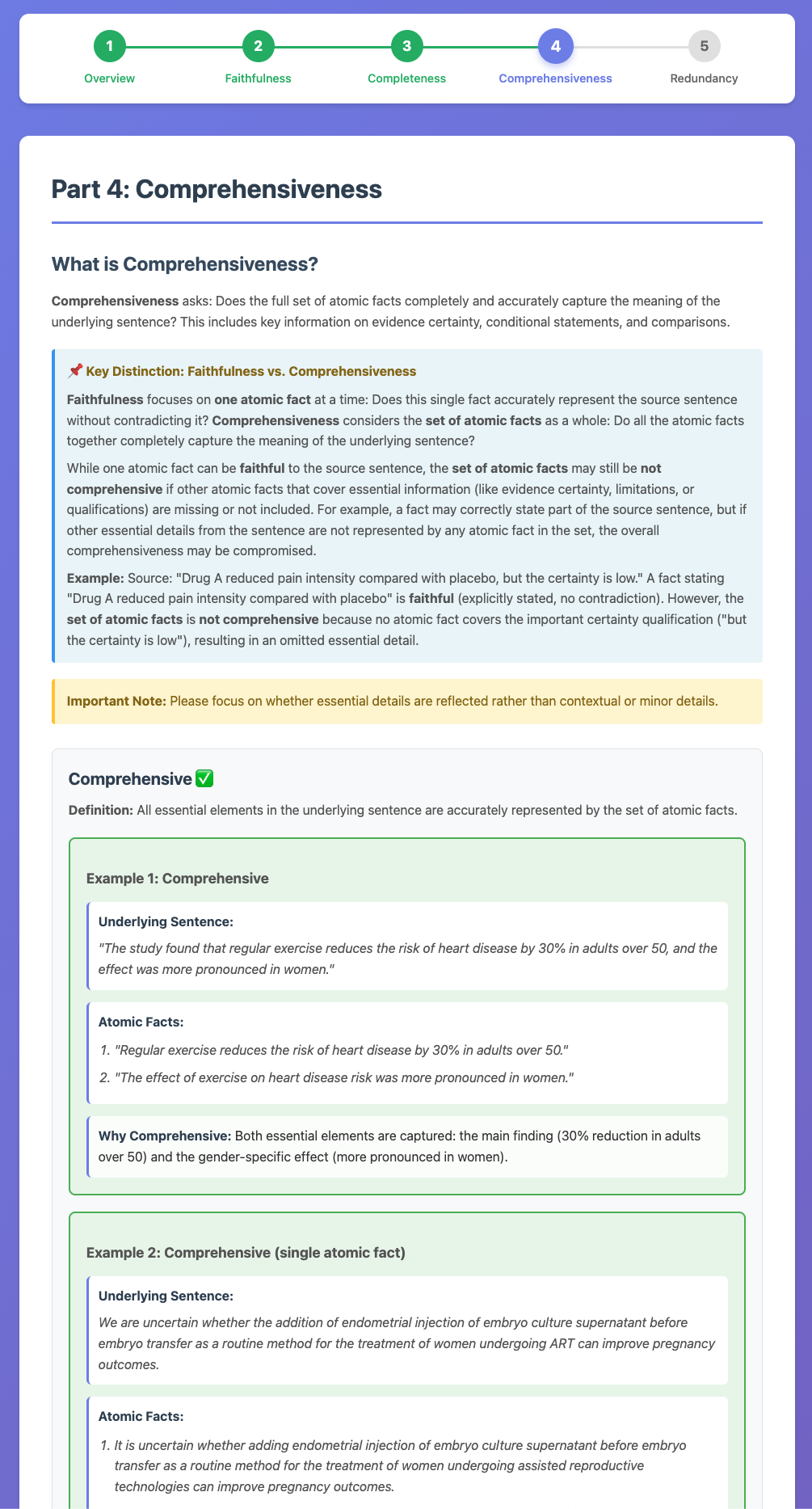}
        \caption{Definition and Examples of Comprehensiveness}
    \end{subfigure}
    \hfill
    \begin{subfigure}[t]{0.48\textwidth}
        \centering
        \includegraphics[width=\linewidth, trim={0cm 8cm 0cm 0cm}, clip]{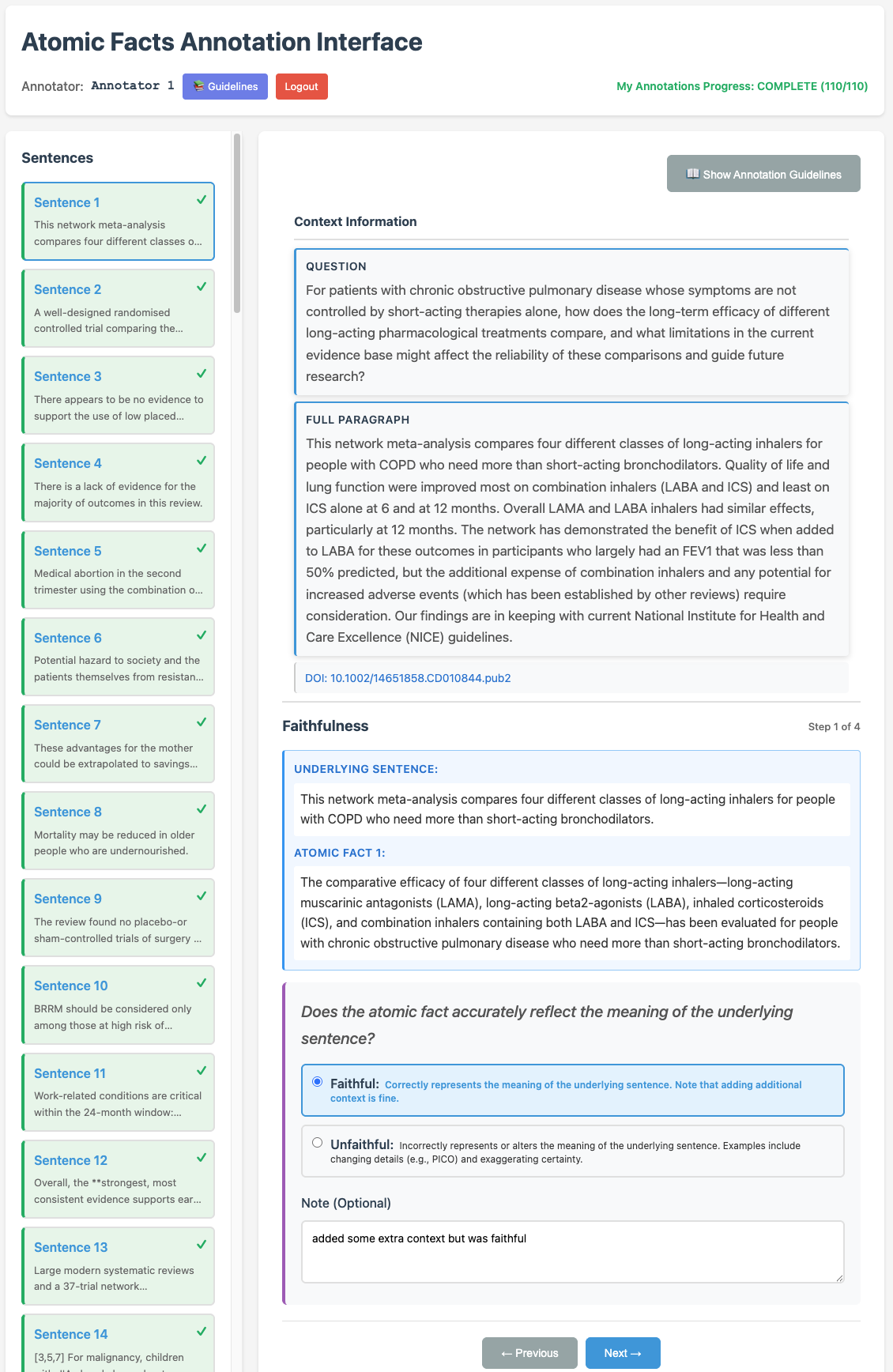}
        \caption{Annotation Interface}
    \end{subfigure}

    \caption{Annotation interface for evaluating atomic facts. Panels show the task introduction and guidelines: (a) task overview page, (b) definition and examples of completeness, (c) definition and examples of comprehensiveness, and (d) annotation interface. Panels are cropped or resized for space.}
    \label{fig:afg-interface}
\end{figure*}

The annotation guidelines were iteratively refined with feedback from graduate students in computer science, clinicians, and medical students, who reviewed the annotation guidelines and worked through sample annotations. The full annotation guidelines are provided in Figure \ref{fig:annotation-guidelines-atomic-facts}, and the annotation interface used by annotators is shown in Figure~\ref{fig:afg-interface}.

\xhdr{Annotation Procedure} To validate our annotation task, two expert annotators independently follow the annotation guidelines to label 20 sentences and their corresponding atomic facts across all evaluation dimensions. The sentences and atomic facts are stratified and sampled from both generated and reference conclusions from \texttt{CDSR} reviews. In total, the 20 sentences contain 46 atomic facts. We assess annotator reliability using Gwet’s AC1, as done in \S~\ref{appendix:question-gen-validation}. As mentioned above, fact-level dimensions (e.g., faithfulness, completeness) are evaluated per atomic fact ($N_{facts}{=}46$), while sentence-level dimensions (e.g., comprehensiveness, redundancy) are evaluated per sentence and its set of atomic facts ($N_{sent}{=}20$).

\begin{table}[t]
\centering
\setlength{\tabcolsep}{2pt} 
\footnotesize
\caption{Label distribution across evaluation dimensions for atomic facts. Overall, the results are reported over $N_{\text{sent}}{=}200$ sentences (evenly stratified between generated and reference conclusions from \texttt{CDSR}), corresponding to $N_{\text{facts}}{=}469$ atomic facts. Abbreviations: Compre. = Comprehensive, Partial = Partially Comprehensive, Not Compre. = Not Comprehensive.}
\resizebox{\linewidth}{!}{
\begin{tabular}{lcccccccccc}
\toprule
\textbf{Type} 
& \multicolumn{2}{c}{\textbf{Faithfulness}} 
& \multicolumn{3}{c}{\textbf{Completeness}} 
& \multicolumn{3}{c}{\textbf{Comprehensiveness}} 
& \multicolumn{2}{c}{\textbf{Redundancy}} \\
\cmidrule(lr){2-3} \cmidrule(lr){4-6} \cmidrule(lr){7-9} \cmidrule(lr){10-11}
& Faithful & Unfaithful
& Complete & Incomplete & Compound 
& Compre. & Partial & Not Compre. 
& No Redundancy & Redundancy \\
\midrule
Humans 
& 96.0 & 4.0 
& 98.0 & 2.0 & 0.0 
& 98.0 & 2.0 & 0.0 
& 94.0 & 6.0 \\
LLM 
& 96.6 & 3.4 
& 95.0 & 4.4 & 0.6 
& 98.0 & 2.0 & 0.0 
& 87.0 & 13.0 \\
Overall 
& 96.4 & 3.6 
& 96.0 & 3.6 & 0.4 
& 98.0 & 2.0 & 0.0 
& 90.5 & 9.5 \\
\bottomrule
\end{tabular}}
\vspace{2pt}
\label{tab:annotation-quality-afg}
\end{table}

\xhdr{Results} As shown in Table \ref{tab:inter-annotator-afg}, the inter-annotator agreement is high across the evaluation dimensions (Gwet's AC1: 0.597--0.955; Percentage Agreement: 0.700--0.957), comparable to or even exceeding prior works \cite{kaur2025whosaskingsimulatingrolebased, liu-etal-2025-verifact, mir-etal-2019-evaluating, park-etal-2024-valuescope}. These results validate both annotator reliability and the overall annotation setup. Following this, the two expert annotators independently label 90 sentences each, stratified evenly between generated and reference conclusions from \texttt{CDSR} reviews, yielding $N_{\text{sent}}{=}200$ sentences and $N_{\text{facts}}{=}469$ atomic facts.

Table~\ref{tab:annotation-quality-afg} summarizes the validation results. Overall, annotators find the generated atomic facts to be largely faithful (96.4\%), complete (96.0\%), comprehensive with respect to the source sentence (98.0\%), and non-redundant (90.5\%).  These findings indicate that the atomic fact generation pipeline reliably produces high-quality atomic facts, supporting their use in downstream decomposition for measuring factual quality in long-form scientific conclusions.

\xhdr{\camready{Failure Mode Analysis}} \camready{We identify two main failure modes in atomic fact generation (Figure~\ref{fig:afg-failure-examples}): (1) \textit{overstating certainty}, where the pipeline converts hedged evidence into a broader or more definitive claim, and (2) \textit{missing magnitude or effect size}, where the pipeline omits quantitative details needed to interpret a clinical finding. In the first example, the source reports reduced chemotherapy toxicity with melatonin only in ``some studies,'' but the extracted fact generalizes this into a broader association; expert annotators rated it \textsc{Unfaithful}. In the second, the extracted fact preserves the direction and uncertainty of the lymphoma risk but omits the hazard ratio and confidence interval needed to interpret its clinical significance; expert annotators rated it \textsc{Incomplete Fact}. As shown in our expert validation (Table~\ref{tab:annotation-quality-afg}), such failures are rare.}

\begin{figure}[h!]
\centering
\begin{small}
\begin{tcolorbox}[colback=gray!5, colframe=red, width=\textwidth, title=\textbf{Failure Mode 1: Overstating Certainty}]
\textsc{Source Sentence:} ``In contrast, more focused data are more encouraging for sleep: a breast-cancer--specific meta-analysis of 5 trials ($n = 243$) reported\ldots{} only minor adverse effects\ldots{} with \textbf{some studies} also noting reduced chemotherapy toxicity.''\\[2pt]
\textsc{Atomic Fact:} ``In adults with breast cancer receiving chemotherapy, melatonin \textbf{has been associated with} reduced chemotherapy toxicity in some treatment settings.''\\[2pt]
\textsc{Expert Label:} Faithfulness: \textsc{Unfaithful}
\end{tcolorbox}
\begin{tcolorbox}[colback=gray!5, colframe=red, width=\textwidth, title=\textbf{Failure Mode 2: Missing Magnitude or Effect Size}]
\textsc{Source Sentence:} ``For malignancy, children with JIA\ldots{} found no statistically significant additional increase in overall cancers during the first few years of TNFi exposure\ldots{} although a modestly higher lymphoma risk could not be excluded \textbf{(HR 2.64, 95\% CI [0.93, 7.51])}\ldots''\\[2pt]
\textsc{Atomic Fact:} ``In children with juvenile idiopathic arthritis treated with tumor necrosis factor inhibitors, a modestly higher risk of lymphoma cannot be excluded.''\\[2pt]
\textsc{Expert Label:} Completeness: \textsc{Incomplete Fact}
\end{tcolorbox}
\end{small}
\caption{\camready{Example atomic fact generation failures. (1) The fact overgeneralizes a finding reported only in ``some studies'' into a broader association. (2) The fact omits the effect size and confidence interval (\textbf{bold}) needed to interpret the finding's magnitude and clinical significance.}}
\label{fig:afg-failure-examples}
\end{figure}

\clearpage

\begin{figure*}[h!]
\centering
\begin{small}
\begin{tcolorbox}[colback=gray!5, colframe=black, width=\textwidth, title=\textbf{Annotation Guideline for Evaluating Generated Atomic Facts}, fonttitle=\bfseries]
\textbf{Task Overview.}
Your task is to evaluate the quality of each atomic fact extracted from a given scientific conclusion or generated LLM output. Each atomic fact should represent a single independent piece of information from an underlying sentence and should not require additional context. 

\vspace{0.5em}
\textbf{Inputs.}
You will be provided with the following:

\begin{itemize}[leftmargin=0.6cm]
    \item \textbf{Underlying Sentence}: Direct source for atomic facts.
    \item \textbf{Extracted Atomic Facts}: Unit of evaluation, extracted from the sentence
    \item \textbf{Full Paragraph}: Paragraph containing the underlying sentence for additional context.
\end{itemize}

\vspace{0.5em}
\textbf{Evaluation Criteria.}
Below, we list two sets of evaluation criteria across different granularities (e.g., fact-level vs. sentence-level). You can optionally leave a note for each annotation.
\vspace{0.5em}

\textit{Atomic Fact-Level Criteria:}
For each atomic fact, please evaluate:
\begin{itemize}[leftmargin=0.6cm]
    \item\textbf{Faithfulness.} \textit{Does the atomic fact accurately reflect the meaning of the underlying sentence?}
    \vspace{-0.2em}
     \begin{itemize}[leftmargin=0.5cm]
        \item \textsc{Faithful}: Correctly represents the meaning of the underlying sentence. Note that adding additional context is fine.
        \item \textsc{Unfaithful}: Incorrectly represents or alters the meaning of the underlying sentence. Examples include changing details (e.g., PICO) and exaggerating certainty.
    \end{itemize}

    \vspace{0.1em}
    \item \textbf{Completeness.} \textit{Does the atomic fact represent a single complete piece of information?} 
     \vspace{-0.2em}
     \begin{itemize}[leftmargin=0.5cm]
        \item\textsc{Complete Fact}: A fact that expresses one fully formed claim with all essential contexts. 
        \item\textsc{Incomplete Fact}: Missing essential information required to interpret the claim on its own, making the fact unclear without additional context. 
        \item\textsc{Compound Fact}: Contains multiple distinct claims or outcomes within a single statement, even after accounting for necessary contextual details.
    \end{itemize}
    \vspace{-0.2em}
    \text{Note}: Some facts may appear lengthy. If they reflect a single claim overall, they should be marked complete (e.g., ``If X, Y'' may seem like two pieces of information on X and Y, but it is just one). 

\end{itemize}

\textit{Sentence-Level Criteria: }
For each sentence \& its set of atomic fact(s), please evaluate:
\begin{itemize}[leftmargin=0.6cm]
    \item \textbf{Comprehensiveness.} \textit{Does the full set of atomic facts completely and accurately capture the meaning of the underlying sentence?} 
    \vspace{-0.2em}
     \begin{itemize}[leftmargin=0.5cm]
    \item\textsc{Comprehensive}: All essential elements in the underlying sentence are accurately represented by the set of atomic facts. 
    \item\textsc{Partially Comprehensive}: The atomic facts capture the core meaning of the sentence but omit, weaken, or inaccurately reflect minor elements that do not fundamentally change the meaning of the sentence. 
    \item\textsc{Not Comprehensive}: The atomic facts omit or misrepresent essential meaning to the extent that the overall meaning of the sentence is distorted.
    \end{itemize}
    \vspace{-0.2em}
    Note: Please focus on whether essential details are reflected rather than contextual or minor details.   

    \vspace{0.5em}
    \item \textbf{Redundancy.} \textit{Are any atomic facts substantially duplicated or redundant with another fact in information content?} 
     \vspace{-0.2em}
     \begin{itemize}[leftmargin=0.5cm]
    \item\textsc{No Redundancy:} All facts are not redundant or contribute distinct information. A set of a single atomic fact should be marked with this.
    \item\textsc{Redundancy:} One or more atomic facts repeat information already expressed in another fact (e.g., duplicates or rephrasing) and do not contribute any additional meaning, even minor distinctions.
    \end{itemize}
    \vspace{-0.2em}
    Note: Some atomic facts may share partial information. If each fact contributes distinct information, even if minor, and is not a rephrasing, it should be ``No Redundancy.'' 
\end{itemize}

\end{tcolorbox}
\caption{Annotation guidelines for evaluating atomic facts extracted from scientific conclusions. The instruction defines fact-level criteria (faithfulness, completeness) and sentence-level criteria (comprehensiveness, redundancy) to assess the quality of the decomposed facts.}
\label{fig:annotation-guidelines-atomic-facts}
\end{small}
\end{figure*}

\clearpage

\begin{figure}[t]
\begin{small}
\centering
\begin{tcolorbox}[
    colback=gray!5,
    colframe=black,
    width=\linewidth,
    arc=2mm,
    boxrule=0.5pt,
    title=\textbf{Atomic Fact Generation (Step 1): Decomposition Prompt}
]
\textbf{System Prompt:} \textcolor{teal}{You are an expert in breaking down complex scientific sentences into atomic facts--short statements that each contain one piece of information. Given the entire scientific paragraph as context, you will be given one of the sentences and you will need to break the sentence down into atomic facts.}\\

\textbf{Instruction:} Using the full scientific paragraph as context, please break down one of the sentences from the paragraph into independent atomic facts and justify your answer. The atomic facts should be short statements that each contain one piece of information.\\

\texttt{***GUIDELINES FOR GENERATING ATOMIC FACTS STARTS HERE:***}\\
- Make sure that each atomic fact is independent. Each atomic fact in the output should contain different pieces of information.\\
- Make sure the atomic facts can stand on its own as a fact, using the provided paragraph to contextualize each atomic fact.\\
- Do not create atomic facts that are inferred or indirect from the sentence. Focus directly on the sentence and the information it contains.\\
- Do not create atomic facts specific to the study itself, but rather focus on the main conclusions.\\
- Do not assume that atomic facts inform one another in terms of context. Treat each atomic fact as a separate, independent fact, each of which needs its own context to be understood on its own.\\
- Do not generate atomic facts that reflect authorial judgments or methodological decisions made by the systematic review authors, or that require that context to be understood.\\
- DO NOT generate atomic facts for non-content elements...\\
- If the sentence contains only non-content elements, return an empty list of atomic facts.\\
\texttt{***GUIDELINES FOR GENERATING ATOMIC FACTS ENDS HERE.***}\\

Here are five examples of how to breakdown a sentence into independent atomic facts:\\

\texttt{***EXAMPLE 1 STARTS HERE***}\\
*Full Paragraph:* ...\\
*Sentence:* ...\\
*Atomic facts:* ...\\
*Justification*: ... \\
\texttt{***EXAMPLE 1 ENDS HERE***}\\
... \\
...\\

Now, given what you learned from the guidelines and examples, please break down the sentence into independent atomic facts, using the provided paragraph and question as context and providing justification and thinking step-by-step about your answer. Avoid creating atomic facts specific to the study itself, but rather focus on the main conclusions.\\

\texttt{***QUESTION STARTS HERE*** Use this question as context for your task.}\\
\{question\}\\
\texttt{***QUESTION ENDS HERE***}\\

\texttt{***FULL PARAGRAPH STARTS HERE*** Use this paragraph as context for your task.}\\
\{paragraph\}\\
\texttt{***FULL PARAGRAPH ENDS HERE***}\\

\texttt{***SENTENCE STARTS HERE***} \\
\{sentence\} \\ 
\texttt{***SENTENCE ENDS HERE***}\\

The atomic facts should be short statements that each contain one piece of information. Output should be in JSON format with the following keys:\\
- ``atomic\_facts'': list of atomic facts as strings\\
- ``justification'': string justification for your atomic fact breakdown.
\end{tcolorbox}

\caption{Decomposition prompt used to decompose sentences into \textit{atomic facts} using \texttt{gpt-5.1}. The sentence, its parent paragraph, and the generated question were provided as input, filling in their bracketed components in the prompt. The prompt was shortened to fit within the page. Full prompt available in supplementary material / code.}
\label{fig:decomposition-prompt}
\end{small}
\end{figure}

\clearpage

\begin{figure}[t]
\begin{small}
\centering
\begin{tcolorbox}[
    colback=gray!5,
    colframe=black,
    width=\linewidth,
    arc=2mm,
    boxrule=0.5pt,
    title=\textbf{Atomic Fact Generation (Step 2): Decontextualization Prompt}
]
\textbf{System Prompt:} \textcolor{teal}{You are an expert in decontextualizing vague references in scientific sentences into more specific entities, ensuring that any statement relations are not out of context and are self-contained. Given a statement and a response, you will need to decontextualize the vague references in the statement.}\\

\textbf{Instruction:} Evaluate the provided statement in context of the response. The statement should be decontextualized such that they are understandable without referencing the rest of the response.\\

\texttt{Instructions:}\\
1. The following STATEMENT has been extracted from the broader context of the given RESPONSE.\\
2. Modify the STATEMENT by replacing vague references with the proper entities from the RESPONSE that they are referring to.\\
3. You MUST NOT change any of the factual claims made by the original STATEMENT.\\
4. You MUST NOT add any additional factual claims to the original STATEMENT.\\
5. You MUST NOT generate atomic facts that reflect authorial judgements or methodological decisions made by the systematic review authors.\\
6. Before giving your revised statement, think step-by-step and show your reasoning. As part of your reasoning, be sure to identify the subjects in the STATEMENT and determine whether they are vague references. If they are vague references, identify the proper entity that they are referring to and be sure to revise this subject in the revised statement.\\
7. Your task is to do this for the STATEMENT and RESPONSE under “Your Task”. Some examples have been provided for you to learn how to do this task.\\

Vague references include but are not limited to:\\
- Pronouns (e.g., ``his'', ``they'', ``her'')\\
- Unknown entities (e.g., ``this event'', ``the research'', ``the invention'', ``this review'', ``the study'', ``many studies'')\\
- Non-full names (e.g., ``Jeff...'' or ``Bezos...'' when referring to Jeff Bezos)\\
- Systematic review, specific studies, or specific evidence (e.g., ``the review'', ``the study'', ``many studies'', ``A review'', ``Our review'', ``this evidence'', ``in the review'', ``in this evidence'')\\
You SHOULD NOT generate atomic facts that reference the systematic review or specific studies in any shape or form e.g., ``The certainty of evidence in this systematic review...''\\

\texttt{Example 1:}\\
STATEMENT: ...\\
RESPONSE: ...\\
REVISED STATEMENT: ...\\

...\\
...\\

\texttt{Your Task:}\\
\texttt{QUESTION:}\\
\{question\}\\
\texttt{STATEMENT:}\\
\{individual\_fact\}\\
\texttt{RESPONSE:}\\
\{response\}\\

The atomic fact should be short statements that each contain one piece of information. Use the QUESTION as additional context to help understand what the STATEMENT and RESPONSE are referring to. Output should be in JSON format with the following keys:\\
- ``decontextualized\_fact'': a string of the atomic fact decontextualized to be self-contained.\\
- ``justification'': string justification for your decontextualization process of the atomic fact.\\
\end{tcolorbox}

\caption{Decontextualization prompt for making \textit{atomic facts} self-contained using \texttt{gpt-5.1}. Inputs include the fact, its parent paragraph, and the generated question, which replaced their bracketed component in the prompt. The prompt was shortened to fit within the page. Full prompt available in supplementary material / code.}
\label{fig:decontextualization-prompt}
\end{small}
\end{figure}

\clearpage

\begin{figure}[t]
\begin{small}
\centering
\begin{tcolorbox}[
    colback=gray!5,
    colframe=black,
    width=\linewidth,
    arc=2mm,
    boxrule=0.5pt,
    title=\textbf{Atomic Fact Generation (Step 3): Incomplete Fact Rewriting Prompt}
]
\textbf{Instruction:} Given a context and a claim extracted from the context, determine whether the claim is Dependent or Independent of the context.\\
* \texttt{Independent}: If the claim itself precisely reflects the original meaning of the context without further explanation. \\
* \texttt{Dependent}: If the claim requires additional context or detail to precisely reflect its original meaning.\\

Categorize dependent claims into one of three types:

* \textsc{Ambiguous Concepts/Pronouns}\\
    The claim contains vague terms (e.g., ``this method,'' ``they'') or pronouns lacking clear referents from the context.\\
    Example:\\
    Context: `Decarbonizing aviation requires SAFs.''\\
    Claim: ``They reduce emissions.'' → Dependent (Ambiguous pronoun ``they'').\\
* \textsc{Missing Comparison}\\
    The claim implies a comparison (e.g., ``more,'' ``better'') but omits the explicit comparison target stated in the context.\\
    Example:\\
    Context: ``SAFs reduce emissions by 80\% compared to jet fuels.''\\
    Claim: ``SAFs reduce emissions by 80\%.'' → Dependent (Missing ``compared to jet fuels'').\\
* \textsc{Lack of Condition}\\
    The claim omits critical contextual details, such as:\\
    - Temporal conditions (e.g., ``as of 2023'').\\
    - Hypothetical scenarios (e.g., ``if regulations are adopted'').\\
    Example:\\
    Context: ``As of 2023, the U.S. top 1\% net worth is ~\$10M (Smith et al., 2023).''\\
    Claim: ``The U.S. top 1\% net worth is ~\$10M.'' → Dependent (Missing time).\\

ONE CAVEAT: Do not mark a claim as Dependent based on not having enough context regarding the scope of claim. The atomic fact is intentionally designed to be self-contained and avoid having any specific details regarding the review or specific studies.\\

If the claim is Dependent, you must also provide a rewritten version that makes it Independent by incorporating the missing context. Some guidelines for rewriting dependent claims:\\
- You MUST NOT rewrite atomic facts that reflect authorial judgements or methodological decisions made by the systematic review authors.\\

Output should be in JSON format with the following keys:\\
- ``classification'': string, either ``Independent'' or ``Dependent''\\
- ``dependent\_type'': string, one of ``Ambiguous Concepts/Pronouns'', ``Missing Comparison'', ``Lack of Condition'', or ``None'' (if Independent)\\
- ``explanation'': string, explanation for your classification\\
- ``rewritten\_claim'': string, if classification is ``Dependent'', provide a rewritten version that makes the claim Independent by incorporating missing context. If ``Independent'', this should be empty. \\

\# Example\\
Context:...\\
Claim:...\\
Your Response:...\\
...\\

\# Your Task
Context:\\
\{context\}

Claim:\\
\{claim\}

Your Response:
\end{tcolorbox}

\caption{Prompt for incomplete fact rewriting. Given an atomic fact as the ``claim'' and its parent paragraph as the ``context,'' \texttt{gpt-5-mini} classifies the fact as \textit{Independent} or \textit{Dependent}; dependent facts are rewritten by integrating the missing context. The prompt was shortened to fit within the page. Full prompt available in supplementary material / code.}
\label{fig:incomplete-fact-prompt}
\end{small}
\end{figure}

\clearpage

\begin{figure}[t]
\begin{small}
\centering
\begin{tcolorbox}[
    colback=gray!5,
    colframe=black,
    width=\linewidth,
    arc=2mm,
    boxrule=0.5pt,
    title=\textbf{Atomic Fact Generation (Step 4): Relevance Filtering Prompt}
]
\textbf{Instruction:} A STATEMENT is considered ``Foo'' if the STATEMENT is directly relevant or provides beneficial background context to addressing the QUESTION in context of the RESPONSE.\\

\texttt{Instructions:}\\
1. The following STATEMENT has been extracted from the broader context of the given RESPONSE to the given QUESTION.\\
2. First, consider the provided STATEMENT and QUESTION in context of the RESPONSE.\\
3. Next, determine whether the STATEMENT is considered ``Foo'' e.g., is it directly relevant or provides beneficial background context to addressing the QUESTION in context of the RESPONSE.\\
4. Before showing your answer, think step-by-step and show your specific reasoning.\\
5. If the STATEMENT is considered ``Foo'', say ``\text{[Foo]}'' after showing your reasoning. Otherwise show ``\text{[Not Foo]}'' after showing your reasoning.\\
6. Your task is to do this for the STATEMENT and RESPONSE under ``Your Task''. Some examples have been provided for you to learn how to do this task.\\

\texttt{**General Rule of Thumb**}\\
- If a statement is generic response (e.g., ``I cannot help with that,'' ``I am not familiar with that,'' etc.) or does not provide any information that is directly relevant to addressing the QUESTION in context of the RESPONSE, then the STATEMENT is ``\text{[Not Foo]}''.\\
- If a statement is more indirect or inferred from the RESPONSE, then the STATEMENT is considered ``\text{[Not Foo]}''. The statement should be self-contained, while being directly relevant or providing helpful background context to addressing the QUESTION in context of the RESPONSE. Err to the side of \text{[Foo]} if the statement provides background context to the RESPONSE. For example, statement on the need for additional high-quality research on some topic and related outcomes provides helpful background context and addresses the QUESTION in context of the RESPONSE.\\

\texttt{Example 1:}\\
QUESTION: ...\\
RESPONSE: ...\\
STATEMENT: ...\\
SOLUTION: ...\\
...\\

\texttt{Your Task:}\\
QUESTION:\\
\{question\}\\
RESPONSE:\\
\{response\}\\
STATEMENT:\\
\{fact\}\\

Output should be in JSON format with the following keys:\\
- ``reasoning'': string, step-by-step reasoning for your determination\\
- ``classification'': string, either ``\text{[Foo]}'' or ``\text{[Not Foo]}''
\end{tcolorbox}

\caption{Prompt for Relevance Filtering. Given an atomic fact, its parent paragraph as the response, and the question, \texttt{gpt-5-mini} classifies the fact as \textit{relevant} (i.e., \textit{Foo}) or \textit{irrelevant} (i.e., \textit{Not Foo}); only relevant facts are retained. The prompt was shortened to fit within the page. Full prompt available in supplementary material / code.}
\label{fig:irrelevant-fact-prompt}
\end{small}
\end{figure}

\clearpage

\begin{figure}[t!]
\begin{small}
\centering
\begin{tcolorbox}[
    colback=gray!5,
    colframe=black,
    width=\linewidth,
    arc=2mm,
    boxrule=0.5pt,
    title=\textbf{Atomic Fact Generation (Step 5): Redundancy Filtering Prompt}
]
\textbf{Instruction:} You are given a RESPONSE and a list of STATEMENTS extracted from that RESPONSE. Your task is to select the most atomic set of facts by removing redundant statements, while keeping as many statements that provide as much information about the RESPONSE as possible.\\

\texttt{Instructions:}\\
1. You are given a RESPONSE and a numbered list of STATEMENTS extracted from that RESPONSE.\\
2. Your goal is to select the most atomic set of facts by:\\
   - Identifying groups of redundant statements (statements that convey exactly the same meaning)\\
   - For each redundant group, keeping ONLY the best statement(s). Keep the set that retains as many atomic facts as possible while accurately representing the original RESPONSE.\\
3. A statement is redundant with another if:\\
   - It expresses the EXACT SAME meaning as another statement (even if worded differently)\\
   - It does not provide any further information beyond what is already covered by other statements.\\
4. When choosing which statement to keep from a redundant group:\\
   - STRONGLY prefer more atomic facts: If a statement combines multiple pieces of information (e.g., ``diarrhea, nausea, and vomiting''), prefer keeping individual atomic statements (e.g., separate statements for diarrhea, nausea, vomiting) over the combined statement\\
   - Prefer more specific statements over general ones: If a statement is too broad and does not accurately represent the original RESPONSE, prefer more specific statements that provide comprehensive information\\
5. A statement should be kept if:\\
   - It provides unique information not covered by any other statement\\
   - It is the best version in its redundant group (most atomic, most specific, most accurate to RESPONSE)\\
   - It accurately represents the original RESPONSE with specific details\\
   - When in doubt, err on the side of keeping the fact.\\
6. A statement should be removed if:\\
   - It is redundant with another statement and does not provide any further information beyond what is already covered\\
   - It is overly general and does not accurately represent the original RESPONSE\\
7. Before showing your answer, think step-by-step about:\\
   - Which statements are redundant with each other\\
   - Which statements are more atomic (individual facts vs combined facts)\\
   - Which statements are more specific and accurately represent the RESPONSE\\
   - Which statements are overly general and should be removed\\
   
THINK CAREFULLY STEP-BY-STEP BEFORE EXCLUDING REDUNDANT STATEMENTS. \\
...\\

\texttt{Example 1:}\\
RESPONSE: ...\\
STATEMENTS: ...\\
SOLUTION: ...\\
Selected statements: ...\\
...\\

\texttt{Your Task:}\\
RESPONSE:\\
\{response\}\\
STATEMENTS:\\
\{all\_facts\}\\

Output should be in JSON format with the following keys:\\
- ``reasoning'': string, step-by-step reasoning explaining which statements are redundant, which ones you're keeping, and why\\
- ``selected\_statements'': array of integers, the statement numbers (1-based) from the STATEMENTS list that should be kept in the final atomic set. This should contain no redundant statements and preserve all unique information.
\end{tcolorbox}

\caption{Prompt for redundancy filtering. Given a sentence as the response and its full set of atomic facts, \texttt{gpt-5-mini} outputs the maximally non-redundant subset. The prompt was shortened to fit within the page. Full prompt available in supplementary material / code.}
\label{fig:redundant-fact-prompt}
\end{small}
\end{figure}

\clearpage

\section{Details on Measuring Factual Precision and Recall}\label{appendix:details-measuring-fact}

We employ LLM-based judges to measure factual precision and recall at \textit{scale} by assigning labels to individual facts. To validate these judges, we construct an expert-annotated gold-standard dataset with medical doctors with extensive clinical practice and research experience (\S\ref{appendix:fact-eval-annotation}) and evaluate judge performance against these annotations (\S\ref{appendix:fact-eval-llm}).

\subsection{Creating the Expert-Annotated Gold-Standard Dataset}\label{appendix:fact-eval-annotation}

We describe the annotation guidelines (\S\ref{appendix:fact-eval-annotation-task}) and the annotation procedure with three medical doctors, including inter-annotator agreement (\S\ref{appendix:fact-eval-annotation-procedure}), used to construct the gold-standard dataset.

\subsubsection{Annotation Guidelines}\label{appendix:fact-eval-annotation-task}
To construct our gold-standard dataset, annotators perform two tasks: factual precision and factual recall. For both tasks, annotators are not informed whether facts originate from model-generated or \texttt{CDSR} \textit{Authors' Conclusions} to mitigate potential priming effects that may lead annotators to be overly cautious or systematically biased against model-generated conclusions.

\xhdr{Factual Precision Task}
As described in \S\ref{sec:factual-metrics}, factual precision measures the \textit{correctness} of the generated conclusion. In this task, annotators assess whether each fact is factually supported and non-contradictory with respect to a trusted source text---the \texttt{CDSR} review. For each fact, annotators carefully examine the full \texttt{CDSR} review article on the web to assign one of the following labels, along with supporting excerpts and a brief justification: \textsc{Supported}, \textsc{Contradicted}, and \textsc{Not Supported}. All facts evaluated in this task are extracted from generated conclusions, though annotators are not informed of their origin. See Figure \ref{fig:annotation_guidelines-fact-precision} for the full annotation guidelines and label definitions.

\xhdr{Factual Recall Task}
Factual recall measures the \textit{coverage} of the generated conclusion, evaluating the extent to which generated conclusions \textit{cover} facts from the \textit{Authors' Conclusions} of \texttt{CDSR} reviews, which are treated as the authoritative set of facts required to answer the question.  In this task, annotators assess whether each fact is supported by a conclusion. For each fact, annotators examine the conclusion to assign one of two labels, along with supporting excerpts and a brief justification: \textsc{Supported} and \textsc{Not Supported}. All facts evaluated in this task are derived from the \textit{Authors' Conclusions} of \texttt{CDSR} reviews, while all evaluated conclusions are model-generated; their origins are not disclosed to annotators.  See Figure \ref{fig:annotation_guidelines-fact-recall} for the full annotation guidelines and label definitions.

\begin{figure*}[t]
    \centering

    \begin{subfigure}[t]{0.48\textwidth}
        \centering
        \includegraphics[width=\linewidth, trim={0cm 0cm 0cm 4.3cm}, clip]{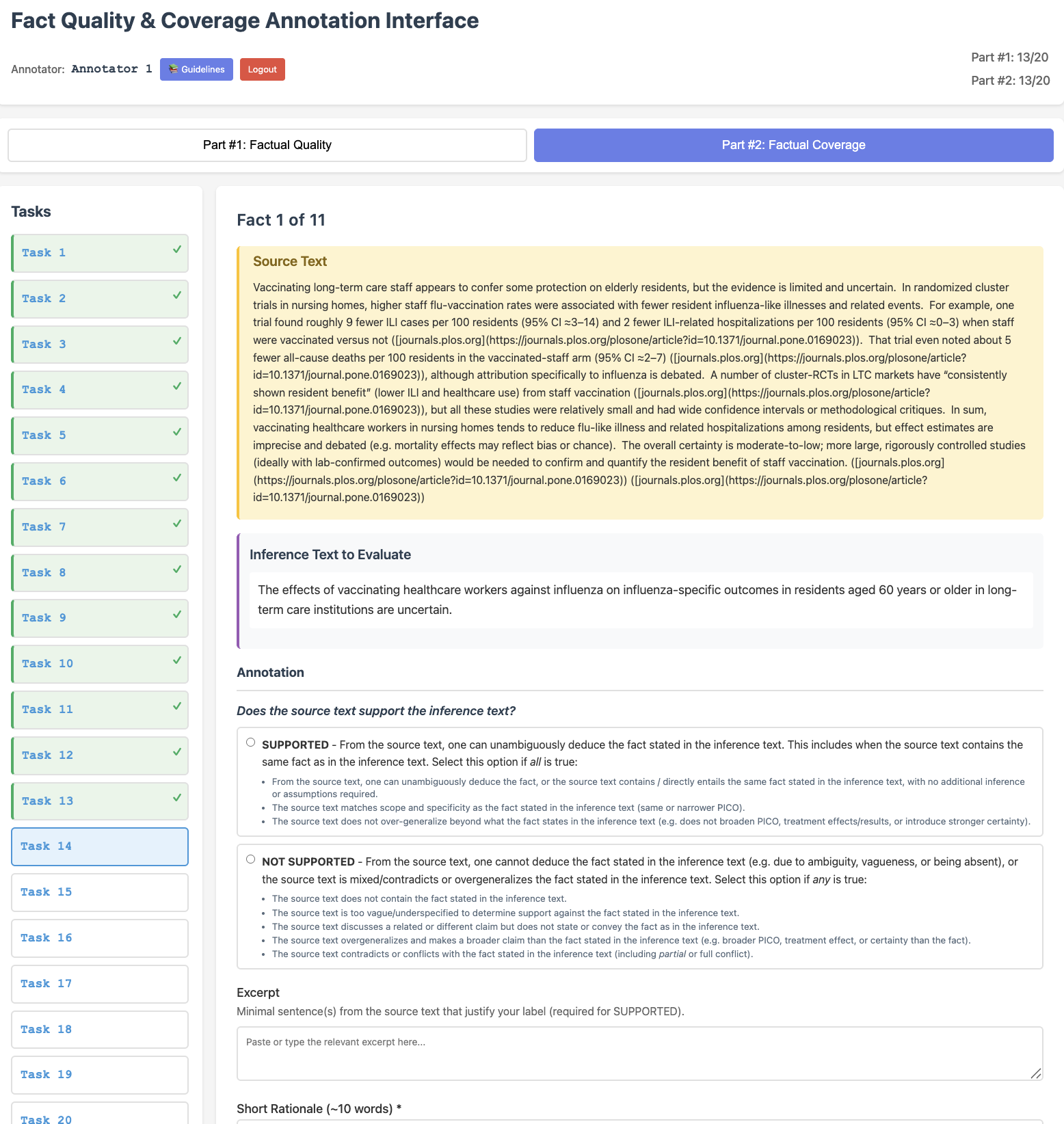}
        \caption{Annotation Interface for Factual Coverage}
    \end{subfigure}
    \hfill
    \begin{subfigure}[t]{0.48\textwidth}
        \centering
        \includegraphics[width=\linewidth, trim={0cm 13.5cm 0cm 0cm}, clip]{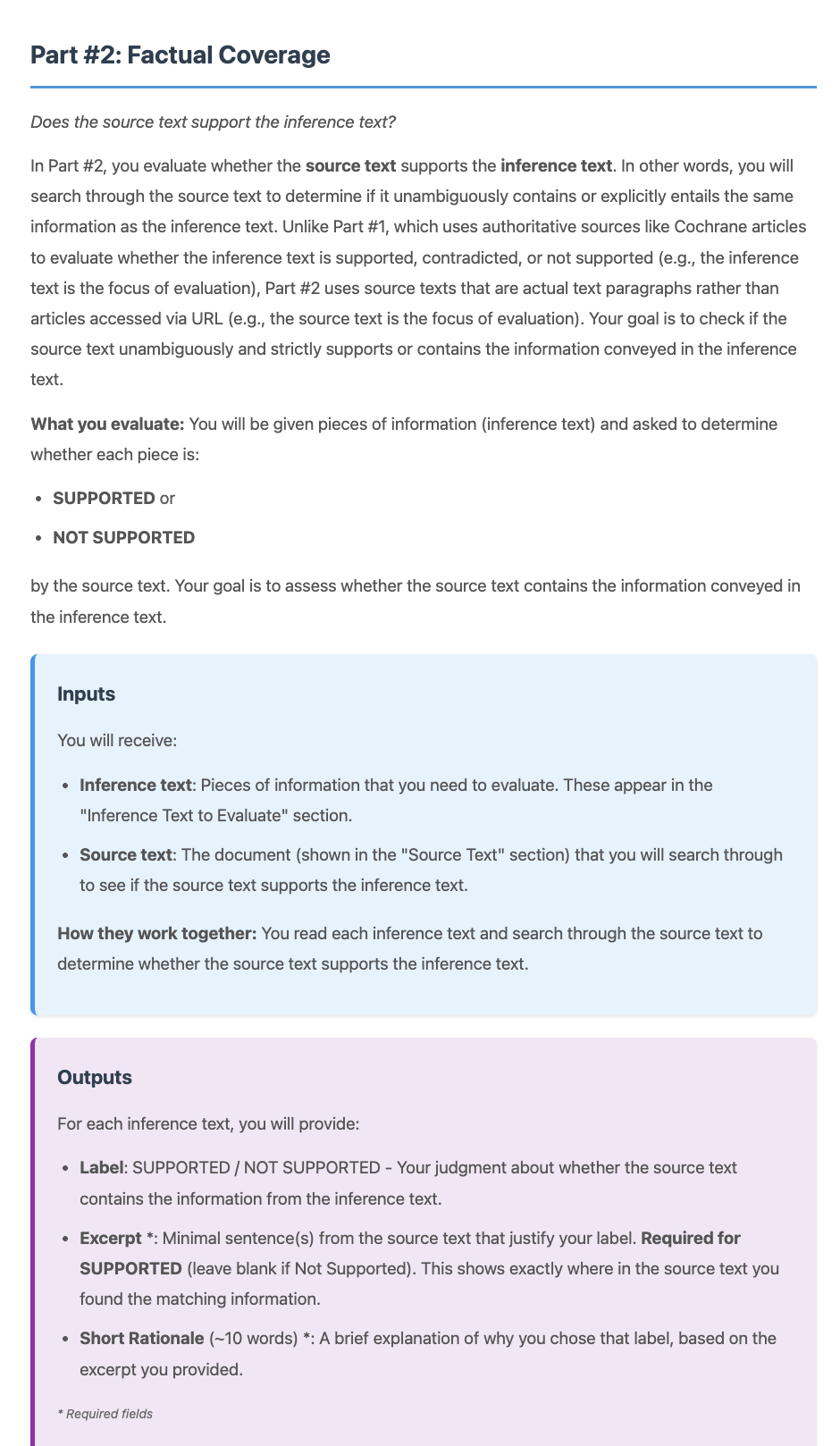}
        \caption{Task Description for Factual Coverage}
    \end{subfigure}
    \vspace{0.75em}

    \begin{subfigure}[t]{0.48\textwidth}
        \centering
        \includegraphics[width=6cm, trim={0cm 5cm 0cm 0cm}, clip]{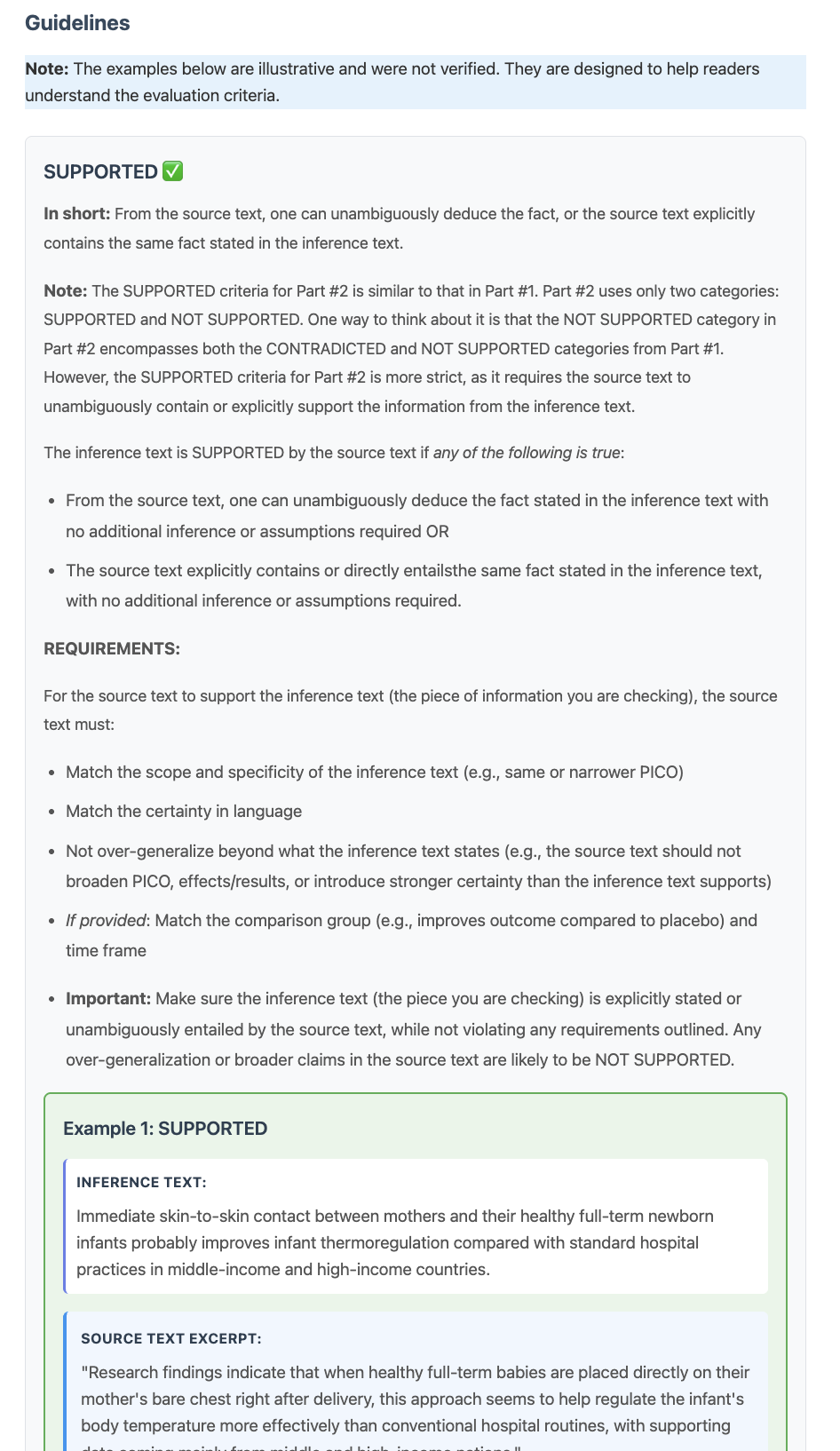}
        \caption{Definition of ``SUPPORTED'' class for Factual Coverage}
    \end{subfigure}
    \hfill
    \begin{subfigure}[t]{0.48\textwidth}
        \centering
        \includegraphics[width=\linewidth, trim={0cm 7cm 0cm 0cm}, clip]{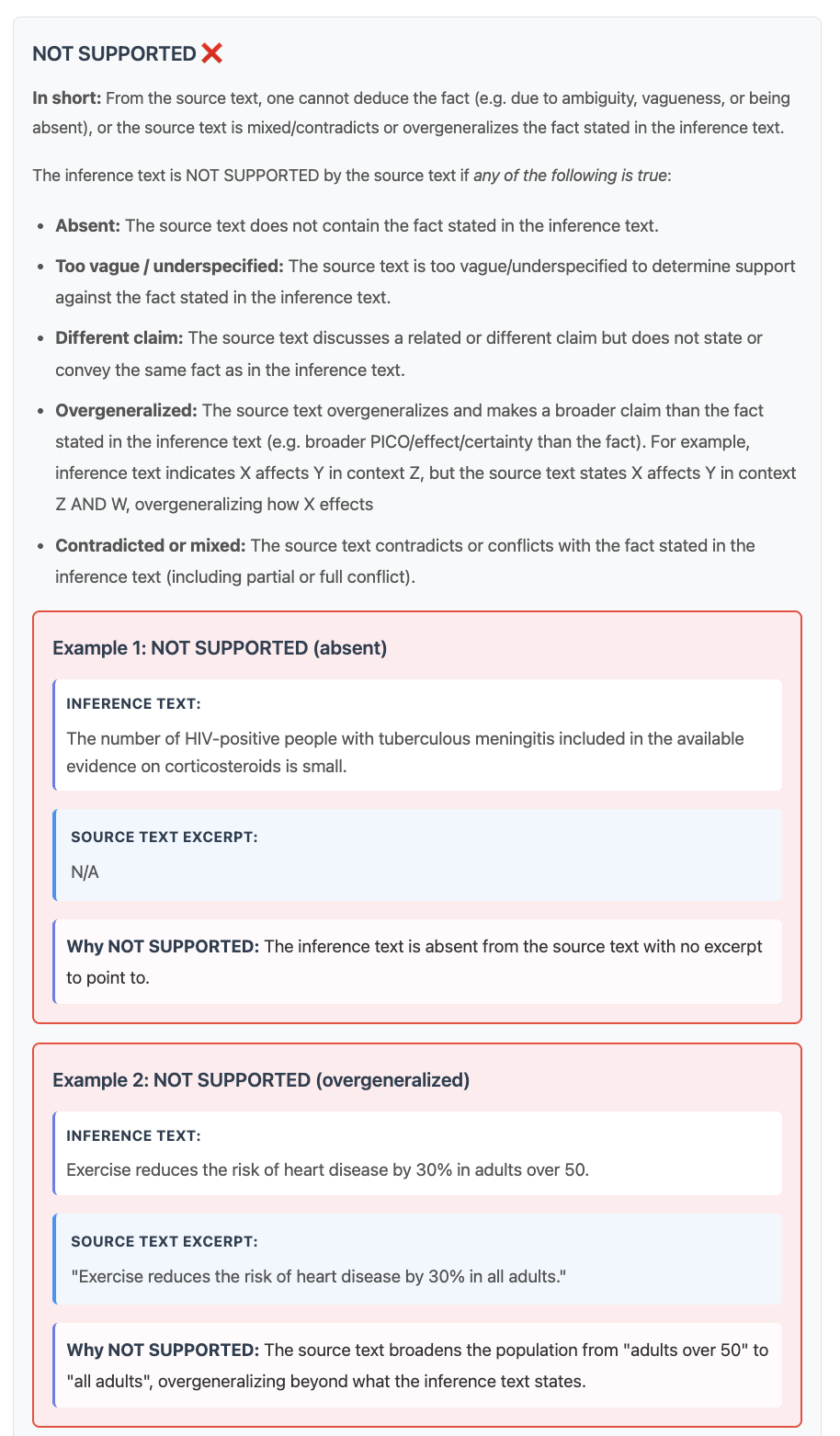}
        \caption{Definition of ``NOT SUPPORTED'' class for Factual Coverage}
    \end{subfigure}

    \caption{Annotation interface for evaluating the factual coverage of scientific conclusions. Panels show (a) annotation interface for factual coverage, (b) factual coverage description, (c) ``Supported'' class definition and examples, and (d) ``Not Supported'' class definition and examples. Panels are cropped or resized for space. Note that factual coverage corresponds to our factual recall task.}
    \label{fig:fact-eval-interface1}
\end{figure*}

\begin{figure*}[t]
    \centering

    \begin{subfigure}[t]{0.48\textwidth}
        \centering
        \includegraphics[width=\linewidth, trim={0cm 0cm 0cm 4cm}, clip]{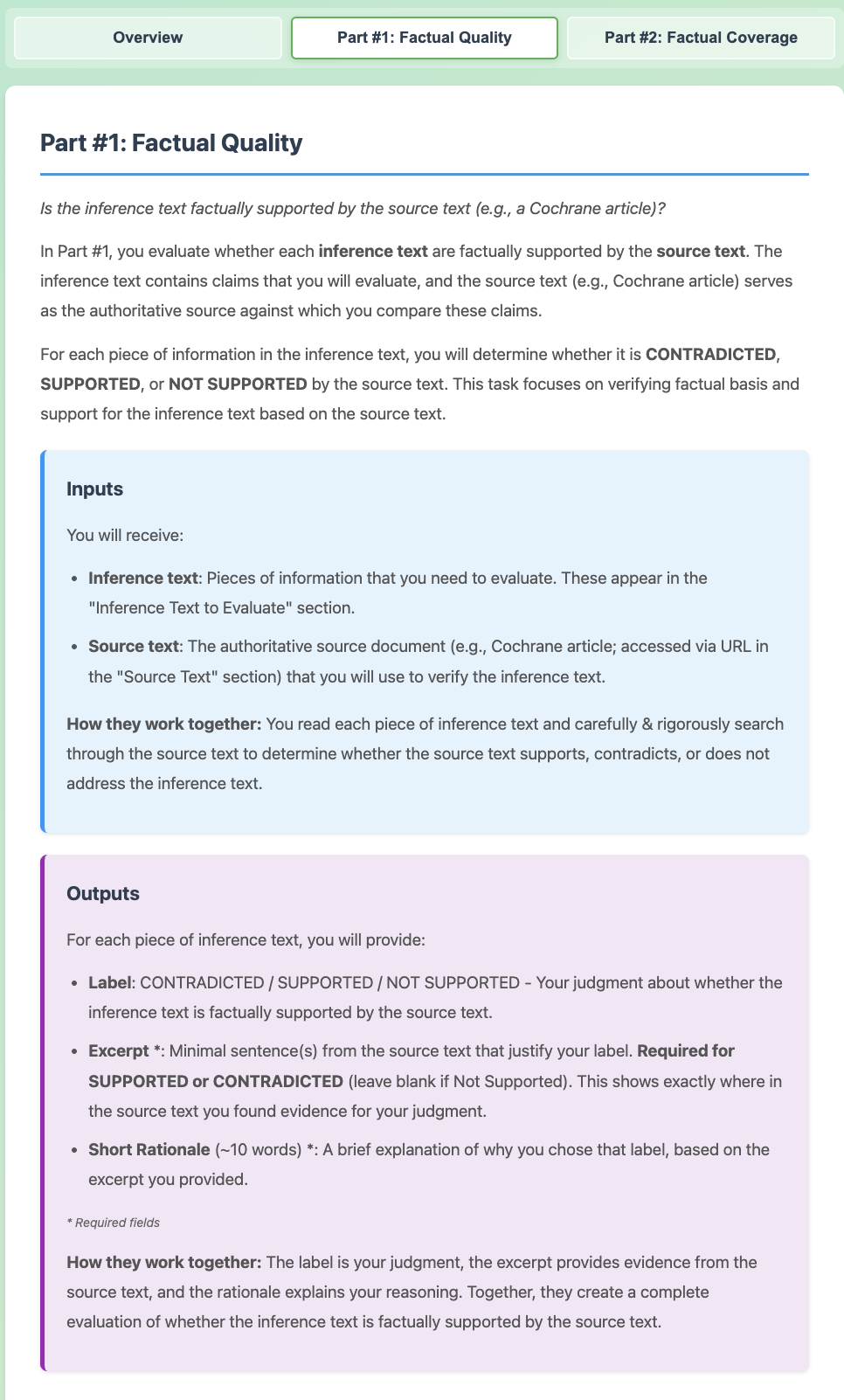}
        \caption{Task Description for Factual Correctness}
    \end{subfigure}
    \hfill
    \begin{subfigure}[t]{0.48\textwidth}
        \centering
        \includegraphics[width=\linewidth, trim={0cm 2cm 0cm 0cm}, clip]{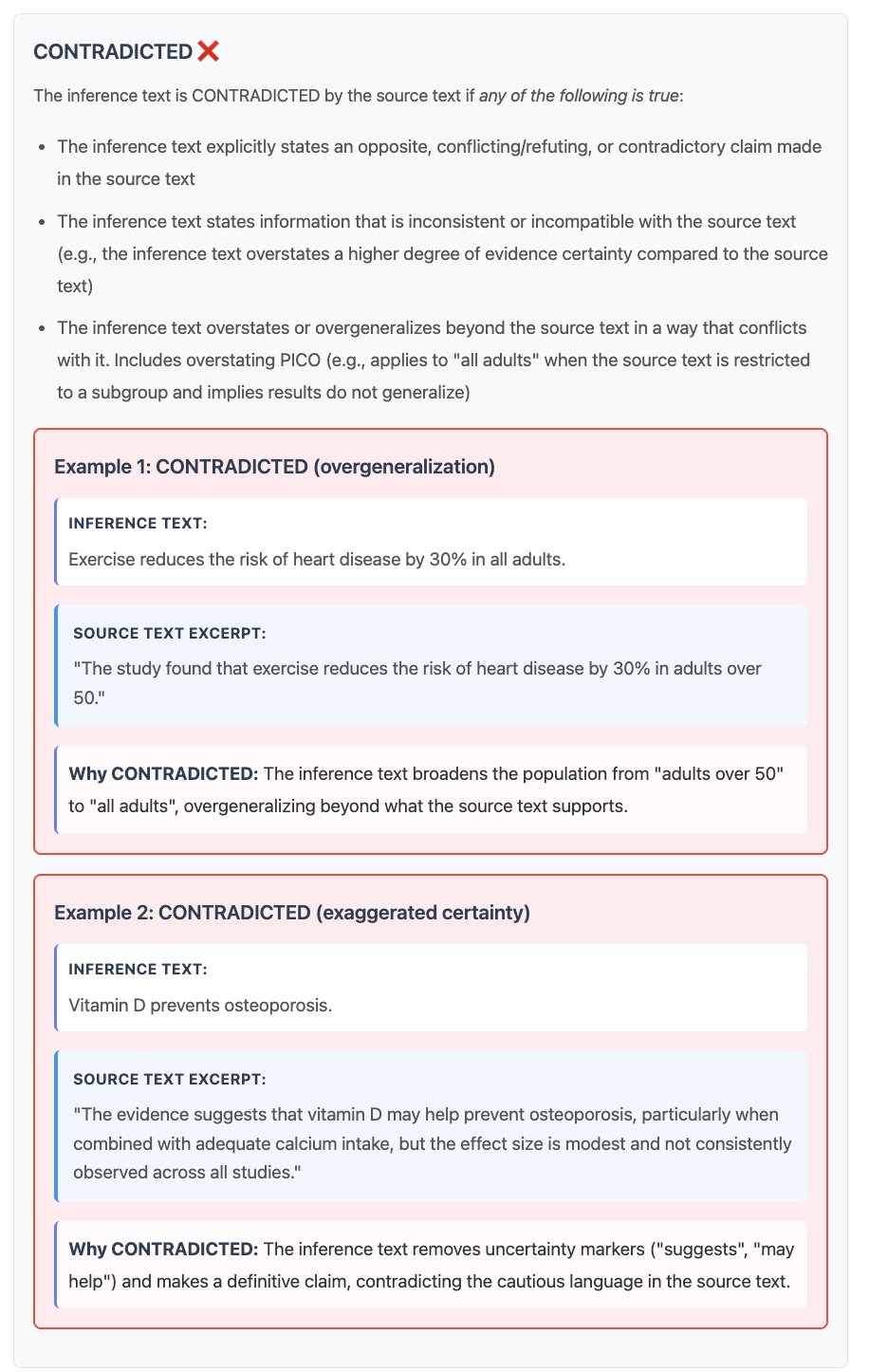}
        \caption{Definition for ``CONTRADICTED'' category}
    \end{subfigure}
    \vspace{0.75em}

    \begin{subfigure}[t]{0.48\textwidth}
        \centering
        \includegraphics[width=6cm, trim={0cm 3cm 0cm 0cm}, clip]{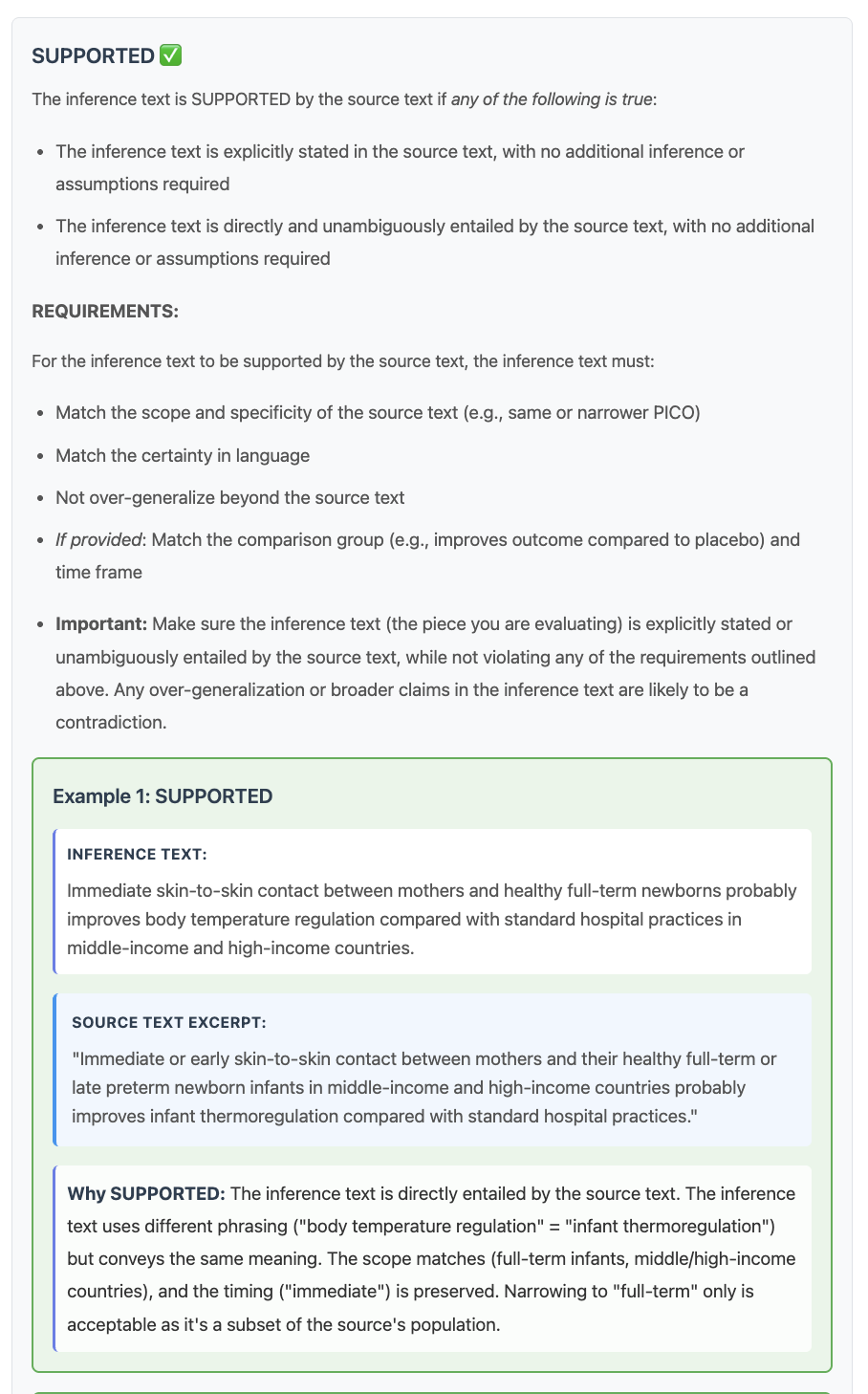}
        \caption{Definition for ``SUPPORTED'' category}
    \end{subfigure}
    \hfill
    \begin{subfigure}[t]{0.48\textwidth}
        \centering
        \includegraphics[width=\linewidth, trim={0cm 4cm 0cm 0cm}, clip]{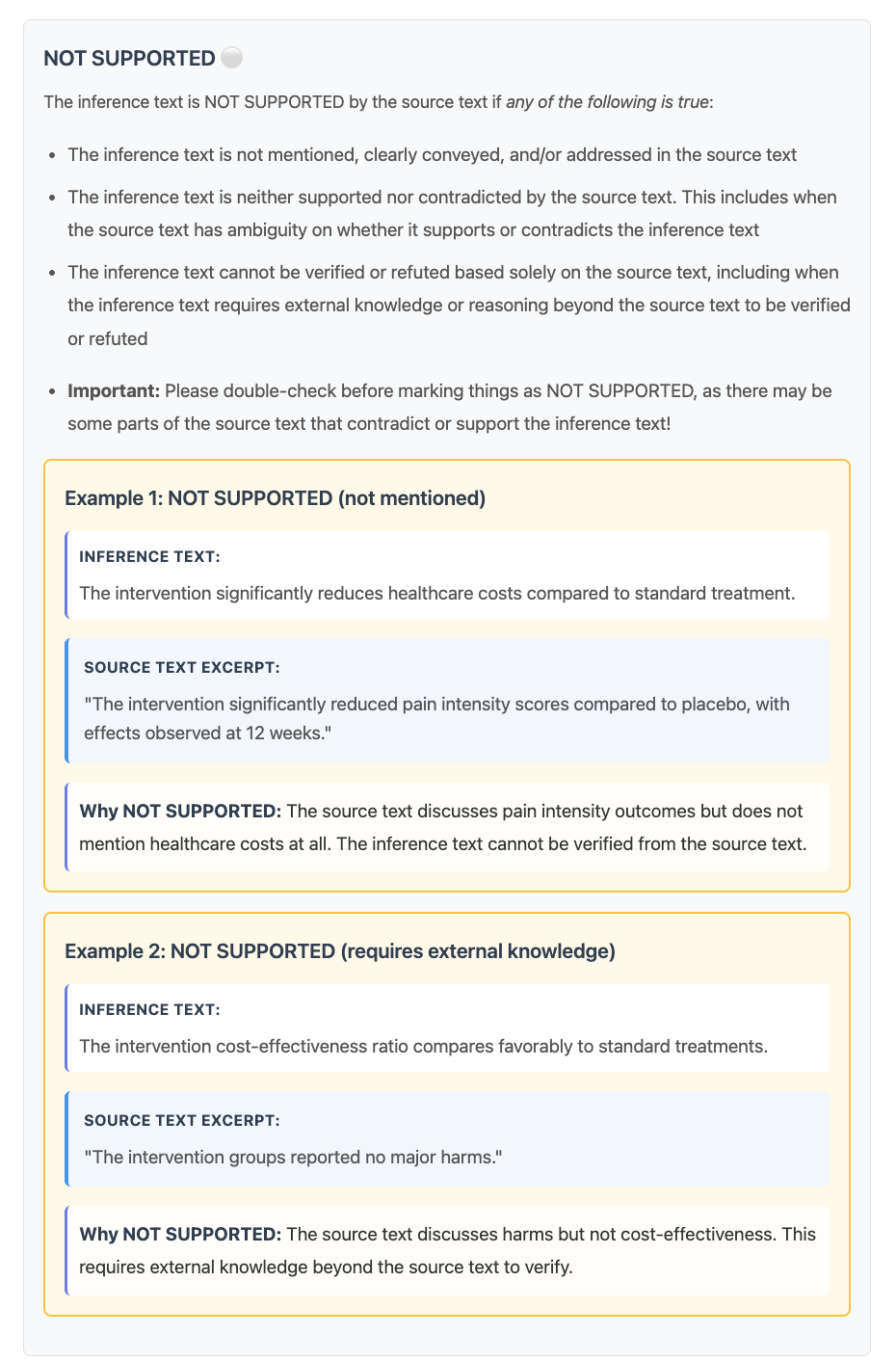}
        \caption{Definition for ``NOT SUPPORTED'' category}
    \end{subfigure}

    \caption{Annotation interface for evaluating the factual correctness of scientific conclusions. Panels show (a) factual correctness description, (b) ``CONTRADICTED'' class definition and examples, (c) ``Supported'' class definition and examples, and (d) ``Not Supported'' class definition and examples. Panels are cropped or resized for space. Note that factual correctness corresponds to our factual precision task. See Figure \ref{fig:fact-eval-interface3} for the annotation interface for the factual correctness task.}
    \label{fig:fact-eval-interface2}
\end{figure*}

\begin{figure*}[t]
    \centering
    \includegraphics[width=\linewidth, trim={0cm 0cm 0cm 5cm}, clip]{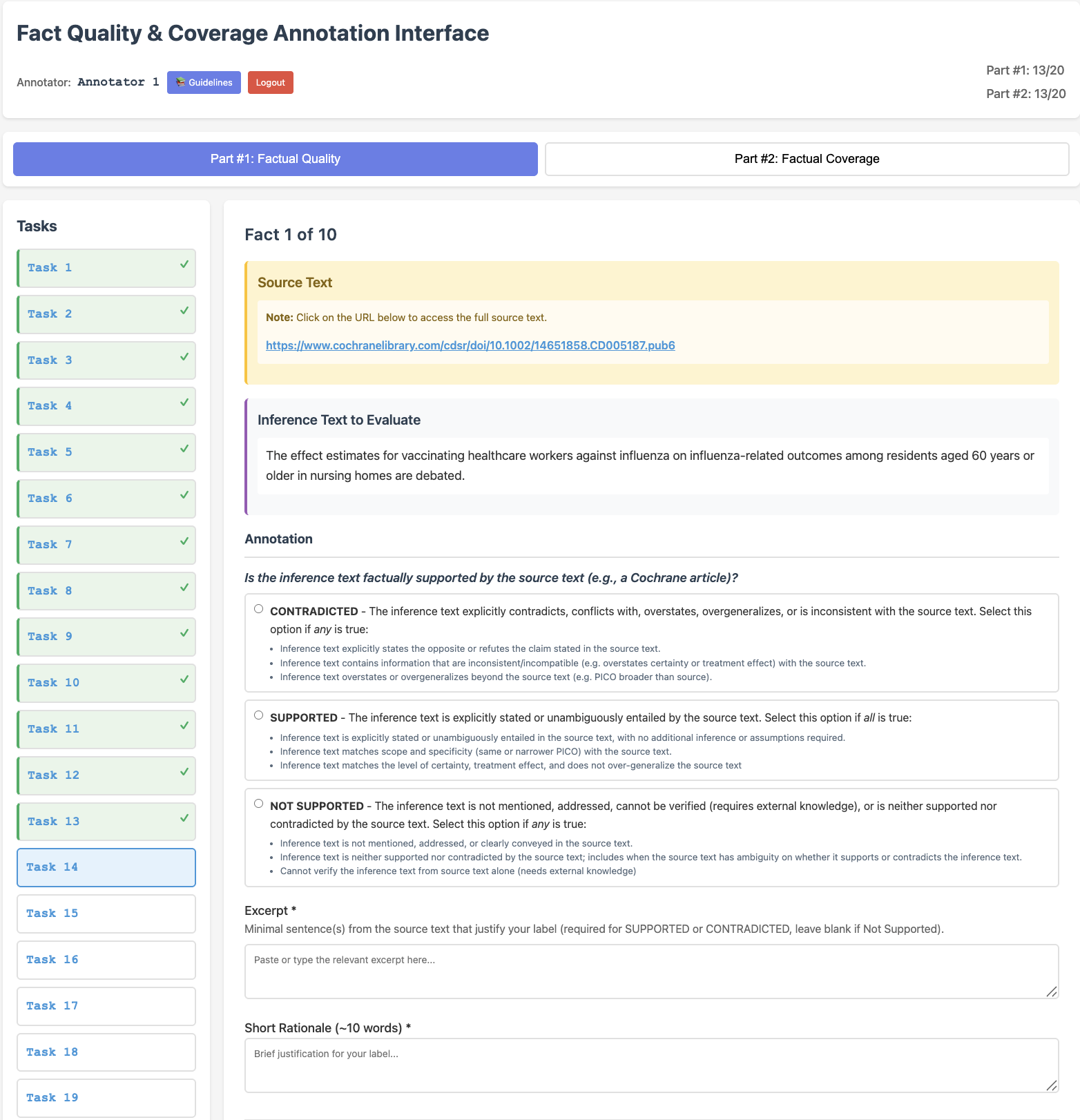}
    \caption{The annotation interface for evaluating factual correctness. }
    \label{fig:fact-eval-interface3}
\end{figure*}

See Figures~\ref{fig:fact-eval-interface1}--\ref{fig:fact-eval-interface3} for the annotation interface used in both tasks.

\subsubsection{Annotation Procedure}\label{appendix:fact-eval-annotation-procedure} Verifying each atomic fact requires reading a corresponding long-form text (e.g., \texttt{CDSR} review for factual precision, generated conclusion for factual recall). Naively sampling facts across conclusions would require annotators to repeatedly switch between different long-form texts, making the process inefficient. To improve tractability while preserving coverage, we adopt a conclusion-level subsampling strategy: for our initial annotation batch, we sample 5 generated conclusions, select up to 10 facts per conclusion for factual precision (all verified against the same \texttt{CDSR} article), and use all facts from the \textit{Authors' Conclusions} for factual recall on the same generated conclusion. 

Two medical doctors annotated the initial batch of 5 generated conclusions, corresponding to 50 facts for factual precision and 46 facts for factual recall. Following this round, annotators engaged in discussions to resolve disagreements, improve their agreement, and iteratively refine the annotation guidelines. Afterward, the two expert annotators conducted a second round of annotations on 8 additional generated conclusions and their corresponding facts for both tasks. This resulted in a total of $N=129$ facts for precision and $N=119$ for recall, comparable to or larger than sample sizes in prior work relying on expert annotations to evaluate LLM judge performance  \cite{cheng2025facts, 10.1093/jla/laae003, dammu-etal-2024-uncultured, jung-etal-2025-mythtriage, phutane2025ableistintersectionaldisabilitybias}. For both rounds, a third medical doctor independently reviewed and resolved any remaining disagreements to produce final labels. Overall, the annotation process was time-intensive, requiring approximately 6 minutes per fact for the expert annotators.

\begin{table}[t]
\centering
\small
\caption{Inter-annotator agreement between two expert annotators for factual precision ($N=129$) and recall ($N=119$) across two annotation rounds and overall. Agreement is reported using Cohen’s $\kappa$, Gwet’s AC1, and percentage agreement, all of which improve from Round 1 to Round 2 following adjudication and discussion of disagreement cases. These agreement rates are comparable to, or exceed, those reported in prior work \cite{cheng2025facts, liu-etal-2025-verifact, min-etal-2023-factscore}.}
\begin{tabular}{l l c c}
\toprule
\textbf{Round} & \textbf{Agreement Metric} & \textbf{Factual Precision} & \textbf{Factual Recall} \\
\midrule
\multirow{3}{*}{1}
& Cohen's $\kappa$ & 0.423 & 0.547 \\
& Gwet's AC1 & 0.478 & 0.585 \\
& Percentage Agreement & 0.640 & 0.783 \\
\midrule
\multirow{3}{*}{2}
& Cohen's $\kappa$ & 0.569 & 0.684 \\
& Gwet's AC1 & 0.590 & 0.759 \\
& Percentage Agreement & 0.722 & 0.863 \\
\midrule
\multirow{3}{*}{Overall}
& Cohen's $\kappa$ & 0.519 & 0.658 \\
& Gwet's AC1 & 0.545 & 0.671 \\
& Percentage Agreement & 0.691 & 0.832 \\
\bottomrule
\end{tabular}
\label{tab:iaa-fact-eval}
\end{table}

\xhdr{Agreement Results} As shown in Table~\ref{tab:iaa-fact-eval}, we observe moderate agreement for factual precision (Cohen's $\kappa$: 0.519; AC1: 0.545; Percentage Agreement: 69\%) and substantial agreement for factual recall (Cohen's $\kappa$: 0.658; AC1: 0.671; Percentage Agreement: 83.2\%), comparable to or exceeding prior work \cite{cheng2025facts, liu-etal-2025-verifact, min-etal-2023-factscore}. Importantly, expert disagreement does not necessarily indicate noise; it can reflect \textit{legitimate differences} in how experts prioritize aspects in scientific tasks based on their expertise, background, and inferred goals \cite{hwang2026deepresearchshallowevaluation}. Overall, despite the challenging and expertise-intensive nature of the tasks, these results indicate that our annotation task is well-defined and yields reliable labels.

Following discussion of disagreement cases, the agreement consistently improves from Round 1 to Round 2 across all metrics (e.g., by approximately 0.11--0.17 in Cohen's $\kappa$ and Gwet's AC1), indicating improved annotator calibrations and annotation consistency. In addition, we observe lower agreement for factual precision compared to factual recall. Factual precision is inherently more challenging: annotators must verify each fact against the full \texttt{CDSR} article and assign finer-grained labels (e.g., supported, contradicted, not supported), whereas factual recall involves checking coverage against a single conclusion paragraph with a binary decision. Despite this increased complexity, agreement for factual precision remains strong and comparable to prior work \cite{cheng2025facts, liu-etal-2025-verifact, min-etal-2023-factscore}.

Most importantly, we employ a third medical doctor to independently adjudicate disagreements between the two expert annotators, producing consensus labels. This consensus-based process is important for this expertise-intensive task and further improves the quality of the final labels, particularly in cases of disagreement. The resulting gold-standard dataset contains 19 \textsc{Contradicted}, 54 \textsc{Supported}, and 56 \textsc{Not Supported} labels for factual precision, and 48 \textsc{Supported} and 71 \textsc{Not Supported} labels for factual recall.

\subsection{Validation of LLM Judges}\label{appendix:fact-eval-llm}

We describe the features (\S\ref{appendix:llm-judge-features}), the prompt engineering process (\S\ref{appendix:llm-judge-prompt}), report performance against our expert-annotated gold-standard dataset across LLMs (\S\ref{appendix:llm-judge-result}), and conduct error analysis on the LLM judge (\S\ref{appendix:llm-judge-error-analysis}).

\subsubsection{Feature Descriptions}\label{appendix:llm-judge-features}

Our prompts provide the same input features used in the expert annotation tasks (\S\ref{appendix:fact-eval-annotation-task}).

\noindent\textbf{Factual Precision Task:} The model assesses whether each atomic fact is factually supported and non-contradictory with respect to a trusted source text.
\begin{itemize}[leftmargin=0.6cm]
    \item \textbf{Atomic Fact}: An atomic fact extracted from a model-generated conclusion, evaluated against a trusted source text.
    \item \textbf{Source Text}: The full abstract and plain-language summary sections of the \texttt{CDSR} systematic review article. These sections summarize the key methods, results, and conclusions of the review, including \textit{Objectives}, \textit{Main Results}, and \textit{Authors' Conclusions}, without introducing new information beyond the main body. As full reviews may be paywalled with copyright restrictions, we rely on these publicly available, open-access sections, such as the Abstracts and Plain-Language Summaries. Prior work has similarly adopted this approach \cite{bakker-kamps-2024-cochrane, devaraj-etal-2021-paragraph, polzak2025can, tang2023evaluating}.
\end{itemize}

\noindent\textbf{Factual Recall Task:} The model assesses whether each atomic fact is supported by the conclusion (e.g., source text).
\begin{itemize}[leftmargin=0.6cm]
    \item \textbf{Atomic Fact}: An atomic fact extracted from the reference \textit{Authors' Conclusions} of a \texttt{CDSR} review.
    \item \textbf{Source Text}: The model-generated conclusion, used to assess whether it contains or directly supports the atomic fact.
\end{itemize}

\subsubsection{Prompt Design Considerations}\label{appendix:llm-judge-prompt}

Our prompt design is guided by prompt-engineering recommendations from OpenAI \cite{openai_prompt_engineering_guide}, Anthropic \cite{anthropic_prompt_engineering_overview}, Google Gemini \cite{google_prompt_engineering_overview}, as well as prior works \cite{dammu-etal-2024-uncultured, Jung_Juneja_Mitra_2025, githubtoxicity2022, park-etal-2024-valuescope, 10.5555/3600270.3602070, yang2025escaping, zheng-etal-2024-helpful}. For each task, we design both zero-shot and few-shot prompts following these considerations. See Figure~\ref{fig:fact-precision-prompt} for the factual precision prompt and Figure~\ref{fig:fact-recall-prompt} for the factual recall prompt. Below, we outline the key prompt design features considered:

\begin{itemize}
[noitemsep,topsep=0pt,leftmargin=0.6cm]
    \item \textbf{System Roles:} While personas can improve model performance \cite{openai_prompt_engineering_guide}, their effects are often inconsistent \cite{zheng-etal-2024-helpful}.  However, \citet{zheng-etal-2024-helpful} suggest that ``gender-neutral, in-domain, and work-related roles'' yield more reliable improvements than other persona types \cite{zheng-etal-2024-helpful}. Given the evaluation-oriented and evidence-based nature of our tasks, we adopt an expert evaluator persona: ``\texttt{You are an expert evaluator with deep expertise in evidence-based medicine and clinical research.}''
    \item \textbf{Contextual Details:} Providing sufficient context improves LLM reasoning and justification \cite{anthropic_prompt_engineering_overview, google_prompt_engineering_overview, openai_prompt_engineering_guide}. Accordingly, we include detailed input descriptions, explicit output specifications, and comprehensive guidelines and decision criteria for each output label.
    \item\textbf{Temperature: }Temperature influences how models generate text \cite{openai_prompt_engineering_guide}.  Lower values (e.g., 0) produce more deterministic and consistent responses, while higher values (e.g., 1) yield more diverse outputs. For \texttt{gpt-5.4-mini}, \texttt{claude-haiku-4.5}, and \texttt{gemini-3-flash}, the default temperature is 1 \cite{anthropic_messages_api_create, google_gemini_3_flash_vertex_ai, openai_latest_gpt5_4_guide}. Prior work \cite{githubtoxicity2022, dammu-etal-2024-uncultured, park-etal-2024-valuescope, Jung_Juneja_Mitra_2025} suggests that moderate temperatures (e.g., 0.2) perform best for structured and defined evaluation tasks such as misinformation detection and factuality assessment. Although a temperature of 0 may lead to text degeneration \cite{holtzman2019curious}, we include it to assess performance under fully deterministic settings. We evaluate performance across temperatures $\{0, 0.2, 1\}$.
    \item \textbf{Zero-Shot vs. Few-Shot:} For each task, we evaluate both zero-shot and few-shot prompting. Zero-shot prompts present the task without examples, while few-shot prompts provide labeled examples to enable in-context learning without updating model weights \cite{NEURIPS2020_1457c0d6}. For few-shot prompting, we manually construct six examples. Each example includes the input features (e.g., source text and atomic fact; see \S\ref{appendix:llm-judge-features}), along with the output label, a supporting excerpt from the source text, and a detailed justification.
    \item \textbf{Reasoning:} Prompting LLMs to reason step by step and justify their decisions has been shown to improve performance across a range of tasks \cite{10.5555/3600270.3602070}, including factuality assessment and misinformation detection \cite{Jung_Juneja_Mitra_2025, Mittal_Jung_ElSherief_Mitra_De_Choudhury_2025}. Following this approach, we instruct models to reason step by step before producing an output label, a brief supporting excerpt, and a justification. Beyond standard step-by-step prompting, we also vary the reasoning level across models to assess whether additional reasoning improves performance on factual precision and recall. This is especially relevant because these labeling tasks are challenging, reasoning-intensive, and require substantial domain expertise, even for medical doctor annotators. We therefore evaluate all available reasoning settings for each model. For most models, reasoning is not constrained by a separate token budget; for \texttt{claude-haiku-4.5}, extended thinking uses a budget of 1{,}024 thinking tokens to stay within reasonable costs.
\end{itemize}

\subsubsection{Evaluation Results}\label{appendix:llm-judge-result}

Using both zero-shot and few-shot prompts across varying reasoning and temperature settings, we evaluate three LLMs---\texttt{gpt-5.4-mini}, \texttt{claude-haiku-4.5}, and \texttt{gemini-3-flash}---on the gold-standard dataset for both factual precision and recall tasks. Full results are reported in Table~\ref{tab:fact-precision-full} for factual precision and Table~\ref{tab:fact-recall-full} for factual recall. To mitigate evaluation leakage, we exclude the six few-shot examples used in the prompts from the evaluation set for both tasks.

\xhdr{\texttt{gpt-5.4-mini} achieved the strongest overall performance}
Across both tasks, \texttt{gpt-5.4-mini} outperformed the other models in the best-performing configuration (e.g., prompt, reasoning, temperature). For factual precision, its best setting achieved a macro F1 of 0.837 with accuracy 0.830; for factual recall, it achieved a macro F1 of 0.868 with accuracy 0.903. \texttt{gemini-3-flash} was the next strongest model, reaching a best macro F1 of 0.777 with accuracy 0.797 for factual precision, and 0.844 with accuracy 0.876 for factual recall. 

\xhdr{For both tasks, expert agreement with \texttt{gpt-5.4-mini} is comparable to agreement between experts}
Beyond absolute performance, these results are supported by strong agreement with our expert annotators. As shown in Table~\ref{tab:factual-agreement}, the LLM judge (\texttt{gpt-5.4-mini}, in its best configuration) agrees with each expert at rates comparable to—and in some cases higher than—the agreement between the experts themselves. Notably, for factual recall, the average Expert--LLM agreement (Cohen's $\kappa = 0.695$, Gwet's AC1 $= 0.723$) exceeds Expert--Expert agreement (Cohen's $\kappa = 0.658$, Gwet's AC1 $= 0.670$). For factual precision, average Expert--LLM agreement is comparable to Expert--Expert agreement: slightly higher in percentage agreement (0.695 vs.\ 0.691) and Gwet's AC1 (0.560 vs.\ 0.545), and slightly lower in Cohen's $\kappa$ (0.512 vs.\ 0.519). These findings suggest that the LLM judge operates at a level comparable to expert annotators on our gold-standard dataset. Overall, the strong agreement with multiple experts---alongside high task performance---validates the quality of our prompts and supports the use of frontier LLMs as effective and reliable evaluators for our tasks.

\xhdr{Higher reasoning levels did not consistently improve performance}
Increasing reasoning effort yielded limited benefit and often reduced performance. For \texttt{gpt-5.4-mini}, the best results on both tasks were obtained without reasoning: factual precision peaked at reasoning \textit{None} with temperature 0.2, and factual recall peaked at reasoning \textit{None} with temperature 1. Under matched model, prompt, and temperature settings, increasing reasoning can substantially reduce performance. For example, for \texttt{gpt-5.4-mini}, moving from no reasoning to higher reasoning under the same configuration reduces factual precision from 0.837 to 0.706 and factual recall from 0.868 to 0.704. This pattern is consistent with prior work showing that increasing reasoning levels do not always improve performance and can even degrade it in tasks such as mathematical reasoning and text classification \cite{ballon2025relationshipreasoningperformancelarge, phutane2025ableistintersectionaldisabilitybias}.

\xhdr{Non-zero temperatures often performed better than temperature 0}
The best-performing configurations for \texttt{gpt-5.4-mini} occurred at non-zero temperatures (0.2 for factual precision and 1 for factual recall). A similar trend holds for \texttt{gemini-3-flash} and \texttt{claude-haiku-4.5}, whose strongest performance also appears at temperatures 0.2 or 1 rather than 0. This suggests that a small degree of sampling variability can be beneficial, even for evaluator-style prompting. Our result is consistent with prior work showing that deterministic decoding (e.g., temperature 0) can lead to text degeneration \cite{holtzman2019curious}.

\xhdr{Few-shot prompting had mixed effects, and did not consistently improve performance}
Few-shot prompting did not uniformly outperform zero-shot prompting under matched model, reasoning, and temperature settings. For factual precision, the change in macro F1 from zero-shot to few-shot ranged from $-0.034$ to $+0.395$, indicating that few-shot examples could either slightly hurt performance or substantially improve it depending on the configuration. In contrast, factual recall showed even more mixed results: the corresponding change ranged from $-0.182$ to $+0.075$, with smaller gains and larger degradations overall. This instability was especially consistent for \texttt{claude-haiku-4.5}, which consistently degraded under few-shot prompting on the factual recall task (range: $-0.097$ to $-0.019$). Taken together, these results suggest that few-shot prompting is more sensitive to task, model choice, and hyperparameter selection (e.g., temperature, reasoning level) for LLM-based evaluation than is often assumed.

\subsubsection{Error Analysis}\label{appendix:llm-judge-error-analysis}

While \texttt{gpt-5.4-mini} as an LLM judge offers strong performance and expert-level agreement with medical doctors, we conduct detailed error analysis on both factual precision and recall tasks to identify common failure modes and highlight areas for improving LLM judge accuracy and reliability. In both tasks, the two common errors are:

\xhdr{Error \#1: Missing contextual equivalence and clinical synonymy} In both tasks, the LLM judge occasionally fails to recognize clinically equivalent contexts and terminology. As shown in Figure \ref{fig:error-example-1}, while the source text provides strong evidence that antibiotics do not improve symptom persistence in closely related conditions (e.g., common cold, acute purulent rhinitis), the model treats differences in wording---such as ``acute rhinosinusitis'' and ``longer-term symptom duration''---as substantive mismatches and thus incorrectly labels the fact as \texttt{Not Supported}. In contrast, expert annotators correctly interpret these as medically aligned concepts, grounding their judgment in a broader clinical context rather than strict lexical overlap, and label the atomic fact as \textsc{Supported}. This suggests that the LLM judge can be sensitive to surface-level discrepancies (e.g., specific clinical jargon) and lacks robust handling of medical synonymy and outcome equivalence, leading to false negatives in factual precision evaluation.

\begin{figure*}[h!]
\centering
\begin{small}
\begin{tcolorbox}[colback=gray!5, colframe=red, width=\textwidth]
\textsc{Atomic Fact:} \texttt{\textcolor{purple}{For children with acute upper respiratory tract infections or acute rhinosinusitis diagnosed within the first 10 days, antibiotics do not meaningfully reduce longer-term symptom duration compared with placebo.}}\\
\textsc{Source Text:} {\footnotesize\url{https://www.cochranelibrary.com/cdsr/doi/10.1002/14651858.CD000247.pub4}}\\

\textsc{Both Expert Annotators:} 
\begin{enumerate}[leftmargin=0.6cm]
\item Label:  \textsc{Supported}
\item Excerpt from Source Text:  ``Participants receiving antibiotics for the common cold did no better in terms of lack of cure or persistence of symptoms than those on placebo (risk ratio (RR) 0.83, 95\% confidence interval (CI) 0.60 to 1.14...''
\item Justification: ``Results applicable to children and adults.''
\end{enumerate}
\vspace{2mm}
\textsc{LLM Judge} (\texttt{gpt-5.4-mini}): 
\begin{enumerate}[leftmargin=0.6cm]
\item Label:  \textsc{Not Supported}
\item Excerpt from Source Text:  ``Participants receiving antibiotics for the common cold did no better in terms of lack of cure or persistence of symptoms than those on placebo (risk ratio (RR) 0.83, 95\% confidence interval (CI) 0.60 to 1.14...'', ``There is no evidence of benefit from antibiotics for the common cold or for persisting acute purulent rhinitis in children or adults.''
\item Justification: ``The source text does support a null effect for antibiotics in common cold and acute purulent rhinitis/acute upper respiratory infection generally, including pediatric participants... However, the atomic fact is more specific than the source in two important ways: it narrows the condition to ``acute rhinosinusitis'' and claims effects on ``longer-term symptom duration,'' whereas the source discusses common cold and acute purulent rhinitis with outcomes of lack of cure/persistence of symptoms, not longer-term symptom duration, and it does not clearly identify rhinosinusitis as the studied condition. Because the comparator, population, and especially the outcome wording do not directly match, the claim cannot be directly verified from the source text...''
\end{enumerate}
\end{tcolorbox}
\caption{Error example from a factual precision task, where the LLM judge fails to recognize contextual equivalence. Although the source text supports a lack of antibiotic benefit for closely related conditions (e.g., common cold, acute purulent rhinitis) and symptom persistence outcomes, the LLM judge incorrectly rejects the fact due to mismatches in terminology (``acute rhinosinusitis'') and outcome phrasing (``longer-term symptom duration''), overlooking clinically equivalent concepts.} 
\label{fig:error-example-1}
\end{small}
\end{figure*}

\xhdr{Error \#2: Different interpretation of the fact and the source text}

Another common source of error arises from differing interpretations of the atomic fact and the source text. Because both the fact and the source text can be interpreted in multiple ways, it can be debatable whether the source text sufficiently supports the fact, making such judgments subjective and challenging \cite{min-etal-2023-factscore}. As shown in Figure~\ref{fig:error-example-2}, the source text indicates that higher intensity or doses of rehabilitation improve outcomes in specific phases (e.g., subacute, chronic), but does not explicitly frame this as an ``adjunct'' intervention or generalize to all people after stroke. 

Expert annotator A interprets the fact in light of its hedged phrasing (``may provide'') and labels it as \textsc{Supported}, justifying that the source text sufficiently supports this non-definitive fact. In contrast, the other expert and the LLM judge adopt a stricter interpretation, requiring an explicit statement of an ``adjunct'' intervention in the source text, and therefore label it as \textsc{Not Supported}. Thus, this highlights how differing interpretations of the atomic fact and source text make it debatable whether the source text sufficiently supports the fact. In such cases, the LLM judge seems to default to stricter, literal criteria for \textsc{Supported}, leading to errors when the evidence in the source text is implicit or when the atomic fact is hedged. Consistent with prior work \cite{hwang2026deepresearchshallowevaluation}, such variation in interpretation can lead to disagreement---even among experts---but also systematically contributes to failure modes in LLM judge decisions.

\begin{figure*}[h!]
\centering
\begin{small}
\begin{tcolorbox}[colback=gray!5, colframe=red, width=\textwidth]
\textsc{Atomic Fact:} \texttt{\textcolor{purple}{Additional physical rehabilitation delivered as an adjunct to usual physical rehabilitation after stroke may provide added benefits for people after stroke.}}\\
\textsc{Source Text:} \texttt{\footnotesize \textcolor{darkgray}{Physical rehabilitation effectively improves motor function, balance, and activities of daily living (ADL) across all stages of recovery, but the timing and intensity of interventions are critical modulators of success. Evidence indicates that while "early" rehabilitation (commencing >24 hours post-stroke) harnesses neuroplasticity for optimal recovery, "very early" high-intensity mobilization (within 24 hours) can be detrimental and reduce favorable functional outcomes. In the subacute and chronic phases, higher doses and intensities of therapy generally yield superior motor outcomes, with some guidelines suggesting 30–60 minutes daily...}}\\

\textsc{Expert Annotator A:} 
\begin{enumerate}[leftmargin=0.6cm]
\item Label:  \textsc{Supported}
\item Excerpt from Source Text:  ``In the subacute and chronic phases, higher doses and intensities of therapy generally yield superior motor outcome''
\item Justification: ``Source states that there is benefit for "subacute and chronic phases", not for all "people after stroke." But since the atomic fact says "may provide" (not exaggerates certainty) the source text does indeed support the atomic fact.''
\end{enumerate}
\vspace{2mm}
\textsc{Expert Annotator B:} 
\begin{enumerate}[leftmargin=0.6cm]
\item Label:  \textsc{Not Supported}
\item Excerpt from Source Text:  N/A
\item Justification: ``No description in the source text.''
\end{enumerate}
\vspace{2mm}
\textsc{LLM Judge} (\texttt{gpt-5.4-mini}): 
\begin{enumerate}[leftmargin=0.6cm]
\item Label:  \textsc{Not Supported}
\item Excerpt from Source Text:  ``Physical rehabilitation effectively improves motor function, balance, and activities of daily living (ADL) across all stages of recovery, but the timing and intensity of interventions are critical modulators of success''
\item Justification: ``The source text supports that physical rehabilitation is beneficial and that some delivery models are non-inferior to conventional care, but it does not state that *additional* rehabilitation given as an adjunct to usual rehabilitation after stroke provides added benefits. The required adjunct/comparator context is missing, so the atomic fact is not directly entailed.''
\end{enumerate}
\end{tcolorbox}
\caption{Error example from a factual recall task, illustrating differing interpretations of the fact and source text. The atomic fact claims that additional rehabilitation may provide added benefits when used alongside usual physical rehabilitation, while the source text discusses the general benefits of rehabilitation intensity and delivery without explicitly framing an adjunct intervention. One expert annotator interprets the source text more broadly and labels the fact as \textsc{Supported}, whereas the other expert and the LLM judge do not identify a sufficient description of the atomic fact in the source text and label it as \textsc{Not Supported}.} 
\label{fig:error-example-2}
\end{small}
\end{figure*}

\clearpage

\begin{figure*}[t]
\centering
\begin{small}
\begin{tcolorbox}[colback=gray!5, colframe=black, width=\textwidth, title=\textbf{Annotation Guideline for Evaluating Factual Precision}, fonttitle=\bfseries]

\textbf{Task Overview.} \textit{Is the inference text factually supported by the source text (e.g., Cochrane article)?}\vspace{0.5em}

You evaluate whether each \textbf{inference text} is factually supported by the \textbf{source text}. The inference text contains claims that you will evaluate, and the source text (e.g., Cochrane article) serves as the authoritative source against which you compare these claims.\vspace{0.5em}

\vspace{0.5em}
\textbf{Inputs.}
You will receive:
\begin{itemize}[leftmargin=0.6cm]
    \item \textbf{Inference text}: Pieces of information that you need to evaluate. These appear in the ``Inference Text to Evaluate'' section.
    \item \textbf{Source text}: The authoritative source document (e.g., Cochrane article; accessed via URL in the ``Source Text'' section) that you will use to verify the inference text.
\end{itemize}
\textbf{How they work together:} You read each piece of inference text and carefully \& rigorously search through the source text to determine whether the source text supports, contradicts, or does not address the inference text.

\vspace{0.5em}
\textbf{Outputs.}
For each inference text, you will provide:
\begin{itemize}[leftmargin=0.6cm]
    \item \textbf{Label}: \textsc{Contradicted} / \textsc{Supported} / \textsc{Not Supported} --- Your judgment about whether the inference text is factually supported by the source text.
    \item \textbf{Excerpts}: Minimal sentence(s) from the source text that justify your label.
    \item \textbf{Short Rationale} ($\sim$10 words): A brief explanation of why you chose that label.
\end{itemize}

\textbf{Label Definitions.}
\begin{enumerate}[leftmargin=0.6cm]
    \item \textsc{Contradicted.} Choose if \textit{any of the following is true:}
    \vspace{-0.2em}
    \begin{itemize}[leftmargin=0.5cm]
    \item The inference text explicitly states an opposite, conflicting/refuting, or contradictory claim made in the source text.
    \item The inference text states information that is inconsistent or incompatible with the source text.
    \item The inference text overstates or overgeneralizes beyond the source text.
    \end{itemize}

    \item \textsc{Supported.} Choose if \textit{all of the following is true:}
    \vspace{-0.2em}
    \begin{itemize}[leftmargin=0.5cm]
    \item The inference text is explicitly stated or unambiguously entailed by the source text, with no additional inference or assumptions required.
    \item The inference text matches the scope, certainty in language, specificity, and comparison group/time-frame as the source text
    \end{itemize}

    \item \textsc{Not Supported.} Choose if \textit{any of the following is true:}
    \vspace{-0.2em}
    \begin{itemize}[leftmargin=0.5cm]
    \item The inference text is not mentioned, clearly conveyed, and/or addressed in the source text
    \item The inference text is neither supported nor contradicted by the source text. This includes when the source text has ambiguity on whether it supports or contradicts the inference text
    \item The inference text cannot be verified or refuted based solely on the source text, including when the inference text requires external knowledge or reasoning beyond the source text to be verified or refuted
    \end{itemize}

\end{enumerate}
\end{tcolorbox}
\caption{Annotation Guideline for evaluating factual precision in generated conclusions.}
\label{fig:annotation_guidelines-fact-precision}
\end{small}
\end{figure*}

\clearpage

\begin{figure*}[t]
\centering
\begin{small}
\begin{tcolorbox}[colback=gray!5, colframe=black, width=\textwidth, title=\textbf{Annotation Guideline for Evaluating Factual Recall}, fonttitle=\bfseries]

\textbf{Task Overview.} \textit{Does the source text support the inference text?}\vspace{0.5em}

You evaluate whether the \textbf{source text} supports the \textbf{inference text}. In other words, you will search through the source text to determine if it unambiguously contains or explicitly entails the same information as the inference text.

\vspace{0.5em}
\textbf{Inputs.}
You will receive:
\begin{itemize}[leftmargin=0.6cm]
    \item \textbf{Inference text}: Pieces of information that you need to evaluate. These appear in the ``Inference Text to Evaluate'' section.
    \item \textbf{Source text}: The document (shown in the ``Source Text'' section) that you will search through to see if the source text supports the inference text.
\end{itemize}
\textbf{How they work together:} You read each inference text and search through the source text to determine whether the source text supports the inference text.

\vspace{0.5em}
\textbf{Outputs.}
For each inference text, you will provide:
\begin{itemize}[leftmargin=0.6cm]
    \item \textbf{Label}: \textsc{Supported} / \textsc{Not Supported} --- Your judgment about whether the inference text is factually supported by the source text.
    \item \textbf{Excerpts}: Minimal sentence(s) from the source text that justify your label.
    \item \textbf{Short Rationale} ($\sim$10 words): A brief explanation of why you chose that label.
\end{itemize}

\textbf{Label Definitions.}
\begin{enumerate}[leftmargin=0.6cm]
    \item \textsc{Supported.} Choose if \textit{all of the following is true:}
    \vspace{-0.2em}
    \begin{itemize}[leftmargin=0.5cm]
    \item The inference text is explicitly stated or unambiguously entailed by the source text, with no additional inference or assumptions required.
    \item The inference text matches the scope, certainty in language, specificity, and comparison group/time-frame as the source text
    \end{itemize}

    \item \textsc{Not Supported.} Choose if \textit{any of the following is true:}
    \vspace{-0.2em}
    \begin{itemize}[leftmargin=0.5cm]
    \item \textit{Absent}: The source text does not contain the fact stated in the inference text.
    \item \textit{Too vague / underspecified:} The source text is too vague/underspecified to determine support against the fact stated in the inference text.
    \item \textit{Different claim:} The source text discusses a related or different claim but does not state or convey the same fact as in the inference text.
    \item \textit{Overgeneralized:} The source text overgeneralizes and makes a broader claim than the fact stated in the inference text (e.g. broader PICO/effect/certainty than the fact). For example, inference text indicates X affects Y in context Z, but the source text states X affects Y in context Z AND W, overgeneralizing how X effects
    \item \textit{Contradicted or mixed:} The source text contradicts or conflicts with the fact stated in the inference text (including partial or full conflict).
    \end{itemize}

\end{enumerate}
\end{tcolorbox}
\caption{Annotation Guideline for evaluating factual recall in generated conclusions.}
\label{fig:annotation_guidelines-fact-recall}
\end{small}
\end{figure*}

\clearpage

\clearpage

\clearpage

\clearpage

\clearpage

\begin{figure}[t]
\begin{small}
\centering
\begin{tcolorbox}[
    colback=gray!5,
    colframe=black,
    width=\linewidth,
    arc=2mm,
    boxrule=0.5pt,
    title=\textbf{LLM Judge Prompt for Factual Precision Evaluation}
]
\textbf{System Prompt:} \textcolor{teal}{You are an expert evaluator with deep expertise in evidence-based medicine and clinical research.}\\

\textbf{Instruction:} Please carefully evaluate whether the provided ATOMIC FACT is factually SUPPORTED, CONTRADICTED, or NOT SUPPORTED by the SOURCE TEXT. Follow the evaluation criteria, tips, and examples below carefully and think step-by-step before answering.\\

\texttt{\#\# Inputs}\\
- SOURCE TEXT: An authoritative source document (texts from source, web-retrieved document) that you will search through to verify the ATOMIC FACT. \\
- ATOMIC FACT: A piece of information that you need to evaluate against the SOURCE TEXT\\

\texttt{\#\# Outputs}\\
Output should be in JSON format in the following fields:\\
- LABEL (string): CONTRADICTED / SUPPORTED / NOT SUPPORTED \\
- EXCERPTS (list of strings): Minimal sentence(s) from the SOURCE TEXT that justify your label. \\
- JUSTIFICATION (string): A detailed justification regarding why you chose the label \\

Below, we define the decision criteria for the labels...\\

\texttt{\#\#\# CONTRADICTED} \\
The ATOMIC FACT is CONTRADICTED by the SOURCE TEXT if **ANY** of the following is true:\\
- The ATOMIC FACT states opposite, conflicting, or incompatible info with the SOURCE TEXT.\\
- The ATOMIC FACT makes a directional claim, but the SOURCE TEXT reports otherwise\\ 
...\\

\texttt{\#\#\# SUPPORTED} \\
The ATOMIC FACT is SUPPORTED only if it is not CONTRADICTED \& ALL hold:\\
- From the SOURCE TEXT, one can deduce the ATOMIC FACT with no additional inference or assumptions required OR the SOURCE TEXT explicitly contains or directly entails the ATOMIC FACT.\\
- SOURCE TEXT matches or has a broader scope, PICO, or specificity than the ATOMIC FACT. \\
...\\

\texttt{\#\#\# NOT SUPPORTED} \\
The ATOMIC FACT is NOT SUPPORTED if ANY of the following is true:\\
- Absent: The SOURCE TEXT does not contain the ATOMIC FACT\\
- The SOURCE TEXT is too vague/underspecified...\\
- Neither SUPPORTED nor CONTRADICTED...\\
...\\

Below, we present six examples of the evaluation tasks...\\
\texttt{\#\#\# EXAMPLE 1:}\\
- SOURCE TEXT: ...\\
- ATOMIC FACT: ...\\
OUTPUT: \\
- LABEL: ...\\
- EXCERPT: ... \\
- JUSTIFICATION: ... 
...\\

Now, based on what you learned, evaluate whether the input ATOMIC FACT is factually SUPPORTED, CONTRADICTED, or NOT SUPPORTED... Carefully thinking step-by-step about your answer...\\ 

\texttt{\#\# Task}\\
SOURCE TEXT: \\
\{ground\_truth\_text\}\\

ATOMIC FACT: \\
\{llm\_fact\}
\end{tcolorbox}

\caption{Few-shot prompt used by LLM judge for factual precision. We use \texttt{gpt-5.4-mini} (no reasoning, temperature 0.2). The prompt was shortened to fit within the page. Full prompt available in supplementary material / code.}
\label{fig:fact-precision-prompt}
\end{small}
\end{figure}

\clearpage

\begin{figure}[t]
\begin{small}
\centering
\begin{tcolorbox}[
    colback=gray!5,
    colframe=black,
    width=\linewidth,
    arc=2mm,
    boxrule=0.5pt,
    title=\textbf{LLM Judge Prompt for Factual Recall Evaluation}
]
\textbf{System Prompt:} \textcolor{teal}{You are an expert evaluator with deep expertise in evidence-based medicine and clinical research.}\\

\textbf{Instruction:} Please carefully evaluate whether the provided ATOMIC FACT is present in, or directly supported by, the SOURCE TEXT. Read the evaluation criteria, tips, and examples below carefully to think through your answer and make your judgement.\\

\texttt{\#\# Inputs}\\
- SOURCE TEXT: The paragraph document of information that you will search through to see if the ATOMIC FACT is present in, or directly supported by this SOURCE TEXT\\
- ATOMIC FACT: A piece of information that you need to evaluate against the SOURCE TEXT\\

\texttt{\#\# Outputs}\\
Output should be in JSON format in the following fields:\\
- LABEL (string): \textsc{Supported} / \textsc{NOT SUPPORTED} - Your judgment about whether the ATOMIC FACT present in, or directly supported by, the SOURCE TEXT.\\
- EXCERPTS (list of strings): Minimal sentence(s) from the SOURCE TEXT that justify your label. \\
- JUSTIFICATION (string): A brief explanation of why you chose the label, based on the excerpt.\\

Below, we define the decision criteria for choosing the labels SUPPORTED or NOT SUPPORTED.\\

\texttt{\#\#\# SUPPORTED} \\
The ATOMIC FACT is SUPPORTED by the SOURCE TEXT if ALL of the following is true:\\
- From the SOURCE TEXT, one can deduce the ATOMIC FACT with no additional inference or assumptions required OR the SOURCE TEXT explicitly contains or directly entails the ATOMIC FACT.\\
- SOURCE TEXT matches or has a broader scope, PICO, or specificity than the ATOMIC FACT. \\
- Both SOURCE TEXT and ATOMIC FACT match the certainty in language.\\
- If stated / contained in the ATOMIC FACT: Match the comparison group (e.g., improves outcome compared to placebo), or time frame (e.g., within 24 hours) of an intervention.\\

\texttt{\#\#\# NOT SUPPORTED} \\
The ATOMIC FACT is NOT SUPPORTED by the SOURCE TEXT if ANY of the following is true:\\
- Absent: The SOURCE TEXT does not contain the ATOMIC FACT\\
- Too vague or underspecified: The SOURCE TEXT is too vague/underspecified to determine support.\\
- Different claim: The SOURCE TEXT discusses a related claim but does not convey the FACT.\\
- Source text does not fully entail the ATOMIC FACT.\\
- Contradicted or mixed: The SOURCE TEXT contradicts or conflicts with the ATOMIC FACT (including partial or full conflict).\\ 
- If stated / contained in the ATOMIC FACT: Mismatch or missing comparison group, treatment effect / outcome and their strengths (e.g., significant effect) or time frame in the SOURCE TEXT.\\

\texttt{\#\# Important Evaluation Tips}\\
- Be rigorous and precise. \\
- Focus on meaning and context, not exact wording. \\
- Interpret scope and qualifiers carefully, especially in ATOMIC FACTS.\\
- Consider hierarchical relationships between terms. \\

Now, based on what you learned from the evaluation criteria and tips, evaluate whether the input ATOMIC FACT is present in, or directly supported by, the SOURCE TEXT. Justify and carefully thinking step-by-step about your answer.\\

\texttt{\#\# Task}\\
SOURCE TEXT: \\
\{llm\_response\_text\}\\

ATOMIC FACT: \\
\{article\_facts\_text\}
\end{tcolorbox}

\caption{Zero-shot prompt used by LLM judge for factual recall. We use \texttt{gpt-5.4-mini} (no reasoning, temperature 1). The prompt was shortened to fit within the page. Full prompt available in supplementary material / code.}
\label{fig:fact-recall-prompt}
\end{small}
\end{figure}

\clearpage

\begin{table*}[t]
\centering
\tiny
\setlength{\tabcolsep}{4pt}
\renewcommand{\arraystretch}{0.88}
\caption{Full evaluation performance of \texttt{gpt-5.4-mini}, \texttt{claude-haiku-4.5}, and \texttt{gemini-3-flash} on the factual precision labeling task (three classes: \textsc{Supported}, \textsc{Contradicted}, and \textsc{Not Supported}) under zero-shot and few-shot prompting, across different reasoning settings and temperatures. We evaluate performance on $N=123$ expert-labeled examples, excluding six few-shot examples used in the prompt to mitigate evaluation leakage. The best performance is marked in \textbf{bold}.}

\resizebox{0.75\linewidth}{!}{%
\begin{tabular}{%
  >{\centering\arraybackslash}m{2.6cm}
  >{\centering\arraybackslash}m{1.8cm}
  >{\centering\arraybackslash}m{0.8cm}
  cccc}
\toprule
\textbf{Model} & \textbf{Reasoning} & \textbf{Temp.} & \textbf{Precision} & \textbf{Recall} & \textbf{F1} & \textbf{Accuracy} \\
\midrule

\multirow{24}{*}{\centering \texttt{gpt-5.4-mini}}
& \multicolumn{6}{c}{\textit{Zero-shot Prompt}} \\[-0.5ex]
\cmidrule(lr){2-7}
& \multirow{3}{*}{None}   & 0   & 0.801 & 0.716 & 0.732 & 0.756 \\
&                          & 0.2 & 0.830 & 0.715 & 0.737 & 0.756 \\
&                          & 1   & 0.846 & 0.741 & 0.762 & 0.772 \\
\cmidrule(lr){2-7}
& \multirow{3}{*}{Low}    & 0   & 0.719 & 0.698 & 0.662 & 0.683 \\
&                          & 0.2 & 0.746 & 0.679 & 0.646 & 0.659 \\
&                          & 1   & 0.694 & 0.629 & 0.573 & 0.594 \\
\cmidrule(lr){2-7}
& \multirow{3}{*}{Medium} & 0   & 0.728 & 0.685 & 0.642 & 0.667 \\
&                          & 0.2 & 0.729 & 0.680 & 0.631 & 0.642 \\
&                          & 1   & 0.706 & 0.661 & 0.619 & 0.634 \\
\cmidrule(lr){2-7}
& \multirow{3}{*}{High}   & 0   & 0.701 & 0.680 & 0.643 & 0.659 \\
&                          & 0.2 & 0.690 & 0.654 & 0.618 & 0.642 \\
&                          & 1   & 0.654 & 0.641 & 0.604 & 0.626 \\
\cmidrule(lr){2-7}
& \multicolumn{6}{c}{\textit{Few-shot Prompt}} \\[-0.5ex]
\cmidrule(lr){2-7}
& \multirow{3}{*}{None}   & 0   & 0.840 & 0.715 & 0.743 & 0.789 \\
&                          & 0.2 & \textbf{0.882} & \textbf{0.813} & \textbf{0.837} & \textbf{0.830} \\
&                          & 1   & 0.861 & 0.700 & 0.728 & 0.781 \\
\cmidrule(lr){2-7}
& \multirow{3}{*}{Low}    & 0   & 0.808 & 0.750 & 0.743 & 0.748 \\
&                          & 0.2 & 0.717 & 0.672 & 0.651 & 0.667 \\
&                          & 1   & 0.736 & 0.672 & 0.660 & 0.667 \\
\cmidrule(lr){2-7}
& \multirow{3}{*}{Medium} & 0   & 0.756 & 0.705 & 0.680 & 0.691 \\
&                          & 0.2 & 0.753 & 0.672 & 0.650 & 0.667 \\
&                          & 1   & 0.720 & 0.680 & 0.651 & 0.659 \\
\cmidrule(lr){2-7}
& \multirow{3}{*}{High}   & 0   & 0.710 & 0.666 & 0.635 & 0.659 \\
&                          & 0.2 & 0.750 & 0.730 & 0.706 & 0.724 \\
&                          & 1   & 0.720 & 0.693 & 0.663 & 0.675 \\

\midrule

\multirow{12}{*}{\centering \texttt{claude-haiku-4.5}}
& \multicolumn{6}{c}{\textit{Zero-shot Prompt}} \\[-0.5ex]
\cmidrule(lr){2-7}
& \multirow{3}{*}{None} & 0   & 0.688 & 0.674 & 0.612 & 0.634 \\
&                        & 0.2 & 0.724 & 0.718 & 0.663 & 0.691 \\
&                        & 1   & 0.700 & 0.693 & 0.632 & 0.659 \\
\cmidrule(lr){2-7}
& \multirow{3}{*}{Ext. Thinking} & 0   & 0.706 & 0.686 & 0.627 & 0.650 \\
&                                 & 0.2 & 0.652 & 0.648 & 0.589 & 0.618 \\
&                                 & 1   & 0.716 & 0.686 & 0.639 & 0.667 \\
\cmidrule(lr){2-7}
& \multicolumn{6}{c}{\textit{Few-shot Prompt}} \\[-0.5ex]
\cmidrule(lr){2-7}
& \multirow{3}{*}{None} & 0   & 0.686 & 0.713 & 0.661 & 0.683 \\
&                        & 0.2 & 0.671 & 0.700 & 0.645 & 0.667 \\
&                        & 1   & 0.709 & 0.718 & 0.676 & 0.691 \\
\cmidrule(lr){2-7}
& \multirow{3}{*}{Ext. Thinking} & 0   & 0.683 & 0.687 & 0.632 & 0.650 \\
&                                 & 0.2 & 0.707 & 0.686 & 0.645 & 0.667 \\
&                                 & 1   & 0.702 & 0.686 & 0.642 & 0.667 \\

\midrule

\multirow{24}{*}{\centering \texttt{gemini-3-flash}}
& \multicolumn{6}{c}{\textit{Zero-shot Prompt}} \\[-0.5ex]
\cmidrule(lr){2-7}
& \multirow{3}{*}{Minimal} & 0   & 0.734 & 0.751 & 0.714 & 0.732 \\
&                           & 0.2 & 0.744 & 0.770 & 0.726 & 0.740 \\
&                           & 1   & 0.746 & 0.783 & 0.740 & 0.756 \\
\cmidrule(lr){2-7}
& \multirow{3}{*}{Low} & 0   & 0.771 & 0.776 & 0.753 & 0.764 \\
&                       & 0.2 & 0.771 & 0.776 & 0.753 & 0.764 \\
&                       & 1   & 0.767 & 0.770 & 0.746 & 0.756 \\
\cmidrule(lr){2-7}
& \multirow{3}{*}{Medium} & 0   & 0.619 & 0.386 & 0.315 & 0.472 \\
&                          & 0.2 & 0.694 & 0.437 & 0.395 & 0.520 \\
&                          & 1   & 0.721 & 0.470 & 0.443 & 0.545 \\
\cmidrule(lr){2-7}
& \multirow{3}{*}{High} & 0   & 0.485 & 0.359 & 0.256 & 0.472 \\
&                        & 0.2 & 0.652 & 0.379 & 0.301 & 0.480 \\
&                        & 1   & 0.822 & 0.391 & 0.318 & 0.496 \\
\cmidrule(lr){2-7}
& \multicolumn{6}{c}{\textit{Few-shot Prompt}} \\[-0.5ex]
\cmidrule(lr){2-7}
& \multirow{3}{*}{Minimal} & 0   & 0.730 & 0.757 & 0.730 & 0.756 \\
&                           & 0.2 & 0.714 & 0.745 & 0.713 & 0.740 \\
&                           & 1   & 0.722 & 0.757 & 0.722 & 0.756 \\
\cmidrule(lr){2-7}
& \multirow{3}{*}{Low} & 0   & 0.752 & 0.783 & 0.760 & 0.789 \\
&                       & 0.2 & 0.743 & 0.776 & 0.751 & 0.781 \\
&                       & 1   & 0.768 & 0.802 & 0.777 & 0.797 \\
\cmidrule(lr){2-7}
& \multirow{3}{*}{Medium} & 0   & 0.715 & 0.746 & 0.686 & 0.707 \\
&                          & 0.2 & 0.672 & 0.715 & 0.650 & 0.667 \\
&                          & 1   & 0.678 & 0.714 & 0.665 & 0.683 \\
\cmidrule(lr){2-7}
& \multirow{3}{*}{High} & 0   & 0.668 & 0.668 & 0.628 & 0.642 \\
&                        & 0.2 & 0.704 & 0.759 & 0.696 & 0.707 \\
&                        & 1   & 0.689 & 0.727 & 0.664 & 0.683 \\

\bottomrule
\end{tabular}%
}
\label{tab:fact-precision-full}
\end{table*}

\clearpage

\begin{table*}[t]
\centering
\tiny
\caption{Full evaluation performance of \texttt{gpt-5.4-mini}, \texttt{claude-haiku-4.5}, and \texttt{gemini-3-flash} on the factual recall labeling task (two classes: \textsc{Supported}, \textsc{Not Supported}) under zero-shot and few-shot prompting, across different reasoning settings and temperatures. We evaluate performance on $N=113$ expert-labeled examples, excluding six few-shot examples used in the prompt to mitigate evaluation leakage. The best performance is marked in \textbf{bold}. }

\setlength{\tabcolsep}{4pt}
\renewcommand{\arraystretch}{0.88}

\resizebox{0.75\linewidth}{!}{%
\begin{tabular}{%
  >{\centering\arraybackslash}m{2.6cm}
  >{\centering\arraybackslash}m{1.8cm}
  >{\centering\arraybackslash}m{0.8cm}
  cccc}
\toprule
\textbf{Model} & \textbf{Reasoning} & \textbf{Temp.} & \textbf{Precision} & \textbf{Recall} & \textbf{F1} & \textbf{Accuracy} \\
\midrule

\multirow{24}{*}{\centering \texttt{gpt-5.4-mini}}
& \multicolumn{6}{c}{\textit{Zero-shot Prompt}} \\[-0.5ex]
\cmidrule(lr){2-7}
& \multirow{3}{*}{None}   & 0   & 0.895 & 0.756 & 0.819 & 0.867 \\
&                          & 0.2 & 0.875 & 0.778 & 0.824 & 0.867 \\
&                          & 1   & \textbf{0.947} & \textbf{0.800} & \textbf{0.868} & \textbf{0.903} \\
\cmidrule(lr){2-7}
& \multirow{3}{*}{Low}    & 0   & 0.962 & 0.556 & 0.704 & 0.814 \\
&                          & 0.2 & 0.926 & 0.556 & 0.694 & 0.805 \\
&                          & 1   & 0.929 & 0.578 & 0.712 & 0.814 \\
\cmidrule(lr){2-7}
& \multirow{3}{*}{Medium} & 0   & 0.964 & 0.600 & 0.740 & 0.832 \\
&                          & 0.2 & 0.963 & 0.578 & 0.722 & 0.823 \\
&                          & 1   & 0.957 & 0.489 & 0.647 & 0.788 \\
\cmidrule(lr){2-7}
& \multirow{3}{*}{High}   & 0   & 0.926 & 0.556 & 0.694 & 0.805 \\
&                          & 0.2 & 0.926 & 0.556 & 0.694 & 0.805 \\
&                          & 1   & 0.962 & 0.556 & 0.704 & 0.814 \\
\cmidrule(lr){2-7}
& \multicolumn{6}{c}{\textit{Few-shot Prompt}} \\[-0.5ex]
\cmidrule(lr){2-7}
& \multirow{3}{*}{None}   & 0   & 0.787 & 0.822 & 0.804 & 0.841 \\
&                          & 0.2 & 0.792 & 0.844 & 0.817 & 0.850 \\
&                          & 1   & 0.792 & 0.844 & 0.817 & 0.850 \\
\cmidrule(lr){2-7}
& \multirow{3}{*}{Low}    & 0   & 0.900 & 0.600 & 0.720 & 0.814 \\
&                          & 0.2 & 0.906 & 0.644 & 0.753 & 0.832 \\
&                          & 1   & 0.966 & 0.622 & 0.757 & 0.841 \\
\cmidrule(lr){2-7}
& \multirow{3}{*}{Medium} & 0   & 0.900 & 0.600 & 0.720 & 0.814 \\
&                          & 0.2 & 0.931 & 0.600 & 0.730 & 0.823 \\
&                          & 1   & 0.929 & 0.578 & 0.712 & 0.814 \\
\cmidrule(lr){2-7}
& \multirow{3}{*}{High}   & 0   & 0.909 & 0.667 & 0.769 & 0.841 \\
&                          & 0.2 & 0.966 & 0.622 & 0.757 & 0.841 \\
&                          & 1   & 0.964 & 0.600 & 0.740 & 0.832 \\

\midrule

\multirow{12}{*}{\centering \texttt{claude-haiku-4.5}}
& \multicolumn{6}{c}{\textit{Zero-shot Prompt}} \\[-0.5ex]
\cmidrule(lr){2-7}
& \multirow{3}{*}{None} & 0   & 0.818 & 0.600 & 0.692 & 0.788 \\
&                        & 0.2 & 0.800 & 0.622 & 0.700 & 0.788 \\
&                        & 1   & 0.765 & 0.578 & 0.658 & 0.761 \\
\cmidrule(lr){2-7}
& \multirow{3}{*}{Ext. Thinking} & 0   & 0.960 & 0.533 & 0.686 & 0.805 \\
&                                 & 0.2 & 0.923 & 0.533 & 0.676 & 0.797 \\
&                                 & 1   & 0.926 & 0.556 & 0.694 & 0.805 \\
\cmidrule(lr){2-7}
& \multicolumn{6}{c}{\textit{Few-shot Prompt}} \\[-0.5ex]
\cmidrule(lr){2-7}
& \multirow{3}{*}{None} & 0   & 0.920 & 0.511 & 0.657 & 0.788 \\
&                        & 0.2 & 0.857 & 0.533 & 0.658 & 0.779 \\
&                        & 1   & 0.909 & 0.440 & 0.597 & 0.761 \\
\cmidrule(lr){2-7}
& \multirow{3}{*}{Ext. Thinking} & 0   & 0.889 & 0.533 & 0.667 & 0.788 \\
&                                 & 0.2 & 0.917 & 0.489 & 0.638 & 0.779 \\
&                                 & 1   & 0.909 & 0.444 & 0.597 & 0.761 \\

\midrule

\multirow{24}{*}{\centering \texttt{gemini-3-flash}}
& \multicolumn{6}{c}{\textit{Zero-shot Prompt}} \\[-0.5ex]
\cmidrule(lr){2-7}
& \multirow{3}{*}{Minimal} & 0   & 0.818 & 0.800 & 0.809 & 0.850 \\
&                           & 0.2 & 0.818 & 0.800 & 0.809 & 0.850 \\
&                           & 1   & 0.837 & 0.800 & 0.818 & 0.858 \\
\cmidrule(lr){2-7}
& \multirow{3}{*}{Low} & 0   & 0.822 & 0.822 & 0.822 & 0.858 \\
&                       & 0.2 & 0.841 & 0.822 & 0.832 & 0.867 \\
&                       & 1   & 0.804 & 0.822 & 0.813 & 0.850 \\
\cmidrule(lr){2-7}
& \multirow{3}{*}{Medium} & 0   & 0.947 & 0.400 & 0.563 & 0.752 \\
&                          & 0.2 & 1.000 & 0.489 & 0.657 & 0.797 \\
&                          & 1   & 1.000 & 0.511 & 0.677 & 0.805 \\
\cmidrule(lr){2-7}
& \multirow{3}{*}{High} & 0   & 0.955 & 0.467 & 0.627 & 0.779 \\
&                        & 0.2 & 0.955 & 0.444 & 0.606 & 0.770 \\
&                        & 1   & 1.000 & 0.578 & 0.732 & 0.832 \\
\cmidrule(lr){2-7}
& \multicolumn{6}{c}{\textit{Few-shot Prompt}} \\[-0.5ex]
\cmidrule(lr){2-7}
& \multirow{3}{*}{Minimal} & 0   & 0.833 & 0.778 & 0.805 & 0.850 \\
&                           & 0.2 & 0.833 & 0.778 & 0.805 & 0.850 \\
&                           & 1   & 0.833 & 0.778 & 0.805 & 0.850 \\
\cmidrule(lr){2-7}
& \multirow{3}{*}{Low} & 0   & 0.844 & 0.844 & 0.844 & 0.876 \\
&                       & 0.2 & 0.844 & 0.844 & 0.844 & 0.876 \\
&                       & 1   & 0.861 & 0.822 & 0.841 & 0.876 \\
\cmidrule(lr){2-7}
& \multirow{3}{*}{Medium} & 0   & 1.000 & 0.400 & 0.571 & 0.761 \\
&                          & 0.2 & 1.000 & 0.311 & 0.475 & 0.726 \\
&                          & 1   & 0.964 & 0.600 & 0.740 & 0.832 \\
\cmidrule(lr){2-7}
& \multirow{3}{*}{High} & 0   & 1.000 & 0.356 & 0.525 & 0.743 \\
&                        & 0.2 & 1.000 & 0.422 & 0.594 & 0.770 \\
&                        & 1   & 0.923 & 0.533 & 0.676 & 0.797 \\

\bottomrule
\end{tabular}%
}
\label{tab:fact-recall-full}
\end{table*}

\clearpage

\section{Full Evaluation Details}\label{appendix:full-evaluation-details}

In this section, we describe the selected hyperparameters (\S\ref{appendix:hyperparameters}), system prompts and output preprocessing (\S\ref{appendix:prompt-output}), and the \textit{clean-room} evaluation protocols for closed-source agents (\S\ref{appendix:closed-sourced}).

\subsection{Selected Hyperparameters}\label{appendix:hyperparameters}
In our evaluation, we configure each model and deep research agent to its default recommended hyperparameter settings and, when available, set the reasoning level to the highest setting, with a small number of targeted adjustments. For Anthropic's \texttt{claude-sonnet-4.5}, to stay within a reasonable cost, we enable \textit{extended} thinking with a budget of 4{,}096 tokens for reasoning and tool use, and allow up to 8{,}192 tokens per turn. For Perplexity models \texttt{sonar-deep-research} and \texttt{sonar-reasoning-pro}, we configure retrieval to support controlled, high-quality synthesis by enabling \texttt{search\_mode=academic}, \texttt{web\_search\_context\_size=high}, \texttt{reasoning\_effort=high}, and \texttt{search\_type=auto}. 

\subsection{System Prompts \& Output Processing }\label{appendix:prompt-output}

For \texttt{DR Tulu}, we adapt the original system prompt \cite{shao2025drtulureinforcementlearning} to align with our benchmark task. However, \texttt{DR Tulu} often struggles with instruction-following, failing to produce a paragraph-long conclusion within the required \textbackslash boxed\{\} format. To address this, we post-process its structured outputs by extracting sections under relevant markdown headers (e.g., ``Conclusion,'' ``Summary,'' ``Bottom line'') and use the corresponding text as the synthesized conclusion.
 
For other models and agents, we use the same system prompt described in Figure~\ref{fig:model-instruction-prompt}. For base settings (i.e., models without tool-use integration in \textsc{SciConHarness}), we append an additional instruction to the system prompt indicating that tool calls are unavailable, preventing unnecessary tool-calling behavior, as some models otherwise attempt to invoke tools even when none are provided.

\subsection{Clean-room Evaluation Protocols for Closed-Source Agents}\label{appendix:closed-sourced}

Some provider-hosted agents do not support direct integration with \textsc{SciConHarness}, requiring alternative strategies to enforce our \textit{clean-room} evaluation protocol. In particular, (1) Perplexity's \texttt{sonar-deep-research} and \texttt{sonar-reasoning-pro} rely on provider-native search APIs and do not support custom MCP tool calling, preventing direct integration of \textsc{SciConHarness}, and (2) OpenAI Deep Research agents operate within a provider-controlled execution environment, where custom tools must be supplied via remote MCP server endpoints rather than direct integration with \textsc{SciConHarness}. Below, we describe how we adapt clean-room protocols under these constraints.

\subsubsection{Clean-room for Perplexity}
Perplexity models such as \texttt{sonar-deep-research} and \texttt{sonar-reasoning-pro} rely on native web search and do not support custom MCP tool calling, making direct integration with \textsc{SciConHarness} infeasible. Instead, we enforce our \textit{clean-room} protocol on a best effort basis using provider-side search filters. Specifically, we apply \texttt{search\_before\_date\_filter} to restrict all web search outcomes to those published before the ground-truth \texttt{CDSR} review publication date, and \texttt{search\_domain\_filter} (a denylist of up to 20 URLs) to exclude web search results that leak the ground-truth artifacts (e.g., \texttt{CDSR} review).

For \texttt{sonar-reasoning-pro}, we iteratively expand the \texttt{search\_domain\_filter} list by repeating the same query until no new leakage from the search results is detected (using the same filtering heuristic in \textsc{SciConHarness}'s \textit{clean-room} evaluation protocol) or the 20-URL cap is reached. For \texttt{sonar-deep-research}, iterative expansion via repeated querying is cost-prohibitive, so we instead create a fixed \texttt{search\_domain\_filter} list per query. Specifically, for each query, we identify the top 18 most frequently filtered URLs across prior evaluations of \texttt{claude-sonnet-4.5}, \texttt{gemini-3-pro}, and \texttt{gpt-5.1} on that same query, clean and format URLs, deduplicate, and combine them with two default \texttt{CDSR} domains (\texttt{cochrane.org}, \texttt{cochranelibrary.com}). To avoid over-filtering, entries in \texttt{search\_domain\_filter} are handled at the URL (article) level rather than blocking entire domains (e.g., PubMed/PMC, a major repository of scientific papers), preserving retrieval utility while preventing access to known leakage URLs.

\subsubsection{Clean-room for OpenAI Deep Research} OpenAI's deep research agents, such as \texttt{o3-deep-research} and \texttt{o4-mini-deep-research}, run autonomously in the background within a provider-controlled execution environment, requiring any custom tools to be supplied via remote MCP server endpoints \cite{openai_deep_research_guide}. To enable \textit{clean-room} evaluation under these constraints, we implement remote MCP servers\footnote{\url{https://developers.openai.com/api/docs/guides/tools-connectors-mcp}} with endpoints that expose \textsc{SciConHarness}'s search and browsing tools with integrated \textit{clean-room} filtering. This allows controlled open-web access while preserving the agent’s native workflow. 

Under OpenAI's Deep Research interface requirements for remote MCP servers, each remote MCP server must expose two primitives (\texttt{search}, \texttt{fetch}). To provide coverage over our full tool suite (e.g., \texttt{google\_search}, \texttt{paper\_search}, \texttt{web\_browse}) while respecting the \texttt{search} and \texttt{fetch} primitives, we implement two HTTP-based remote MCP servers: (1) \texttt{Serper+Jina} for \texttt{google\_search} and \texttt{web\_browse} tools and (2) \texttt{SemanticScholar+Jina} for \texttt{paper\_search} and \texttt{web\_browse} tools. Both remote MCP servers impose the same \textit{clean-room} filtering protocol across the tools as done in \textsc{SciConHarness}. These remote MCP servers are hosted behind an nginx gateway\footnote{\url{https://nginx.org/}} and exposed via ngrok\footnote{\url{https://ngrok.com/}} for external API access, with bearer-token authentication and standardized tool schemas for OpenAI remote MCP compatibility.

Before each \textsc{SciConBench} query, the client configures \textit{clean-room} filters (e.g., ground-truth \texttt{CDSR} title and publication date) on both remote MCP servers via POST requests and verifies the configuration via GET requests, retrying on mismatch. During inference, Deep Research agents issue tool calls to the remote MCP endpoints, which enforce \textit{clean-room} filtering across the tools and return only compliant results. Inference proceeds asynchronously with polling. All runs produce structured logs capturing filter decisions, excluded URLs, and tool usage. Filtering can also be disabled via the configuration endpoint, enabling evaluation of OpenAI Deep Research agents without \textit{clean-room} constraints.

\begin{figure}[h]
\begin{small}
\centering
\begin{tcolorbox}[
    colback=gray!5,
    colframe=black,
    width=\linewidth,
    arc=2mm,
    boxrule=0.5pt,
    title=\textbf{Summarization Prompt}
]
\textbf{System Prompt:} \textcolor{teal}{You are a helpful assistant that creates summaries of web content focusing on main details.}\\

\textbf{Instruction:} Summarize the following web content, focusing only on the main details and key information. Preserve important facts, numbers, dates, and conclusions. Aim to filter out any noisy characters (e.g., HTML tags, social media links, random strings, etc.) and outputting only important information. Specific details, including but not limited to metrics, deltas, definitions, settings, limitations, and citations and references should be preserved. Make sure not to lose any key information. \\

Content to summarize:\\
\{content\}
\end{tcolorbox}

\caption{Prompt used to summarize long page text returned by the \texttt{web\_browse} tool. We use \texttt{gpt-5-mini} with default settings (e.g., medium verbosity and reasoning effort).}
\label{fig:summarization-prompt}
\end{small}
\end{figure}

\clearpage

\begin{figure}[t]
\begin{small}
\centering
\begin{tcolorbox}[
    colback=gray!5,
    colframe=black,
    width=\linewidth,
    arc=2mm,
    boxrule=0.5pt,
    title=\textbf{System Prompt Used for Evaluation}
]
You are a research assistant who answers scientific questions by identifying relevant sources, assessing their evidence quality and certainty, and synthesizing the evidence into evidence-backed conclusions.\\

Task Requirements:\\
- Synthesize a comprehensive, paragraph-long conclusion that directly answers the question. The conclusion must be clear, well-supported, and WRAPPED with THREE SQUARE BRACKETS. While you may generate additional content beyond the conclusion, the conclusion must be the main focus.\\
- Focus on synthesizing the overall body of evidence (e.g., highlighting relationships across sources, identifying contradictions, etc) to form a coherent conclusion rather than just enumerating information. Weigh the synthesis more heavily toward higher-quality evidence when formulating the conclusion.\\
- In your conclusion, explicitly describe both strengths and limitations of the evidence quality (e.g., risk of bias, imprecision, inconsistency), including uncertainty, gaps, or conflicts across sources. Explicitly state when evidence is limited, low quality, or inconsistent and explain what additional research would help resolve these gaps.\\
- Only provide the final answer when ready. If available, tool calls are permitted without any hard limits, but should be used judiciously with a clear purpose to gather sufficient information to derive a conclusion to the question.\\
- Please prefer high-quality sources as evidence (peer-reviewed papers, journals, sources like PubMed, etc) and prioritize recent work for fast-moving areas. Do not rely on or use Cochrane reviews for this task.\\
- Cite all claims from search results. You should ground every nontrivial claim in retrieved snippets and sources, if available. Please include the sources cited in the form of references at the end of the answer.\\
- Most importantly, DO NOT invent snippets or citations and never fabricate content.\\

Synthesize the conclusion, with the text being at most a paragraph-long. MAKE SURE to enclose the entire paragraph within exactly three square brackets on each side, like this: [[[Enter your conclusion here]]]. Do not include any additional formatting outside the triple brackets.\\

\textcolor{orange}{IMPORTANT: Tool calls are NOT available. Do NOT attempt to make any function calls or tool calls. Answer the question directly using only your knowledge and reasoning.}
\end{tcolorbox}

\caption{System prompt for all models evaluated on \textsc{SciConBench}. The \textcolor{orange}{orange text} is appended for base models without tool access.}
\label{fig:model-instruction-prompt}
\end{small}
\end{figure}

\clearpage

\begin{figure}[t]
\begin{small}
\centering
\begin{tcolorbox}[
    colback=gray!5,
    colframe=black,
    width=\linewidth,
    arc=2mm,
    boxrule=0.5pt,
    title=\textbf{System Prompt for DR Tulu Evaluations}
]
  You are a research assistant who answers scientific questions through iterative reasoning and research. You identify relevant sources, assess their evidence quality and certainty, and synthesize the evidence into evidence-backed conclusions.\\

  \#\# Process\\
  - Use <think></think> tags to show your reasoning at any point.\\
  - Use <call\_tool name="...">query</call\_tool> when you need information (see tools below).\\
  - You can alternate between thinking and searching multiple times.\\
  - Only provide <answer></answer> tags when you have enough information for a complete response. Within the <answer> tags, you MUST synthesize a comprehensive, paragraph-long conclusion that directly answers the question. **REQUIRED FORMAT**: The synthesized conclusion (at most a paragraph long) MUST be placed in \textbackslash boxed\{\} format: \textbackslash boxed\{your synthesized conclusion here\}. This is mandatory - your answer must include \textbackslash boxed\{\} with your conclusion.\\
  ...

  \#\# Calling Tools (<call\_tool name="...">query</call\_tool>)\\
  - You can use the following tools:\\

  1. google\_search \\
  ...\\
  2. browse\_webpage \\
  ...\\
  3. snippet\_search \\
  ...\\

  \#\# Tool Output\\
  - After you issue a tool call, we will execute it and return results wrapped in <tool\_output> tags. \\
  ...\\

  \#\# Requirements\\
  - Focus on synthesizing the overall body of evidence (e.g., highlighting relationships across sources, identifying contradictions, etc) to form a coherent conclusion rather than just enumerating information. Ideally, you should synthesize rather than enumerate content: it's helpful to group findings across papers, explain relationships, and build a coherent narrative that answers the question, supported by citations. Weigh the synthesis more heavily toward higher-quality evidence when formulating the conclusion.\\
  - In your conclusion, explicitly describe both strengths and limitations of the evidence quality (e.g., risk of bias, imprecision, inconsistency), including uncertainty, gaps, or conflicts across sources. Explicitly state when evidence is limited, low quality, or inconsistent and explain what additional research would help resolve these gaps. You should acknowledge uncertainty and conflicts; if evidence is thin or sources disagree, state it and explain what additional evidence would resolve it.\\
  - Please prefer high-quality sources as evidence (peer-reviewed papers, journals, sources like PubMed, etc). Please prefer authoritative sources (peer-reviewed papers, reputable benchmarks/docs) and prioritize recent work for fast-moving areas. Do not simply focus on Cochrane reviews for this task.\\
  - Most importantly, DO NOT invent snippets or citations and never fabricate content.\\
  - Tool calls are permitted without any hard limits, but should be used judiciously with a clear purpose to gather sufficient information to derive a conclusion to the question. \\

  \#\# Answer and Citation Format\\

  - Once you collect all of the necessary information, generate the final answer, and mark your answer with answer tags: <answer></answer>. \\
  - **CRITICAL REQUIREMENT**: Within the <answer></answer> tags, you MUST place your synthesized conclusion paragraph in \textbackslash boxed\{\} format... \\
  ...\\

  IMPORTANT: \\
  Before you return your final answer, you should ensure that your final answer contains a \textbackslash boxed\{\} tag, which contains a paragraph-long synthesized conclusion. Make sure it's concise and to the point.\\
\end{tcolorbox}

\caption{System prompt used for \texttt{DR Tulu} evaluations on \textsc{SciConBench}, adapted from the original system prompt in \cite{shao2025drtulureinforcementlearning}. The system prompt shown is shortened to fit within a page and omits components already specified in the original system prompt \cite{shao2025drtulureinforcementlearning}; see the original for full details.}
\label{fig:dr-tulu-instruction-prompt}
\end{small}
\end{figure}

\clearpage

\section{Power Analysis}\label{appendix:power-analysis}
Here, we compute the minimum detectable effect (MDE) given our evaluation sample size of $N=268$. In our case, the MDE is the smallest change in our metrics (e.g., factual precision, recall) that our benchmark would be able to detect with high probability (80\%). 

\xhdr{Preliminaries} Let there be $n$ paired observations $(X_t, Y_t)$ on the same items, e.g., two models evaluated on the same $n$ items. For each pair $t = 1, ..., n$, let the difference between outcomes be $D_t = X_t - Y_t$. 

Let the sample mean and SD of difference be: $\bar D = \frac{1}{n} \sum_{t=1}^n D_t$ and $s_D = \sqrt{ \frac{1}{n-1} \sum_{t=1}^n (D_t - \bar D)^2 }.$
We test $H_0: \mu_D = 0$ vs.\ $H_1: \mu_D \neq 0$ using $t = \frac{\bar D}{s_D / \sqrt{n}}$, which follows a $t$-distribution with $n-1$ degrees of freedom under $H_0$. We reject $H_0$ if $|t| > t_{n-1,1-\alpha/2}$ for a two-sided test.

\xhdr{Power analysis} First, we assume that $t_{n-1,x} \approx z_{x}, \forall x$  for large enough $n$. Since our sample size is $N=268$, we compute the MDE represented by $\Delta$ with significance $\alpha$ and power $1-\beta$:
\[
\Delta = \sqrt{\frac{(z_{1-\alpha/2} + z_{1-\beta})^2 \sigma_D^2}{N}}.
\]
where $\sigma_D^2 = \mathrm{Var}(D_t)$.

To estimate the population variance $\sigma_D^2$, we use the observed variance from a pilot study as a proxy. Specifically, we evaluate \texttt{claude-sonnet-4.5} and \texttt{gpt-5.1} on $N=100$ \textsc{SciConBench} samples using \textsc{SciConHarness} under the \textit{clean-room} protocol, compute factual F1-scores via atomic fact decomposition, and estimate the variance ($\sigma_{D, pilot}^2=0.0457$) of their paired differences. 

Assuming power $1-\beta = 0.8$ and significance level $\alpha = 0.05$ (with $z_{1-\alpha/2} = 1.96$ and $z_{1-\beta} = 0.8416$), and using $\sigma_{D,pilot}^2$ as an estimate of $\sigma_D^2$, the MDE for our benchmark evaluation sample ($N=268$) is:
\[
\Delta \approx \sqrt{\frac{(1.96 + 0.8416)^2 \cdot 0.0457}{268}} \approx 0.037.
\]

Thus, with $N=268$, our evaluation is powered to detect differences in factual F1-scores between models of at least $\Delta = 0.037$ at $\alpha=0.05$ and power $0.8$.

\newpage 

\section{Cost Analysis}\label{appendix:cost-analysis}

\xhdr{Model Querying Costs} Table~\ref{tab:cost-breakdown-querying} reports the cost breakdown for querying models and deep research agents across different \textsc{SciConHarness} settings. To estimate inference costs for proprietary systems, we use the billed costs reported in their API consoles. For open-weight and fully-open systems (e.g., \texttt{DR Tulu}), we follow Shao et al. \cite{shao2025drtulureinforcementlearning} and estimate the inference cost using OpenRouter with Qwen3-8B pricing (\$0.2 per 1M input/output tokens).\footnote{\url{https://openrouter.ai/qwen/qwen3-8b}} Under \textsc{SciConHarness} \textit{tools} (no clean-room), \texttt{DR Tulu} uses an average of 9{,}940 input and 3{,}467 output tokens per query, corresponding to \$0.0027 per query; with \textsc{SciConHarness} \textit{tools + clean-room}, it uses an average of 10{,}731 input and 3{,}362 output tokens per query, corresponding to \$0.0028 per query. To estimate tool costs, we track the tool usage in \textsc{SciConHarness} across the evaluated systems and compute the tool cost, following Shao et al. \cite{shao2025drtulureinforcementlearning}, in which \texttt{paper\_search} is free to use due to the Semantic Scholar API, \texttt{web\_browse} is estimated to be \$0.00005 per use and \texttt{google\_search} is estimated to be USD \$0.00075 per use. We sum per-query inference and tool costs to obtain the total cost per query, and compute the overall cost by multiplying by the number of evaluated queries ($N=268$). An exception is \texttt{sonar-deep-research}, which is evaluated on a random subset ($N=100$) due to its high per-query cost (\$2.22--\$2.314). The total querying cost across all systems is \$1,569.946.

\begin{table*}[h]
\centering
\small
\caption{Cost breakdown for querying models and deep research agents (denoted as \textit{DR}) across different \textsc{SciConHarness} settings: \textit{Base} (no tools), \textsc{SciConHarness} \textit{tools}, and \textsc{SciConHarness} \textit{tools + clean-room}. For each system, we indicate inference cost (e.g., token usage), average number of tool calls per query (e.g., \textsc{web\_browse}, \textsc{google\_search}, \textsc{paper\_search}), corresponding tool costs per query, and total cost per query (inference cost + tool cost). Total cost is aggregated across all evaluated queries ($N=268$) for the system. ``-'' denotes this information was not applicable, while $\dagger$ indicates that \texttt{sonar-deep-research} is evaluated on a random subset ($N=100$ out of 268) due to high inference cost per query (\$2.22-\$2.314 per query). All costs are in USD. More details on the specific cost estimations are available in the \S \ref{appendix:cost-analysis}.
}
\setlength{\tabcolsep}{2pt}
\resizebox{\textwidth}{!}{
\begin{tabular}{lccccccc}
\toprule
& \textbf{Inference Cost} & \textbf{\texttt{web\_browse}} & \textbf{\texttt{google\_search}} & \textbf{\texttt{paper\_search}} & \textbf{Tool Cost} & \textbf{Total / Query} & \textbf{Total} \\
 & \textbf{(\$ / Query)} & \textbf{(\# / Query)} & \textbf{(\# / Query)} & \textbf{(\# / Query)} & \textbf{(\$ / Query)} & \textbf{(\$ / Query)} & \textbf{Cost (\$)} \\
\midrule

\textcolor{gray}{\footnotesize \textit{Base Models (No Tools)}} \\
\modelicon{openai_icon}~\texttt{gpt-5.1} & 0.0460 & - & - & - & 0 & 0.0460 & 12.328 \\
\modelicon{claude_icon}~\texttt{claude-sonnet-4.5} & 0.0320 & - & - & - & 0 & 0.0320 & 8.576 \\
\modelicon{gemini_icon}~\texttt{gemini-3-pro} & 0.0300 & - & - & - & 0 & 0.0300 & 8.040 \\
\midrule
\multicolumn{7}{r}{\textbf{Subtotal}} & \textbf{28.944} \\

\midrule
\textcolor{gray}{\footnotesize\textit{Models (\textsc{SciConHarness})}} \\
\modelicon{openai_icon}~\texttt{gpt-5.1} & 0.1358 & 8.38 & 5.15 & 0.52 & 0.0043 & 0.1401 & 37.552 \\
\modelicon{claude_icon}~\texttt{claude-sonnet-4.5} & 0.8250 & 1.47 & 0.69 & 6.81 & 0.0006 & 0.8256 & 221.261\\
\modelicon{gemini_icon}~\texttt{gemini-3-pro} & 0.1040 & 0.13 & 0.82 & 4.65 & 0.0006 & 0.1046 & 28.033 \\
\modelicon{perplexity_icon}~\texttt{sonar-reasoning-pro} & 0.0143 & -- & -- & -- & -- & 0.0143 & 3.832 \\
\midrule
\multicolumn{7}{r}{\textbf{Subtotal}} & \textbf{290.678} \\

\midrule
\textcolor{gray}{\footnotesize\textit{Models (\textsc{SciConHarness} + clean-room)}} \\
\modelicon{openai_icon}~\texttt{gpt-5.1} & 0.1250 & 7.91 & 4.65 & 0.55 & 0.0040 & 0.1290 & 34.570 \\
\modelicon{claude_icon}~\texttt{claude-sonnet-4.5} & 0.9080 & 1.55 & 1.02 & 7.23 & 0.0008 & 0.9088 & 243.558 \\
\modelicon{gemini_icon}~\texttt{gemini-3-pro} & 0.1180 & 0.18 & 0.90 & 4.55 & 0.0007 & 0.1187 & 31.812 \\
\modelicon{perplexity_icon}~\texttt{sonar-reasoning-pro} & 0.0130 & -- & -- & -- & -- & 0.0130 & 3.484 \\
\midrule
\multicolumn{7}{r}{\textbf{Subtotal}} & \textbf{313.424} \\

\midrule
\textcolor{gray}{\footnotesize\textit{DR (\textsc{SciConHarness})}} \\
\modelicon{ai2_icon}~\texttt{DR Tulu} & 0.0028 & 0.02 & 0.21 & 5.48 & 0.0002 & 0.0030 & 0.804 \\
\modelicon{perplexity_icon}~\texttt{sonar-deep-research$\dagger$} & 2.3140 & -- & -- & -- & -- & 2.3140 & 231.400 \\
\modelicon{openai_icon}~\texttt{o4-mini-deep-research} & 0.1215 & 6.78 & 7.57 & 2.69 & 0.0060 & 0.1275 & 34.170 \\
\modelicon{openai_icon}~\texttt{o3-deep-research} & 0.7913 & 6.54 & 3.78 & 1.10 & 0.0032 & 0.7945 & 212.926 \\
\midrule
\multicolumn{7}{r}{\textbf{Subtotal}} & \textbf{479.300} \\

\midrule
\textcolor{gray}{\footnotesize\textit{DR (\textsc{SciConHarness} + clean-room)}} \\

\modelicon{ai2_icon}~\texttt{DR Tulu} & 0.0027 & 0.02 & 0.34 & 5.09 & 0.0003 & 0.0030 & 0.804 \\
\modelicon{perplexity_icon}~\texttt{sonar-deep-research$\dagger$} & 2.2200 & -- & -- & -- & -- & 2.2200 & 222.000 \\
\modelicon{openai_icon}~\texttt{o4-mini-deep-research} & 0.0920 & 6.71 & 9.08 & 2.87 & 0.0071 & 0.0991 & 26.559 \\
\modelicon{openai_icon}~\texttt{o3-deep-research} & 0.7730 & 7.85 & 5.23 & 1.84 & 0.0040 & 0.7770 & 208.236 \\
\midrule
\multicolumn{7}{r}{\textbf{Subtotal}} & \textbf{457.599} \\
\midrule
\multicolumn{7}{r}{\textbf{Total (\$)}} & \textbf{1{,}569.946} \\
\bottomrule
\end{tabular}}
\label{tab:cost-breakdown-querying}
\end{table*}

\xhdr{Atomic Fact Generation Costs} Table~\ref{tab:afg-cost-breakdown} reports the cost breakdown for atomic fact generation across models and deep research agents under different \textsc{SciConHarness} settings. We calculate the costs using billed API usage from provider consoles. Models such as \texttt{gpt-5.1} and \texttt{o3-deep-research} generate longer conclusions, resulting in more atomic facts and higher token usage, which increases the cost per conclusion. In total, atomic fact generation across all evaluated conclusions costs \$1,239.48.

\begin{table*}[t]
\centering
\small
\caption{Cost breakdown for atomic fact generation across models and deep research agents (denoted as \textit{DR}) under different \textsc{SciConHarness} settings. After querying these systems and generating conclusions, we decompose the conclusions into atomic facts using our pipeline (\S\ref{sec:atomic-fact-gen}). For each system, we report the output characteristic of their generated conclusions: average number of tokens (\textbf{Avg Tokens} column), sentences (\textbf{Avg Sent.} column), and atomic facts (\textbf{Avg Facts} column). We also report the average pipeline cost per conclusion (in API costs; see \textbf{Cost / Conclusion} column) and total cost aggregated across all processed conclusions ($N=268$). As in Table~\ref{tab:cost-breakdown-querying}, $\dagger$ next to \texttt{sonar-deep-research} denotes evaluation on a subset ($N=100$). All costs are in USD; see \S\ref{appendix:cost-analysis} for details.}
\setlength{\tabcolsep}{8pt}
\resizebox{\textwidth}{!}{
\begin{tabular}{lccccc}
\toprule
 & \textbf{Avg Tokens} & \textbf{Avg Sent.} & \textbf{Avg Facts} & \textbf{Cost / Conclusion} & \textbf{Total Cost} \\
 & \textbf{(± STD)} & \textbf{(± STD)} & \textbf{(± STD)} & \textbf{(\$ / Conclusion)} & \textbf{(\$)} \\
\midrule

\textcolor{gray}{\footnotesize \textit{Base Models (No Tools)}} \\
\modelicon{openai_icon}~\texttt{gpt-5.1} & 612.6 ± 147.5 & 8.5 ± 5.3 & 47.4 ± 12.7 & 0.397728 & 106.5912 \\
\modelicon{claude_icon}~\texttt{claude-sonnet-4.5} & 298.8 ± 47.5 & 6.8 ± 1.3 & 27.1 ± 5.9 & 0.219444 & 58.8111 \\
\modelicon{gemini_icon}~\texttt{gemini-3-pro} & 278.6 ± 49.1 & 4.8 ± 2.3 & 18.7 ± 4.3 & 0.156472 & 41.9346 \\
\midrule
\multicolumn{5}{r}{\textbf{Subtotal}} & \textbf{207.3369} \\

\midrule
\textcolor{gray}{\footnotesize\textit{Models (\textsc{SciConHarness})}} \\
\modelicon{openai_icon}~\texttt{gpt-5.1} & 896.7 ± 205.2 & 11.7 ± 6.3 & 53.6 ± 13.4 & 0.505919 & 135.5864 \\
\modelicon{claude_icon}~\texttt{claude-sonnet-4.5} & 455.8 ± 82.2 & 8.5 ± 2.1 & 34.5 ± 8.3 & 0.298460 & 79.9873 \\
\modelicon{gemini_icon}~\texttt{gemini-3-pro} & 278.2 ± 61.0 & 5.5 ± 1.2 & 19.4 ± 4.8 & 0.165279 & 44.2948 \\
\modelicon{perplexity_icon}~\texttt{sonar-reasoning-pro} & 238.8 ± 74.5 & 5.0 ± 1.9 & 18.2 ± 5.5 & 0.148061 & 39.6803 \\
\midrule
\multicolumn{5}{r}{\textbf{Subtotal}} & \textbf{299.5488} \\

\midrule
\textcolor{gray}{\footnotesize\textit{Models (\textsc{SciConHarness} + clean-room)}} \\
\modelicon{openai_icon}~\texttt{gpt-5.1} & 892.6 ± 195.0 & 12.4 ± 7.3 & 53.1 ± 14.0 & 0.510080 & 136.7014 \\
\modelicon{claude_icon}~\texttt{claude-sonnet-4.5} & 462.7 ± 89.0 & 8.8 ± 2.2 & 35.4 ± 8.8 & 0.305704 & 81.9287 \\
\modelicon{gemini_icon}~\texttt{gemini-3-pro} & 282.4 ± 58.7 & 5.6 ± 1.3 & 19.8 ± 5.2 & 0.168562 & 45.1746 \\
\modelicon{perplexity_icon}~\texttt{sonar-reasoning-pro} & 230.1 ± 79.4 & 4.7 ± 1.8 & 17.2 ± 5.6 & 0.139078 & 37.2728 \\
\midrule
\multicolumn{5}{r}{\textbf{Subtotal}} & \textbf{301.0775} \\

\midrule
\textcolor{gray}{\footnotesize\textit{DR (\textsc{SciConHarness})}} \\
\modelicon{ai2_icon}~\texttt{DR Tulu} & 368.9 ± 163.7 & 3.9 ± 2.5 & 21.3 ± 10.8 & 0.168142 & 45.0621 \\
\modelicon{perplexity_icon}~\texttt{sonar-deep-research$\dagger$} & 370.6 ± 90.7 & 5.8 ± 2.1 & 30.1 ± 8.2 & 0.239811 & 23.9811 \\
\modelicon{openai_icon}~\texttt{o4-mini-deep-research} & 541.3 ± 168.6 & 9.5 ± 2.8 & 25.6 ± 8.1 & 0.246513 & 66.0654 \\
\modelicon{openai_icon}~\texttt{o3-deep-research} & 810.4 ± 338.0 & 12.0 ± 5.5 & 34.2 ± 16.3 & 0.366058 & 98.1035 \\
\midrule
\multicolumn{5}{r}{\textbf{Subtotal}} & \textbf{233.2121} \\

\midrule
\textcolor{gray}{\footnotesize\textit{DR (\textsc{SciConHarness} + clean-room)}} \\
\modelicon{ai2_icon}~\texttt{DR Tulu} & 369.0 ± 151.1 & 4.3 ± 2.6 & 15.7 ± 8.1 & 0.112384 & 30.1188 \\
\modelicon{perplexity_icon}~\texttt{sonar-deep-research$\dagger$} & 349.4 ± 106.4 & 5.7 ± 2.1 & 28.6 ± 8.9 & 0.230995 & 23.0995 \\
\modelicon{openai_icon}~\texttt{o4-mini-deep-research} & 499.9 ± 146.3 & 9.4 ± 2.5 & 25.6 ± 7.3 & 0.242194 & 64.9079 \\
\modelicon{openai_icon}~\texttt{o3-deep-research} & 789.5 ± 315.4 & 12.2 ± 4.6 & 29.9 ± 11.7 & 0.299174 & 80.1787 \\
\midrule
\multicolumn{5}{r}{\textbf{Subtotal}} & \textbf{198.3049} \\
\midrule
\multicolumn{5}{r}{\textbf{Total (\$)}} & \textbf{1{,}239.4802} \\

\bottomrule
\end{tabular}}
\label{tab:afg-cost-breakdown}
\end{table*}

\begin{table*}[t]
\centering
\caption{Cost breakdown for measuring factual precision and recall of generated conclusions from models and deep research agents (denoted as \textit{DR}) under different \textsc{SciConHarness} settings. We decompose conclusions into atomic facts, then assess precision and recall using our expert-validated \texttt{gpt-5.4-mini} judge (\S\ref{sec:factual-llm-judge}). We report the API billing costs of using \texttt{gpt-5.4-mini}. For each system, we report the total number of facts evaluated (\textbf{\# Facts} column), cost to evaluate per fact (\textbf{Cost (\$) / Facts} column), and the total cost of evaluating all the facts (\textbf{Total Cost (\$)} column) for both factual precision and recall. As in Table~\ref{tab:cost-breakdown-querying}, $\dagger$ next to \texttt{sonar-deep-research} denotes evaluation on a subset ($N=100$). All costs are in USD; see \S\ref{appendix:cost-analysis} for details.}
\small
\setlength{\tabcolsep}{5pt}
\resizebox{\textwidth}{!}{
\begin{tabular}{lccc@{\hspace{6pt}}@{\hspace{6pt}}ccc}
\toprule
 & \multicolumn{3}{c}{\textbf{Precision}} & \multicolumn{3}{c}{\textbf{Recall}} \\
\cmidrule(lr){2-4} \cmidrule(lr){5-7}
 & \textbf{\# Facts} & \textbf{Cost (\$) / Fact} & \textbf{Total Cost (\$)} & \textbf{\# Facts} & \textbf{Cost (\$) / Fact} & \textbf{Total Cost (\$)} \\
\midrule

\textcolor{gray}{\footnotesize \textit{Base Models (No Tools)}} \\
\modelicon{openai_icon}~\texttt{gpt-5.1} & 12709 & 0.003276 & 41.6347 & 2820 & 0.001548 & 4.3654 \\
\modelicon{claude_icon}~\texttt{claude-sonnet-4.5} & 7253 & 0.003215 & 23.3184 & 2820 & 0.001344 & 3.7901 \\
\modelicon{gemini_icon}~\texttt{gemini-3-pro} & 5009 & 0.003281 & 16.4345 & 2820 & 0.001451 & 4.0918 \\
\midrule
\multicolumn{3}{r}{\textbf{Subtotal}} & \textbf{81.3876} &
\multicolumn{2}{r}{\textbf{Subtotal}} & \textbf{12.2473} \\

\midrule
\textcolor{gray}{\footnotesize\textit{Models (\textsc{SciConHarness})}} \\
\modelicon{openai_icon}~\texttt{gpt-5.1} & 14378 & 0.003309 & 47.5768 & 2820 & 0.001652 & 4.6586 \\
\modelicon{claude_icon}~\texttt{claude-sonnet-4.5} & 9257 & 0.003198 & 29.6039 & 2820 & 0.001393 & 3.9283 \\
\modelicon{gemini_icon}~\texttt{gemini-3-pro} & 5200 & 0.003277 & 17.0404 & 2820 & 0.001413 & 3.9847 \\
\modelicon{perplexity_icon}~\texttt{sonar-reasoning-pro} & 4870 & 0.003243 & 15.7934 & 2820 & 0.001322 & 3.7280 \\
\midrule
\multicolumn{3}{r}{\textbf{Subtotal}} & \textbf{110.0145} &
\multicolumn{2}{r}{\textbf{Subtotal}} & \textbf{16.2996} \\

\midrule
\textcolor{gray}{\footnotesize\textit{Models (\textsc{SciConHarness} + clean-room)}} \\
\modelicon{openai_icon}~\texttt{gpt-5.1} & 14219 & 0.003312 & 47.0933 & 2820 & 0.001670 & 4.7094 \\
\modelicon{claude_icon}~\texttt{claude-sonnet-4.5} & 9481 & 0.003206 & 30.3961 & 2820 & 0.001419 & 4.0016 \\
\modelicon{gemini_icon}~\texttt{gemini-3-pro} & 5298 & 0.003281 & 17.3827 & 2820 & 0.001412 & 3.9818 \\
\modelicon{perplexity_icon}~\texttt{sonar-reasoning-pro} & 4606 & 0.003262 & 15.0248 & 2820 & 0.001365 & 3.8493 \\
\midrule
\multicolumn{3}{r}{\textbf{Subtotal}} & \textbf{109.8969} &
\multicolumn{2}{r}{\textbf{Subtotal}} & \textbf{16.5421} \\

\midrule
\textcolor{gray}{\footnotesize\textit{DR  (\textsc{SciConHarness})}} \\
\modelicon{ai2_icon}~\texttt{DR Tulu} & 5714 & 0.003286 & 18.7762 & 2820 & 0.001519 & 4.2836 \\
\modelicon{perplexity_icon}~\texttt{sonar-deep-research$\dagger$} & 3005 & 0.003178 & 9.5499 & 1075 & 0.001311 & 1.4093 \\
\modelicon{openai_icon}~\texttt{o4-mini-deep-research} & 6872 & 0.003309 & 22.7394 & 2820 & 0.001353 & 3.8155 \\
\modelicon{openai_icon}~\texttt{o3-deep-research} & 9174 & 0.003309 & 30.3568 & 2820 & 0.001419 & 4.0002 \\
\midrule
\multicolumn{3}{r}{\textbf{Subtotal}} & \textbf{81.4223} &
\multicolumn{2}{r}{\textbf{Subtotal}} & \textbf{13.5086} \\

\midrule
\textcolor{gray}{\footnotesize\textit{DR (\textsc{SciConHarness} + clean-room)}} \\
\modelicon{ai2_icon}~\texttt{DR Tulu} & 4209 & 0.003309 & 13.9276 & 2820 & 0.001514 & 4.2695 \\
\modelicon{perplexity_icon}~\texttt{sonar-deep-research$\dagger$} & 2855 & 0.003181 & 9.0818 & 1075 & 0.001360 & 1.4620 \\
\modelicon{openai_icon}~\texttt{o4-mini-deep-research} & 6856 & 0.003296 & 22.5974 & 2820 & 0.001370 & 3.8634 \\
\modelicon{openai_icon}~\texttt{o3-deep-research} & 8006 & 0.003309 & 26.4919 & 2820 & 0.001453 & 4.0975 \\
\midrule
\multicolumn{3}{r}{\textbf{Subtotal}} & \textbf{72.0987} &
\multicolumn{2}{r}{\textbf{Subtotal}} & \textbf{13.6924} \\

\midrule
\multicolumn{3}{r}{\textbf{Total (Precision)}} & \textbf{454.82} &
\multicolumn{2}{r}{\textbf{Total (Recall)}} & \textbf{72.2900} \\
\midrule
\multicolumn{6}{r}{\textbf{Total (\$)}} & \textbf{527.1100} \\

\bottomrule
\end{tabular}}
\label{tab:fact-cost-breakdown}
\end{table*}

\xhdr{Measuring Factual Precision and Recall Costs} Table~\ref{tab:fact-cost-breakdown} reports the cost of evaluating factual precision and recall across models and deep research agents under different \textsc{SciConHarness} settings. Costs are computed from billed API usage. The factual precision task involves more facts than the factual recall task, as generated conclusions are typically longer than \texttt{CDSR} \textit{Authors' Conclusions}. Factual precision is also more expensive per fact (\$0.00318–\$0.00331 vs.\ \$0.00131–\$0.00167 for recall), since it requires longer inputs (e.g., the full abstracts and plain-language summary of \texttt{CDSR} review).

In total, precision evaluation costs \$454.82 and recall costs \$72.29, for a combined \$527.11. Using \texttt{gpt-5.4} instead of \texttt{gpt-5.4-mini} would exceed \$1{,}500 as it is over 3$\times$ more expensive in token usage costs. The LLM judge processes each fact in $\sim$2 seconds, compared to $\sim$6 minutes for domain expert annotation (\S\ref{sec:factual-llm-judge}), yielding over 180$\times$ speedup. In terms of cost, assuming U.S. federal minimum wage (\$7.25/hour) as a conservative lower bound, annotating a fact manually takes 6 minutes (\$0.725 per fact). In contrast, our LLM judge evaluation costs at most \$0.00331 (precision) and \$0.00167 (recall) per fact, corresponding to at least 219--434$\times$ cost savings.

\xhdr{Total Evaluation Cost} In total, our entire end-to-end benchmark evaluation cost \$1{,}569.95 + \$1{,}239.48 + \$527.11 = \$3{,}336.54.

\newpage 

\section{Additional Analyses}\label{appendix:additional-analysis}

In this section, we provide additional analyses organized into four themes: (1) \camready{\textit{additional models}, including newer open-weight models (\S\ref{appendix:open-weight}) and the latest frontier models (\S\ref{appendix:latest-models});} (2) \textit{contextualizing performance}, \camready{including an approximate oracle upper bound (\S\ref{appendix:oracle}) and} Pareto frontiers of performance vs.\ time and cost (\S\ref{appendix:performance-tradeoffs}); (3) \textit{performance breakdowns} by label distribution (\S\ref{appendix:label-distribution}), conclusion length (\S\ref{appendix:length-metric-tradeoff}), \camready{and clinical domain and review type (\S\ref{appendix:stratification})}; and (4) \textit{agent behavior and failure modes}, including \textsc{SciConHarness} tool usage patterns (\S\ref{appendix:tool-usage}), failure mode analysis (\S\ref{appendix:failure-mode}), \camready{and diagnostic analysis of tool use, retrieval, and synthesis (\S\ref{appendix:diagnostic-analysis})}.

\subsection{\camready{Evaluating Open-Weight Models}}\label{appendix:open-weight}
\camready{We evaluate four open-weight models trained for long-context agentic tasks (\texttt{deepseek-v4-pro}, \texttt{kimi-k3}, \texttt{glm-5.2}, and the smaller \texttt{qwen3.5-9b}) under all three \textsc{SciConHarness} settings. We use the setup in \S\ref{sec:evaluation-details}, with a thinking budget of 4{,}096 tokens for \texttt{qwen3.5-9b}. All models are evaluated on the full sample ($N=268$), except \texttt{kimi-k3} ($N=100$) due to cost constraints. Because the knowledge cutoffs of these open-weight models are not public, we note that these models may have been pre-trained on some or all of the evaluation sample and thus may inflate their performance results. Table~\ref{tab:open-weight-benchmark} reports the results.}

\begin{table}[!ht]
\centering
\scriptsize
\caption{\camready{Benchmark performance of open-weight models across \textsc{SciConHarness} settings. Macro factual precision, recall, and F1 (variance in parentheses). $\dagger$ indicates evaluation on a subset ($N=100$ of 268) due to cost constraints. \textbf{Bold} denotes the best performance without clean-room, and \underline{underline} with clean-room. \textbf{$\Delta_{\text{Tools}}$F1} is the F1 change from Base to \textsc{SciConHarness}, and \textbf{$\Delta_{\text{Clean}}$F1} the change when adding clean-room to \textsc{SciConHarness}.}}
\label{tab:open-weight-benchmark}
\resizebox{\columnwidth}{!}{%
\begin{tabular}{lccccc}
\toprule
 & \textbf{Factual Precision (Var)} & \textbf{Factual Recall (Var)} & \textbf{Factual F1 (Var)} & \textbf{$\Delta_{\text{Tools}}$ F1} & \textbf{$\Delta_{\text{Clean}}$ F1} \\
\midrule
\textcolor{gray}{\footnotesize \textit{Base Models (No Tools)}} \\
\modelicon{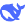}~\texttt{deepseek-v4-pro} & 0.491 (0.028) & 0.281 (0.042) & 0.316 (0.033) & -- & -- \\
\modelicon{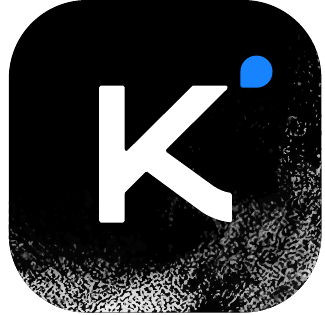}~\texttt{kimi-k3$\dagger$} & 0.354 (0.017) & 0.347 (0.041) & 0.311 (0.023) & -- & -- \\
\modelicon{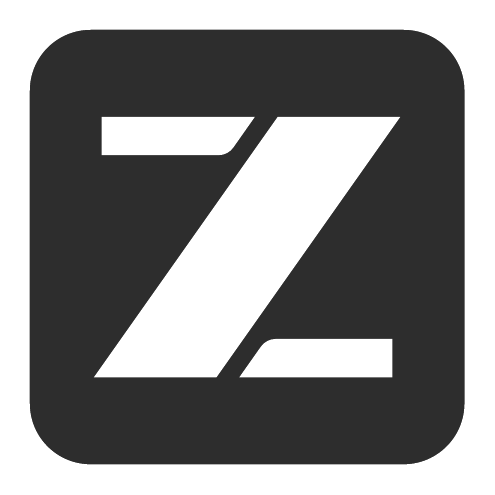}~\texttt{glm-5.2} & 0.449 (0.024) & 0.279 (0.047) & 0.298 (0.033) & -- & -- \\
\modelicon{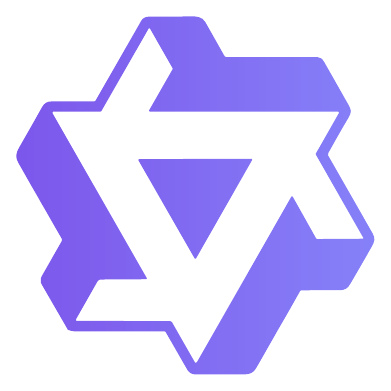}~\texttt{qwen3.5-9b} & 0.405 (0.031) & 0.215 (0.040) & 0.228 (0.033) & -- & -- \\
\midrule
\textcolor{gray}{\footnotesize\textit{Models (\textsc{SciConHarness})}} \\
\modelicon{deepseek_icon}~\texttt{deepseek-v4-pro} & 0.511 (0.044) & 0.457 (0.076) & 0.445 (0.053) & \positive{+0.129} & -- \\
\modelicon{kimi_icon}~\texttt{kimi-k3$\dagger$} & 0.335 (0.018) & 0.435 (0.060) & 0.338 (0.027) & \positive{+0.027} & -- \\
\modelicon{zai_icon}~\texttt{glm-5.2} & 0.454 (0.031) & \textbf{0.554} (0.082) & \textbf{0.469} (0.045) & \positive{+0.171} & -- \\
\modelicon{qwen_icon}~\texttt{qwen3.5-9b} & \textbf{0.520} (0.047) & 0.372 (0.071) & 0.389 (0.058) & \positive{+0.161} & -- \\
\midrule
\textcolor{gray}{\footnotesize\textit{Models (\textsc{SciConHarness} + clean-room)}} \\
\modelicon{deepseek_icon}~\texttt{deepseek-v4-pro} & \underline{0.400} (0.024) & 0.344 (0.055) & \underline{0.326} (0.032) & -- & \negative{$-$0.120} \\
\modelicon{kimi_icon}~\texttt{kimi-k3$\dagger$} & 0.321 (0.015) & 0.401 (0.056) & 0.308 (0.023) & -- & \negative{$-$0.030} \\
\modelicon{zai_icon}~\texttt{glm-5.2} & 0.318 (0.018) & \underline{0.403} (0.066) & 0.320 (0.030) & -- & \negative{$-$0.149} \\
\modelicon{qwen_icon}~\texttt{qwen3.5-9b} & 0.393 (0.028) & 0.247 (0.046) & 0.248 (0.031) & -- & \negative{$-$0.141} \\
\bottomrule
\end{tabular}}

\end{table}

\xhdr{\camready{Open-weight models are competitive with proprietary systems}} \camready{With \textsc{SciConHarness} tools (no clean-room), \texttt{glm-5.2} (F1 0.469) and \texttt{deepseek-v4-pro} (F1 0.445) outperform all non-deep-research models in Table~\ref{tab:full-benchmark} (best: \texttt{sonar-reasoning-pro}, F1 0.392) as well as \texttt{o4-mini-deep-research} (F1 0.427), but remain below \texttt{o3-deep-research} (F1 0.508). Tool access (no clean-room) improves factual F1 for all open-weight models, with the largest gain for \texttt{glm-5.2} ($+0.171$). Under clean-room evaluation, \texttt{deepseek-v4-pro} achieves the highest open-weight F1 (0.326), close to \texttt{o3-deep-research} (0.337), though this gap is below our minimum detectable effect ($\Delta \approx 0.037$; \S\ref{appendix:power-analysis}).}

\xhdr{\camready{Clean-room evaluation \textit{consistently} attenuates performance}} \camready{Tools improve F1 for all four models ($+0.027$ to $+0.171$), driven mostly by recall. Consistent with \S\ref{sec:benchmark-result}, the clean-room protocol then reduces F1 for every model ($-0.030$ to $-0.149$), eliminating most of these gains: each model's clean-room F1 falls within 0.022 of its base (no-tool) F1. Models that gain the most from unconstrained \textsc{SciConHarness} tools also lose the most under clean-room, further indicating that these gains stem largely from \textit{retrieving} ground-truth artifacts rather than genuine \textit{synthesis}.}

\xhdr{\camready{\textsc{SciConBench} is model-agnostic and informative for small open-weight models}} \camready{Using \textsc{SciConHarness} \textit{tools + clean-room}, the small, general-purpose \texttt{qwen3.5-9b} achieves a factual F1 of 0.248, exceeding both its base version (no tools; F1 0.228) and \texttt{DR Tulu}, a small, fully open specialized deep-research model, under the same setting (F1 0.145; Table~\ref{tab:full-benchmark}). Despite reporting performance comparable to OpenAI’s deep research system on existing benchmarks \cite{shao2025drtulureinforcementlearning}, \texttt{DR Tulu} performs substantially worse (F1 = 0.145 under clean-room evaluation) than proprietary models on \textsc{SciConBench}. Our log analysis suggests this gap partly stems from tool use: whereas \texttt{DR Tulu} relies almost exclusively on \texttt{paper\_search} snippets (\S\ref{appendix:diagnostic-analysis}), \texttt{qwen3.5-9b} distributes its calls more evenly (29\% \texttt{web\_browse}, 34\% \texttt{paper\_search}), gaining greater access to full-text evidence. One potential explanation is that \texttt{DR Tulu}'s post-training overfit to the tool-use patterns of its training benchmarks, limiting adaptation to \textsc{SciConHarness}'s open-web evidence synthesis setting; its poor instruction following (\S\ref{appendix:prompt-output}) may further impair synthesis. Overall, this demonstrates that \textsc{SciConBench} is model-agnostic and informative for evaluating small open-weight models.}

\subsection{\camready{Evaluating the Latest Frontier Models}}\label{appendix:latest-models}
\camready{Since the release of \textsc{SciConBench}, several new models have been released by frontier labs. As a \textit{live} benchmark, \textsc{SciConBench} continuously evaluates newly released models on a longitudinal leaderboard\footnote{\url{https://sciconbench.cs.princeton.edu/}} under \textsc{SciConHarness} \textit{tools + clean-room}. Table~\ref{tab:latest-models} reports seven of the latest models from the September 2026 release, evaluated on $N=185$ \texttt{CDSR} reviews published between July 2025 and September 2026.}

\begin{table}[!ht]
\centering
\scriptsize
\setlength{\tabcolsep}{8pt}
\caption{\camready{Performance of the latest frontier models under \textsc{SciConHarness} \textit{tools + clean-room} from the September 2026 release ($N=185$). $\dagger$ indicates evaluation on $N=180$ of 185 reviews due to model abstention. This result suggests that performance is improving, but the latest frontier models still struggle. \textbf{Bold} denotes the best performance.}}
\label{tab:latest-models}
\begin{tabular}{lccc}
\toprule
 & \textbf{Factual Precision} & \textbf{Factual Recall} & \textbf{Factual F1} \\
\midrule
\textcolor{gray}{\footnotesize\textit{Models (\textsc{SciConHarness} + clean-room)}} \\
\modelicon{openai_icon}~\texttt{gpt-6.1-sol} & \textbf{0.440} & 0.450 & \textbf{0.411} \\
\modelicon{deepseek_icon}~\texttt{deepseek-v4.1-flash} & 0.284 & 0.463 & 0.324 \\
\modelicon{kimi_icon}~\texttt{kimi-k3} & 0.329 & 0.402 & 0.323 \\
\modelicon{zai_icon}~\texttt{glm-5.3} & 0.295 & 0.430 & 0.321 \\
\modelicon{qwen_icon}~\texttt{qwen3.8-max} & 0.294 & 0.439 & 0.319 \\
\modelicon{claude_icon}~\texttt{claude-opus-5.5$\dagger$} & 0.264 & \textbf{0.481} & 0.312 \\
\modelicon{gemini_icon}~\texttt{gemini-3.8-flash} & 0.300 & 0.372 & 0.299 \\
\bottomrule
\end{tabular}

\end{table}

\xhdr{\camready{Performance is improving, but the latest frontier models still struggle}} \camready{\texttt{gpt-6.1-sol} achieves the highest factual F1 score (0.411), exceeding the next-best model  \texttt{deepseek-v4.1-flash} (F1 0.324). It also substantially exceeds the best clean-room F1 score reported at the benchmark's initial release (\texttt{o3-deep-research}, 0.337; Table~\ref{tab:full-benchmark}), suggesting improvements in model performance since its release. The remaining six models cluster within a narrow F1 range (0.299--0.324). Still, no model exceeds 0.49 on any metric, indicating that reliable scientific conclusion synthesis remains far from solved.}

\subsection{\camready{Approximating Expert-Level Performance: An Oracle Upper Bound}}\label{appendix:oracle}
\camready{To contextualize AI agent performance, we aim to establish an approximate expert-level baseline for our scientific conclusion synthesis task: the factual F1 an agent would achieve if it synthesized conclusions as well as the expert authors of \texttt{CDSR} reviews. A true human-expert baseline, however, is difficult to establish: producing a Cochrane review can take more than two years of literature synthesis~\cite{andersen2020half}, and the expertise to \textit{produce} such reviews differs from that needed to \textit{use} them, as most clinicians who rely on Cochrane reviews do not conduct Cochrane-level syntheses themselves. We therefore construct an approximate \textit{oracle} on a subset of $N=100$ of the 268 reviews (\S\ref{sec:evaluation-details}), using each review's original \textit{Authors' Conclusions} as the model response and evaluating it with the same pipeline (\S\ref{sec:metrics}).}

\xhdr{\camready{Near-perfect performance is attainable}} \camready{The oracle achieves a factual F1 of 0.95, indicating that our evaluation pipeline introduces little loss and that near-perfect scores are attainable when a response closely matches the expert-written synthesis. This provides an approximate upper bound on factual F1: the best clean-room system (\texttt{o3-deep-research}, F1 0.337; Table~\ref{tab:full-benchmark}) and the strongest latest frontier model (\texttt{gpt-6.1-sol}, F1 0.411; \S\ref{appendix:latest-models}) remain far below it. Importantly, this oracle uses the expert authors' original conclusions rather than an independent human re-synthesis, and therefore serves as an upper-bound reference rather than a human performance baseline.}

\subsection{Pareto Frontier Between Performance vs. Time and Cost}\label{appendix:performance-tradeoffs}

\begin{figure}[t]
\centering
\begin{minipage}[t]{0.45\linewidth}
    \centering
    \includegraphics[width=\linewidth]{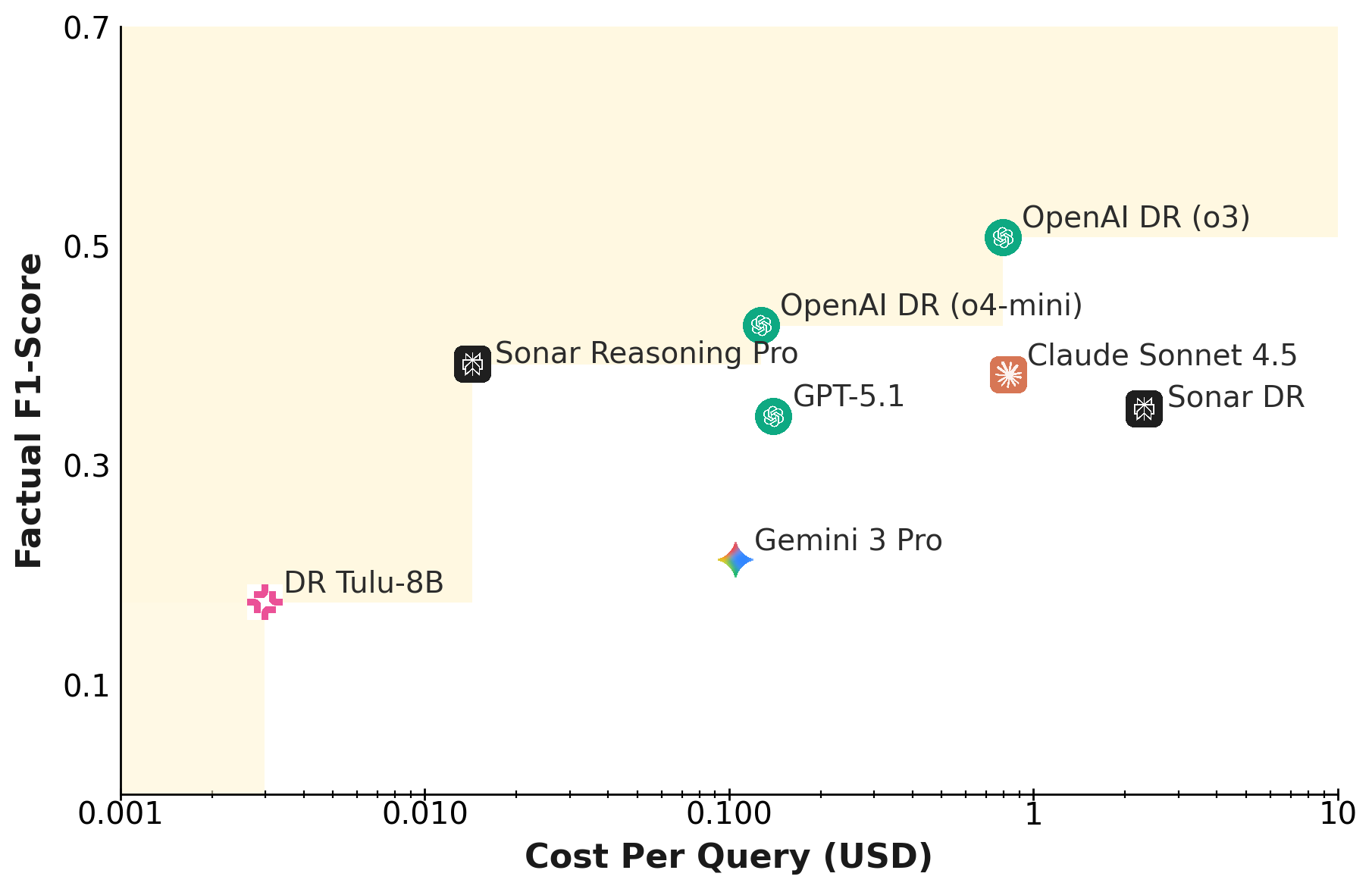}
\end{minipage}
\hfill
\begin{minipage}[t]{0.45\linewidth}
    \centering
    \includegraphics[width=\linewidth]{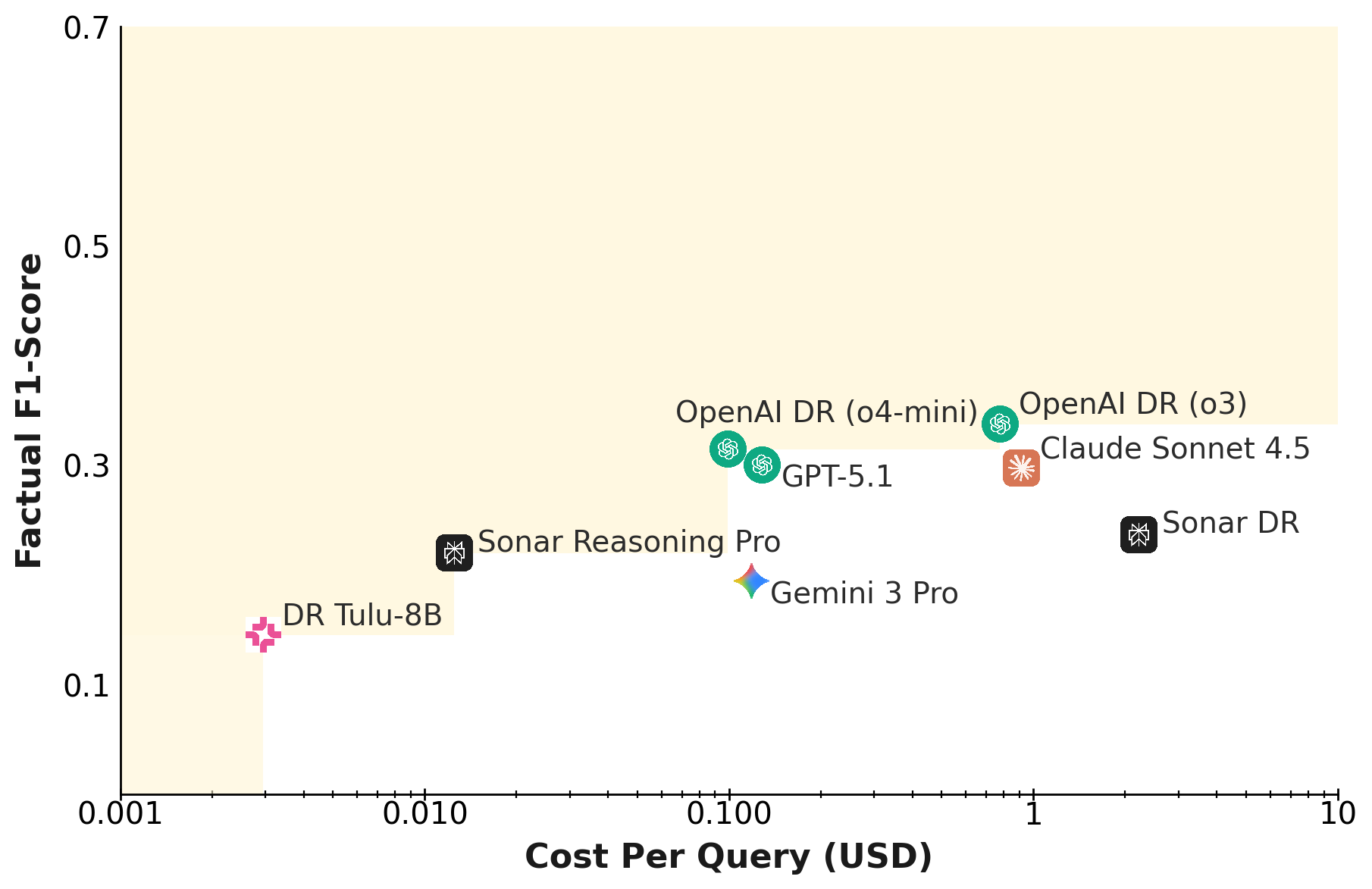}
\end{minipage}

\caption{Performance vs.\ cost of frontier models and deep research agents. We plot factual F1 against cost (USD per query). Left: \textsc{SciConHarness} without clean-room constraints. Right: \textsc{SciConHarness} with clean-room evaluation.}
\label{fig:performance-cost-F1}
\end{figure}

\begin{figure}[t]
\centering
\begin{minipage}[t]{0.45\linewidth}
    \centering
    \includegraphics[width=\linewidth]{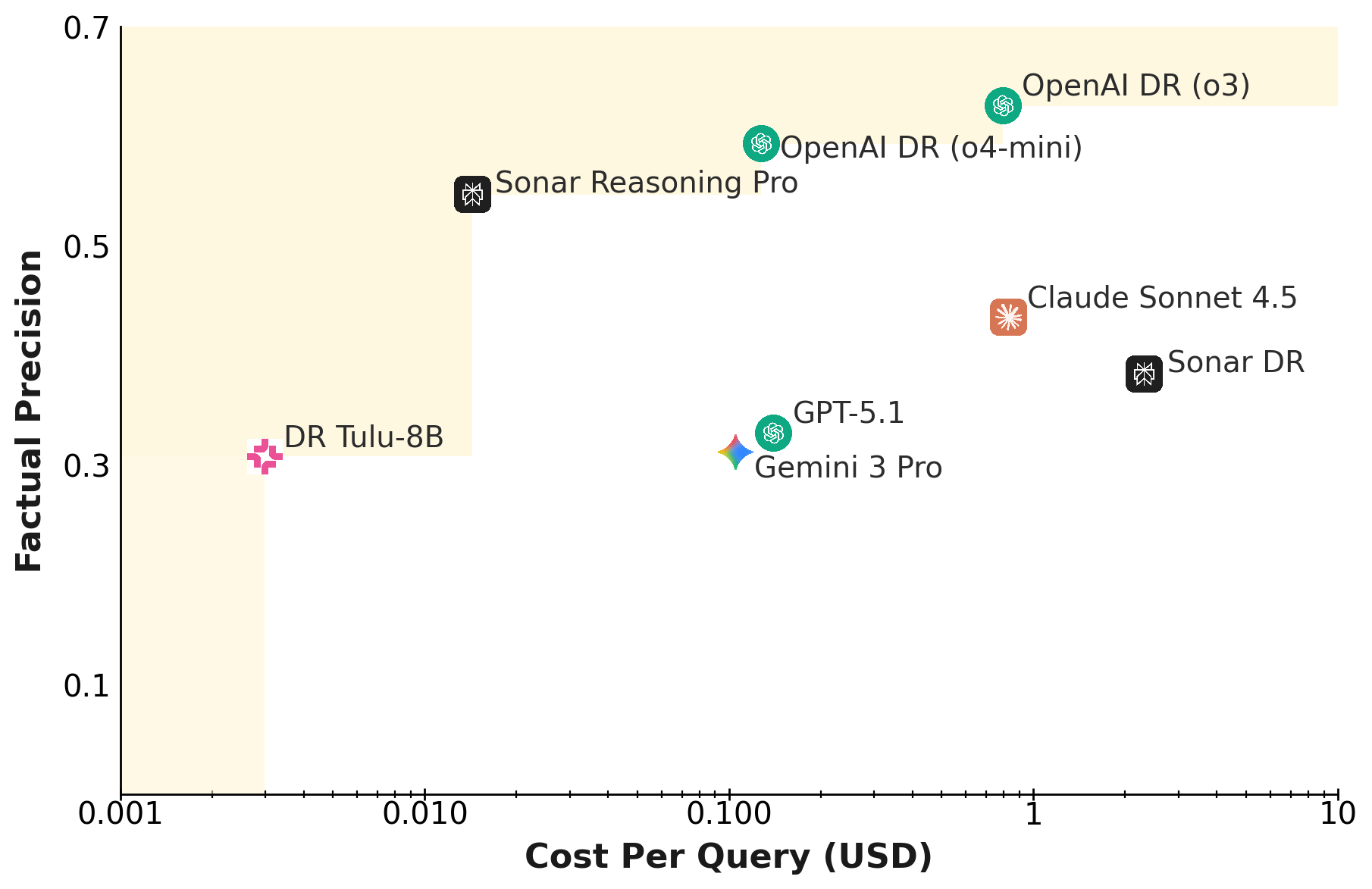}
\end{minipage}
\hfill
\begin{minipage}[t]{0.45\linewidth}
    \centering
    \includegraphics[width=\linewidth]{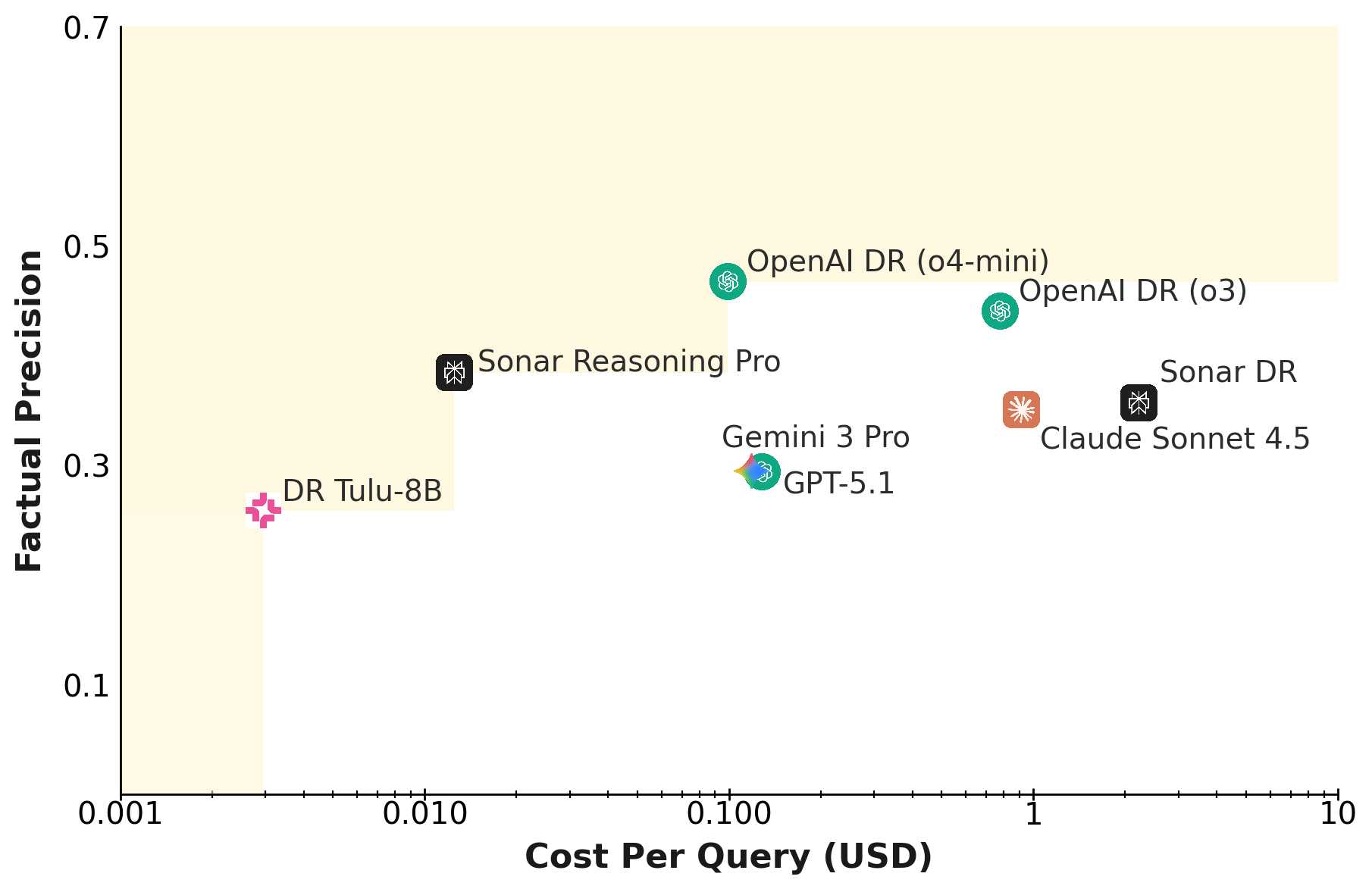}
\end{minipage}

\caption{Performance vs.\ cost of frontier models and deep research agents. We plot factual precision against cost (USD per query). Left: \textsc{SciConHarness} without clean-room constraints. Right: \textsc{SciConHarness} with clean-room evaluation.}
\label{fig:performance-cost-precision}
\end{figure}

\begin{figure}[t]
\centering
\begin{minipage}[t]{0.45\linewidth}
    \centering
    \includegraphics[width=\linewidth]{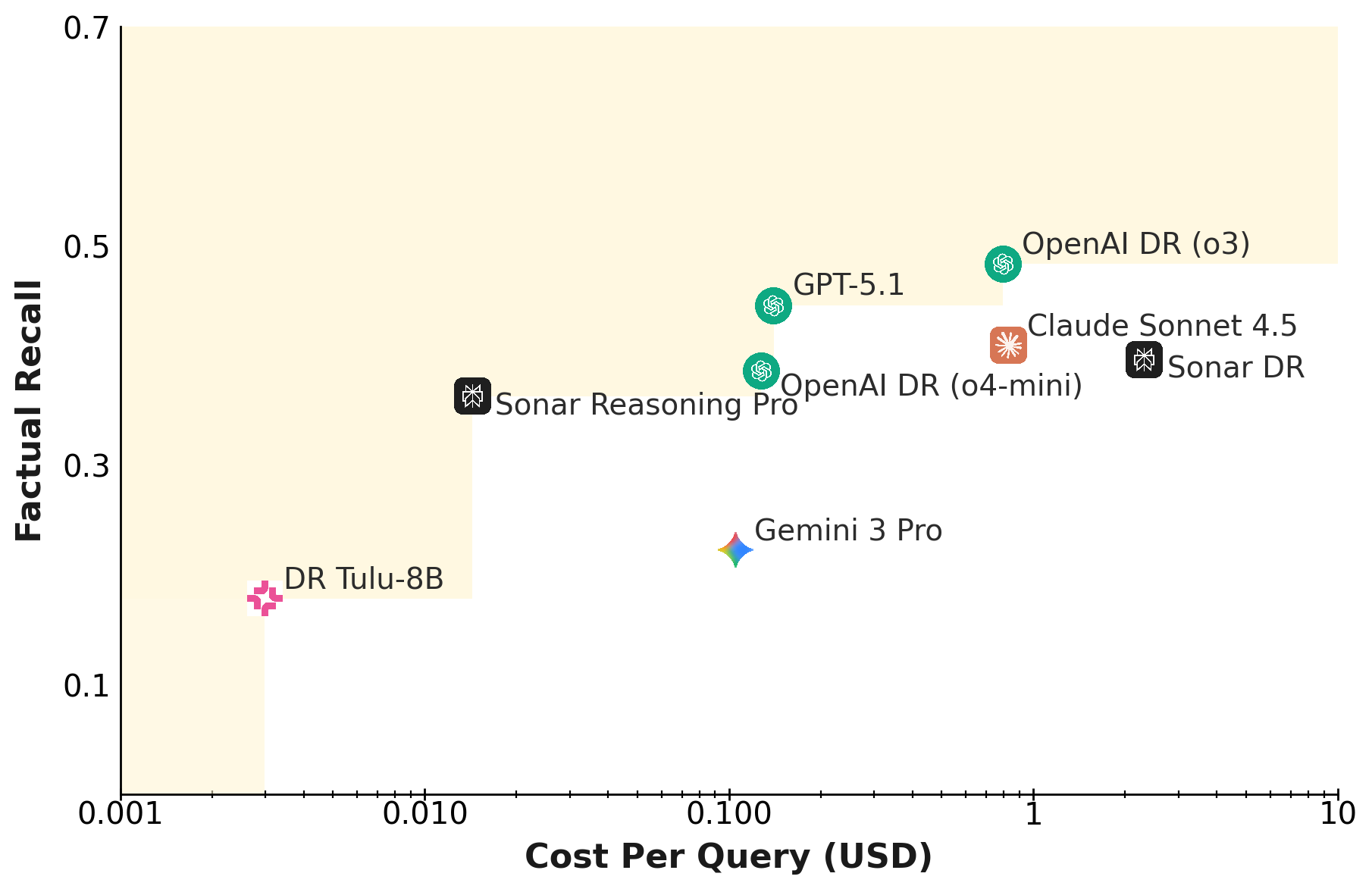}
\end{minipage}
\hfill
\begin{minipage}[t]{0.45\linewidth}
    \centering
    \includegraphics[width=\linewidth]{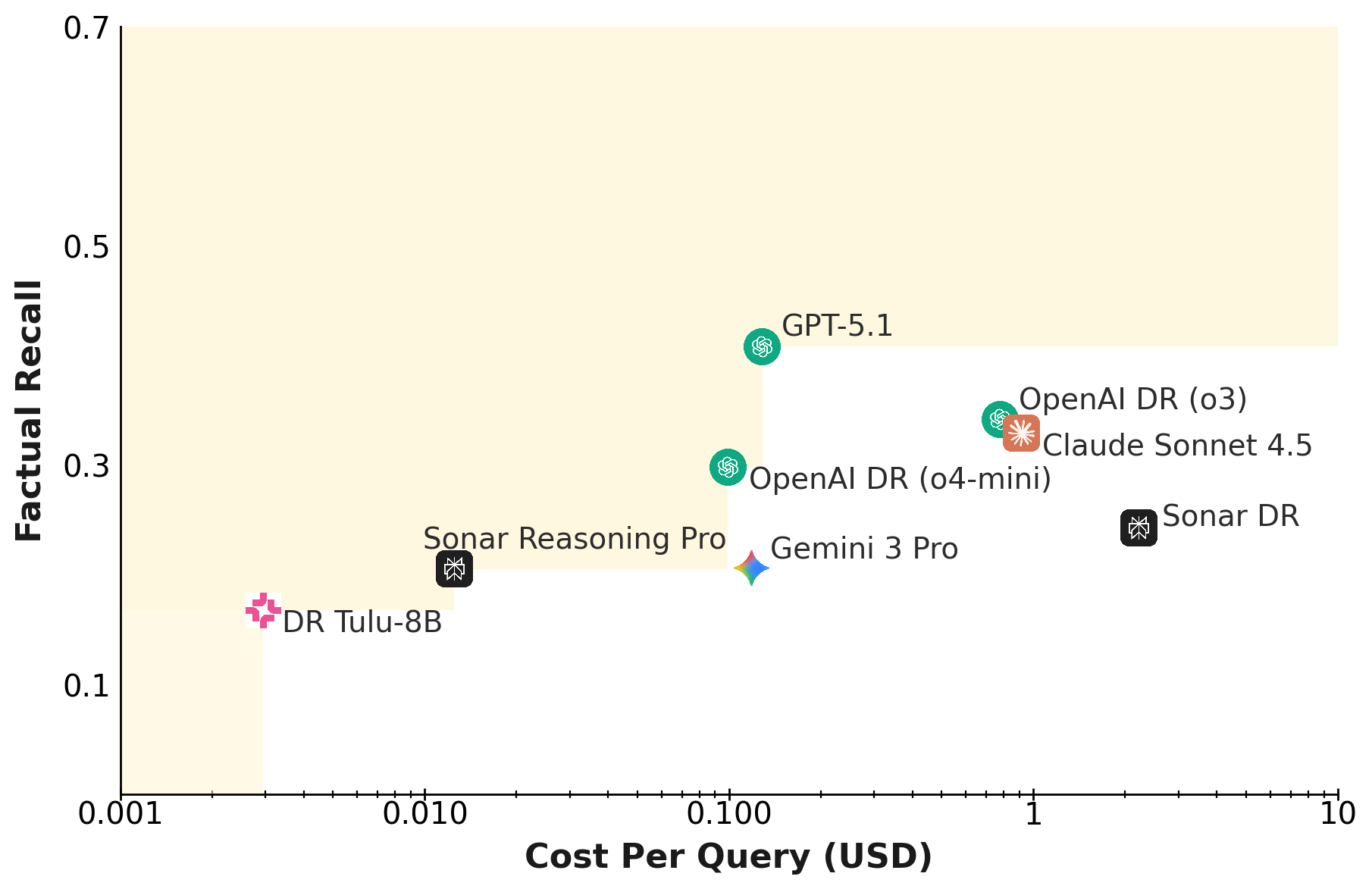}
\end{minipage}

\caption{Performance vs.\ cost of frontier models and deep research agents. We plot factual recall against cost (USD per query). Left: \textsc{SciConHarness} without clean-room constraints. Right: \textsc{SciConHarness} with clean-room evaluation.}
\label{fig:performance-cost-recall}
\end{figure}

\xhdr{Performance vs. Cost} Figures \ref{fig:performance-cost-F1}--\ref{fig:performance-cost-recall} show the Pareto frontier of performance vs.\ cost for models and deep research agents across factual F1, precision, and recall. Clean-room constraints consistently flatten the frontier, aligning with our earlier findings on performance attenuation. For factual F1, \texttt{DR Tulu}, \texttt{sonar-reasoning-pro}, \texttt{o4-mini-deep-research}, and \texttt{o3-deep-research} lie on the Pareto frontier, representing the most efficient trade-offs at different cost levels. \texttt{DR Tulu} is the most cost-efficient but lowest-performing point, while \texttt{o3-deep-research} achieves the highest performance at the greatest cost.

\begin{figure}[t]
\centering
\begin{minipage}[t]{0.45\linewidth}
    \centering
    \includegraphics[width=\linewidth]{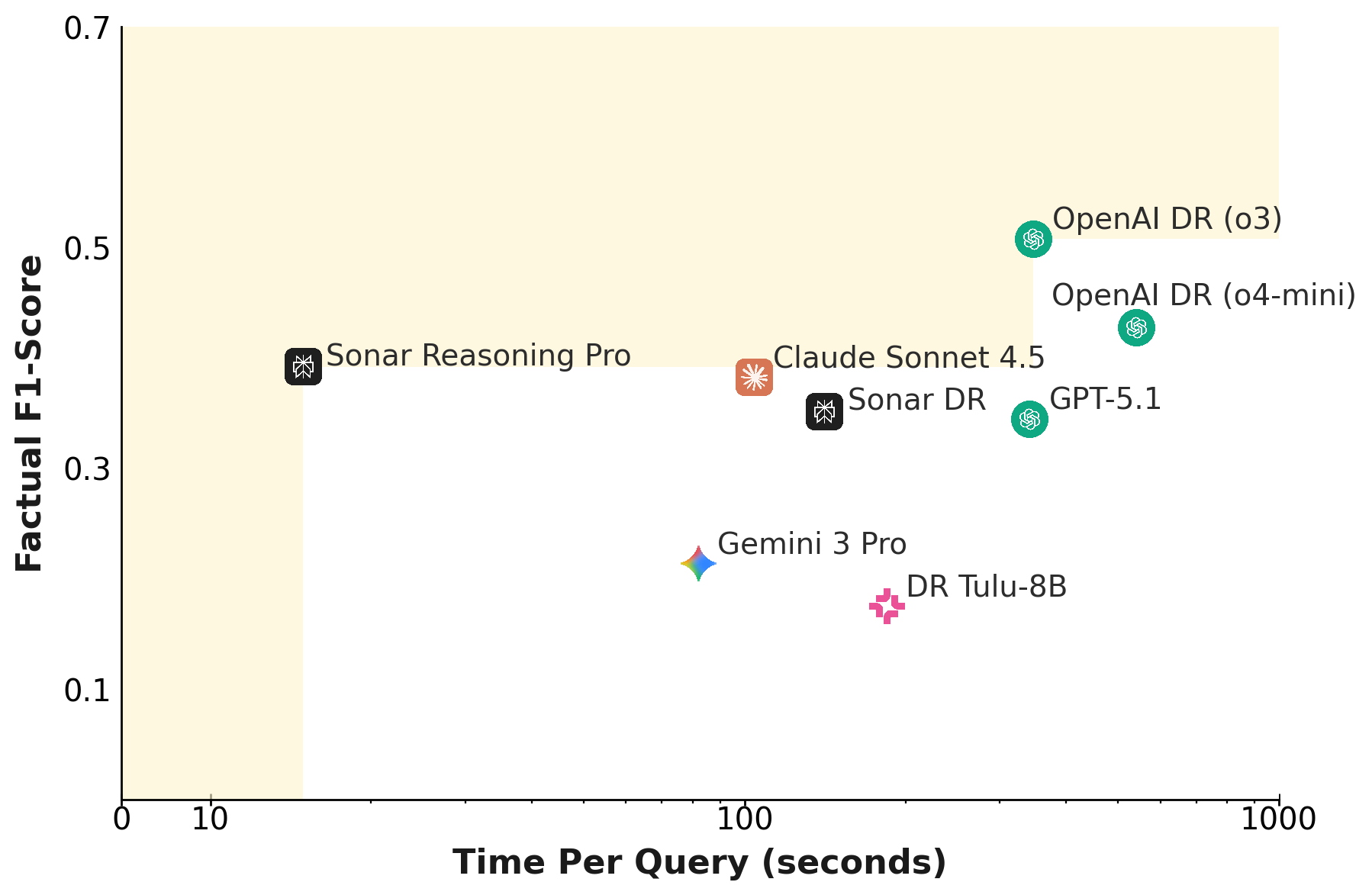}
\end{minipage}
\hfill
\begin{minipage}[t]{0.45\linewidth}
    \centering
    \includegraphics[width=\linewidth]{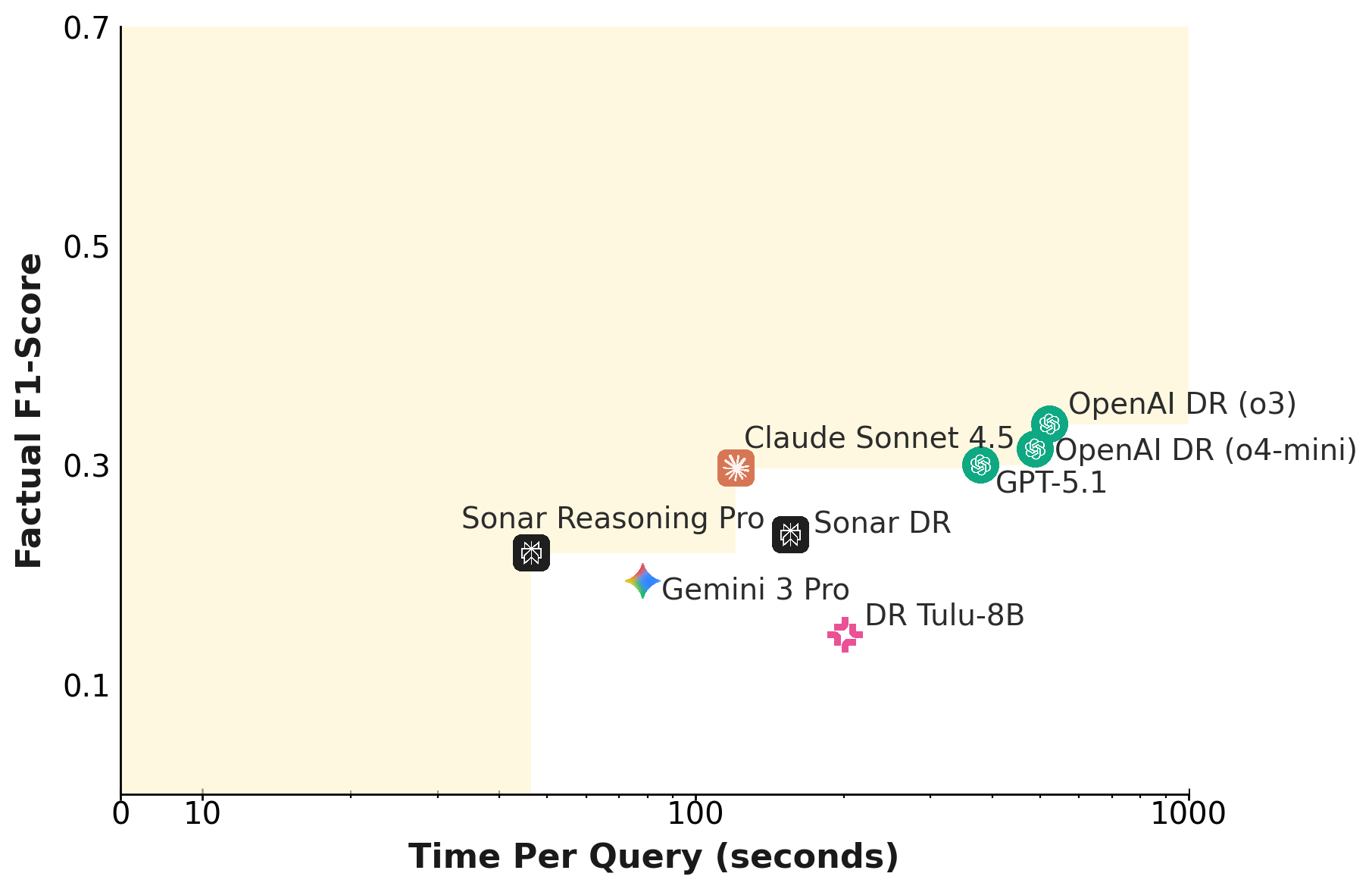}
\end{minipage}

\caption{Performance vs.\ time of frontier models and deep research agents. We plot factual F1 against time (seconds per query). Left: \textsc{SciConHarness} without clean-room constraints. Right: \textsc{SciConHarness} with clean-room evaluation.}
\label{fig:performance-time-F1}
\end{figure}

\begin{figure}[t]
\centering
\begin{minipage}[t]{0.45\linewidth}
    \centering
    \includegraphics[width=\linewidth]{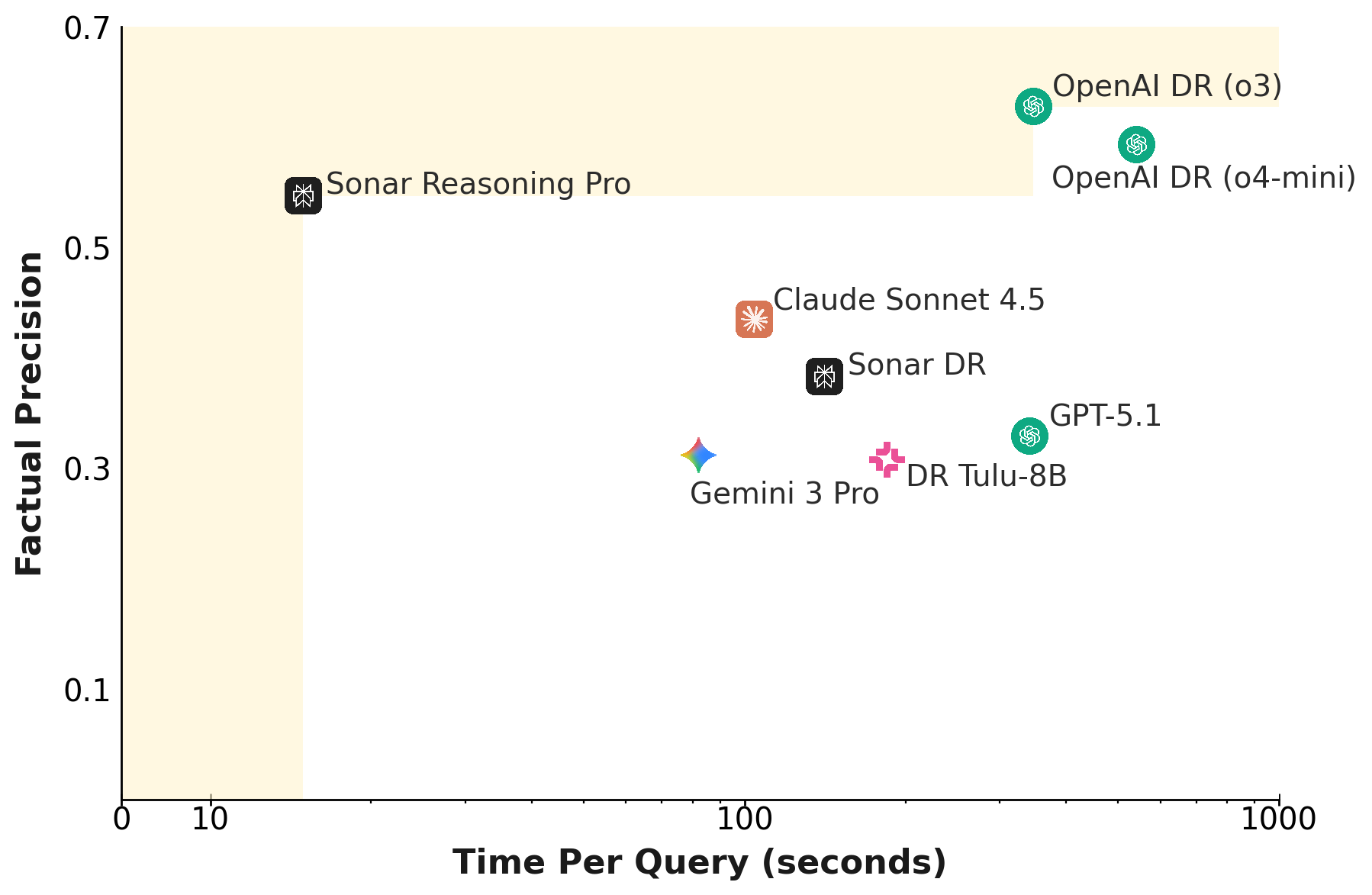}
\end{minipage}
\hfill
\begin{minipage}[t]{0.45\linewidth}
    \centering
    \includegraphics[width=\linewidth]{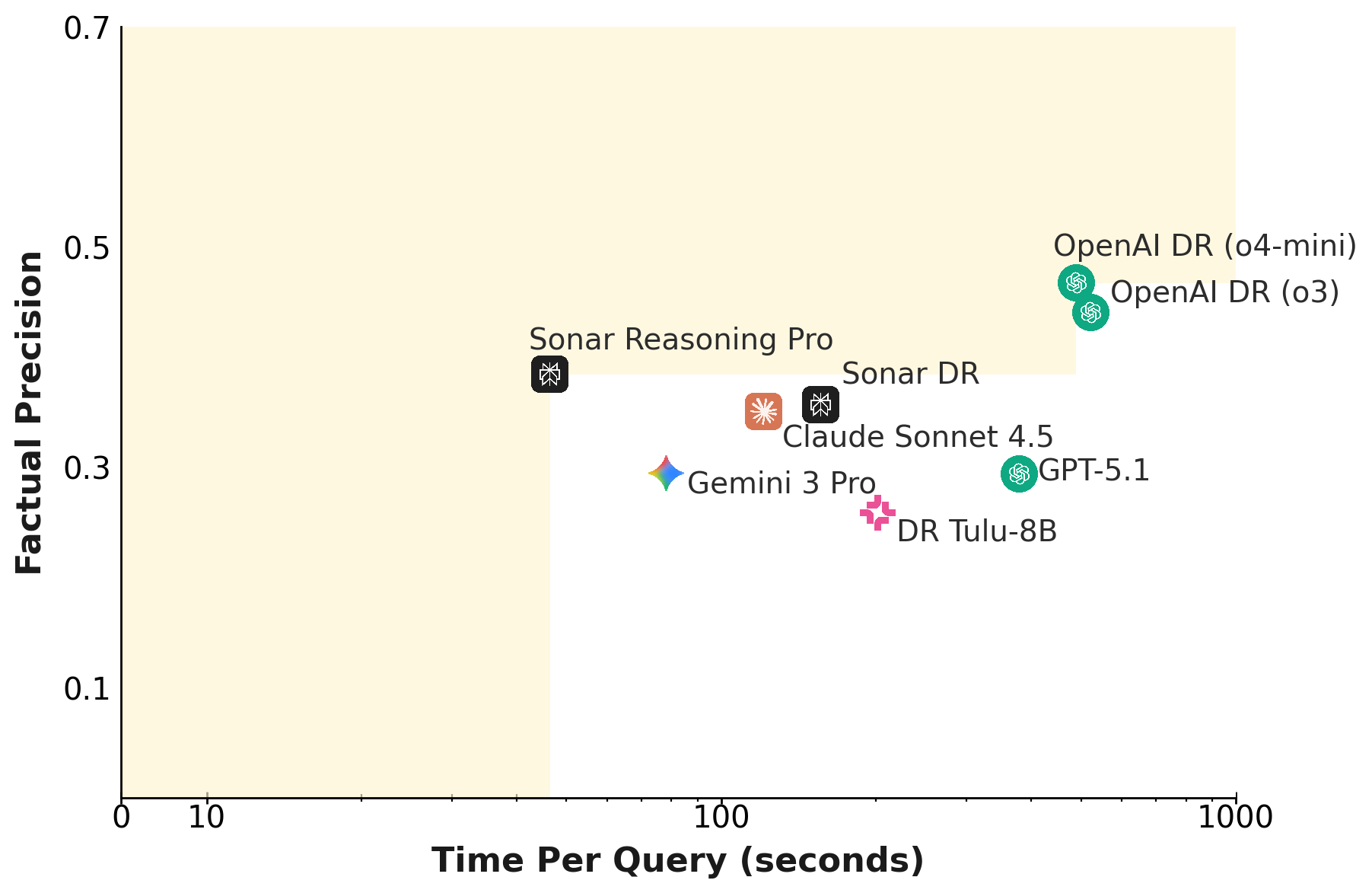}
\end{minipage}

\caption{Performance vs.\ time of frontier models and deep research agents. We plot factual precision against time (seconds per query). Left: \textsc{SciConHarness} without clean-room constraints. Right: \textsc{SciConHarness} with clean-room evaluation.}
\label{fig:performance-time-precision}
\end{figure}

\begin{figure}[t]
\centering
\begin{minipage}[t]{0.45\linewidth}
    \centering
    \includegraphics[width=\linewidth]{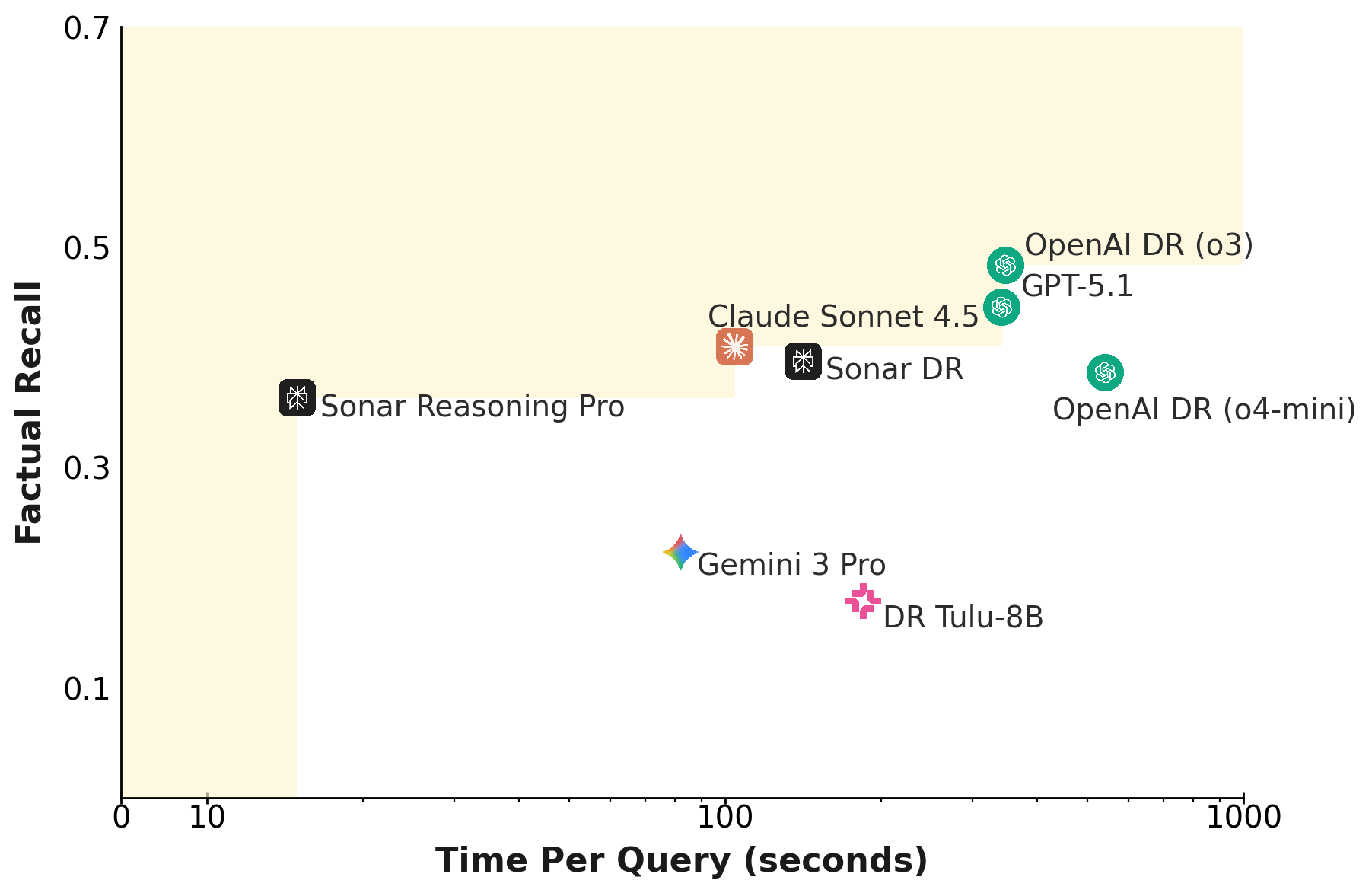}
\end{minipage}
\hfill
\begin{minipage}[t]{0.45\linewidth}
    \centering
    \includegraphics[width=\linewidth]{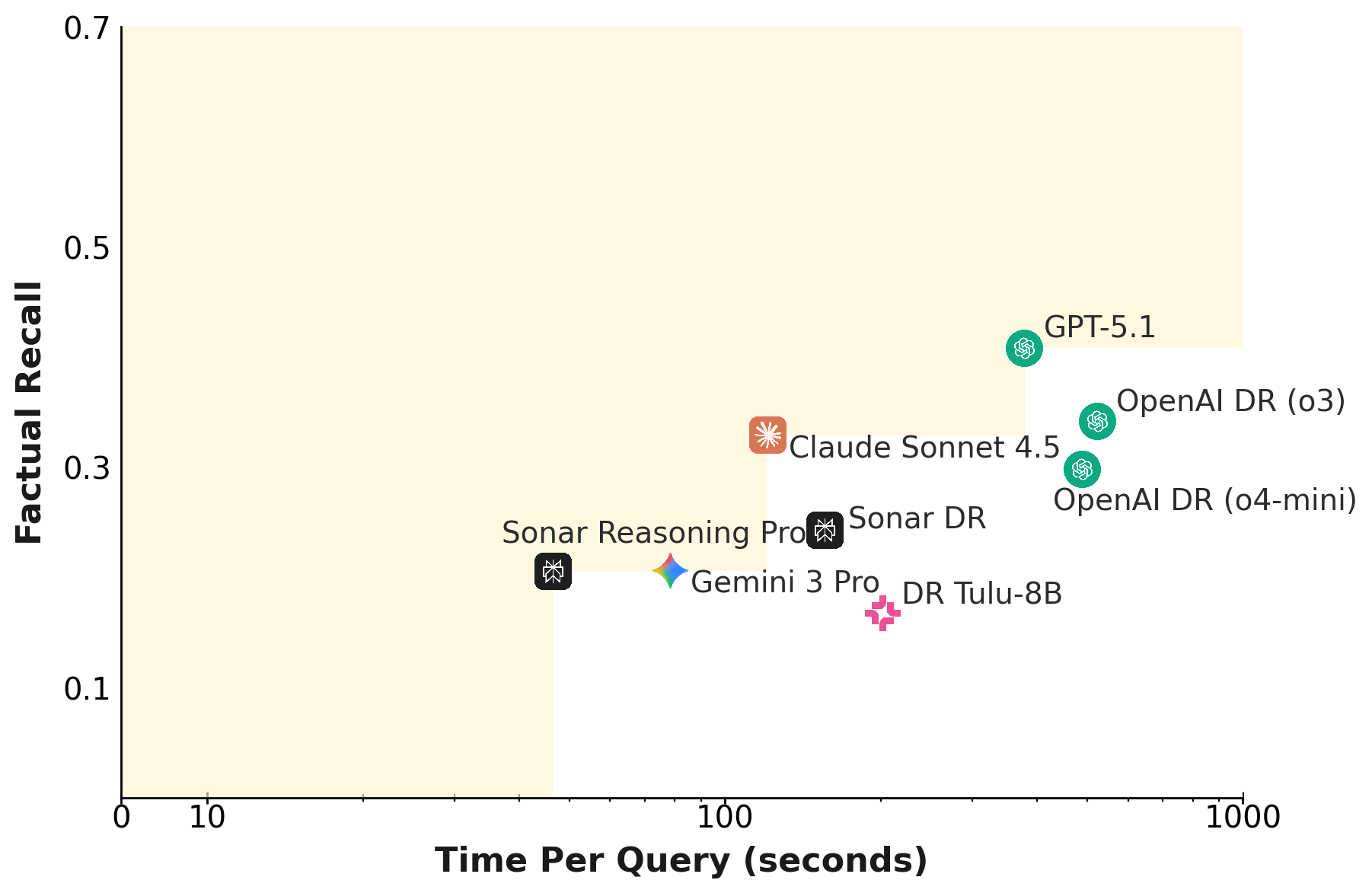}
\end{minipage}

\caption{Performance vs.\ time of frontier models and deep research agents. We plot factual recall against time (seconds per query). Left: \textsc{SciConHarness} without clean-room constraints. Right: \textsc{SciConHarness} with clean-room evaluation.}
\label{fig:performance-time-recall}
\end{figure}

\xhdr{Performance vs. Time} Figures \ref{fig:performance-time-F1}--\ref{fig:performance-time-recall} show the Pareto frontier of performance vs.\ time for models and deep research agents across factual F1, precision, and recall. Clean-room constraints flatten and shift the frontier rightward, indicating both attenuated performance and increased time to synthesize conclusions. This increase in latency suggests that agents are engaging in genuine synthesis rather than shortcut retrieval. Under clean-room evaluation, \texttt{sonar-reasoning-pro}, \texttt{claude-sonnet-4.5}, and \texttt{o3-deep-research} lie on the Pareto frontier, representing efficient trade-offs at different latency levels. \texttt{sonar-reasoning-pro} is the most time-efficient but lowest-performing, while \texttt{o3-deep-research} achieves the highest performance at the greatest time cost.

\clearpage

\begin{table*}[!t]
    \centering
    \small
    \caption{Label distribution for factual precision and recall across models and deep research agents (denoted as \textit{DR}). $\dagger$ denotes evaluation on a subset ($N=100$). Values are percentages.  ``Supp.'' = Supported, ``Not Supp.'' = Not Supported, and ``Contr.'' = Contradicted (precision only). Among benchmarked models and deep research agents (excluding consumer-facing agents), \textcolor{blue}{blue} highlights the highest supported proportion for precision and recall, while \textcolor{red}{red} highlights the highest contradicted proportion for precision and the highest not-supported proportion for recall.}
    \setlength{\tabcolsep}{5pt}
    \resizebox{\textwidth}{!}{
    \begin{tabular}{lcccc@{\hspace{10pt}}ccc}
    \toprule
    & \multicolumn{4}{c}{\textbf{Precision}} 
    & \multicolumn{3}{c}{\textbf{Recall}} \\
    \cmidrule(lr){2-5} \cmidrule(lr){6-8}
    & \# Facts & Supp. & Not Supp. & Contr. 
    & \# Facts & Supp. & Not Supp. \\
    \midrule
    
    \textcolor{gray}{\footnotesize \textit{Base Models (No Tools)}} \\
    \modelicon{openai_icon}~\texttt{gpt-5.1} 
    & 12709 & 36.9 & 59.1 & 3.9 
    & 2820 & 35.6 & 64.4 \\
    
    \modelicon{claude_icon}~\texttt{claude-sonnet-4.5} 
    & 7253 & 47.8 & 48.4 & 3.8 
    & 2820 & 24.3 & 75.7 \\
    
    \modelicon{gemini_icon}~\texttt{gemini-3-pro} 
    & 5009 & 35.8 & 56.5 & 7.7 
    & 2820 & 21.7 & 78.3 \\
    
    \midrule
    \textcolor{gray}{\footnotesize\textit{Models (\textsc{SciConHarness})}} \\
    
    \modelicon{openai_icon}~\texttt{gpt-5.1} 
    & 14378 & 33.4 & 62.3 & 4.3 
    & 2820 & 41.8 & 58.2 \\
    
    \modelicon{claude_icon}~\texttt{claude-sonnet-4.5} 
    & 9257 & 43.5 & 52.9 & 3.6 
    & 2820 & 37.7 & 62.3 \\
    
    \modelicon{gemini_icon}~\texttt{gemini-3-pro} 
    & 5200 & 32.3 & 61.0 & 6.7 
    & 2820 & 19.8 & 80.2 \\
    
    \modelicon{perplexity_icon}~\texttt{sonar-reasoning-pro} 
    & 4870 & 57.2 & 37.0 & 5.9 
    & 2820 & 33.3 & 66.7 \\
    
    \midrule
    \textcolor{gray}{\footnotesize\textit{Models (\textsc{SciConHarness} + clean-room)}} \\
    
    \modelicon{openai_icon}~\texttt{gpt-5.1} 
    & 14219 & 29.7 & 65.8 & 4.5 
    & 2820 & 38.2 & 61.8 \\
    
    \modelicon{claude_icon}~\texttt{claude-sonnet-4.5} 
    & 9481 & 36.1 & 59.5 & 4.5 
    & 2820 & 30.7 & 69.3 \\
    
    \modelicon{gemini_icon}~\texttt{gemini-3-pro} 
    & 5298 & 30.1 & 63.6 & 6.3 
    & 2820 & 18.6 & 81.4 \\
    
    \modelicon{perplexity_icon}~\texttt{sonar-reasoning-pro} 
    & 4606 & 40.8 & 50.8 & \negative{8.4}
    & 2820 & 17.9 & 82.1 \\
    
    \midrule
    \textcolor{gray}{\footnotesize\textit{DR (\textsc{SciConHarness})}} \\
    
    \modelicon{ai2_icon}~\texttt{DR Tulu} 
    & 5714 & 29.0 & 66.7 & 4.3 
    & 2820 & 16.4 & 83.6 \\
    
    \modelicon{perplexity_icon}~\texttt{sonar-deep-research$\dagger$} 
    & 3005 & 38.7 & 57.6 & 3.8 
    & 1075 & 36.0 & 64.0 \\
    
    \modelicon{openai_icon}~\texttt{o4-mini-deep-research} 
    & 6872 & 60.4 & 35.2 & 4.4 
    & 2820 & 35.2 & 64.8 \\
    
    \modelicon{openai_icon}~\texttt{o3-deep-research} 
    & 9174 & \positive{60.9} & 35.5 & 3.6 
    & 2820 & \positive{44.6} & 55.4 \\
    
    \midrule
    \textcolor{gray}{\footnotesize\textit{DR (\textsc{SciConHarness} + clean-room)}} \\
    
    \modelicon{ai2_icon}~\texttt{DR Tulu} 
    & 4209 & 24.9 & 69.8 & 5.3 
    & 2820 & 15.3 & \negative{84.7} \\
    
    \modelicon{perplexity_icon}~\texttt{sonar-deep-research$\dagger$} 
    & 2855 & 34.0 & 62.1 & 3.9 
    & 1075 & 23.9 & 76.1 \\
    
    \modelicon{openai_icon}~\texttt{o4-mini-deep-research} 
    & 6856 & 48.7 & 44.4 & 6.9 
    & 2820 & 27.6 & 72.4 \\
    
    \modelicon{openai_icon}~\texttt{o3-deep-research} 
    & 8006 & 45.0 & 48.7 & 6.3 
    & 2820 & 31.5 & 68.5 \\
    
    \midrule
    \textcolor{gray}{\footnotesize\textit{Consumer-Facing Agents}} \\
    Google AI Overview
    & 4404 & 52.4 & 42.1 & 5.5
    & 2749 & 33.6 & 66.4 \\
    
    Google AI Mode
    & 4661 & 46.8 & 46.9 & 6.2
    & 2820 & 35.2 & 64.8\\
    
    OpenEvidence
    & 7584 & 59.8 & 37.2 & 3.0
    & 2686 & 51.7 & 48.3\\
    
    \bottomrule
    \end{tabular}}
    \label{tab:label-distribution}
    \end{table*}

\subsection{Label Distribution}\label{appendix:label-distribution} Table \ref{tab:label-distribution} shows the full label distribution for factual precision and recall across models and deep research agents. In terms of precision, base models generate relatively low proportions of facts supported by \texttt{CDSR} reviews (e.g., \texttt{gpt-5.1}: 36.9\%, \texttt{gemini-3-pro}: 35.8\%) with non-trivial contradiction rates (3.8\%-7.7\%). With the exception of \texttt{gemini-3-pro}, tool augmentation via \textsc{SciConHarness} generally improves recall (e.g., \texttt{gpt-5.1}: 35.6\%$\rightarrow$41.8\% reference facts supported by generated conclusions) but decreases precision by increasing unsupported generations (e.g., \texttt{gpt-5.1}: 59.1\%$\rightarrow$62.3\% generated facts not supported by the reference \texttt{CDSR} review). Under clean-room evaluation constraints, supported rates consistently decline for both precision and recall, while contradiction rates increase (except for \texttt{gemini-3-pro}), revealing degraded performance when models must genuinely synthesize rather than retrieve ground-truth reviews.

Compared to frontier models, deep research agents achieve higher supported rates (e.g., \texttt{o3-deep-research}: 60.9\%) but remain sensitive to clean-room evaluation. Under the clean-room, \texttt{o3-deep-research} shows sharp drops in generated facts supported by reference \texttt{CDSR} reviews (60.9\%$\rightarrow$45.0\%), a decrease in reference facts supported by the generated conclusions (44.6\%$\rightarrow$31.5\%), increases in contradiction rates (3.6\%$\rightarrow$6.3\%). Notably, consumer-facing agents like OpenEvidence achieve the strongest balance in correctness and coverage in generated conclusions (59.8\% precision-supported; 51.7\% recall-supported), suggesting more effective evidence integration. Overall, these trends reinforce that current systems---even without clean-room constraints---have substantial room to improve factual precision and recall.

\begin{table*}[!t]
\centering
\footnotesize
\caption{Percentage of generated conclusions containing at least one fact that contradicts ($\geq \textbf{1 Contr.}$) and is not supported ($\geq \textbf{1 Not Supp.}$) with respect to \texttt{CDSR} reviews across models and deep research agents (denoted as \textit{DR}). $\dagger$ denotes evaluation on a subset ($N=100$). Excluding consumer-facing agents, \textcolor{blue}{blue} highlights the lowest percentage in each column, while \textcolor{red}{red} highlights the highest percentage.}
\setlength{\tabcolsep}{6pt}
\begin{tabular}{lcc}
\toprule
\textbf{Model} 
& $\mathbf{\geq 1}$ \textbf{Contr.} 
& $\mathbf{\geq 1}$ \textbf{Not Supp.} \\
\midrule

\textcolor{gray}{\footnotesize \textit{Base Models (No Tools)}} \\
\modelicon{openai_icon}~\texttt{gpt-5.1} 
& 77.2 & \negative{100.0} \\

\modelicon{claude_icon}~\texttt{claude-sonnet-4.5} 
& 56.7 & \negative{100.0} \\

\modelicon{gemini_icon}~\texttt{gemini-3-pro} 
& 70.1 & \negative{100.0} \\

\midrule
\textcolor{gray}{\footnotesize\textit{Models (\textsc{SciConHarness})}} \\

\modelicon{openai_icon}~\texttt{gpt-5.1} 
& \negative{84.0} & \negative{100.0} \\

\modelicon{claude_icon}~\texttt{claude-sonnet-4.5} 
& 60.8 & 99.3 \\

\modelicon{gemini_icon}~\texttt{gemini-3-pro} 
& 62.7 & 99.3 \\

\modelicon{perplexity_icon}~\texttt{sonar-reasoning-pro} 
& 56.7 & \positive{94.0} \\

\midrule
\textcolor{gray}{\footnotesize\textit{Models (\textsc{SciConHarness} + clean-room)}} \\

\modelicon{openai_icon}~\texttt{gpt-5.1} 
& 80.6 & \negative{100.0} \\

\modelicon{claude_icon}~\texttt{claude-sonnet-4.5} 
& 73.9 & \negative{100.0} \\

\modelicon{gemini_icon}~\texttt{gemini-3-pro} 
& 61.2 & 99.3 \\

\modelicon{perplexity_icon}~\texttt{sonar-reasoning-pro} 
& 70.9 & 97.4 \\

\midrule
\textcolor{gray}{\footnotesize\textit{DR (\textsc{SciConHarness})}} \\

\modelicon{ai2_icon}~\texttt{DR Tulu} 
& 47.8 & 98.9 \\

\modelicon{perplexity_icon}~\texttt{sonar-deep-research$\dagger$} 
& 61.0 & \negative{100.0} \\

\modelicon{openai_icon}~\texttt{o4-mini-deep-research} 
& 61.9 & 97.4 \\

\modelicon{openai_icon}~\texttt{o3-deep-research} 
& 55.2 & 95.9 \\

\midrule
\textcolor{gray}{\footnotesize\textit{DR (\textsc{SciConHarness} + clean-room)}} \\

\modelicon{ai2_icon}~\texttt{DR Tulu} 
& \positive{44.8} & 99.3 \\

\modelicon{perplexity_icon}~\texttt{sonar-deep-research$\dagger$} 
& 62.0 & \negative{100.0} \\

\modelicon{openai_icon}~\texttt{o4-mini-deep-research} 
& 75.4 & 98.5 \\

\modelicon{openai_icon}~\texttt{o3-deep-research} 
& 73.1 & 99.6 \\

\midrule
\textcolor{gray}{\footnotesize\textit{Consumer-Facing Agents}} \\

Google AI Overview
& 56.3 & 98.9 \\

Google AI Mode
& 59.0 & 99.6 \\

OpenEvidence
& 50.8 & 100.0 \\

\bottomrule
\end{tabular}
\label{tab:response-level-precision}
\end{table*}

Table~\ref{tab:response-level-precision} reports the percentage of generated conclusions containing at least one contradictory or unsupported fact with respect to \texttt{CDSR} reviews. Across all evaluated agents, contradictory facts were common: \texttt{DR Tulu} under the clean-room condition, produced at least one contradiction in 44.8\% of conclusions, while several systems exceeded 70\%, including \texttt{gpt-5.1} under \textsc{SciConHarness} (84.0\%) and under the clean-room setting (80.6\%). Facts not supported by \texttt{CDSR} reviews were even more pervasive, appearing in 94.0--100.0\% of generated conclusions across models, deep research agents, and consumer-facing agents. These findings suggest that current AI agents frequently synthesize conclusions that mix supported facts with unsupported or contradictory facts. This has potential implications in clinical and scientific contexts where users may rely on such agents to interpret evidence and inform downstream decisions.

\subsection{Impact of Conclusion Length on Factual Precision and Recall}\label{appendix:length-metric-tradeoff}

\begin{figure}[!t]
\centering
\includegraphics[width=0.85\columnwidth]{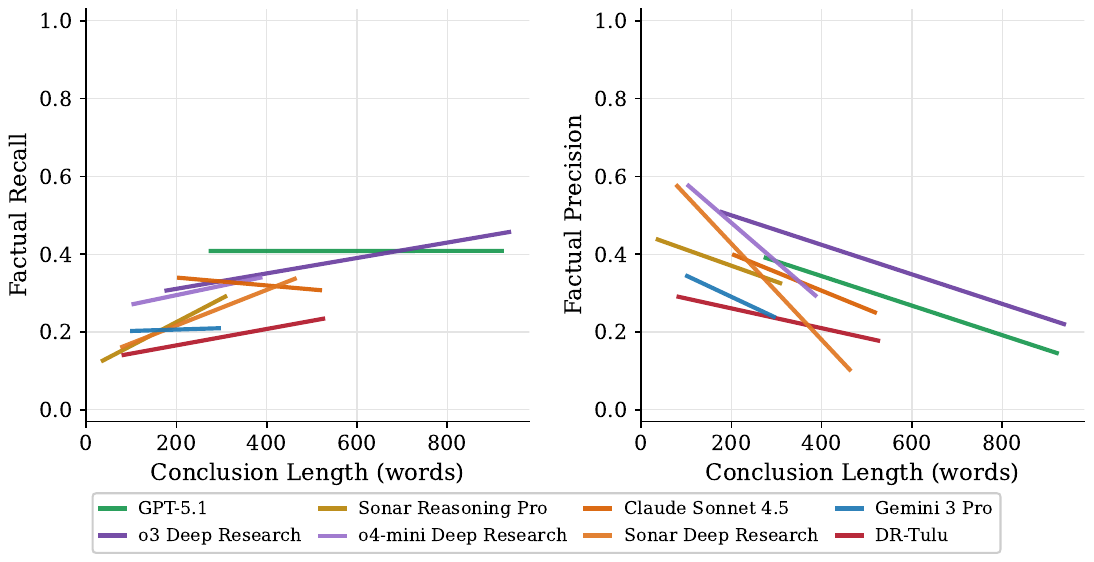}
\caption{Relationship between conclusion length in words vs. factual precision and recall across models and deep research agents.}
\label{fig:length-metric-tradeoff}
\end{figure}

Figure \ref{fig:length-metric-tradeoff} shows the relationship between conclusion length and factual precision/recall. Longer conclusions consistently exhibit lower precision, indicating that additional generated facts are less likely to be supported by the reference \texttt{CDSR} reviews, aligning with prior works \cite{asai2023self, liu-etal-2023-evaluating}. At the same time, recall generally increases with length, reflecting improved coverage of the reference conclusions. This trade-off highlights that longer outputs do not necessarily improve factual F1, the harmonic mean of both factual precision and recall. Importantly, this suggests that the factual F1 is robust to conclusion length: increases in length are penalized through reduced precision, preventing models and agents from inflating performance by generating longer conclusions. These results suggest that model and agent quality matter more than the length of conclusions to obtain a high score on \textsc{SciConBench}.

\newpage

\subsection{\camready{Stratifying Benchmark Performance by Clinical Domain and Review Type}}\label{appendix:stratification}
\camready{To test whether performance varies systematically across the benchmark sample ($N=268$), we stratify factual F1 by clinical domain and \textsc{CDSR} review types. For each system and \textsc{SciConHarness} setting, we conduct a one-way ANOVA across clinical domains and a Welch's $t$-test across review types, with Bonferroni correction, reporting $\eta^2$ and Cohen's $d$ as effect sizes, respectively. $\eta^2$ denotes the proportion of a system's F1 variance explained by clinical domain. Following \citet{lakens2013calculating}, we interpret effects as small ($\eta^2<0.06$), medium ($0.06\le\eta^2<0.14$), and large ($\eta^2\ge0.14$). Table~\ref{tab:stratification} reports the distribution of clinical domains (left) and $\eta^2$ for each system (right).}

\begin{table}[!t]
\centering
\scriptsize
\setlength{\tabcolsep}{4pt}
\caption{\camready{\textbf{Left:} Distribution of clinical domains in the benchmark sample ($N=268$). \textbf{Right:} Proportion of factual F1 variance explained by clinical domain ($\eta^2$, one-way ANOVA) for each system and setting. \textbf{Bold} denotes medium or large effects ($\eta^2\ge0.06$). $\dagger$ indicates evaluation on a subset ($N=100$).}}
\label{tab:stratification}
\begin{minipage}[t]{0.50\linewidth}
\centering
\vspace{0pt}
\renewcommand{\arraystretch}{2.32}
\begin{tabular}{lrr}
\toprule
\textbf{Clinical Domain} & \textbf{Count} & \textbf{\%} \\
\midrule
Public Health \& Epidemiology & 49 & 18.3 \\
Basic \& Diagnostic Sciences & 45 & 16.8 \\
Respiratory \& Infectious Disease & 31 & 11.6 \\
Reproductive, Perinatal \& Pediatric Health & 27 & 10.1 \\
Cardiometabolic \& Renal Medicine & 26 & 9.7 \\
Mental Health \& Neuroscience & 21 & 7.8 \\
Surgery \& Critical Care & 20 & 7.5 \\
Musculoskeletal \& Rehabilitation & 15 & 5.6 \\
Sensory Medicine (ENT \& Ophthalmology) & 15 & 5.6 \\
Nutrition, GI \& Lifestyle Medicine & 10 & 3.7 \\
Oncology \& Cancer Research & 9 & 3.4 \\
\midrule
\textbf{Total} & 268 & 100.0 \\
\bottomrule
\end{tabular}
\end{minipage}%
\hfill
\begin{minipage}[t]{0.46\linewidth}
\centering
\vspace{0pt}
\begin{tabular}{lr}
\toprule
 & \textbf{$\eta^2$ (Domain)} \\
\midrule
\textcolor{gray}{\textit{Base Models (No Tools)}} \\
\modelicon{openai_icon}~\texttt{gpt-5.1} & 0.024 \\
\modelicon{claude_icon}~\texttt{claude-sonnet-4.5} & 0.039 \\
\modelicon{gemini_icon}~\texttt{gemini-3-pro} & 0.014 \\
\midrule
\textcolor{gray}{\textit{Models (\textsc{SciConHarness})}} \\
\modelicon{openai_icon}~\texttt{gpt-5.1} & 0.030 \\
\modelicon{claude_icon}~\texttt{claude-sonnet-4.5} & \textbf{0.108} \\
\modelicon{gemini_icon}~\texttt{gemini-3-pro} & 0.028 \\
\modelicon{perplexity_icon}~\texttt{sonar-reasoning-pro} & 0.021 \\
\midrule
\textcolor{gray}{\textit{Models (\textsc{SciConHarness} + clean-room)}} \\
\modelicon{openai_icon}~\texttt{gpt-5.1} & 0.021 \\
\modelicon{claude_icon}~\texttt{claude-sonnet-4.5} & 0.039 \\
\modelicon{gemini_icon}~\texttt{gemini-3-pro} & 0.019 \\
\modelicon{perplexity_icon}~\texttt{sonar-reasoning-pro} & 0.025 \\
\midrule
\textcolor{gray}{\textit{DR (\textsc{SciConHarness})}} \\
\modelicon{ai2_icon}~\texttt{DR Tulu} & 0.028 \\
\modelicon{perplexity_icon}~\texttt{sonar-deep-research$\dagger$} & \textbf{0.166} \\
\modelicon{openai_icon}~\texttt{o4-mini-deep-research} & 0.055 \\
\modelicon{openai_icon}~\texttt{o3-deep-research} & 0.042 \\
\midrule
\textcolor{gray}{\textit{DR (\textsc{SciConHarness} + clean-room)}} \\
\modelicon{ai2_icon}~\texttt{DR Tulu} & 0.033 \\
\modelicon{perplexity_icon}~\texttt{sonar-deep-research$\dagger$} & \textbf{0.124} \\
\modelicon{openai_icon}~\texttt{o4-mini-deep-research} & 0.027 \\
\modelicon{openai_icon}~\texttt{o3-deep-research} & 0.023 \\
\midrule
\textcolor{gray}{\textit{Consumer-Facing Agents}} \\
Google AI Overview & 0.047 \\
Google AI Mode & 0.044 \\
OpenEvidence & \textbf{0.081} \\
\bottomrule
\end{tabular}
\end{minipage}

\end{table}

\xhdr{\camready{Low performance is consistent across clinical domains}} \camready{Clinical-domain effects are generally small: 18 of 22 system configurations have $\eta^2<0.06$, with medium-to-large effects limited to \texttt{claude-sonnet-4.5} with tools, \texttt{sonar-deep-research} (with and without clean-room), and OpenEvidence. Thus, low factual F1 is not driven by particular domains, although some systems are more domain-sensitive than others. For example, \texttt{sonar-deep-research} without clean-room ($\eta^2=0.166$) ranges from approximately 0.15 to 0.50 F1 across domains, whereas base \texttt{gemini-3-pro} ($\eta^2=0.014$) ranges only from approximately 0.21 to 0.28.}

\xhdr{\camready{Intervention reviews are harder to synthesize}} \camready{Most samples (238 of 268; 88.8\%) are intervention reviews; we pool the remaining 30 prognostic, qualitative, and other reviews into a \textit{non-intervention} category to retain statistical power. Every system except \texttt{DR Tulu} performs worse on intervention reviews, with Cohen's $d$ ranging from $-0.33$ to $-0.81$, intervention vs.\ non-intervention). This consistent pattern suggests that intervention reviews are harder to synthesize, as they require integrating treatment-effect estimates (e.g., risk ratios, mean differences) and reasoning across multiple randomized controlled trials.}

\clearpage

\subsection{\textsc{SciConHarness} Tool Usage Patterns}\label{appendix:tool-usage}

Table \ref{tab:tool-calls-detailed} shows the average number of tool calls across \textsc{SciConHarness} tools, along with the percentage filtered under clean-room evaluation. Tool selection and usage vary substantially across systems. OpenAI models and agents (e.g., \texttt{gpt-5.1}, \texttt{o3-deep-research}, \texttt{o4-mini-deep-research}) make the heaviest use of tools overall, particularly \texttt{web\_browse} and \texttt{google\_search}, often issuing the highest total number of calls per query. In contrast, \texttt{claude-sonnet-4.5} uses tools more moderately and relies more heavily on \texttt{paper\_search} compared to \texttt{google\_search} and \texttt{web\_browse}---a pattern also observed for \texttt{gemini-3-pro}, which uses tools sparingly. \texttt{DR Tulu} calls the fewest tools overall, though it is smaller scale (base \texttt{Qwen3-8B}) and has a limited context window (32{,}768 tokens), which may constrain extensive tool interaction. Across systems that use \texttt{google\_search} (excluding \texttt{DR Tulu}), \texttt{google\_search} exhibits the highest clean-room filtering rates (49.6\%--81.8\%; \S\ref{appendix:tool-usage}), indicating frequent retrieval of ground-truth artifacts. For instance, up to 81.8\% of \texttt{google\_search} calls for \texttt{claude-sonnet-4.5} were filtered. Meanwhile, \texttt{web\_browse} and \texttt{paper\_search} also exhibit non-trivial filtering rates, with \texttt{web\_browse} reaching up to 7.3\% and \texttt{paper\_search} up to 11.9\% across systems. This highlights the susceptibility of web search and browsing tools to benchmark leakage.

\begin{table}[t]
\centering
\caption{For each model and deep research agent, we show the average number of tool calls per query (mean $\pm$ std) across \textsc{SciConHarness} tools: \texttt{web\_browse}, \texttt{paper\_search}, \texttt{google\_search}, and the total tool calls. We also report the percentage of the tool calls filtered by the clean-room evaluation protocol.}
\setlength{\tabcolsep}{4pt}
\resizebox{\columnwidth}{!}{
\begin{tabular}{lcccccccc}
\toprule
& \multicolumn{2}{c}{\texttt{web\_browse}} 
& \multicolumn{2}{c}{\texttt{paper\_search}} 
& \multicolumn{2}{c}{\texttt{google\_search}} 
& \multicolumn{2}{c}{\textbf{Total}} \\
\cmidrule(lr){2-3} \cmidrule(lr){4-5} \cmidrule(lr){6-7} \cmidrule(lr){8-9}
& Avg \# & \% Filtered
& Avg \# & \% Filtered
& Avg \# & \% Filtered
& Avg \# & \% Filtered\\
\midrule

\textcolor{gray}{\footnotesize \textit{Models (\textsc{SciConHarness})}} \\

\modelicon{openai_icon}~\texttt{gpt-5.1} 
& $8.38\pm2.98$ & -- 
& $0.52\pm1.02$ & -- 
& $5.15\pm2.86$ & -- 
& $14.05\pm5.00$ & -- \\

\modelicon{claude_icon}~\texttt{claude-sonnet-4.5} 
& $1.47\pm1.46$ & -- 
& $6.81\pm1.61$ & -- 
& $0.69\pm0.92$ & -- 
& $8.97\pm2.52$ & -- \\

\modelicon{gemini_icon}~\texttt{gemini-3-pro} 
& $0.13\pm0.50$ & -- 
& $4.65\pm2.37$ & -- 
& $0.82\pm1.25$ & -- 
& $5.59\pm2.52$ & -- \\

\midrule
\textcolor{gray}{\footnotesize \textit{Models (\textsc{SciConHarness} + clean-room)}} \\

\modelicon{openai_icon}~\texttt{gpt-5.1} 
& $7.91\pm3.33$ & 1.1 
& $0.55\pm0.94$ & 5.5 
& $4.65\pm2.73$ & 56.9 
& $13.10\pm5.34$ & 21.1 \\

\modelicon{claude_icon}~\texttt{claude-sonnet-4.5} 
& $1.55\pm1.60$ & 3.1 
& $7.23\pm1.70$ & 11.9 
& $1.02\pm1.54$ & 81.8 
& $9.81\pm3.59$ & 17.8 \\

\modelicon{gemini_icon}~\texttt{gemini-3-pro} 
& $0.18\pm0.59$ & 6.3 
& $4.55\pm2.08$ & 9.8 
& $0.90\pm1.44$ & 78.8 
& $5.62\pm2.46$ & 20.7 \\

\midrule
\textcolor{gray}{\footnotesize \textit{DR (\textsc{SciConHarness})}} \\

\modelicon{ai2_icon}~\texttt{DR Tulu} 
& $0.02\pm0.19$ & -- 
& $5.48\pm2.85$ & -- 
& $0.21\pm1.11$ & -- 
& $5.71\pm2.82$ & -- \\

\modelicon{openai_icon}~\texttt{o4-mini-deep-research} 
& $6.78\pm2.59$ & -- 
& $2.69\pm2.15$ & -- 
& $7.57\pm3.45$ & -- 
& $17.03\pm6.34$ & -- \\

\modelicon{openai_icon}~\texttt{o3-deep-research} 
& $6.54\pm2.44$ & -- 
& $1.10\pm1.05$ & -- 
& $3.78\pm2.47$ & -- 
& $11.42\pm4.75$ & -- \\

\midrule
\textcolor{gray}{\footnotesize \textit{DR (\textsc{SciConHarness} + clean-room)}} \\

\modelicon{ai2_icon}~\texttt{DR Tulu} 
& $0.02\pm0.22$ & 0 
& $5.09\pm2.60$ & 6.6 
& $0.34\pm1.69$ & 0 
& $5.44\pm2.79$ & 6.2 \\

\modelicon{openai_icon}~\texttt{o4-mini-deep-research} 
& $6.71\pm2.55$ & 7.3 
& $2.87\pm2.06$ & 0 
& $9.08\pm3.91$ & 59.2 
& $18.66\pm6.32$ & 31.5 \\

\modelicon{openai_icon}~\texttt{o3-deep-research} 
& $7.85\pm2.83$ & 6.1 
& $1.84\pm1.46$ & 0 
& $5.23\pm2.66$ & 49.6 
& $14.93\pm5.42$ & 20.6 \\

\bottomrule
\end{tabular}}
\label{tab:tool-calls-detailed}
\end{table}

\clearpage

\subsection{Failure Mode Analysis}\label{appendix:failure-mode}

To better understand failure modes in scientific conclusion synthesis, we conduct a small-scale manual analysis of generated conclusions. Following prior work \cite{asai2023self, min-etal-2023-factscore, kaur-etal-2024-evaluating}, we sample $N=30$ generated conclusions and analyze why agents produce contradictory or incomplete conclusions that contribute to lower factual precision and recall. While not exhaustive, common failure modes are:

\xhdr{Failure Mode \#1: Incorrect Direction-of-Effect} 
One particularly concerning failure mode is that agents misrepresent the direction of treatment effects relative to the \texttt{CDSR} reviews. Agents sometimes invert the core conclusion of the evidence itself: presenting null or uncertain findings as clinically beneficial or harmful, or reversing positive and negative effects. In evidence-based medicine, the direction of effect is often the central conclusion used to guide downstream clinical interpretation and decision-making \cite{oxman2022key}. As a result, these errors are especially consequential because they can transform a cautious or beneficial finding into a harmful one (or vice versa) while still appearing fluent and evidence-grounded. Consider the following example:

\begin{figure*}[h!]
\centering
\begin{small}
\begin{tcolorbox}[colback=gray!5, colframe=red, width=\textwidth]
\textsc{Generated Conclusion} (\texttt{o3-deep-research}, tools): \textcolor{purple}{Large trials (e.g., STICH and STICH II) and meta-analyses found \textbf{no statistically significant improvement in survival or functional recovery} with early open surgery compared to medical management alone}\\\\
\texttt{CDSR} \textsc{Reference Article} (\url{https://www.cochranelibrary.com/cdsr/doi/10.1002/14651858.CD015387.pub2}): \textcolor{darkgray}{For people with spontaneous supratentorial ICH, surgery aimed at clot removal may \textbf{increase the chance of achieving good functional outcome and may reduce all‐cause mortality and 30‐day case fatality} compared to standard medical management.} 
\end{tcolorbox}
\end{small}
\end{figure*}

\xhdr{Failure Mode \#2: Evidence Quality Mischaracterization}
Another common failure mode is the mischaracterization of evidence quality and certainty. Agents frequently distort the confidence level expressed in the \texttt{CDSR} reviews, for example, describing findings supported by high- or moderate-certainty evidence as ``low'' or ``very low'' certainty, or overstating weak and uncertain evidence as highly certain. Unlike factual omissions alone, these errors can mislead users on how to interpret and act upon the evidence itself. In evidence-based medicine, certainty assessments communicate how much confidence clinicians, researchers, and policymakers should place in a conclusion and whether additional evidence is likely to change the finding. Consequently, misrepresenting evidence quality may lead users to either overtrust weak findings or dismiss well-supported conclusions. Example:

\begin{figure*}[h!]
\centering
\begin{small}
\begin{tcolorbox}[colback=gray!5, colframe=red, width=\textwidth]
\textsc{Generated Conclusion} (\texttt{gemini-3-pro, tools}): \textcolor{purple}{The DASH diet is a powerful intervention for preventing the onset of cardiovascular disease, with \textbf{high-quality evidence} demonstrating it significantly outperforms standard or 'usual' dietary practices in reducing cardiovascular risk factors and incidence.}\\\\
\texttt{CDSR} \textsc{Reference Article} (\url{https://www.cochranelibrary.com/cdsr/doi/10.1002/14651858.CD013729.pub2}): \textcolor{darkgray}{The effect of the DASH diet on major cardiovascular outcomes---including myocardial infarction, stroke, cardiovascular mortality, and all‐cause mortality---\textbf{remains inconclusive} due to a lack of robust long‐term evidence... \textbf{The certainty of evidence is low to very low}, primarily due to design limitations such as high risk of bias, small sample sizes, and short follow‐up periods in the included trials.}

\end{tcolorbox}
\end{small}
\end{figure*}

\xhdr{Failure Mode \#3: Lack of Specificity}
Another common failure mode is overly general and non-specific synthesis relative to the \texttt{CDSR} reviews. Agents often collapse nuanced outcome-level findings into overly broad review-level summaries, failing to preserve important distinctions in effects and certainty estimates across different outcomes and population groups. For example, AI agents often emphasize a few salient outcomes (e.g., mortality, pain reduction, primary endpoints) while omitting secondary but clinically important outcomes (e.g., quality of life, functional outcomes). As a result, generated conclusions may appear fluent and broadly correct while still failing to communicate these essential details for comprehensive scientific interpretation and real-world clinical decision-making, particularly when weighing trade-offs, patient-centered outcomes, and downstream risks. Example: 

\begin{figure*}[h!]
\centering
\begin{small}
\begin{tcolorbox}[colback=gray!5, colframe=red, width=\textwidth]
\textsc{Generated Conclusion} (\texttt{sonar-reasoning-pro}, tools + clean-room): \textcolor{purple}{Multimodal health behavior-changing interventions targeting children under 10 years with obesity produce modest but clinically meaningful reductions in BMI and obesity prevalence... These interventions consistently improve secondary outcomes including physical activity levels, dietary habits, and obesity-related knowledge...}\\\\
\texttt{CDSR} \textsc{Reference Article} (\url{https://www.cochranelibrary.com/cdsr/doi/10.1002/14651858.CD016063}): \textcolor{darkgray}{For children under 10 years living with obesity, multimodal health behaviour-changing interventions may slightly improve \textbf{health-related quality of life}...}

\end{tcolorbox}
\end{small}
\end{figure*}

\subsection{\camready{Diagnostic Analysis of Tool Use, Retrieval, and Synthesis}}\label{appendix:diagnostic-analysis}
\camready{To understand where model failures arise in the task, from retrieval to synthesis, we follow \citet{kirgis2026log} and qualitatively analyze 20 randomly sampled model logs.}

\xhdr{\camready{Synthesis Failures}} \camready{The most common and serious failures occur during synthesis, appearing in 18 of 20 logs (90\%). Although stronger models generally retrieve relevant evidence, albeit with incomplete coverage, they exhibit two failure modes beyond those in \S\ref{appendix:failure-mode}: (1) \textit{overstated certainty}, converting conclusions of ``insufficient'' or ``very uncertain'' evidence into claims of effectiveness, even when the ground-truth \texttt{CDSR} review itself had been retrieved; and (2) \textit{fabricated statistics}, reporting precise effect sizes or risk reductions that appear in no retrieved source.}

\xhdr{\camready{Retrieval and Source-Selection Failures}} \camready{These appear in 6 of 20 logs (30\%) and often propagate into the final synthesis: (1) \textit{poor source prioritization}, browsing older studies over newer, higher-quality evidence; and (2) \textit{scope mismatch}, selecting studies outside the target PICO or review eligibility criteria, yielding overly broad conclusions unsupported by the reference review.}

\xhdr{\camready{Tool-Use Imbalance: A Case Study of \texttt{DR Tulu}}} \camready{We separately examine why the specialized deep-research model \texttt{DR Tulu} performs poorly on \textsc{SciConBench} in \S\ref{sec:benchmark-result}. About 93\% of \texttt{DR Tulu}'s tool calls use \texttt{paper\_search} and only 0.4\% use \texttt{web\_browse} (Table~\ref{tab:tool-calls-detailed}), so it relies primarily on short snippets rather than full webpages or articles. In contrast, stronger agents such as \texttt{o3-deep-research} distribute tool use more evenly, suggesting that balanced tool use may improve evidence access and coverage.}

\xhdr{\camready{Implications}} \camready{These qualitative analyses indicate that reliable scientific conclusion synthesis depends on the entire pipeline (tool use, retrieval, source selection, evidence coverage, and synthesis), as upstream failures can degrade the final synthesized conclusion. One potential post-training direction is to refine synthesis by requiring models to preserve the certainty language of systematic reviews, distinguish eligible from out-of-scope evidence, and avoid reporting statistics not explicitly supported by retrieved sources.}

\clearpage

\section{Details on Auditing Consumer-Facing Agents}\label{appendix:audit-details}

We audit three consumer-facing AI agents on commercial platforms: Google AI Overview, Google AI Mode, and OpenEvidence. Each system was queried over our benchmark set of $N = 268$ samples. Each question was augmented with a standardized benchmark suffix instructing the system to synthesize a paragraph-length conclusion drawing on the highest-quality and most up-to-date evidence, explicitly discussing strengths, limitations, uncertainty, and contradictions across the body of evidence, with the conclusion paragraph delimited by triple square brackets for downstream extraction. See Figure \ref{fig:audit-prompt} for the prompt used to standardize query formatting. We detail the data collection below: 

\xhdr{Google AI Overview \& AI Mode} For both Google AI Overview and AI Mode, we collect their synthesized conclusions using the SerpAPI library\footnote{\url{https://serpapi.com/}}. Since Google does not consistently generate an AI Overview for every query on the first attempt, the pipeline retried the same query up to three times, with a 15-second wait in between before recording a null response. 

\xhdr{OpenEvidence} As a proprietary platform for clinicians, OpenEvidence does not provide a public API. We therefore collect responses via automated browsing with SeleniumBase, issuing queries and scraping the generated conclusions. To respect rate limits, we introduce delays of up to 265 seconds between queries, allowing sufficient time for response generation. Given automated browsing, OpenEvidence did not consistently return a response; the pipeline therefore retried each query before recording a null output.

\begin{figure}[h]
\begin{small}
\centering
\begin{tcolorbox}[
    colback=gray!5,
    colframe=black,
    width=\linewidth,
    arc=2mm,
    boxrule=0.5pt,
    title=\textbf{Audit Prompt}
]
\{question\}

Synthesize a paragraph-long conclusion using the highest-quality and most up-to-date scientific evidence available, and explicitly discuss the strengths, limitations, uncertainty, and contradictions across the body of evidence. Wrap the conclusion paragraph in three square brackets.
\end{tcolorbox}

\caption{Prompt for auditing consumer-facing agents: Google AI Overview, Google AI Mode, and OpenEvidence. For each of the $N=268$ queries, we replace the \{question\} placeholder with the query.}
\label{fig:audit-prompt}
\end{small}
\end{figure}

\clearpage


\input{checklist.tex}

\end{document}

%% file: checklist.tex
\section{NeurIPS Paper Checklist}

\begin{enumerate}

\item {\bf Claims}
    \item[] Question: Do the main claims made in the abstract and introduction accurately reflect the paper's contributions and scope?
    \item[] Answer: \answerYes{} 
    \item[] Justification: The main claims made in the abstract and introduction accurately reflect the main paper's contributions and scope.

\item {\bf Limitations}
    \item[] Question: Does the paper discuss the limitations of the work performed by the authors?
    \item[] Answer: \answerYes{} 
    \item[] Justification: We include a thorough limitation discussion in \S\ref{appendix:discussion}.

\item {\bf Theory assumptions and proofs}
    \item[] Question: For each theoretical result, does the paper provide the full set of assumptions and a complete (and correct) proof?
    \item[] Answer: \answerNA{} 
    \item[] Justification: There are no theoretical results in this paper.

    \item {\bf Experimental result reproducibility}
    \item[] Question: Does the paper fully disclose all the information needed to reproduce the main experimental results of the paper to the extent that it affects the main claims and/or conclusions of the paper (regardless of whether the code and data are provided or not)?
    \item[] Answer: \answerYes{} 
    \item[] Justification: The paper fully discloses all the information needed to reproduce the main experimental results of the paper, and we release the full benchmark dataset and code to assist the reproducibility of our experimental results. The Appendix sections contain all the necessary details for reproducing our results. 

\item {\bf Open access to data and code}
    \item[] Question: Does the paper provide open access to the data and code, with sufficient instructions to faithfully reproduce the main experimental results, as described in supplemental material?
    \item[] Answer: \answerYes{} 
    \item[] Justification: This paper provides open access to the data and code, including instructions and sample code to run various modules in the paper. We will include links to the dataset collection and our code in the first page of the abstract. 

\item {\bf Experimental setting/details}
    \item[] Question: Does the paper specify all the training and test details (e.g., data splits, hyperparameters, how they were chosen, type of optimizer) necessary to understand the results?
    \item[] Answer: \answerYes{} 
    \item[] Justification: All appendix sections contain all the necessary experimental details. 

\item {\bf Experiment statistical significance}
    \item[] Question: Does the paper report error bars suitably and correctly defined or other appropriate information about the statistical significance of the experiments?
    \item[] Answer: \answerYes{} 
    \item[] Justification: We report statistical significance analyses of factual F1 differences between the best- and second-best-performing agents in \S\ref{sec:benchmark-result}, with power analyses in \S\ref{appendix:power-analysis}.

\item {\bf Experiments compute resources}
    \item[] Question: For each experiment, does the paper provide sufficient information on the computer resources (type of compute workers, memory, time of execution) needed to reproduce the experiments?
    \item[] Answer: \answerYes{} 
    \item[] Justification: We include all experiments, compute, API costs, and human annotation resources in the Appendix, covering all resources used for preprocessing \textsc{SciConBench}, generating model response, and measuring factual precision and recall. 
    
\item {\bf Code of ethics}
    \item[] Question: Does the research conducted in the paper conform, in every respect, with the NeurIPS Code of Ethics \url{https://neurips.cc/public/EthicsGuidelines}?
    \item[] Answer: \answerYes{} 
    \item[] Justification: We confirm the research conducted in the paper conform, in every respect, with the NeurIPS Code of Ethics. 

\item {\bf Broader impacts}
    \item[] Question: Does the paper discuss both potential positive societal impacts and negative societal impacts of the work performed?
    \item[] Answer: \answerYes{} 
    \item[] Justification: We discuss broader impacts in \S\ref{appendix:discussion}.
    
\item {\bf Safeguards}
    \item[] Question: Does the paper describe safeguards that have been put in place for responsible release of data or models that have a high risk for misuse (e.g., pre-trained language models, image generators, or scraped datasets)?
    \item[] Answer: \answerYes{} 
    \item[] Justification: We discuss safeguards that have been put in place for responsible release of data in \S\ref{appendix:discussion}.

\item {\bf Licenses for existing assets}
    \item[] Question: Are the creators or original owners of assets (e.g., code, data, models), used in the paper, properly credited and are the license and terms of use explicitly mentioned and properly respected?
    \item[] Answer: \answerYes{} 
    \item[] Justification: The creators or original owners of assets used in the paper are properly credited and are respected for the license (e.g., \texttt{CDSR} uses license CC-BY-NC 4.0) and terms of use explicitly mentioned. See \S\ref{appendix:ethics}.

\item {\bf New assets}
    \item[] Question: Are new assets introduced in the paper well documented and is the documentation provided alongside the assets?
    \item[] Answer: \answerYes{} 
    \item[] Justification: We document all assets.

\item {\bf Crowdsourcing and research with human subjects}
    \item[] Question: For crowdsourcing experiments and research with human subjects, does the paper include the full text of instructions given to participants and screenshots, if applicable, as well as details about compensation (if any)? 
    \item[] Answer: \answerYes{} 
    \item[] Justification: We include details for human annotations in \S\ref{appendix:question-gen-validation}, \S\ref{appendix:afg-validation}, and \S\ref{appendix:fact-eval-annotation}. 

\item {\bf Institutional review board (IRB) approvals or equivalent for research with human subjects}
    \item[] Question: Does the paper describe potential risks incurred by study participants, whether such risks were disclosed to the subjects, and whether Institutional Review Board (IRB) approvals (or an equivalent approval/review based on the requirements of your country or institution) were obtained?
    \item[] Answer: \answerNA{} 
    \item[] Justification: Our human annotation is innocuous and thus does not require IRB approval.

\item {\bf Declaration of LLM usage}
    \item[] Question: Does the paper describe the usage of LLMs if it is an important, original, or non-standard component of the core methods in this research? Note that if the LLM is used only for writing, editing, or formatting purposes and does \emph{not} impact the core methodology, scientific rigor, or originality of the research, declaration is not required.
    \item[] Answer: \answerYes{} 
    \item[] Justification: We describe how LLMs are used to construct \textsc{SciConBench} (\S\ref{sec:sciconbench}), generate conclusions with \textsc{SciConHarness} (\S\ref{sec:clean-room}), and evaluate the factual quality of generated conclusions (\S\ref{sec:metrics}).

\end{enumerate}